\documentclass[12pt]{book}

\usepackage[table]{xcolor}

\usepackage[a-1b]{pdfx} 

\usepackage{pdfpages}

\newdimen \jot \jot=5mm
\usepackage[T1]{fontenc}
\usepackage{lmodern} 
\usepackage{times}
\usepackage{tocbibind} 
\usepackage{setspace}
\usepackage{blindtext,mdframed,pifont}
\usepackage{amsmath,amssymb,array}
\usepackage{bbm}
\usepackage{booktabs}
\usepackage{adjustbox}

\usepackage{graphicx}
\usepackage{float}
\usepackage{tikz,pgfplots,tikzscale}
\usepackage{graphics} 
\usetikzlibrary{bayesnet,shapes,arrows,shadows,fit,positioning,decorations.pathmorphing}
\usepackage{dcolumn}
\usepackage{caption}
\usepackage{subcaption}
\usepackage{capt-of}
\usepackage{latexsym}
\usepackage{colortbl}
\usepackage{tabularx}
\usepackage{multirow}
\usepackage{algorithm}
\usepackage{algorithmic}
\usepackage{framed}
\usepackage{enumitem}
\usepackage{longtable}

\usepackage{nicefrac}       
\usepackage{microtype}      
\usepackage{wrapfig}
\usepackage[most]{tcolorbox}

\usepackage[toc,page]{appendix}

\usepackage{footnote}
\makesavenoteenv{tabular}
\makesavenoteenv{table}

\usepackage[
style = authoryear,
sorting = nty,
maxbibnames = 399,
backend = bibtex,
natbib = true
]{biblatex}
\bibliography{concatenated}

\newcommand*{\blankpage}{%
\vspace*{\fill}
{\centering Intended to be blank.\par}
\vspace{\fill}}

\renewcommand*{\cleardoublepage}{\clearpage\if@twoside \ifodd\c@page\else
\blankpage
\thispagestyle{empty}
\newpage
\if@twocolumn\hbox{}\newpage\fi\fi\fi}

\AtEveryBibitem{
\clearfield{note}
\clearfield{month}
}

\let\oldInclude=\include
\def\include#1{\bgroup\def\clearpage{\relax}\oldInclude{#1}\egroup}

\newcolumntype{R}[2]{%
    >{\adjustbox{angle=#1,lap=\width-(#2)}\bgroup}%
    l%
    <{\egroup}%
}
\newcolumntype{P}[1]{>{\centering\arraybackslash}p{#1}}
\newcolumntype{M}[1]{>{\centering\arraybackslash}m{#1}}
\newcolumntype{N}{@{}m{0pt}@{}}

\newcommand{\citetemp}[1]{\textbf{\texttt{(#1)}}}

\newcommand*\rot{\multicolumn{1}{R{45}{.01em}}}

\newcommand{\Comment}[1]{\hfill {\footnotesize $\lhd$ #1}}

\usepackage{todonotes}
\newcommand{\Note}[2]{}
\renewcommand{\Note}[2]{\todo[color=#1,size=\small, inline=true]{#2}} 

\usepackage{soul}
\newcommand{\NoteSoul}[1]{}
\definecolor{brilliantlavender}{rgb}{0.96, 0.73, 1.0}
\definecolor{thistle}{rgb}{0.847, 0.749, 0.847}

\newcommand{\removed}[1]{}

\DeclareMathOperator*{\E}{\mathbb{E}}

\definecolor{warningbackground}{RGB}{252,226,158}

\newcommand{\LR}{LR}
\newcommand{\STL}{STL}

\newcommand{\MTL}{MTL}

\newcommand{\dgcca}{\emph{dGCCA}} 
\newcommand{\dgccaandbow}{\emph{dGCCA + BOW}} 
\newcommand{\lascca}{\emph{LasCCA}} 
\newcommand{\wgcca}{\emph{wGCCA}} 
\newcommand{\gcca}{\emph{GCCA}} 
\newcommand{\gccasv}{\emph{GCCA-sv}} 
\newcommand{\gccasvandbow}{\emph{GCCA-sv + BOW}}
\newcommand{\gccaandbow}{\emph{GCCA + BOW}}
\newcommand{\gccawnet}{\emph{GCCA-net}} 
\newcommand{\bow}{\emph{BOW}} 
\newcommand{\bowpca}{\emph{BOW-PCA}} 
\newcommand{\bowpcawnet}{\emph{NetSim-PCA}} 
\newcommand{\bowpcaandbow}{\emph{BOW-PCA + BOW}}
\newcommand{\wordtovec}{\emph{Word2Vec}}
\newcommand{\collab}{\emph{NetSim}} 
\newcommand{\rand}{\emph{Random}} 
\newcommand{\netpop}{\emph{NetSize}} 

\newcolumntype{H}{>{\setbox0=\hbox\bgroup}c<{\egroup}@{}}

\newcommand{\LDA}{{\sc LDA}}

\newcommand{\dsprite}[0]{\emph{dDMR}}
\newcommand{\ddmr}[0]{\emph{dDMR}}

\newcommand{\dmr}[0]{\emph{DMR}}
\newcommand{\dmrpca}[0]{\emph{DMR-PCA}}

\newcommand{\lda}[0]{\emph{LDA}}

\newcommand{\redden}[1]{\textcolor{red}{#1}}

\DeclareMathOperator{\tr}{Tr}

\DeclareMathOperator*{\argmax}{arg\,max}
\DeclareMathOperator*{\argmin}{arg\,min}

\DeclareRobustCommand{\rchi}{{\mathpalette\irchi\relax}}
\newcommand{\irchi}[2]{\raisebox{\depth}{$#1\chi$}} 

\tcbset{
    frame code={}
    center title,
    left=0pt,
    right=0pt,
    top=0pt,
    bottom=0pt,
    colback=gray!30,
    colframe=black,
    width=\dimexpr\textwidth\relax,
    enlarge left by=0mm,
    boxsep=5pt,
    arc=0pt,outer arc=0pt,
}

\usepackage[hypcap=true]{caption}
\hypersetup{linktocpage}

\begin{document}
\pagestyle{plain}
\pagenumbering{roman}
\setcounter{page}{1}

\newcommand{\bm}[1]{ \mbox{\boldmath $ #1 $} }
\newcommand{\bin}[2]{\left(\begin{array}{@{}c@{}} #1 \\ #2
             \end{array}\right) }
\renewcommand{\contentsname}{Table of Contents}

\baselineskip=24pt

\pagenumbering{gobble}
\thispagestyle{empty}
\begin{center}
\vspace*{.25in}
{\bf\LARGE{ Learning Representations of Social Media Users }}\\ 
\vspace*{.75in}
by \\*[18pt]
\vspace*{.2in}
Adrian Benton\\ 
\vspace*{1in}
A dissertation submitted to The Johns Hopkins University\\
in conformity with the requirements for the degree of\\
Doctor of Philosophy\\
\vspace*{.75in}
Baltimore, Maryland \\
October, 2018 \\     
\vspace*{.5in}
\begin{small}
\copyright{ }2018 by Adrian Benton \\ 
All rights reserved
\thispagestyle{empty}
\end{small}
\end{center}

\chapter*{Abstract}

\pagenumbering{roman}
\setcounter{page}{1}
\thispagestyle{empty}

Social media users routinely interact by posting text
updates, sharing images and videos, and establishing connections with other
users through friending.
User representations are routinely used in recommendation systems by
platform developers, targeted advertisements by marketers, and by
public policy researchers to gauge public opinion across demographic
groups.  Computer scientists consider the problem of inferring user
representations more abstractly; how does one extract a stable user
representation -- effective for many downstream tasks -- from a medium
as noisy and complicated as social media?

The quality of a user representation is ultimately
task-dependent (e.g. does it improve classifier performance, make
more accurate recommendations in a recommendation system)
but there are also proxies that are less sensitive to the specific task.
Is the representation predictive of latent properties such as a
person's demographic features, socio-economic class, or mental health
state?  Is it predictive of the user's future behavior?

In this thesis, we begin by showing how user representations
can be learned from multiple types of user behavior on social
media.  We apply several extensions of generalized canonical
correlation analysis
to learn these representations and evaluate them at three tasks:
predicting future hashtag mentions, friending behavior, and demographic
features.  We then show how user features can be employed as distant
supervision to improve topic model fit.  We extend a standard supervised
topic model, Dirichlet Multinomial Regression (DMR), to make better use
of high-dimensional supervision.  Finally, we show how user
features can be integrated into and improve existing classifiers in
the multitask learning framework.  We treat user representations --
ground truth gender and mental health features -- as
auxiliary tasks to improve mental health state prediction.  We also use
distributed user representations learned in the first chapter to improve
tweet-level stance classifiers, showing that distant user information
can inform classification tasks at the granularity of a single message.

{\bf Committee:}

Mark Dredze

Raman Arora

David Yarowsky

Dirk Hovy

\cleardoublepage

\chapter*{Acknowledgments}

I owe the completion of this document, and the years of work poured into
it to the mentorship, collaboration, and friendship of many people.
First, I am grateful to my committee for reading this document and
giving me feedback after my oral exam.

Raman Arora mentored me for the multiview representation learning in
this thesis.  I value his patient explanations of basic linear algebra
concepts and for sharing his interpretations of multiview
representation learning techniques.
Dirk Hovy showed me how good research should be presented, both
as publications and talks.  I appreciate his faith in my coding ability,
his camaraderie, and his devotion to freshly-baked bread.
David Yarowsky convinced me to come to Hopkins with a hard pitch for JHU
during a meal at the Helmand.  The pitch was effective.

Despite being my first and only Ph.D. advisor, I can confidently say that
Mark Dredze is an excellent advisor.  He is inexplicably optimistic when
presented with lukewarm results and has always given me the freedom to
pursue questions that interest me.  I value the time he gave me to implement
and understand the models I worked with, and the opportunity to help manage
other student projects. I am grateful to him not only because he is my advisor,
but also because he is a pretty good one.  I hope that one day Mark will have a
bubble soccer-amenable lab.

The CLSP is one of the strongest NLP groups in the world and the quality
of the other students and faculty continually impressed me; it was an honor
to be accepted as a student here.
Adam Teichert and Michael Paul were responsible for mentoring me on
several projects and gave me a crash course in machine learning debugging.
Michael Paul, in particular, taught me the dark art of repairing a
faulty Gibbs sampler and to rejoice when a bug is spotted.  I also worked closely
with Huda Khayrallah while working on applying deep generalized
CCA.  I appreciate our conversations and her witty reflections on grad school
life.  Her resourcefulness was also critical in salvaging my thesis defense.

My decision to pursue a Ph.D. was encouraged by many researchers at the
University of Pennsylvania including Lyle Ungar and John H. Holmes.
I am particularly grateful to Delphine Dahan for trusting a lowly
undergraduate with a handful of computer science classes under his
belt to join her lab, and Shawndra Hill with whom I had the closest working
relationship while at Penn.  Shawndra is an example of what true
dedication to research looks like, always insightful and indefatigable.  She
continually drove me to expect more of myself.

I owe my persistence in the Ph.D. program in large part to the
2016 JSALT Social Media/Mental Health summer workshop group: Kristy,
Andy, Glen, Meg, Jeff, Fatemeh, Leo, and Bu Sun.  Joining this group
was the best academic decision I made in the past five years.  Not only
was this a productive summer, it was also the most enjoyable -- a
refreshing blend of research and socializing.  I am not sure how I would
have gotten past the third-year doldrums without them.

Leo Razoumov mentored me while interning at Amazon Research
and gave me the freedom to craft my own project with few constraints.  He
was an extremely knowledgeable sounding board in all things
mathematically rigorous and career-related.  He has been a kind and
supportive friend and I hope to see more of him in New York.

Many non-research relationships sustained me as well.  My chess team has
been a fundamental constant in my life.  My parents have always supported
my education and pushed me to always do my best.  The O.S. \& Spike Wright
foundation sent pounds of trail mix and stacks of dog photos in support of
thesis writing.  The foundation's support was an essential component to
thesis completion.

Most of all, I owe this whole grad school stint to Kika's patience,
emotional support, and ultimately gentle nudging towards graduation.
Thank you for sticking around this long.  I love you and I hope you
decide to stick around a while longer.

\cleardoublepage


\baselineskip=24pt

\tableofcontents
\pagebreak

\listoftables
\pagebreak

\listoffigures

\pagebreak
\pagenumbering{arabic}

\chapter{Introduction}
\label{chap:intro}

\section{Motivation}

Social media platforms offer researchers and data scientists
a massive source of user-generated data including not only what users say, but
who they are friends with, their self-reported descriptions, and which
posts they like. Social media data is valuable to two major groups of stakeholders:
{\bf Technologists} and social science {\bf Researchers}.
\textbf{Technologists} are focused on engineering and are concerned with either maintaining and augmenting 
the social media platforms themselves, or
building tools that perform well on social media data.  Social science \textbf{researchers} treat social media data as a lens on society, an imperfect version of how humans communicate with each other naturally.  They use social media data to answer deep questions about people and the world at large, and only care about building strong tools insofar as these tools can help judge hypotheses.  Each of these groups has a different set of tasks to complete and questions they want to answer.

\paragraph*{Technologists}

Maintainers of social media platforms routinely use user-generated data to improve
their products.  These include improving the platform's friend recommendation
system \parencite{hannon2010recommending,kywe2012survey,konstas2009social} and content
recommendation or feed optimization
\parencite{kramer2014experimental,chen2012collaborative,yan2012tweet,guy2010social}.  These
features are tuned to retain users and increase the ``addictiveness'' of the platform.
Advertising revenue is the foundation of many social media platforms' business models.
Platforms such as Facebook specifically attract advertisers by using user data to better
predict advertisement click-through rate.

Natural language processing (NLP) can be conceptually decomposed into an array of subtasks
around automatically extracting information from human-generated text.  Practitioners
build tools to address each of these subtasks: part-of-speech taggers,
syntactic parsers, sentiment analyzers, semantic parsers, etc.  These tools were traditionally
trained to perform well on standard text such as newswire, and
extending them to perform well on social media posts is an active area of research
\parencite{gimpel2011part,rosenthal2017semeval,strauss2016results,daiber2016denoised}.

\paragraph*{Researchers}

Social media data can also be used to test theories of how information flows through
social networks \parencite{wu2011says}, how these networks are structured
\parencite{martin2016exploring}, and how to identify and quantify social influence
\parencite{bakshy2011everyone}.  Hypotheses which were theoretically motivated,
or were empirically validated by painstakingly compiling word-of-mouth data can
be tested at scale in observational studies on social media \parencite{jansen2009twitter}.
The persuasiveness of these observational studies hinges on arguing that the
online behavior is evidence of a causal relationship, and this causal argument often relies on
controlling for potential confounds in social media data \parencite{tan2014effect}.

Showing that social media data merely has predictive power for real-world happenings may also be sufficient
when building predictive models.
Predictive models of real-world trends such as disease incidence, stock market prices, and
sentiment on public policy issues based on messages people post to Twitter can be used as
surrogates for more traditional surveys
\parencite{tumasjan2010predicting,paul2011you,oconnor2010tweets,bollen2011twitter}.

In general, user-generated social media data is attractive to both groups for the following properties:

\begin{enumerate}
  \item Social media data can be used as a proxy for actual human interactions.  This is most
        relevant to Researchers but also to Technologists who would like to
        extend their systems to noisier, more naturally-produced language than news
        articles.
  \item The data are also multi-modal and offer several views of these interactions.  Take for
        example users' friending and messaging behavior against images or videos they post.
        Multiple views of user behavior can be used to build stronger tools,
        but can also suggest different hypotheses to test.
  \item Social media platforms have a vast user base which has engorged platform servers
        with text and other interaction data. As-of the end of 2017,
        Facebook reported over 2 billion registered users, with over 17 billion video and 94
        billion text chats initiated that year through their messaging
        feature\footnote{\url{https://newsroom.fb.com/news/2017/12/messengers-2017-year-in-review/}}.
        Models trained on many examples can generalize better to out-of-sample data, which is important
        for both Technologists and Researchers.
  \item Social media data is produced and effectively updated in realtime.  Consequently,
        models can be frequently updated to stay fresh and relevant.  Hypotheses can also be
        tested as time progresses to validate that past findings hold in the present.
\end{enumerate}

Some benefits such as quantity and rate of updating are shared with
other online data sources such as server and web search logs.
However, these other data sources do not capture natural human
interactions.

\removed{
\textbf{In General:}
The most obvious\footnote{And potentially nefarious \parencite{guardian2018cambridgeanalytica}.}
use of these data is to predict unobserved (\emph{latent}) stable user features based on what
can be observed from their online activity.  These include standard demographic features
such as gender, age, ethnicity, and socioeconomic status
\parencite{rao2010classifying,volkova2014inferring,volkova2015inferring}, but
also encompass traits such as political orientation/belief set a user
may hold \parencitetemp{TODO}, user personality type \parencite{park2015automatic}, \citetemp{TODO},
or mental condition \citetemp{TODO}.  Predicting temporary latent state such as whether
a person has an illness \citetemp{mpaul paper}, or has recently experienced a significant
life event, or their current location \citetemp{carmen}.
}

\subsection{Problems Associated with Social Media Data}

\begin{figure}
\centering
\includegraphics[height=.2\linewidth]{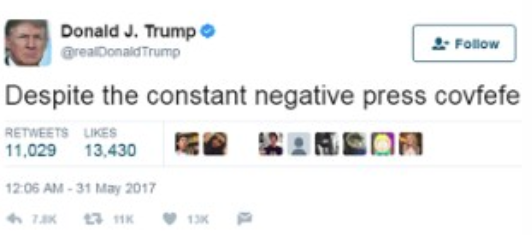}
\caption{Message on Twitter from the 45th President of the United States.}
\label{fig:introduction:covfefe}
\end{figure}

Social media data can be a source for building robust NLP tools,
predictive models of real-world trends on online activity, and
testing social scientific theories.  However, there are several fundamental
problems that anyone who wants to build tools using these data must overcome.
Figure \ref{fig:introduction:covfefe} is an example of a message posted on Twitter
(tweet) that demonstrates many of these
problems\footnote{\url{https://www.cnn.com/2017/05/31/politics/covfefe-trump-coverage/index.html}}.

\paragraph*{Feature Sparsity} NLP models often rely on token n-gram features
to make predictions.  Longer document lengths allow these models to generalize
well.  On social media, messages with only a handful of tokens are common,
leading to very sparse feature vectors.  Additional sparsity arises from the fact
that conversations on social media span many domains, even when one is restricting
to messages made in English, for instance.  This is further exacerbated by
frequent typos, butter fingers, and intentionally alternate spellings.  The vocabulary
size is larger for social media messages than restricted domains such as Wall
Street Journal articles.

\begin{figure}
\centering
\includegraphics[height=.2\linewidth]{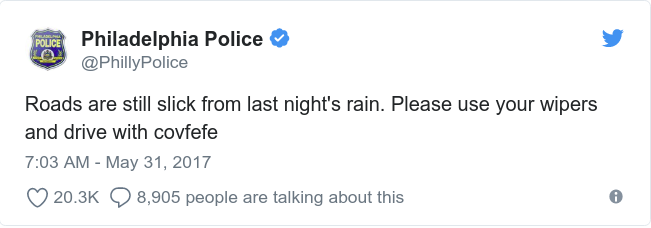}
\caption{Following tweet with the mysterious ``covfefe''.}
\label{fig:introduction:covfefe2}
\end{figure}

\paragraph*{Context} Most importantly, the \textbf{context} of this tweet is
absent from the tweet itself.  ``\emph{Despite the negative press covfefe}''
is not a complete English sentence.  However, knowing that the user posting this
tweet is the current President of the United States of America who has a confrontational history
with the press, one can infer that ``covfefe'' was meant to be the word
``coverage'' and that this was meant to be followed by some self-aggrandizing statement.
Context can also include previous activity within the social media platform.  In this
case, context can be previous messages within a discussion, or other messages posted with
similar content.  Figure \ref{fig:introduction:covfefe2} shows another message where humor is dependent
on awareness of the original presidential \emph{covfefe}.  Lack of context is a fundamental problem
in NLP, since natural language regularly refers to events and entities in the real world, but it is exacerbated on
social media primarily because of short message length.

\subsection{User Features to Alleviate Problems with Social Media}

Knowing the context around a social media message is key to understanding its
meaning.  Author demographic features, such as age, gender, and socio-economic class
are an old and critical piece of this context.  Demographic features are traditionally treated as
categorical, where users that fall in the same demographic category can be thought of as
belonging to the same ``hard cluster''. 

\paragraph*{Hard User Clusters}   Partitioning a population by some
subset of stable properties and then describing the behavior of each of these subgroups is one way to use user features to improve social media systems.
This idea has been applied in several fields:

\begin{enumerate}
  \item \textbf{Marketing}: \emph{Market segmentation}, particularly
        \emph{demographic segmentation} is a classic marketing strategy.
        Smith \parencite{smith1956product} described market segmentation in 1956, and
        contrasted this strategy with product differentiation, which refers to supply
        side heterogeneity (distinguishing one's product from the competition). Although
        Smith considered market segmentation abstractly: ``Segmentation
        is based upon developments on the demand side of the market and represents a rational
        and more precise adjustment of product and marketing effort to consumer or user
        requirements``, demographic quantities such as gender and age typically defined
        the boundaries between different market segments.  This was because these
        boundaries aligned with dominant stereotypes (e.g. men buy lawnmowers and women buy
        dish soap) and these characteristics could be reliably quantified.  This strategy
        has been transplanted to political campaigns as well --
        instead of hawking soap, campaigns sell a candidate or policy
        platform.  More fine-grained targeting has also been used to better appeal to
        consumers.  These include using psychographic properties to define groups or
        identifying consumers with specific interests
        or habits (gardeners, bicyclists, latte-drinkers), as is offered by the Facebook
        advertising platform\footnote{\url{https://www.facebook.com/business/products/ads}}.
        Nevertheless, partitioning the market into broad clusters based on a set of
        categorical indicators is the norm.
  \item \textbf{Computational social science/policy}: Social scientists and public policy researchers are
        interested in characterizing a population by different groups 
        for essentially the same reason as marketing scientists: policy opinions or beliefs are not homogeneously
        distributed throughout a population.  Public policy surveys reflect this by
        disaggregating public opinion by different groups, often along
        demographic features.  Exploration of the best subset of features to segment
        a population into homogeneous subgroups is an active area of research in public health
        \parencite{boslaugh2004comparing}.
  \item \textbf{NLP}: There has also been interest in using author features to
        improve performance at standard NLP tasks including sentiment analysis and
        part-of-speech tagging \parencite{hovy2015demographic}.  Inferring latent user features from
        social media has been thoroughly explored in NLP, either predicting typical demographic
        properties \parencite{volkova2015inferring} or less typical features such as profession
        or interests \parencite{beller2014believer}.
\end{enumerate}

\paragraph*{Problems with Using Hard User Clusters in Social Media}
Although this is an attractive solution when demographic features
are available, many social media platforms do not share
user demographics.  Even when demographic features are available,
by hand annotation for example, the labels may be influenced by the
annotators' biases (e.g. a user is coded as black because of how
they tweet) in hand annotation, or reporting bias when demographics are
self-reported.  Fitting classifiers and tuning systems on
disjoint subsets of users will not work well on small training
sets, harming performance compared to engineering systems on
the full training set.  More importantly, it is not clear a priori
which user features we should be conditioning on for each task.

\section{Proposal: User Embeddings}

We instead propose learning distributed user embeddings based
on a user's online behavior.  A user embedding,
in this context, is a vector of real numbers, that
succinctly captures properties of that user.  Users with similar
embeddings should behave similarly on social media.

The process
of learning user embeddings is inspired by work in NLP on learning
word and generally text embeddings.  Word embeddings are vector
representations of words where closeness is able to capture semantic
and syntactic relationships between words.  The key component in learning
word embeddings is that they are trained to be predictive of
the word's context, where context is defined as the identity of
surrounding words (collocations) \parencite{mikolov2013distributed}.
Surrounding words make for good context when learning word embeddings,
but there is no clear analog for ``user context''.

Social media
user activity is arguably much richer than the appearance of a
word in documents.  A user can be characterized by the friends they connect to, the messages they post, or the articles and images they share.
In order to incorporate multiple behaviors, we propose using multiview representation
learning techniques to learn user representations that capture multiple
views of online activity simultaneously.
We also learn user embeddings with standard dimensionality reduction techniques and evaluate their
effectiveness at several downstream social media tasks.

\paragraph*{Continuous Relaxations of Other Hard Clusters}
Using vector-valued user embeddings rather than categorical user features
to improve classifiers is an old idea and has analogs in
several other domains:

\begin{itemize}
  \item Recommendation systems can be broadly categorized into those that
    make recommendations based on which subgroup a user belongs to
    (content-based filtering) vs. those that make recommendations based
    on preferences of similar users, where similarity is defined as
    ``rating items similarly'' (collaborative filtering).
    Content-based methods cluster users by provided features, such as
    given demographics or stated genre preferences, and make
    recommendations based on similarity in this feature space, although
    these user representations may also be distributed (e.g. a user
    embedding learned based on a free text description field)
    \parencite{adomavicius2005toward}.  Collaborative filtering approaches
    based on factorization of a large user-item matrix are similar in
    spirit to our distributed user embeddings.
  \item In NLP, distributed word embeddings were preceded by Brown
    clusters, which are hierarchical cluster representations of words
    \parencite{brown1992class}.   Words that share more parent nodes in
    this hierarchy tend to be more similar semantically than those
    that do not.
\end{itemize}

\section{Contributions}

In this thesis we learn distributed social media user embeddings and
evaluate how well these embeddings and traditional user features improve downstream tasks.  In
the process, we present machine learning models to both learn and
use user features.
There are two main thrusts of this thesis: methodological contributions in
learning user embeddings, and evaluating these user features at improving
downstream tasks.

We learn the following user embeddings:

\begin{itemize}
  \item Principal component analysis (PCA) embeddings of the
        \emph{ego} user's message text.  We also consider PCA reductions of
        different user activity views such as the \emph{friend},
        \emph{follower}, \emph{mention}ed user networks, as well as reductions of the
        text of those groups.
  \item Generalized canonical correlation analysis (GCCA) derived
        embeddings, where user views are different types of user activity.
        We present two novel, orthogonal extensions of GCCA: to
        discriminatively weight the reconstruction error of each view,
        and to learn nonlinear transformations from observed to
        latent space.
\end{itemize}

We evaluate user features and pretrained user embeddings at improving performance
at the following tasks:

\begin{itemize}
  \item {\bf User-level hashtag prediction}: Predicting whether or not a Twitter user will use a particular hashtag in a future tweet.
  \item {\bf Friend recommendation}: Predicting whether Twitter users have established a friend relationship with each other.
  \item {\bf Demographic prediction}: Predicting users' age, gender, or political affiliation.
  \item {\bf Topic model fit}: Improving the quality and fit of supervised topic models on corpora of social media posts.
  \item {\bf Tweet-level stance classification}: Predicting the opinion expressed in a tweet with respect to a specific issue.
\end{itemize}

\begin{figure}
\centering
\includegraphics[width=.8\textwidth,trim={0 3cm 2cm 0},clip]{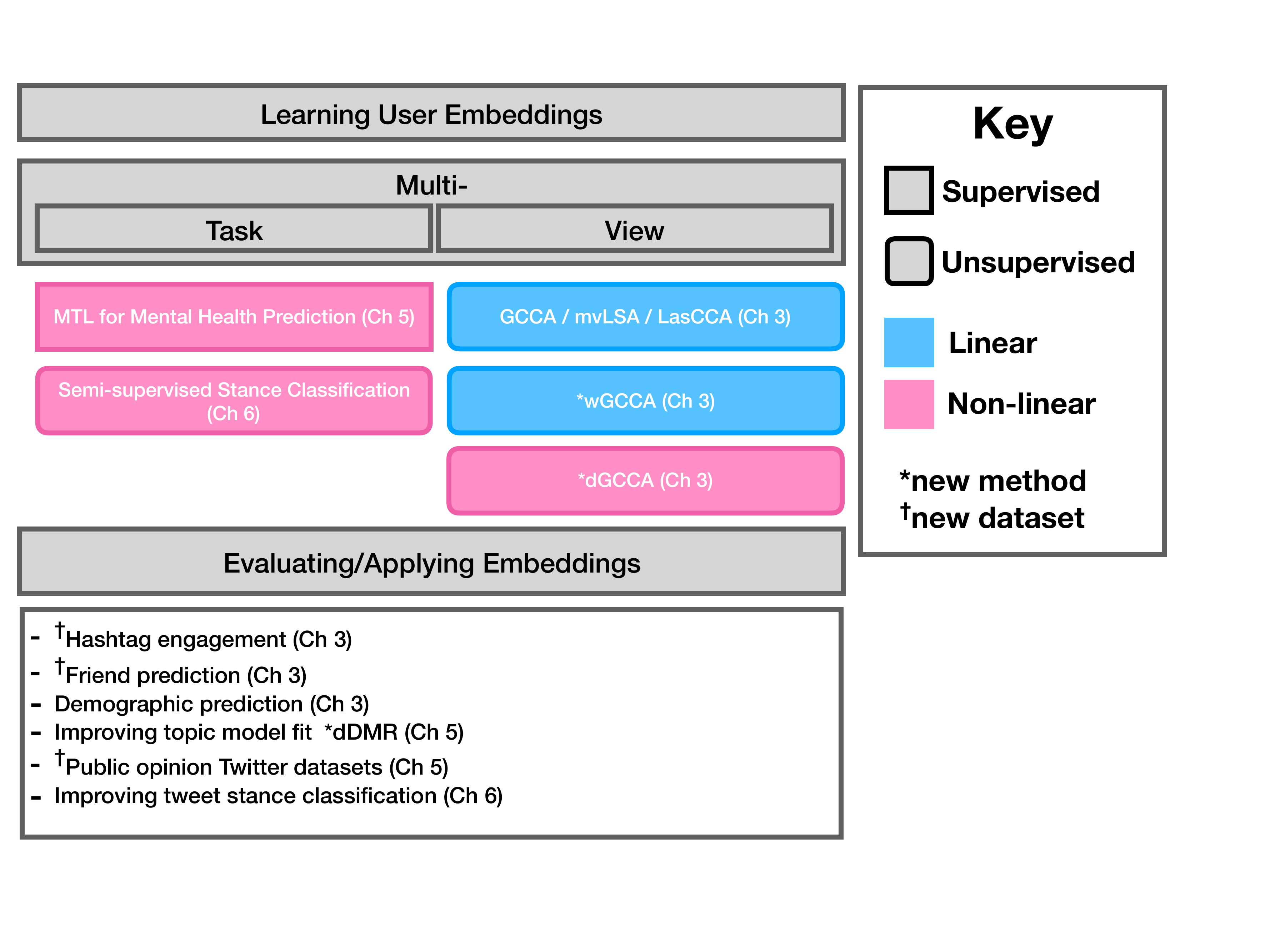}
\caption{Classification scheme of methods we explore for learning user embeddings and how they are evaluated.}
\label{fig:introduction:roadmap}
\end{figure}

\removed{
In the process, we developed the following new models 

\begin{enumerate}
  \item Learn user embeddings on Twitter data according to multiple methods.
  \begin{itemize}
    \item CCA methods
    \item supervised, neural (MTL)
    \item seq2seq/adversarial-trained embeddings conditioned on user profile
  \end{itemize}
  \item Intrinsic + extrinsic evaluation of user embeddings
  \begin{itemize}
    \item Hashtag engagement + friend prediction
    \item Predicting demographic features
    \item Efficacy as priors for topic modeling
    \item Use to generate synthetic examples for small data classifier training for short texts
    \item Compare to baseline representations such as demographic features inferred by
          automatic tools and other embedding methods (word2vec, PCA, autoencoder)
  \end{itemize}
  \item Models
  \begin{itemize}
    \item Many-view canonical correlation analysis: discriminatively weight views (wGCCA),
          neuralize to jointly learn nonlinear maps along with shared embedding / break strong
          data distribution assumptions of CCA (DGCCA)
    \item Topic model: neuralize prior to condition topic distribution on arbitrary,
          high-dimensional metadata (dDMR)
  \end{itemize}
\end{enumerate}
}

\paragraph*{Model Implementations:}
A main contribution of this thesis are the publicly released implementations of
models presented here.
Many of these methods cannot be applied na\"{i}vely to real-world datasets --
scaling them up to the number of examples and feature dimensionality
present in real-world datasets, accounting for missing data, and differentially
weighting the importance of views are all non-trivial challenges.  Table
\ref{tab:introduction:method_implementations} lists the method implementations we
have made public.

\begin{table}
  \small
  \begin{center}
    \begin{tabular}{|c|p{1.25cm}|p{4.25cm}|p{5.6cm}|} \hline
    \bf Method & \bf Model Class & \bf Features & URL \\ \hline
    \hline
    \wgcca{} & multiview method & Supports missing data per example-view and weighting of views & \url{https://github.com/abenton/wgcca} \\ \hline
    \lascca{} & multiview method & Supports missing data per example-view & \url{https://github.com/abenton/MissingView-LasCCA} \\ \hline
    \dgcca{} & multiview method & Supports missing data per example-view and learn neural embedding of views & \url{https://bitbucket.org/adrianbenton/dgcca-py3} \\ \hline
    \ddmr{} & topic model & Learn neural embedding of document supervision & \url{https://github.com/abenton/deep-dmr} \\ \hline
\end{tabular}
\end{center}
  \caption{ Publicly-released implementations of methods presented in this thesis.
    The \textbf{Features} column highlights contributions of this implementation over
    existing implementations.  The \textbf{URL} where each implementation is
    currently hosted is noted in the rightmost column.}
\label{tab:introduction:method_implementations}
\end{table}

Since publication, these implementations have been used by researchers to
learn relationships between many different kinds of data sources such as substance
abuse and social media language \parencite{ding2017multi}, speech and
cognitive impairment features\footnote{The \wgcca{} implementation was
  shared with the
  \emph{Remote Monitoring of Neurodegeneration through Speech} team at the
  Third Frederick Jelinek Memorial Summer Workshop (JSALT 2016).  The \dgcca{}
  implementation was also extended and applied by the
  \emph{Grounded Sequence to Sequence Transduction} team at JSALT 2018.},
as well as to learn multimodal representations of video \parencite{tsai2017detecting}.

\section{Overview}

We present several methods for learning embeddings and evaluate
according to many objectives.  Figure \ref{fig:introduction:roadmap}
classifies the different methods we explore, evaluation tasks we
perform, and what function each serves in this thesis.
Newly-developed models and new datasets are denoted by $^*$ and
$^\dagger$, respectively.

Chapter \ref{chap:background} contains background on user
features and embeddings in social media research as well as
the methods we use in this thesis to learn and utilize
learned user embeddings: multiview representation learning and
multitask learning.

Chapter \ref{chap:mv_twitter_users} describes how user embeddings
can be learned by unsupervised multiview learning techniques, and
analyzes the efficacy of different views on downstream embedding
performance in predicting which hashtags a Twitter user will
mention, who they will friend, and their demographic features.
We present an extension of generalized	
canonical correlation analysis (GCCA): weighted GCCA to learn user embeddings.
Weighted GCCA and the demographic
prediction experiments were presented in \textcite{benton2016learning},
published as a short paper in the Proceedings of the
Conference of the Association for Computational Linguistics (ACL) in 2016.
The experiments with deep GCCA were presented in \textcite{benton2017deep},
an ar$\rchi$iv preprint.
Robust LasCCA was developed and implemented during an internship at Amazon Research.

In Chapter \ref{chap:user_conditioned_topicmodels},
we evaluate using (distant) author-level features to better fit
topic models to social media messages.  We then
describe a supervised topic model that can make effective use
of user embeddings to better fit social media text, in lieu of
explicit author features: deep Dirichlet Multinomial Regression.
This model was originally presented in \textcite{benton2018deep},
published as a long paper in the Proceedings of the 2018 Conference
of the North American Chapter of the Association for Computational
Linguistics (NAACL).
The distant author feature topic model experiments were presented in
\textcite{benton2016collective}, a long paper published in the Proceedings
of the Thirtieth AAAI Conference on Artificial Intelligence (AAAI), 2016.

Chapter \ref{chap:mtl_mentalhealth} describes work in leveraging several
Twitter user mental conditions to better predict suicide risk
from their social media posts.  It also considers how including user
features such as demographics as an auxiliary task can improve
mental condition prediction.  This work was published in
\textcite{benton2017multitask}, a long paper in the Proceedings of
the European Chapter of the Association for Computational
Linguistics (EACL), 2017.

Chapter \ref{chap:mtl_stance} describes a final application of
embeddings where we show how the embeddings learned in chapter
\ref{chap:mv_twitter_users} can be used to learn stronger tweet
stance classification models in a multitask learning framework.
Neural classifiers can be pretrained to predict generic user embedding
features for a general set of users before being finetuned on
a specific task.  This can alternately be read as an
extension of chapter \ref{chap:mtl_mentalhealth} to treating learned
user embeddings as auxiliary tasks, not categorical supervision.
This work was presented at the 4th Workshop on Noisy
User-Generated Text (W-NUT) \parencite{benton2018using}.

Chapter \ref{chap:conclusion} summarizes the contribution of each
chapter and provides direction for future work.

\cleardoublepage

\chapter{Background}
\label{chap:background}

This chapter begins in Section \ref{sec:background:user_application_motivation}
with a discussion of existing work in applying user features to improve
downstream systems, followed by sections on methods we will use to learn
user representations and integrate them into existing models.
Section \ref{sec:background:mv_rep_learning} can be read as a
primer on multiview representation learning that covers the
basics of canonical correlation analysis and extensions to
more than two views and nonlinear mappings.  We primarily use these methods to learn
user embeddings in Chapter \ref{chap:mv_twitter_users}.
Section \ref{sec:background:mtl_neural} finally describes the multitask
learning paradigm, which is used to inject user information into trained models
in Chapters \ref{chap:mtl_mentalhealth} and \ref{chap:mtl_stance}.

\section{Applications of User Features}
\label{sec:background:user_application_motivation}

User features and representations have been shown to help in a variety of
downstream tasks.  Here we give a selection of tasks that benefit from stronger
information about the user.

\subsection{Inferring Latent User Features}

We rarely have direct access to latent user features such as gender,
personality, socioeconomic class, or political affiliation.  Models that
can infer these traits are particularly desirable since the demographics predictions
can be used as proxies for many different kinds of behaviors we may want to predict.

\paragraph*{Demographics}

Most work predicting latent social media user features has been focused on predicting
user demographic features \parencite{volkova2015inferring}.  These include
predicting properties such as age \parencite{rao2010classifying,nguyen2011author,nguyen2013old},
gender
\parencite{zamal2012homophily,culotta2016predicting}, race or ethnicity
\parencite{pennacchiotti2011machine,preotiuc2018user}, socioeconomic category
\parencite{volkova2015inferring,culotta2016predicting}, and political affiliation
\parencite{pennacchiotti2011machine,cohen2013classifying,volkova2014inferring}.
The standard approach is to train a supervised model to predict one of these 
properties given a corpus of annotated, observed user data.
These models either treat the demographic prediction problem as regression
\parencite{nguyen2011author} (e.g. predicting an author's age),
or classification \parencite{pennacchiotti2011machine} (e.g. predicting an
author's gender).

Standard features include word \parencite{rao2010classifying} and character n-gram
features \parencite{pennacchiotti2011machine} of messages
users post, output from NLP systems such as word stems or part of speech tags
\parencite{zamal2012homophily,nguyen2013old,preotiuc2018user}, and topic and word
embedding features \parencite{pennacchiotti2011machine,preotiuc2018user}.
Dictionary features such as the Linguistic Inquiry and Word
Count (LIWC) are also popular, especially since these features
have been shown to correlate with meaningful demographic and psychometric user properties
\parencite{tausczik2010psychological}.  It is also common
to draw on features of the local network such as the identities of neighboring
users or text that they post \parencite{culotta2016predicting,yang2017overcoming}.
This is done either
by aggregating information from friends or followers of the source user into a
feature vector \parencite{zamal2012homophily}, or by sharing predictions made on
neighboring users through the social graph \parencite{yang2017overcoming}.
Work that takes the latter approach exploits \emph{homophily}, the tendency of
similar users to establish connections with others in the social graph.

\paragraph*{Mental Properties}

Although more recent, there is also a community around predicting
less traditional user properties such as mental health \parencite{dechoudhury2013predicting,W15-1204,W16-0311} and user
personality \parencite{schwartz2013personality,W14-3214,preotiuc-pietro:ea:2015personality}.
Similar to predicting demographics, the typical approach is to train supervised models
to predict these features.  A major difficulty with learning mental health classifiers
is that unlike demographic features such as gender which are relatively easy
to annotate, mental health is a particularly sensitive characteristic.
Not only are subjects reticent in divulging this information, but care should
be taken by researchers, even when mental health status is inferred
\parencite{benton2017ethical}.

One clever approach is to consider public messages self-reporting having
a particular condition as genuine \parencite{harman2014measuring,coppersmith2015adhd}.
\textcite{padrez2015linking} assembled a parallel corpus of electronic
health records alongside Twitter and Facebook posts.  However, these
subjects manually opted in from a single hospital emergency department,
and therefore the number of positive examples for any single mental health
condition is small.

Personality is a less sensitive target to predict, since users regularly subject
themselves to online personality tests
\parencite{kosinski2013private,plank2015personality}.
Nevertheless, knowing user personality has implications for predicting future
behavior \parencite{cadwalladr2018cambridge}, and it is unclear how comfortable
users are with personalized inferences made about them without their consent.

\removed{
\paragraph*{Unsupervised Approaches to Inferring User Features}

This thesis does not focus on supervised prediction of distinct user features. Rather
we would like to learn generic user representations that are effective for many tasks.
There has been less work on unsupervised learning of user demographics.  What
has been done falls into one of two camps: latent variable modeling, and
representation learning with post-hoc analysis of clusters.

Some examples of the first category include 

\textcite{bergsma2013broadly} show that clustering Twitter names can
be predictive of demographic categories.

\begin{itemize}
  \item Bayesian latent variable models
  \item Post-hoc clustering of embeddings, align with known features, calculate purity
  \item \citetemp{silvio amir neural user profiles to predict mental properties}
\end{itemize}
}

\removed{
\paragraph*{Location}

Although less relevant to the work in this thesis, there has also been work on predicting
less stable user properties such as their location.  \textcite{hecht2011tweets}
infer a U.S. Twitter user's location based on the description in their location field
and language use, achieving roughly 30\% accuracy at predicting their self-described U.S.
state.  \textcite{carmen} similarly infer user location by also considering
tweet metadata such as geolocation information.
Research has also considered the information that a user's friends can provide
to predicting the ego user's location \parencite{cranshaw2010bridging,cho2011friendship,davis2011inferring,mislove2010you,jurgens2013location}.
In this work a user's unknown location is inferred by or known or inferred locations of users
close in the social graph.
\textcite{cheng2010you} also presents a probabilistic model for a user's location that is
dependent not only on what they say but who they are friends with, combining both text and
network information.

Instead of trying to infer properties about particular users, researchers have also used
social media users as ``sensors'' to detect where and when specific important events
occur, such as disasters \parencite{oh2010exploration,sakaki2010earthquake,earle2012twitter},
as well as more open-ended
events \parencite{becker2011beyond,cataldi2010emerging,weng2011event,ritter2012open}.
}

\subsection{Recommendation Systems}

Recommendation systems can be grouped into
two main classes based on how recommended items are ranked:
{\em collaborative filtering} and {\em content-based}.  Content-based systems are
more strongly dependent on user profile since recommendations are based on
representations of the user and the item being recommended.  Collaborative filtering
systems make
recommendations based on prior consumption.
Collaborative filtering systems have a hard time making useful
recommendations early on because they rely on a history of the user's consumption.
This is known as the cold-start problem
\parencite{adomavicius2005toward}.  Content-based
systems are not as susceptible to the cold-start problem, since they can make
recommendations based on extraneous user factors (e.g. a user description that is
populated at enrollment).  Although systems rarely fall squarely in one category or
the other, this remains a useful dichotomy.

Recommendation systems on social media platforms
either recommend \emph{content} to consume or recommend \emph{friends} to connect with
\parencite{phelan2009using,leskovec2010signed}.
Facebook and Twitter news feeds are examples of content recommendation
systems operating over the space of other users' messages.  Predicting whether or
not a user would click on an
advertisement can also be viewed as a recommendation system, where the items that
are recommended are advertisements for various products
\parencite{lohtia2003impact,dembczynski2008predicting}.
In this case, clicking on an advertisement constitutes consuming the item.  Note
that a content-based approach, modeling the user, is critical since ad clicks are
very rare events \parencite{wang2011click}.

\removed{
\subsubsection{Applications}

\paragraph*{Content Recommendation}

\begin{itemize}
  \item Products to purchase (e.g. Amazon recommendations, Netflix movies)
  \item Stories/posts, feed optimization (e.g. Twitter, Facebook)
  \item Predicting CTR for advertisers
\end{itemize}

\paragraph*{Friend Recommendation}

Based on content that they share or close personal 
}

\subsection{Social Science}

Social media data provides researchers with a platform to
study the effects of human relationships, social networks, on behavior.
One goal of social media analytics is to replace
traditional survey mechanisms \parencite{thacker1988,krosnick2005} by
monitoring messages posted on social media.  Although the surveys
that are simulated are most regularly seen in political polling
\parencite{tumasjan2010predicting}, they also appear in tracking disease and public health
\parencite{paul2011you,culotta2014reducing}, and opinion related to public policy
issues \parencite{oconnor2010tweets,stefanone2015image,benton2016after}.

However, social media users are a biased sample relative to the general population
\parencite{ruths2014social}. This presents difficulties when predicting
survey responses directly from online messages.  One way to account
for difference in the populations is to adjust one's predictions based on demographic
features of the social media population you are measuring.
Inferred demographics can be used to appropriately adjust for bias on social
media \parencite{culotta2014reducing}.

User features are also important to control for as potential confounds when
measuring influence in social networks \parencite{hill2006network}.
\textcite{aral2009distinguishing} find that features such as a user's demographics
explain most of the tendency to adopt a mobile application in an instant messaging
network.  Not controlling for homophily in the network means that the effect of social
influence -- one user adopting the application leads their friends to adopt it --
is overestimated since users with a natural propensity to adopt will share other features
that also make them more likely to be linked.  Global position in the social network
may also be used as a substitute for latent user features \parencite{hill2011cluster}.

\subsection{Message-Level Prediction}

User information can help systems make better predictions for single
messages/documents even when not clearly related to
the message-level prediction task.  Work related to improving NLP
systems by conditioning on user demographics is a key example.
\textcite{hovy2015demographic} show that training separate classifiers for
product reviewers of different gender and age can
improve accuracy at predicting product category and rating.
\textcite{johannsen2015cross} show that author gender is predictive of
certain types of syntactic patterns in online
product reviews.  This suggests that knowing features of the user writing
a review could
improve syntactic parsing of sentences.  Similarly, \textcite{Hovy:Soegaard:2015}
show that author age affects the performance of already trained
part-of-speech taggers, suggesting a disparity in the way younger vs. older
authors use language.

Instead of relying on  ground truth user features, messages can be conditioned
on generic user embeddings.  For example, \textcite{amir2016modelling} finds that tweet
sarcasm detection can be improved by augmenting the input features with pretrained user
embeddings as a source of context.

\section{Multiview Representation Learning}
\label{sec:background:mv_rep_learning}

\subsection{Motivation}

The goal of multiview representation learning is to learn representations
for a class of objects that capture correspondences between multiple feature sets,
\emph{views}, associated with each object.  We learn these representations
because we believe they will be predictive of some latent object property,
a useful component in a downstream system.
These feature sets often correspond to multiple modalities.  Take for example the X-ray
microbeam dataset, a corpus of speech utterances containing acoustic measurements
paired with the position of speech articulators \parencite{westbury1994xray}.  Multiview
learning methods have successfully been applied to this data to learn representations
predictive of what phone a person is uttering at each frame \parencite{Wang_15b}.

Multiview methods are applied under the assumption that each view is sufficient to
predict a target of interest \emph{given enough training data}
\parencite{kakade2007multi}.  However, we almost never have enough training data, so variance
in our small training set will obscure the mapping from input features to target.
By learning a representation of what is common between views, discarding uncorrelated
noise, we ignore uninformative variance in our input features and yield better downstream performance.
Single-view dimensionality reduction techniques such as principal component
analysis may discard variance in the data that happens to be correlated across views,
simply because it treats the input features as a single feature set.

In this thesis, we learn multiview user embeddings derived by applying variants of canonical
correlation analysis (CCA), an old statistical technique for finding linear transformations
of two random variables such that they are maximally correlated
\parencite{hotelling1936relations}.  In this section we describe the CCA problem and
present solution derivations.  We also describe objectives extending CCA to learn
nonlinear mappings between two views, nonlinear kernel CCA and deep CCA.
We finally discuss MAXVAR generalized CCA, an extension to maximizing
correlation between more than two views.  See \textcite{uurtio2017tutorial} for
another CCA tutorial, a discussion about using it as an analysis tool, and interpreting the
learned embeddings.

\paragraph*{Other Multiview Techniques}

Although in this thesis we learn user embeddings with methods related to CCA,
there is a long history of using multiview techniques to learn representations
as well as classifiers.  Below is a selection of related methods.

Co-training is a semi-supervised approach for training a robust classifier from
few labeled examples \parencite{blum1998combining}. In this method, the feature
set is partitioned into two views, and an independent classifier is trained independently
on each view.  An unlabeled dataset is then tagged by each classifier, and the unlabeled
data along with the predicted labels are used to augment the other classifier's training
set.  This entrains each classifier to make similar predictions from different feature
sets.  This framework has applicability beyond learning classifiers, and has also been
applied to the problem of multiview clustering \parencite{kumar2011co}.

Siamese networks are a class of neural models that can be applied to multiview
representation learning \parencite{bromley1994signature}.  In this framework, each view
is passed through a network and network weights are trained to minimize the $\ell_2$
distance between the siamese network output layers.  This is similar in spirit to CCA,
where as we will show, correlation between two views is maximized.

Another related class of models are multiview probabilistic generative models, where a latent
variable is assumed to govern the distribution of several observed views.  Topic models that infer
a shared distribution over topics for multiple document views (e.g. the body and title of
a  news article) are one class of models \parencite{ahmed2010staying}.  CCA has a corresponding
probabilistic model as well (Section \ref{subsubsec:background:probabilistic_cca}).

\subsection{Canonical Correlation Analysis}

CCA is a statistical technique used to
learn a linear relationship between two sets of random variables.
These two sets of variables are referred to as \emph{views}.  CCA is applied
when one wants to maximize correlation between views and discard independent variation
as noise.

\subsubsection{Problem Definition}

Suppose we are given two data matrices $X \in \mathbb{R}^{n \times p}$ and
$Y \in \mathbb{R}^{n \times q}$ where $X$ corresponds to view 1, and $Y$ corresponds to
view 2.  $n$ is the number of examples in your data, $p$ is
the number of features in view 1, and $q$ is the number of features in view 2.

The one-dimensional CCA problem is as follows:

{
  \setlength{\jot}{0ex}
\begin{equation}
\begin{aligned}
  & \max_{u \in \mathbb{R}^{p}, v \in \mathbb{R}^{q}} z_X^T z_Y \span\omit  \\
  \text{where } & z_X = Xu ; & z_Y = Yv \\
  \text{subject to } & \|z_X\|_2 = 1 ; & \|z_Y\|_2 = 1
\end{aligned}
\label{eq:background:onedim_cca_obj}
\end{equation}
}

The solutions to this problem, $u$ and $v$, are points in the
feature space that are mapped to points in $\mathbb{R}^{n}$ by the view data matrices.
$u$ and $v$ are called the \emph{canonical weight vectors} or \emph{canonical weights},
and their images under $X$ and $Y$, $z_X$ and $z_Y$, are called the \emph{canonical variates}.
This is the one-dimensional CCA problem, as we are finding a single pair of canonical
weights.  It can be extended to finding more than one set of canonical weight vectors by
solving for $u^i$ and $v^i$ that satisfy the above problem, with
the additional constraints that $z^{i}_X$ and $z^{i}_Y$ are orthogonal to all other
$z^{j}_X$ and $z^{j}_Y$.  For all $i \in 1 \ldots k$, the $k$-dimensional CCA problem then becomes:

{
  \setlength{\jot}{0ex}
\begin{equation}
\begin{aligned}
  & \max_{u^{i} \in \mathbb{R}^{p}, v^{i} \in \mathbb{R}^{q}} (z^{i}_X)^T z^{i}_Y \span\omit \\
  \text{subject to } & \|z^{i}_X\|_2 = 1;  & \|z^{i}_Y\|_2 = 1  \\
  & \forall j < i, & (z^{i}_{X})^{T} z^{j}_{X} = 0 \\
  & & (z^{i}_{Y})^{T} z^{j}_{Y} = 0
\end{aligned}
\label{eq:background:kdim_cca_obj}
\end{equation}
}

\paragraph*{Why \emph{Correlation} Analysis?}

If we consider $z_X$ and $z_Y$ to be $n$ draws of two scalar-valued random variables,
then the empirical correlation between these variables is
$\frac{1}{n-1} \frac{z_X^T z_Y}{\sqrt{z_X^T z_X} \sqrt{z_Y^T z_Y}}$.

The constraints in the CCA objective ensure that $z_X$ and $z_Y$ are both unit-norm, so:

{
  \setlength{\jot}{0ex}
\begin{equation*}
\begin{aligned}
  {\text \sc corr}(z_X, z_Y)  & = \frac{1}{n-1} \frac{z_X^T z_Y}{\sqrt{z_X^T z_X} \sqrt{z_Y^T z_Y}} \\
  & = \frac{1}{n-1} \frac{z_X^T z_Y}{\sqrt{1} \sqrt{1}} \\
  & = \frac{z_X^T z_Y}{n - 1}
\end{aligned}
\end{equation*}
}

This quantity is maximized when the inner product between $z_X$ and $z_Y$ is maximized.

\subsubsection{Notation and Terminology}

Suppose we are given two data matrices $X \in \mathbb{R}^{n \times p}$ and
$Y \in \mathbb{R}^{n \times q}$ where $X$ corresponds to view 1, and $Y$ corresponds to
view 2.  $n$ is the number of examples in the data, $p$ is
the number of features in view 1, and $q$ is the number of features in view 2.
To simplify the following derivations, we assume that the
columns of each of these matrices are normalized such that their means are zero and have
unit variance\footnote{If you are given views that do not satisfy these constraints, they
  can be normalized simply by subtracting the mean from each column and dividing each
  feature value by the column standard deviation.  Remember to save these data means
  and standard deviations used in preprocessing, so they can be applied to test data.}.
We assume, without loss of generality, that $p \leq q$.

\paragraph*{Useful Definitions}

The sample \emph{auto-covariance} matrices for views 1 and 2:

{
  \setlength{\jot}{0ex}
\begin{equation*}
\begin{aligned}
  C_{XX} & = \frac{1}{n-1} X^T X \\
  C_{YY} & = \frac{1}{n-1} Y^T Y
\end{aligned}
\end{equation*}
}

The sample \emph{cross-covariance} matrices between views 1 and 2:

{
  \setlength{\jot}{0ex}
\begin{equation*}
\begin{aligned}
  C_{XY} & = \frac{1}{n-1} X^T Y \\
  C_{YX} & = \frac{1}{n-1} Y^T X
\end{aligned}
\end{equation*}
}

Note that $C_{XY} = C_{YX}^T$.  Finally, the \emph{joint covariance} matrix for both views:

\[
\left(
\begin{array}{cc}
C_{XX} & C_{XY} \\
C_{YX} & C_{YY}
\end{array}
\right)
\]

This is the covariance matrix in the single-view setting, the auto-covariance matrix
for the concatenation of both views.  However, it will be useful
to consider each block of this matrix separately, since they correspond to the auto-covariance and cross-covariance
matrices of the individual matrices.

\subsubsection{Solution}
\label{sec:background:cca_solution}

Below are two sketches of derivations for solving the CCA problem.  The first
derivation was given in \textcite{hotelling1936relations}, and the second was published
over 50 years later in \textcite{ewerbring1990canonical}.

\paragraph*{Original Hotelling Derivation}

At a high-level, the original solution presented by Hotelling gives the solution for
the first pair of canonical weights and variates.  Iit boils down to the
following steps:

{
\begin{tcolorbox}
  \setstretch{1.0}
  \begin{enumerate}
  \item Form the augmented Lagrangian of the CCA problem.
  \item Take the partial derivatives of the augmented Lagrangian with respect
        to the unknowns.
  \item Mathematically massage these equations to yield an eigenvalue problem.
  \item Show that the eigenvectors solutions to this equation are one set of
        canonical weights, and the eigenvalues are correlations between canonical
        variates of the CCA problem.
  \item Solve for the other set of canonical weights by substitution.
\end{enumerate}
\end{tcolorbox}
}

The first observation is that {\sc corr}$(z_X, z_Y)$ does not change if we scale $z_X$ or
$z_Y$, so let us scale $u$ and $v$ to ensure they are both unit-norm.

{
  \setlength{\jot}{0ex}
\begin{equation*}
\begin{aligned}
  z_X^T z_X & = u^T X^T X u & = u^T C_{XX} u & = 1 \\
  z_Y^T z_Y & = v^T Y^T Y v & = v^T C_{YY} v & = 1
\end{aligned}
\end{equation*}
}

Thus, we can rewrite the CCA problem as:

{
  \setlength{\jot}{0ex}
\begin{equation}
\begin{aligned}
  & \max_{u,v} \langle X u, Y v \rangle = u^T C_{XY} v \\
  \text{subject to } \ & u^T C_{XX} u = 1 \\
  & \ v^T C_{YY} v = 1
\end{aligned}
\label{eq:background:transformed_cca_unitnorm}
\end{equation}
}

First we use the Lagrange multipler technique to fold the constraints into the
objective with Lagrange multipliers $\lambda_X$ and $\lambda_Y$:

\begin{equation}
\begin{aligned}
  \mathcal{L}(u, v, \lambda_X, \lambda_Y) & = u^T C_{XY} v - \lambda_X (u^T C_{XX} u - 1) - \lambda_Y (v^T C_{YY} v - 1)
\end{aligned}
\label{eq:background:cca_lagrangian}
\end{equation}

We take the partial derivative of the righthand side with respect to
$u$ and $v$, and set each equal to 0 --
a solution of the objective must necessarily also be a stationary point of
the Lagrangian for some non-negative values $\lambda_X$ and $\lambda_Y$.

{
  \setlength{\jot}{0ex}
\begin{equation}
\begin{aligned}
  \frac{\delta \mathcal{L}}{\delta u} & = C_{XY} v - 2 \lambda_X C_{XX} u & = 0\\
  \frac{\delta \mathcal{L}}{\delta v} & = C_{YX} u - 2 \lambda_Y C_{YY} v & = 0
\end{aligned}
\label{eq:background:hotelling_lagrangian_deriv}
\end{equation}
}


Multiply each equation on the left by $u^T$ and $v^T$, respectively.

{
  \setlength{\jot}{0ex}
\begin{equation*}
\begin{aligned}
  u^T C_{XY} v - 2 \lambda_X u^T C_{XX} u & = 0\\
  v^T C_{YX} u - 2 \lambda_Y v^T C_{YY} v & = 0
\end{aligned}
\end{equation*}
}

We know that a solution to the CCA problem must satisfy the unit norm constraints
constraints, so we insert these:

{
  \setlength{\jot}{0ex}
\begin{equation}
\begin{aligned}
  u^T C_{XY} v & = 2 \lambda_X \\
  v^T C_{YX} u & = 2 \lambda_Y
\end{aligned}
\end{equation}
}

and since the left-hand side of the first equation is just the transpose of the
second (and is a scalar), we know that the multipliers must be the same value
$\lambda = \lambda_X = \lambda_Y$.  Substituting for $\lambda$ back into Equation
\ref{eq:background:hotelling_lagrangian_deriv} yields:

{
  \setlength{\jot}{0ex}
\begin{equation}
\begin{aligned}
  C_{XY} v & = 2 \lambda C_{XX} u \\
  \frac{C_{XX}^{-1} C_{XY} v}{2 \lambda} & = u \\
  C_{YX} u & = 2 \lambda C_{YY} v \\
  \frac{C_{YY}^{-1} C_{YX} u}{2 \lambda} & = v \\
\end{aligned}
\label{eq:background:hotelling_wrt_weights}
\end{equation}
}

Note that $C_{XX}$ and $C_{YY}$ are invertible because they are both symmetric --
$X^{T}X = (X^{T}X)^{T}$ -- and positive definite --
$\forall w \in R^{p}, (X w)^T (X w) > 0$\footnote{We are assuming the auto-covariance matrices are is full-rank and $\|w\|  > 0$.  This is a reasonable assumption if $n >> p,q$.}.
A final substitution of $u$ from Equation \ref{eq:background:hotelling_wrt_weights} into Equation
\ref{eq:background:hotelling_lagrangian_deriv} yields:

{
  \setlength{\jot}{0ex}
\begin{equation}
\begin{aligned}
 \frac{C_{YX} C_{XX}^{-1} C_{XY} v}{2 \lambda} & = 2 \lambda C_{YY} v \\
 ( C_{YY}^{-1} C_{YX} C_{XX}^{-1} C_{XY} ) v & = 4 \lambda^{2} v
\end{aligned}
\end{equation}
}

This is in the form of an eigenvalue problem, where $v$ is the principal
eigenvector for the left-hand side matrix and $4 \lambda^{2}$ is its associated
eigenvalue.  We can use an eigensolver to solve for $v$.  Once solved,
we can substitute $v$ back into Equation \ref{eq:background:hotelling_lagrangian_deriv}
and finally solve for $u$.

This derivation can be extended to
finding $k$ pairs of canonical weights by replacing $u$ and
$v$ by matrices $U \in \mathbb{R}^{p \times k}$ and $V \in \mathbb{R}^{q \times k}$, where each
successive column $V^{i}$
is the $i^{\text{th}}$ eigenvector of $C_{YY}^{-1} C_{YX} C_{XX}^{-1} C_{XY}$
and $U$ is solved by substitution into Equation \ref{eq:background:hotelling_wrt_weights}.
In summary, assuming that $C_{XX}$ and $C_{YY}$ are
invertible, a solution to the $k$-dimensional CCA problem can be found by:

{
\begin{tcolorbox}
  \begin{equation}
  \begin{aligned}
    V & = k\ \text{top eigenvectors of}\ C_{XX}^{-1} C_{XY} C_{YY}^{-1} C_{YX} \\
    U & = \frac{C_{XX}^{-1} C_{XY} V}{2 \lambda}
  \end{aligned}
  \label{eq:background:hotelling_solution}
  \end{equation}
\end{tcolorbox}
}

\paragraph*{SVD of Joint Covariance Matrix}

A second derivation of the CCA solution expresses the CCA objective
in terms of the data's joint covariance matrix \parencite{ewerbring1990canonical}.  The
constraint that successive canonical variates be orthogonal to each other and unit-norm
can be written as:

{
  \setlength{\jot}{0ex}
\begin{equation*}
\begin{aligned}
 U^T C_{XX} U & = I \\
 V^T C_{YY} V & = I
\end{aligned}
\end{equation*}
}

where if $U \in \mathbb{R}^{p \times k}$ and $V \in \mathbb{R}^{q \times k}$, with $k <= p$, then
$I$ is the $k \times k$ identity matrix.  The objective can also be written as:

\begin{equation*}
\begin{aligned}
 U^T C_{XY} V & = \Lambda
\end{aligned}
\end{equation*}

where $\Lambda \in \mathbb{R}^{k \times k}$ is a diagonal matrix whose diagonal values,
$\Lambda_{1,1} \ldots \Lambda_{k,k}$, are the canonical correlations, $(z^{i}_X)^T z^{i}_Y$.
We can express both the constraints and the objective in a single equation:

\begin{equation}
\begin{aligned}
\left(
\begin{array}{cc}
U^T & 0 \\
0 & V^T
\end{array}
\right)
\left(
\begin{array}{cc}
C_{XX} & C_{XY} \\
C_{YX} & C_{YY}
\end{array}
\right)
\left(
\begin{array}{cc}
U & 0 \\
0 & V
\end{array}
\right) = &
\left(
\begin{array}{cc}
I & \Lambda \\
\Lambda & I
\end{array}
\right)
\end{aligned}
\label{eq:background:svd_eq}
\end{equation}

If we let $\widetilde{U} = C_{XX}^{\frac{1}{2}} U$ and
$\widetilde{V} = C_{YY}^{\frac{1}{2}} V$, we can rewrite Eq.
\ref{eq:background:svd_eq} as:

\begin{equation*}
\begin{aligned}
\left(
\begin{array}{cc}
\widetilde{U}^T & 0 \\
0 & \widetilde{V}^T
\end{array}
\right)
\left(
\begin{array}{cc}
I & C_{XX}^{-\frac{1}{2}} C_{XY} C_{YY}^{-\frac{1}{2}} \\
C_{YY}^{-\frac{1}{2}} C_{YX} C_{XX}^{-\frac{1}{2}}  & I
\end{array}
\right)
\left(
\begin{array}{cc}
\widetilde{U} & 0 \\
0 & \widetilde{V}
\end{array}
\right) = &
\left(
\begin{array}{cc}
I & \Lambda \\
\Lambda & I
\end{array}
\right)
\end{aligned}
\end{equation*}

From the on-diagonal blocks, we know that $\widetilde{U}$ and $\widetilde{V}$ must be
orthogonal matrices, and we can manipulate the off-diagonal elements to show that:

{
  \setlength{\jot}{0ex}
\begin{equation}
\begin{aligned}
\widetilde{U}^T C_{XX}^{-\frac{1}{2}} C_{XY} C_{YY}^{-\frac{1}{2}} \widetilde{V} & = \Lambda \\
C_{XX}^{-\frac{1}{2}} C_{XY} C_{YY}^{-\frac{1}{2}} & = \widetilde{U} \Lambda \widetilde{V}^{T}
\end{aligned}
\label{eq:background:svd_cca_solution}
\end{equation}
}

We can solve for $\widetilde{U}$, $\widetilde{V}$, and $\Lambda$ by a rank-$k$ truncated
singular value decomposition (SVD) of the left-hand matrix, then solve for the canonical
weights by $U = C_{XX}^{-\frac{1}{2}} \widetilde{U}$ and $V = C_{YY}^{-\frac{1}{2}} \widetilde{V}$.

\paragraph*{Why these Derivations are Interesting}

The first derivation was originally presented in \textcite{hotelling1936relations}.
This second derivation is worth seeing since we can learn both pairs of
canonical weights using an SVD of a particularly constructed matrix.  It also relates
the CCA weights to the joint sample covariance matrix.  One can draw connections to other
linear techniques such as principal component analysis, where an eigendecomposition of the
joint covariance matrix yields the principal components:

\begin{equation}
\begin{aligned}
\left(
\begin{array}{cc}
C_{XX} & C_{XY} \\
C_{YX} & C_{YY}
\end{array}
\right) = & U^{T} \Lambda U
\end{aligned}
\label{eq:background:pca_svd_eq}
\end{equation}

\removed{
\begin{itemize}
  \item Hotelling solution 
  \item The SVD solution derives the canonical weights as a function of the joint covariance
        matrix simultaneously (no need for backsubstitution as in the Hotelling solution).
        One can also draw parallels to other dimensionality reduction techniques that rely
        on the 
\end{itemize}
}

\subsubsection{Probabilistic Interpretation}
\label{subsubsec:background:probabilistic_cca}

\textcite{bach2005probabilistic} showed that the solution for the CCA objective
is equivalent to the maximum likelihood solution of latent weights in a particular
generative model.  The generative story of this model is simply as follows:

{
  \setlength{\jot}{0ex}
\begin{equation}
\begin{aligned}
  z \sim \mathcal{N} (0, I_k), \min (p, q) \leq k \leq 1 \\
  x | z \sim \mathcal{N} (W_X z + \mu_X, \psi_X), W_X \in \mathbb{R}^{p \times k} \\
  y | z \sim \mathcal{N} (W_Y z + \mu_Y, \psi_Y), W_Y \in \mathbb{R}^{q \times k}
\end{aligned}
\label{eq:background:prob_cca_story}
\end{equation}
}

where $z$ is the latent vector representation for an example, $W_X$ and $W_Y$ are weight
matrices mapping this embedding to observed views, and $x$ and $y$ are the observed
views of this example.  $\psi_X$, $\psi_Y$ are positive definite covariance matrices and
$\mu_X$, $\mu_Y$ are arbitrary means that parameterize independent noise in each view.
\textcite{bach2005probabilistic} show that the maximum likelihood estimates of $W_X$ and
$W_Y$ are:

{
  \setlength{\jot}{0ex}
\begin{equation}
\begin{aligned}
  \hat{W}_X = & C_{XX} U M \\
  \hat{W}_Y = & C_{YY} V M
\end{aligned}
\label{eq:background:prob_cca_solution}
\end{equation}
}

Here, the $C_{XX}$ and $C_{YY}$ are the sample auto-covariance matrices, $U$ and $V$ are the
left singular vectors of $C_{XX}$ and $C_{YY}$ respectively, and $M$ is the square root of
the diagonal matrix of singular values $C_{XX}^{-\frac{1}{2}} C_{XY} C_{YY}^{-\frac{1}{2}}$.

This probabilistic model demonstrates when CCA is appropriate for learning a
representation: \emph{when variation in observed views are independent conditional on their
  latent representation with independent Gaussian noise applied to each view}.  The model also
suggests that the CCA problem can be approximately solved by iterative algorithms for
estimating latent variables in probabilistic models such as Expectation Maximization.

\subsection{Nonlinear Variants}

One drawback of CCA is it can only uncover linear relationships between views.  Although
less work has been devoted to maximizing correlation between views after nonlinear transformation, there are two prominent methods for doing so: \emph{Kernel CCA} and \emph{Deep CCA}.

\subsubsection{Kernel CCA}
\label{subsubsec:background:kernel_cca}

One method to uncover a nonlinear relationship between two views is to consider kernel CCA
\parencite{lai2001kernel} (KCCA).  Similar to kernel PCA, the practitioner defines kernel
functions that independently define the similarity between points in view 1 and view 2,
and these kernel functions are used in lieu of inner product when computing correlation.

\paragraph*{Problem}

Let $K_{X} \in \mathbb{R}^{n \times n}$ and $K_{Y} \in \mathbb{R}^{n \times n}$ be symmetric
positive semi-definite \emph{Gram} matrices expressing the similarity between examples
according to features from views 1 and 2.  The kernel CCA problem is defined as:

{
  \setlength{\jot}{0ex}
\begin{equation}
\begin{aligned}
  \max_{z_{X} \in \mathbb{R}^{n}, z_{Y} \in \mathbb{R}^{n}} corr(z_X, z_Y) & = \langle z_X, z_Y \rangle = \alpha^T K_X^T K_Y \beta \\
  \text{where } & z_X = K_X \alpha, z_Y = K_Y \beta \\
  \text{subject to } & \|z_X\|_2 = \sqrt{\alpha K_{X}^2 \alpha} = 1 \\
  & \|z_Y\|_2 = \sqrt{\beta K_{Y}^2 \beta} = 1
\end{aligned}
\label{eq:background:kcca_problem}
\end{equation}
}

where $\alpha, \beta \in \mathbb{R}^{n}$ take the place of the canonical weights $u, v$ in
vanilla CCA.  Note also that $K_X$ and $K_Y$ replace $X$ and $Y$.  The similarity between this
problem formulation and linear CCA comes from the fact that the canonical variates, $z_X$ and
$z_y$, lie in the span of the data in both problems.

\paragraph*{Derivation}

Similar to the Hotelling solution, we
can form the augmented Lagrangian, take the derivative with respect to the weights
$\alpha$ and $\beta$, and set them equal to zero:

{
  \setlength{\jot}{0ex}
\begin{equation*}
\begin{aligned}
  \mathcal{L}(\alpha, \beta, \lambda_1, \lambda_2) & = \alpha^T K_X^T K_{Y}^T - \lambda_X (\alpha^T K_X^2 \alpha - 1) - \lambda_Y (\beta^T K_Y^2 \beta - 1) \\
  \frac{\delta \mathcal{L}}{\delta \alpha} & = K_X^T K_Y \beta - 2 \lambda_X K_X^2 \alpha = 0 \\
  \frac{\delta \mathcal{L}}{\delta \beta} & = K_Y^T K_X \alpha - 2 \lambda_Y K_Y^2 \beta = 0
\end{aligned}
\end{equation*}
}

We then right-multiply each derivative by $\alpha^T$ and $\beta^T$ respectively, substitute
in the unit-norm constraints, and find that at the solution $\lambda_X = \lambda_y$:

{
  \setlength{\jot}{0ex}
\begin{equation*}
\begin{aligned}
  \alpha^T K_X^T K_Y \beta - 2 \lambda_X \alpha^T K_X^2 \alpha = & 0 \\
  \beta^T K_Y^T K_X \alpha - 2 \beta^T \lambda_Y K_Y^2 \beta = & 0 \\
  2 \lambda_X \alpha^T K_X^2 \alpha = & 2 \lambda_Y \beta^T K_Y^2 \beta \\
  2 \lambda_X = & 2 \lambda_Y = 2 \lambda;
\end{aligned}
\end{equation*}
}

If we substitute $\lambda$ back in and solve for $\alpha$ in
$\frac{\delta \mathcal{L}}{\delta \alpha} = 0$, we find that:

{
  \setlength{\jot}{0ex}
\begin{equation*}
\begin{aligned}
  K_X^T K_Y \beta - 2 \lambda K_X^2 \alpha = & 0 \\
  \frac{K_X^T K_Y \beta} {2 \lambda} = & K_X^2 \alpha \\
  \frac{K_X^{-1} K_Y \beta} {2 \lambda} = & \alpha
\end{aligned}
\end{equation*}
}

Substituting for $\alpha$ in $\frac{\delta \mathcal{L}}{\delta \beta} = 0$ yields:

{
  \setlength{\jot}{0ex}
\begin{equation*}
\begin{aligned}
  K_Y^T K_x (\frac{K_X^{-1} K_Y\beta}{2 \lambda}) = &2 \lambda K_Y^T K_Y \beta \\
  K_Y^2 \beta = & 4 \lambda^{2} K_Y^2 \beta
\end{aligned}
\end{equation*}
}

This suggests that if $K_Y$ is invertible, then $\beta$ is completely unconstrained, and the
canonical correlation is 1.  A standard solution is to change the unit-norm constraints to be
$\alpha^T (K_X + \epsilon_X I) \alpha = 1$ and $\beta^T (K_X + \epsilon_Y I) \beta = 1$.
The choice of regularization strength can be selected by heldout correlation captured.  This
is 

\paragraph*{Details}

KCCA can be applied to test data by computing the similarity between each test
example to each training example, then mapping this kernel matrix with $\alpha$
or $\beta$ depending on the
view\footnote{This is analogous to the way one applies kernel PCA to test data.}.
For example, say we compute $K_X^{test} \in \mathbb{R}^{n \times m}$ for $m$ test examples:

\begin{equation*}
\begin{aligned}
z_X^{test} = & \alpha K_X^{test}
\end{aligned}
\end{equation*}

The derivation is instructive since we see how closely the kernel CCA derivation follows the
original Hotelling solution, and it is important since it underscores the necessity of
regularizing the Gram matrix regularization (otherwise the problem is not well-defined).
This idea of regularization is also applicable to the original CCA objective, where a small
amount of diagonal weight can be added to the sample auto-covariance matrices, mostly to
ensure invertibility.

\subsubsection{Deep CCA}
\label{subsubsec:background:deep_cca}

Although KCCA maximizes correlation between views subject to an implicit nonlinear mapping,
the nonlinear mapping solely depends on the choice of kernel and the training set.
In addition, computing and inverting the Gram matrices are very expensive operations in
space and computation time.  One model that solves these problems is Deep CCA (DCCA), a
CCA variant that alternates between maximizing correlation between views and updating a
nonlinear mapping from observed views to shared space \parencite{andrew2013}.  The nonlinear mappings for each view
are parameterized by two neural
networks.  In addition to \emph{learning} the nonlinear mappings to
shared space, DCCA also avoids computing and inverting large
$n \times n$ Gram matrices.
The crux of fitting DCCA to a dataset lies in the gradient update to update the per-view
neural networks.

\paragraph*{Problem}

The DCCA problem for the first canonical component is defined as follows:

{
  \setlength{\jot}{0ex}
\begin{equation}
\begin{aligned}
  & \max_{\theta_X, \theta_Y, u, v} z_X^{T} z_Y  \\
  \text{where } & z_X = f_X(X; \theta_X) u \\
    & z_Y = f_Y(X; \theta_Y) v \\
  \text{subject to } & \|z_X\|_2 = 1 \\
  & \|z_Y\|_2 = 1 \\
\end{aligned}
\label{eq:background:dcca_problem}
\end{equation}
}

Here $\theta_X$ and $\theta_Y$ are the set of weights parameterizing fixed neural network
architectures $f_X$ and $f_Y$ respectively.  These are functions that map each example view
to a fixed vector of the dimensionality of the network output layer $p_f$ and $q_f$.

Note that if the network weights are fixed, the solution for canonical weights $u$ and $v$
is just that given by linear CCA with respect to the output layer activations on the
training set.  In addition, if we fix the canonical weights then we can update network
weights $\theta_X$ and $\theta_Y$ by backpropagation (assuming we can differentiate the
objective with respect output layer activations $f_X(X; \theta_X)$ and $f_Y(Y; \theta_Y)$).

This suggests an optimization scheme where at each iteration we alternate between updating
network weights by backpropagation and then solving for canonical weights.  This way the
orthonormality constraint on canonical components is maintained after each iteration.

\paragraph*{DCCA Gradient}

Let the output layer activations of two views passed through their associated networks
be $X_f$ and $Y_f$, and assume that each has zero mean.
The gradient of the correlation objective with respect to $X_f$ is given as:

{
  \setlength{\jot}{0ex}
\begin{equation}
\begin{aligned}
  \frac{\delta {\text \sc corr} (X_f, Y_f)}{\delta X_f} \propto & \nabla_{XY} Y_f - \nabla_{XX} X_f \\
  \text{where } & \nabla_{11} =  C_{XX}^{-\frac{1}{2}} U D U^T C_{XX}^{-\frac{1}{2}} \\
   & \nabla_{12} = C_{XX}^{-\frac{1}{2}} U V^T C_{YY}^{-\frac{1}{2}}
\end{aligned}
\label{eq:background:dcca_gradient}
\end{equation}
}

The partial derivative with respect to $Y_f$ is similar.
Note that here we overload the notation for $C_{XX}$ and $C_{YY}$ to be
the sample auto-covariance matrices of the \emph{output layer activations},
$X_f$ and $Y_f$.  Similarly $C_{XY}$ is the cross-covariance matrix of $X_f$ and $Y_f$.
$U$, $\Lambda$, and $V$ are solved by a singular value decomposition in the
\textcite{ewerbring1990canonical} CCA solution:
$C_{XX}^{-\frac{1}{2}} C_{XY} C_{YY}^{-\frac{1}{2}} = \widetilde{U} \Lambda \widetilde{V}^{T}$
(equation \ref{eq:background:svd_cca_solution}).

\subsection{(Many-View) Generalized Canonical Correlation Analysis}
\label{subsec:background:gcca}

Can we apply CCA to maximize correlation between more than just two views?
Unfortunately there is no single generalization of correlation to more than a
pair of random variables.  The extensions to more than two views, generalized CCA (GCCA),
frame the multiview correlation analysis problem as one of optimizing some function of
the correlation matrix between all pairs of views, $\mathbb{\phi}$:

\begin{equation}
  \begin{aligned}
  \mathbb{\phi} = &
  \left[
\begin{array}{cccc}
 z_1^T z_1 & z_1^T z_2 & \cdots & z_1^T z_V \\
 z_2^T z_1 & z_2^T z_2 & \cdots & z_2^T z_V \\
 \vdots & \vdots & \ddots & \vdots \\
 z_v^T z_1 & z_V^T z_2 & \cdots & z_V^T z_V
\end{array}
\right] \\
  \text{where } & \forall i \in 1 \ldots z_i = X_i u_i \\
  \text{subject to } & \|z_i\| = 1
  \end{aligned}
\label{eq:background:manyview_correlation_matrix}
\end{equation}

where $V$ is the number of views in our data, $X_i \in \mathbb{R}^{n \times p_i}$ is the data
matrix, $u_i \in \mathbb{R}^{p_i}$ are the canonical weights, and $z_i \in \mathbb{R}^{n}$ are
the canonical variates for view $i$.

\subsubsection{Problem Formulations}

\textcite{kettenring1971canonical} gives five different formulations of linear many-view
CCA objective.  In these formulations, the canonical variates
are learned, one-at-a-time, with the constraint that the variates are orthogonal to
each other.  For each canonical variate, we want to find the canonical weights $u_i$ that
satisfy one of the following optimization problems:

\begin{itemize}
  \item {\sc SUMCOR}: maximize the sum of correlations: $\max \sum_{i,j \in 1 \ldots V} \mathbb{\phi}_{i,j}$
  \item {\sc MAXVAR}: maximize the largest eigenvalue of $\mathbb{\phi}$: $\max \lambda_1$ 
  \item {\sc SSQCOR}: maximize sum of squared correlations: $\max \sum_{i,j \in 1 \ldots V} \mathbb{\phi}_{i,j}^{2}$
  \item {\sc MINVAR}: minimize the smallest eigenvalue of $\mathbb{\phi}$: $\min \lambda_V$
  \item {\sc GENVAR}: minimize determinant of $\mathbb{\phi}$: $det(\mathbb{\phi}) = \prod_{i \in \ldots V} \lambda_i$
\end{itemize}


In this thesis we focus on the
{\sc MAXVAR} objective to learn user embeddings\footnote{In Chapter \ref{chap:mv_twitter_users}, subsection
  \ref{subsec:mv_twitter_users:sumcor_gcca} we also consider an iterative algorithm
  to approximately solve for weights satisfying the {\sc SUMCOR} objective .}.
The {\sc MAXVAR} formulation is attractive since the optimal canonical weights $U_i$
can be found by standard linear algebra operations and singular value decompositions,
much like the two-view CCA objective.

\subsubsection{MAXVAR GCCA Problem}

\textcite{kettenring1971canonical} shows that the {\sc MAXVAR} GCCA formulation is equivalent
to a problem presented earlier by \textcite{carroll1968generalization}.  This formulation
introduces an auxiliary variable to the problem, a matrix $G \in \mathbb{R}^{n \times k}$ that
acts as a low-dimensional shared representation across views.  The {\sc MAXVAR} objective with
auxiliary variable formulation is as follows:

\begin{equation}
\begin{aligned}
\argmin_{G,U_i} \sum_{i} \left\|G - X_i U_i \right\|_F^2 \\
\text{such that} \ G^{T} G = I^{k}
\end{aligned}
\label{eq:background:maxvar_gcca_objective}
\end{equation}


Assuming columns of each $X_i$ are centered, the optimal solution for shared
representation $G$ and canonical weights $U_i$ is given as:

\begin{equation}
  \begin{aligned}
    G \Lambda G = \sum_{i=1}^{V} X_i (X_i^{T} X_i)^{-1} X_i^{T} & \text{eigendecomposition of rhs} \\
    \forall i \in 1 \ldots V, U_i = (X_i^{T} X_i)^{-1} X_i^T G
\end{aligned}
\label{eq:background:maxvar_gcca_solution}
\end{equation}

\paragraph*{Multiview LSA}

Unfortunately, the {\sc MAXVAR} GCCA solution given above does not scale well as the number
of examples in one's dataset nor the dimensionality of each view increases.  Note that the
matrix whose eigenvectors are $G$ has $n$ rows and columns.   As the number of examples increases,
this matrix will quickly become impossible to store in RAM on most computers.  Similarly,
inverting the $p_i \times p_i$ auto-covariance matrix, $(X_i^{T} X_i)$, will quickly become
intractable as the dimensionality of view $i$ increases.

\textcite{rastogi2015} offers several important tweaks to make this solution tractable:
the Multiview LSA algorithm.  The key contribution of Multiview LSA is that they consider
a truncated SVD decomposition of each view's data matrix: $X_i = A_i S_i B_i^{T}$.  They
use these low-rank decompositions to avoid
forming the full $n \times n$ matrix in \ref{eq:background:maxvar_gcca_solution}.  They show
that $G$ is approximately the left singular vectors of the following matrix:

\begin{equation*}
  \begin{aligned}
    \left(
      \begin{array}{ccc} A_1 & \cdots & A_V \end{array}
    \right)
  \end{aligned}
\label{eq:background:mvlsa_proj_matrix}
\end{equation*}

In practice, they also regularize the auto-covariance matrices, which leads to
a slight scaling of the columns of each $A_i$.

\paragraph*{Details}

One mild assumption for these methods is that the covariance matrices be invertible, that
they have full rank.  This assumption can be fulfilled by adding a small value to the
diagonal of each of the covariance matrices.  For example,
$\widetilde{C}_{XX} = C_{XX} + \epsilon I^{p \times p}$.  This tweak is known as
Regularized CCA or canonical ridge when applied to the two-view CCA problem and is
a critical component in Multiview LSA.

One potential difficulty with this solution are the orthonormality constraints on the
columns of $G$.  \emph{This constraint applies to all examples across the entire dataset}.
Because of this, it is not clear how one would design a stochastic or minibatch algorithm
to solve the {\sc MAXVAR} GCCA problem that only considers a subset of examples at a time.

It is also important to consider the presence of missing data within views when
applying GCCA to real data.  \textcite{van2006} introduces masking matrices
$\forall i \in 1 \ldots V, K_i$ into the {\sc MAXVAR} objective to address this problem:

\begin{equation}
\begin{aligned}
\argmin_{G,U_i} \sum_{i} \left\|K_i (G - X_i U_i) \right\|_F^2
\end{aligned}
\label{eq:background:maxvar_gcca_missingview_objective}
\end{equation}

Each mask, $K_i \in \mathbb{R}^{n \times n}$, is a diagonal matrix, where diagonal elements
are either 0 or 1.  Examples with data missing in a view are encoded by a zero whereas views
with data present are encoded by a one.
Although this is cosmetic, it is important to include these masking term, otherwise
canonical weights will be artificially forced to map views towards zero (assuming views
with missing data are represented as zero vectors).


\subsubsection{Neural Alternatives to GCCA}

Neural architectures that maximize a correlation objective are popular alternatives
and scale better to large numbers of examples than classic solutions to GCCA
problems.  \textcite{kumar2011co} elegantly outlines two main approaches these methods
take to learn a joint representation from many views: either by (1) explicitly
maximizing pairwise similarity/correlation between views or by (2) alternately optimizing a
shared, ``consensus'' representation and view-specific transformations to maximize
similarity.

Models such as the Siamese network proposed by \textcite{masci2014multimodal},
fall in the former camp, minimizing the squared error between embeddings learned from each
view, leading to a quadratic increase in the terms of the loss function size as the number
of views increase.  \textcite{rajendran2015bridge} extends Correlational Neural
Networks \parencite{chandar2015correlational} to many views and avoid this quadratic explosion
in the loss function by only computing correlation between each view embedding and the
embedding of a pivot view.  Although this model may be appropriate for tasks such as
multilingual image captioning, there are many datasets where there is no clear method
of choosing a pivot view.  The MAXVAR-GCCA objective does not suffer from a quadratic
increase in computational complexity with respect to the number of views, nor does it
require a privileged pivot view, since the shared representation is learned from the
per-view representations.

\subsubsection{Nonlinear (Deep) GCCA}

In spite of encouraging theoretical guarantees, multiview learning techniques cannot freely model nonlinear relationships between arbitrarily many views. Either they are able to model variation across many views, but can only learn linear mappings to the shared space \parencite{GCCA1}, or they simply cannot be applied to data with more than two views using existing techniques based on kernel CCA~\parencite{hardoon2004canonical} and deep CCA~\parencite{andrew2013}.  Deep Generalized Canonical Correlation Analysis (\dgcca) is one recently-introduced model that fills this gap.  Here we briefly describe the \dgcca{} model -- see \textcite{benton2017deep} for further details.


\paragraph*{Model}

\dgcca{} is a model that can benefit from the expressive power of deep neural networks
and can also leverage statistical strength from more than two views in data, unlike Deep CCA
which is limited to only two views.

\begin{figure}
\begin{center}
\includegraphics[height=0.4\textwidth]{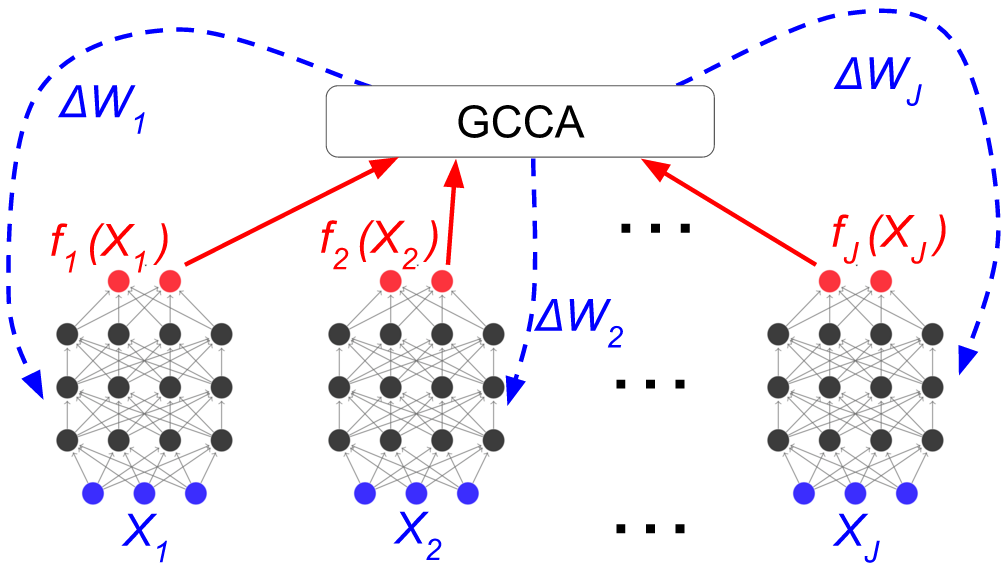}
\end{center}
\caption{A schematic of DGCCA with deep networks for $J$ views.}
\label{fig:mv_twitter_users:dgcca_schematic}
\end{figure}

\dgcca{} learns a nonlinear map for each view in order to maximize the correlation between the learnt representations across views. In training, \dgcca{} passes the input vectors in each view through multiple layers of nonlinear transformations and backpropagates the gradient of the GCCA objective with respect to network parameters to tune each view's network, as illustrated in Figure \ref{fig:mv_twitter_users:dgcca_schematic}.
The objective is to train networks that reduce the GCCA reconstruction error among their outputs. At test time, new data can be projected by feeding them through the learned network for each view.

In the \dgcca{} problem, we consider $J$ views in our data and let $X_j \in \mathbb{R}^{d_j \times N}$ denote the $j^{th}$ input matrix.\footnote{Our notation for this section closely follows that of \textcite{andrew2013}}
The network for the $j^{th}$ view consists of $K_j$ layers. Assume, for simplicity, that each layer in the $j^{th}$ view network has $c_j$ units with a final (output) layer of size $o_j$. 

The output of the $k^{th}$ layer for the $j^{th}$ view is $h^j_k = s(W^j_k h^j_{k-1})$, where  $s :  \mathbb{R} \rightarrow  \mathbb{R} $ is a nonlinear activation function and $W^j_{k} \in  \mathbb{R}^{c_k\times c_{k-1}}$ is the weight matrix for the $k^{th}$ layer of the $j^{th}$ view network. We denote the output of the final layer as $f_j(X_j)$. 

\dgcca{} can be expressed as the following optimization problem: find weight matrices $W^j = \{W^j_{1}, \ldots, W^j_{K_j}\}$ defining the functions $f_j$, and linear transformations $U_j$ (of the output of the $j^{th}$ network), for $j=1,\ldots,J$,  such that

\begin{align}
\label{eq:mv_twitter_users:dgcca}
\begin{split}
 	\argmin_{U_j \in \mathbb{R}^{o_j \times r}, G \in \mathbb{R}^{r
 	\times N}} &\sum_{j=1}^J
    \|G - U_j^\top f_j(X_j)\|_F^2 
    \\ \text{subject to} \qquad &GG^\top = I_r
\end{split},
\end{align}

where $G \in \mathbb{R}^{r\times N}$ is the shared representation we are
interested in learning. 

\paragraph*{Gradient Derivation Sketch}

Next, we show a sketch of the gradient derivation.  See \textcite{benton2017deep}
for the full gradient derivation with respect to network output layer.
It is straightforward to show that the solution to the GCCA problem is given by solving an eigenvalue problem. In particular, define $C_{jj} = f(X_j)f(X_j)^\top \in \mathbb{R}^{o_j \times o_j}$ to be the scaled empirical covariance matrix of the $j^{th}$ network output, and let $P_j = f(X_j)^\top C_{jj}^{-1} f(X_j) \in \mathbb{R}^{N \times N}$ be the corresponding projection matrix that whitens the data; note that $P_j$ is symmetric and idempotent.  We define $M = \sum_{j=1}^J P_j$. Since each $P_j$ is positive semi-definite, so is $M$. Then, it is easy to check that the rows of $G$ are the top $r$ (orthonormal) eigenvectors of $M$, and $U_j = C_{jj}^{-1}f(X_j)G^\top$. Thus, at the minimum of the objective, we can rewrite the reconstruction error as follows: 
\begin{align*}
\sum_{j=1}^J \|G - U_j^\top f_j(X_j)\|_F^2 &= 
\sum_{j=1}^J \|G - G f_j(X_j)^\top C_{jj}^{-1} f_j(X_j)\|_F^2 
= rJ - \tr(GMG^\top)
\end{align*}

Minimizing the GCCA objective (w.r.t. the weights of the
neural networks) means maximizing $\tr(GMG^\top)$, which is the sum of eigenvalues $L=\sum_{i=1}^r \lambda_i(M)$.
Taking the derivative of $L$ with respect to each output layer $f_j(X_j)$ we have:
\[
\frac{\partial L}{\partial f_j(X_j)}
= 2U_jG - 2U_jU_j^\top f_j(X_j)
\]
Thus, the gradient is the difference between the
$r$-dimensional auxiliary representation $G$ embedded into
the subspace spanned by the columns of $U_j$ (the first term) 
and the projection of
the actual data in $f_j(X_j)$ onto the said subspace (the second term).
Intuitively, if the auxiliary representation $G$ is far away from
the view-specific representation $U_j^\top f_j(X_j)$, then the
network weights should receive a large update.  Computing the gradient
descent update has time complexity $O(J N r d)$, where
$d = max (d_1, d_2, \ldots , d_J )$ is the largest dimensionality of the input views.

\paragraph*{Relationship to Semi-supervised Models over Text and Network}

\textcite{rahimi2018semi} describe applying a graph convolutional network (GCN)
to predict Twitter user location.  Although this model merges text and network
information about users it is not directly related to the multiview methods we
analyze here.  The GCN is fundamentally a
semi-supervised model that differentially weights features of neighboring
users within the network graph to better infer node labels.  On the other hand,
multiview methods learn example representations by finding transformations of each
example view to maximize the between-view correlation rather than predicting
supervision.

\emph{However}, there is nothing preventing one from using a GCN
as a view transformation layer, which can subsequently be tuned according to a
\dgcca{} objective.  The \dgcca{} learning algorithm is agnostic to the
architecture of the
transformation network as well as the form of input view representation, so long
as the objective is differentiable with respect to the neural network weights.

\section{Multitask Learning and Neural Models}
\label{sec:background:mtl_neural}

In chapters \ref{chap:mtl_mentalhealth} and \ref{chap:mtl_stance}, we use a machine
learning framework called \emph{multitask learning} (MTL) to inject user
information into classification models.  In this section we present the MTL setting
at a high level and discuss why neural networks are particularly convenient models
to train in this framework.

\subsection{Motivation}

MTL was first presented in \textcite{caruana1993multitask} and is discussed in detail in
Caruana's dissertation \parencite{caruana1997multitask}.  MTL is a machine learning
framework for exploiting \emph{related} auxiliary tasks to improve a classifier's
generalization performance at some main task that the practitioner cares about.  The
classifier is trained to perform well according to these auxiliary tasks along with the
main task, updating weights or a representation common across the tasks.  Caruana describes
MTL as introducing a human ``inductive bias'' to the main task model.  This inductive bias
is encoded by which additional tasks the practitioner believes will serve as useful guides
to a model that must perform well at the single main task.

Consider an example from \textcite{collobert:ea:2011}.  In this paper, the authors want
to improve a semantic role labeling system.  This system takes a sequence of tokens as
input and generates a sequence of labels encoding semantic roles, one for each token,
as output -- this is their main task.
They then consider a related task of language modeling -- the auxiliary task.  The auxiliary
language modeling task is formulated as maximizing the score that the model assigns to real
English sentences, while minimizing the score assigned to fake, generated English sentences.
They hypothesize that a model that can successfully discriminate between real and fake
English text will be better at assigning semantic role labels than models
that have a poor sense of what constitutes well-formed English.  They report reducing
semantic role labeling word error rate from 16.5 to 14.4 after joint MTL training with
the language modeling auxiliary task -- over a $10\%$ reduction in word error rate.

\subsubsection{Benefits}

There are several benefits to the MTL framework aside from improving classifier
generalization over traditional single-task learning.

The first is that auxiliary tasks are only necessary during train, not at test time.
The auxiliary tasks serve as beneficial regularizers for the classifier being learned, and
can be discarded.  This is analogous to the way that neural network weights may be trained
with dropout, weight decay, or other regularization techniques, but those terms only
influence how weights are updated during training, not how predictions are ultimately made.
This was the major motivation behind using multitask learning to improve pneumonia risk
prediction given medical history in \textcite{caruana1996using}.  In this work, they use the
results from lab tests as auxiliary tasks.  These lab tests are
time-consuming, expensive, and are only available
\emph{after a patient has been hospitalized}.  However, they are predictive of pneumonia risk,
so a classifier that can predict these results will also better predict pneumonia
risk.  These lab results are available at train time, but are clearly
unavailable during test.

MTL is especially effective when there are few labeled training
examples for the main task, but many labeled examples when considering
the entire set of auxiliary related tasks.  This effectively expands the
training size of the classifier's training set, reducing the error incurred
by sampling variance. Secondly, MTL allows for datasets where different subsets
of examples are annotated for different tasks.  This is especially important since
it allows the practitioner to combine disparate datasets togther even though
disjoint example sets are annotated in each case.

\subsection{Learning Setting}

In supervised MTL, we want to train a classifier such that it achieves low
expected loss across multiple tasks.  We are given a total of $T$ tasks and
one classifier for each task, $\{f_1, f_2, \ldots f_T\}$.  The classifier for task $t$ maps
examples from domain $\mathcal{X}$ to predictions in domain
$\mathcal{Y}_t$\footnote{The assumption that the domain of each classifier is the same
  is actually a simplification.  In general, the domain of each classifier may be different
  (either subsets of a shared feature set or completely different feature sets),
  so long as there exist parameters shared across classifiers \parencite{zhang2011multi}.}.
Each classifier $f_t$ is determined by two sets of parameters:

\begin{itemize}
  \item $\Theta$: parameters shared by all classifiers
  \item $\theta_t$: the task-specific parameters for task $t$
\end{itemize}

One can also consider all the $f_t$ as a single model that generates vector-valued
predictions : $\langle y_1, y_2, \ldots, y_T \rangle$.

If $\mathcal{L}_t: \mathbb{R}^{\mathcal{Y}_t, \mathcal{Y}_t} \rightarrow \mathbb{R}^{+}$ is the
loss function for task $t$ and examples are drawn from the joint probability distribution
$P_t$, then the MTL objective is as follows:

\begin{equation}
  \begin{aligned}
    \sum_{t=1}^{T} \E_{x,y \sim P_t} \mathcal{L}_t (y, f_t(x ; \Theta, \theta_t))
  \end{aligned}
  \label{eq:background:mtl_objective}
\end{equation}

In other words, we want to learn task-shared parameters, $\Theta$, and task-specific
parameters, $\{\theta_1, \ldots, \theta_T \}$, that minimize the average expected loss
across all tasks\footnote{In practice the objective will also include a regularization
  term to penalize large parameter weights.  This term is orthogonal to the multitask
  objective, though.}.

\subsubsection{Neural Models}

\textcite{caruana1993multitask} first introduces MTL in the context of neural models, and for
good reason: MTL is trivial to implement in a neural model.  This
is because neural models are simple to optimize with respect to multiple loss functions,
so long as each loss function is differentiable with respect to the model parameters.


\begin{figure}
\begin{center}
\includegraphics[page=1,trim={2cm 3cm 5cm 3cm},width=1.0\textwidth]{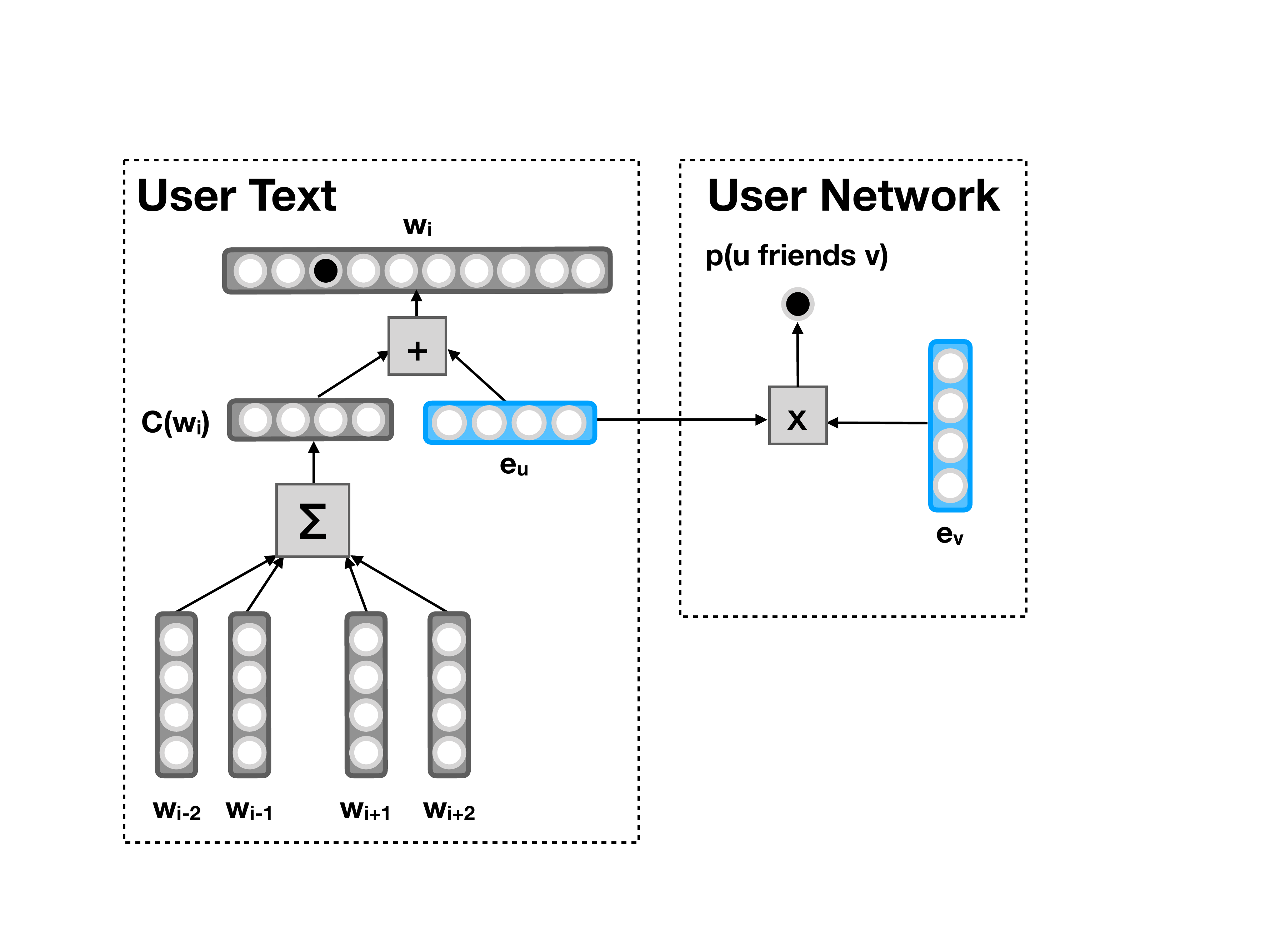}
\end{center}
\caption{Diagram of two tasks presented in \textcite{li2015} to learn user representations in
  an MTL setting.  The user representations that are learned are depicted by the blue
  vectors $e_u$ and $e_v$, and components for the text modeling and friendship prediction
  tasks are separated by dotted lines.  The text prediction task is addressed by a multinomial
  logistic regression model -- the feature set is the mean of average context word
  embeddings and a user's vector representation.  The friendship prediction task is
  defined as a parameterless logistic regression model determined solely by the dot
  product of the two user representations.}
\label{fig:background:li_mtl_user_model}
\end{figure}

Take the model presented in \textcite{li2015} for learning Twitter user representations
in an MTL framework that capture both similarity in posted text and closeness in
the social network \footnote{In their full model, \textcite{li2015} also learn
  representations that are predictive of other attributes such as occupation, location,
  and gender as well.  We omit these additional tasks as we mean to use this as a simple
  model of how neural network models and training is especially conducive to MTL.
  Inferring user representations is also a key problem in this thesis.}.

\paragraph*{Example: Multitask Learning of User Representations}

A simplified version of their model is diagrammed in Figure
\ref{fig:background:li_mtl_user_model}.  Consider two tasks to aid learning vector
representations for Twitter users: a user-conditioned language modeling task and a friend
prediction task. Their language modeling task is described by the following objective:

{
  \setlength{\jot}{0ex}
\begin{equation}
\begin{aligned}
  C(w_i, u) = & \sum_{j=i-k; j \neq i}^{i+k} L(w_j) \\
  s(w_i \| C(w_i), u) = & \frac{C(w_i) + e_u}{2k + 1} W_{\text{text}} \\
  p(w_i \| C(w_i), u) = & \frac{s(w_i, u)_{w_i}}{\sum s(w_i, u)}
\end{aligned}
\label{eq:background:li_mtl_text_obj}
\end{equation}
}

where $L$ is a word embedding lookup table mapping word indices to vectors, $C(w_i)$ is the
$k$-window word embedding context around word $w_i$,  and $e_u$ is the vector representation
for user $u$.  $e_u$ as well as $W_{\text{text}}$ and the word embeddings in $L$ are all
parameters learned during model training to maximize the probability, $p(w_i \| C(w_i), u)$.
This is
equivalent to the Paragraph Vector embedding technique where instead of learning a vector for
each paragraph, we learn a representation for each user \parencite{le2014distributed}.  The model
is trained similarly to paragraph vectors, where positive examples is the instantiated word in
its context and negative examples are generated by sampling uniformly from the remainder of
the vocabulary.

The friend prediction task is modeled more simply:

{
  \setlength{\jot}{0ex}
\begin{equation}
\begin{aligned}
  p(u \ \text{friends}\ v) = & \frac{1}{1 + exp\{-e_u^T e_v\}}
\end{aligned}
\label{eq:background:li_mtl_network_obj}
\end{equation}
}

where $e_u$ and $e_v$ are embeddings for two distinct users $u$ and $v$.  The probability that
user $u$ is friends with $v$ is determined by passing their dot product through a sigmoid
function.

The empirical log-likelihood for both of these tasks for a set of users $U$,
the sequence of words each user posts $w^u$, and pairs of friends $F$ is then:

{
  \setlength{\jot}{0ex}
\begin{equation}
\begin{aligned}
  \log \mathcal{L}_{\text{text}} = & \sum_{u \in U} \sum_{w_i \in w^{u}} p(w_i \| C(w_i), u) \\
  \log \mathcal{L}_{\text{friend}} = & \sum_{(u, v) \in F} p(u \ \text{friends }\ v) \\
  \log \mathcal{L}_{\text{joint}} = & \mathcal{L}_{\text{text}} + \mathcal{L}_{\text{friend}} 
\end{aligned}
\label{eq:background:li_mtl_joint_obj}
\end{equation}
}

Model parameters can be learned by alternately sampling pairs of users for the friend
prediction task, and sampling words in context for the language modeling task to update user
representations.  For each task, user
representations are updated by stochastic gradient descent.  This is a consequence of the
fact that the joint loss for both of these tasks is a linear combination of the per-task
losses, and so the joint loss remains differentiable with respect to the user representations.

\paragraph*{Other Multitask Models}

Although this thesis only considers neural networks trained in multitask
fashion, a wide variety of machine learning models have been extended to
the MTL setting.

Along with neural networks, \textcite{caruana1997multitask} presents MTL extensions
for feature-weighted k-nearest neighbor regression and decision trees.  In this model,
the per-feature weights for computing $\ell_2$ distance are selected to minimize average
mean squared error across tasks rather than for a single task.  Multitask decision trees are
learned by not just maximizing information gain of a single task, but a weighted average of
information gain across all tasks.

\textcite{evgeniou2004regularized} present a generic regularization-based framework
inspired by fitting support vector machines.  In this framework, each task's model is assumed
to have parameters which are close to each other.  This ``closeness'' is enforced by
penalizing large deviations from a set of shared parameters while maximizing the margin
from decision boundary (in the case of max-margin classification).
MTL has also influenced learning of unsupervised tasks such as clustering
\parencite{gu2009learning} and nonlinear regression models such as Gaussian
processes \parencite{yu2005learning}.


\subsubsection{Options for MTL}

The MTL framework is extremely flexible in how models can be trained.  This is both
a blessing and a curse: flexibility means that MTL is applicable to improving just about any
predictive model, but means that there is potentially more space to explore to find the
most effective way to deploy MTL.

\paragraph*{Task Selection and Weighting}

How does one define \emph{related} tasks?  This is the fundamental problem in deriving
generalization improvements from MTL training and one without a clear answer.  Simple
measures such as correlation between class labels are not necessary to identify related tasks as \parencite{caruana1997multitask}.  The best definition of task-relatedness offered in \textcite{caruana1997multitask} is not particularly illuminating:

\begin{quote}
The most precise definition for relatedness we have been able to devise so far is the
following: Tasks $A$ and $B$ are related if there exists an algorithm $M$ such that $M$ learns
better when given training data for $B$ as well, and if there is no modification to M that
allows it to learn A this well when not given the training data for $B$. While precise, this
definition is not very operational.
\end{quote}

This cannot be easily used as a heuristic for task selection since the definition of
relatedness amounts to going ahead and training a model on both tasks (showing tasks
are unrelated is even more difficult, requiring a sweep over all learning algorithms/models).
Similar to selecting tasks, it is typical to weight auxiliary tasks differently within
the loss function based on which are believed to be the most beneficial.  This is akin to a
soft selection of auxiliary tasks.

Although there are methods to jointly infer how ``related'' tasks are as well as
learn weights for each task \parencite{bakker2003task,kang2011learning}, there is no
panacea, since they typically make strong assumptions on how data were generated.
Ultimately, which auxilliary tasks that will lead to best improve generalization
performance must be selected by the practitioner based on their domain knowledge,
the data and tasks available, and empirical evaluation.

\paragraph*{Training Regimen}

Practically, there are a few additional decisions that must be made when training
neural models.  When updating parameters stochastic gradient descent, how should we
sample examples?  Sampling uniformly at random is not ideal when the number of examples is
unbalanced across tasks.  It is typical to experiment with how auxiliary tasks should be
sampled to avoid entraining weights too strongly towards one or another.  In this sense,
deciding how to
sample examples is similar to deciding how to weight the loss -- oversampling examples
from one task will lead to parameter updates that improve that task more.
A final pass of fine-tuning toward the main task is often helpful.  This, however, may be
liable to discarding the benefits of MTL if one does not freeze shared parameters or
limit the number and size of single-task updates.

\subsection{Discussion: Relationship to Multiview Methods}

The models we consider in this thesis integrate auxiliary information into
embeddings and models to improve generalization performance at some
downstream task.  The integration of auxiliary information can come in the
form of a multiview method, learning embeddings that capture correlation
between views, supervision used to condition distribution over topics in
a topic model, or as an auxiliary task for a supervised classifier.

Since many of the approaches we present here are naturally extended to
semi-supervised learning settings (multitask learning in particular), one
might be tempted to think of these methods as \emph{transductive}.  However, unlike
transductive algorithms which infer labels for a batch of unlabeled examples at
training time \parencite{gammerman1998learning}, the multiview and multitask algorithms we
present here do not need to infer labels for unlabeled examples.  Multiview
algorithms have no notion of a target or label
which the embedding is entrained to predict (although they can easily be extended
to have one \parencite{Wang_15b}).  Multitask training (particularly for neural models)
can easily be extended to incorporate new examples labeled for any of the main or
auxiliary tasks -- main task labels can be inferred by the model, but they need
not be used in the training process.

It is more useful to view multiview and multitask learning approaches as ways
of inserting inductive biases into learned model features.  The major
difference between these two approaches, multiview and multitask learning, is
in the kinds of biases that each can express.  Multiview methods suppose that the
latent features one wants to learn are best captured by that which is common between
a set of ``auxiliary'' features, or views.  Under this interpretation, we presume
that observed views are generated conditional on this latent feature (consider the
probabilistic interpretation of CCA \parencite{bach2005probabilistic}).  In neural
multitask learning, one supposes that this latent feature vector (e.g. a hidden
layer in one's network) is \emph{predictive of} the auxiliary tasks, and is
\emph{generated by} a separate set of input features.

To take an example from chapter \ref{chap:mv_twitter_users}, suppose we have
collected the past tweets and list of local network friends for a large set of
Twitter users.  We can take two separate approaches to model
these users.  We could apply a multiview representation learning method such as CCA
to map the text and network views to a shared space.  This approach assumes that the
text and network features are generated independently conditioned on some latent
feature vector.  If we took a supervised multitask learning approach, we would
either predict a user's local network from their text, their text from their
local network, or would predict both text and local network from a completely
separate set of input features.

One critical difference between multiview and multitask learning is the flexibility
afforded by multitask learning setting.  The multiview methods we consider in this
thesis are constrained to maximizing correlation between views, and often make
strong assumptions on the distribution of observations (e.g. observations are
Gaussian distribution).  On the other hand, multitask learning is closer to a
philosophy of model training that happens to be easily translated to neural
network training.

\cleardoublepage

\chapter{Multiview Embeddings of Twitter Users}
\label{chap:mv_twitter_users}

In this chapter we present methods to learn \emph{unsupervised} embeddings for
a general set of
users from different views of their online behavior.  We evaluate these embeddings
both intrinsically according to how well they capture hashtag usage and friending
behavior and extrinsically according to how well they predict demographic features.
This chapter was adapted mainly from \textcite{benton2016learning}, published as a
short paper in ACL 2016.  The deep GCCA
experiments were presented in \textcite{benton2017deep}, an ar$\rchi$iv preprint.
The LasCCA algorithm was implemented and adapted to support missing views during
a 2017 summer internship at Amazon.

Dense, low-dimensional vector embeddings have a long history in NLP, and recent work
on neural models have provided new and  popular algorithms for training representations
for word types \parencite{mikolov2013distributed,faruqui2014}, sentences \parencite{kiros2015}, and entire
documents \parencite{le2014distributed}.  These embeddings exhibit desirable
properties, such
as capturing some aspects of syntax or semantics and outperforming their sparse
counterparts at downstream tasks.

While there are many approaches to generating embeddings of {\em text}, it is not clear how
to learn embeddings for social media {\em users}.  There are several different types of data
(views) we can use to build user representations: the text of messages they post,
neighbors in their local network, articles they link to, images they upload, etc.  Although
user embeddings can always be finetuned for a supervised objective, it is unclear which
unsupervised views and methods perform best across a variety of tasks.

Multiview embedding methods such as Generalized Canonical Correlation Analysis (\gcca)
\parencite{carroll1968generalization,van2006,arora2014multi,rastogi2015}
are attractive methods for simultaneously capturing information from multiple user views.
These methods may be more appropriate for learning user embeddings than concatenating views
into a single vector, since views may correspond to different modalities (image vs. text
data) or have very different distributional properties.  Treating all features as equal in
this concatenated vector would not be appropriate.

In this chapter we present an extension of the {\sc MAXVAR}-GCCA problem that offers
increased flexibility
in learning user embeddings than standard \gcca: weighted \gcca{} (\wgcca).
\wgcca{} allows the practitioner to discriminatively weight the per-view loss,
forcing user embeddings to capture variation in some views more closely than others.
View weighting is chosen based on either a prior notion of which views will be the most
informative or by tuning to improve some downstream metric -- this is up
to the embedder's discretion.
We also consider an algorithm to approximately solve the (linear) {\sc SUMCOR}-GCCA
problem, large-scale CCA \parencite{fu2016efficient} (\lascca), as another multiview user embedding
method.  We adapt the \lascca{} implementation presented in \textcite{fu2016efficient}
to support data with missing views (especially important when considering data compiled from
social media).

We evaluate multiview embeddings at how well they capture hashtag usage and friending
behavior and how well they predict user demographic features.  We compare their
performance at these tasks to single-view baselines and show that the location of users
in embedding space can capture average peoples' notions of what constitutes a similar group
of users.  This is analogous to how word embeddings capture semantic and syntactic
properties of word types.

In Section \ref{sec:mv_twitter_users:learning_user_embeddings} we first describe the
different types of user behavior used to learn embeddings and how this dataset was assembled.
Sections \ref{sec:mv_twitter_users:baseline_embedding_methods}
and \ref{sec:mv_twitter_users:multiview_embedding_methods}
describe the baseline and multiview methods we use to learn embeddings.
Section \ref{sec:mv_twitter_users:experiment_description} describes
how embeddings were evaluated and Section \ref{sec:mv_twitter_users:results} finally contains
both quantitative and qualitative evaluation of user embeddings.

\section{User Behavior Data}
\label{sec:mv_twitter_users:learning_user_embeddings}

What is the best type of behavior to learn user embeddings on?  Although the answer
ultimately depends on how these embeddings will be used, some types of user behavior and
embedding methods will be more appropriate for a variety of tasks.  To answer this question,
we assembled a dataset of general Twitter users, with multiple aspects of user
behavior.  Knowing how the dataset was assembled is critical to understanding
what kind of user behavior is available to each embedding method.

\subsection{Data Collection}

We uniformly sampled 200,000 users from a stream of publicly available tweets from
the 1\% Twitter stream from April 2015.  
We removed users with verified accounts, more than 10,000 followers, or non-English
profiles to restrict to typical, English speaking
users\footnote{Language identified with the python module \texttt{langid} \parencite{lui2012langid}.}.
For each user we collected their 1,000 most recent tweets, and then filtered out
non-English tweets.  We removed users without English tweets in January or February
2015, yielding a total of 102,328 users.  Although limiting tweets to only these
two months restricted the number of tweets we were able to work with, it also
ensured that our data are drawn from a narrow time window, controlling for differences
in user activity over wide stretches of time. This allows us to learn 
distinctions between users, and not temporal distinctions of content.

Next, we collect information about the users' friend and mention networks.
Specifically, for each user \textbf{mentioned} in a tweet by one of the 102,328 users,
we collect their 200 most recent English tweets
from January and February 2015. Similarly, we collected the 5,000 most
recently added friends and followers for each of the 102,328 users. We then sampled
10 friends and 10 followers for each user and collected up to the 200 most recent
English tweets for these followers and friends from January and February 2015.
Limits on the number of users and tweets per user were imposed so that we
could operate within Twitter's API limits\footnote{The Twitter REST API limits on collecting local network information are especially strict.  A non-privileged API key can only pull 5,000 friend/follower IDs per minute for a single user.}.

This data supports our evaluation tasks as well as the four sources of behavior/content
for each user: their tweets, tweets of mentioned users, friends, and
followers.

\subsection{User Views}
\label{subsec:mv_twitter_users:user_views}

We consider four main views/sources of information about a user.  \textbf{ego} information
as represented by the text in public tweets the user posts, \textbf{mention}ed information
represented by messages made by people mentioned in a tweet posted by the ego user,
\textbf{friend} information for those people who the ego user follows, and \textbf{follower}
information for those who follow the ego user.  Although there are other views we could have
collected (e.g. the user description or image), prior work has shown that these four views
are predictive of latent user attributes, and therefore would be useful for learning user
embeddings \parencite{volkova2014}.

Two main representations are considered when constructing views: either
text representations or a direct representation of the friend or follower IDs.

\subsubsection{Text}

For each text source we can aggregate the many tweets into a single document, e.g. all
tweets written by accounts mentioned by a user.
We represent this document as a bag-of-words (\bow) in a vector space model with 
a vocabulary of the 20,000 most frequent word types after stopword removal.
We consider TF-IDF weighted \bow{} vectors.  This was done for tweets made by the \textbf{ego}
user, \textbf{mentions}, \textbf{friends}, and \textbf{followers}.

A common problem with these 
representations is that they suffer from
the curse of dimensionality. A natural solution is to apply a dimensionality
reduction technique to find a compact representation that captures as much information
as possible from the original input. Here, we consider principal components analysis
(PCA), a ubiquitous linear dimensionality reduction technique.   The text views that are
fed into multiview embedding methods are all first reduced by PCA before learning the
embedding.  We run PCA and extract up to the top 1,000 principal components for each
of the above views.  This speeds up fitting multiview embedding methods since the
feature dimensionality of each view is reduced.

\subsubsection{Network}

An alternative to text based representations is to use the social network of users as
a representation. We encode a user's social network
as a vector by treating the set of users in the social graph as a vocabulary, where
users with similar social
networks have similar vector representations (\collab).  This is an $n$-dimensional vector
that encodes the user's social network as a bag-of-words over this vocabulary. In other
words, a user is represented by a summation of the one-hot encodings of each neighboring
friend or follower in their social network. In this representation, the number of friends
two users have
in common is equal to the dot product between their social network vectors. We define the
social network as one's followers or friends.
The motivation behind this representation is that users who have similar networks may behave
in similar ways. Such network features are commonly used to construct user representations
as well as to make user recommendations~\parencite{lu2012,kywe2012}.

The binary representations of local network are reduced to the
top 1,000 principal components, as are the text representations.

\section{Baseline Embedding Methods}
\label{sec:mv_twitter_users:baseline_embedding_methods}

Each of these views can be treated as a user embedding in their own right.
They can also be combined using different methods to yield aggregate user
representations across views. Here we describe baseline user embeddings we evaluate.

\subsection{PCA}

For the following experiments, we consider the PCA representations
as a baseline.  We consider up to the top 1,000 principal
components within each view as the user embedding.  In order to fairly compare
multiview embedding methods to methods that do not maximize correlation between
views, we also consider a na\"{i}ve combination of PCA views as an embedding.

We consider \emph{all} possible combinations of views obtained by concatenating original view features, and
subsequently reducing the dimensionality by PCA.  By considering all possible concatenation of views, we
ensure that this method has access to the same information as multiview methods.
Both the raw \bow{} and \bowpca{} representations have been explored in previous work
for demographic prediction~\parencite{volkova2014,zamal2012homophily} and recommendation
systems \parencite{abel2011,zangerle2013}.
Only the best performing view subset evaluated on the development set is reported on test.


\subsection{Word2Vec}

\bowpca{} is limited to linear representations
of \bow{} features based on global context. Modern
neural network based
approaches to learning word embeddings, including
word2vec continuous bag of words and skipgram models,
can learn representations that capture
local context around each word \parencite{mikolov2013}.
We represent each view as the simple average
of the word embeddings for all tokens within that view
(e.g., all words written by the ego user). Word embeddings
are learned on a sample of 87,755,398 tweets
and profiles uniformly sampled from the 1\% Twitter
stream in April 2015 along with all tweets and profiles
collected for our set of users -- a total of over a billion
tokens.  We use the word2vec tool, select either skipgram
or continuous bag-of-words embeddings on dev
data for each prediction task, and train for 50 epochs.
We use the default settings for all other parameters.

\section{Multiview Embedding Methods}
\label{sec:mv_twitter_users:multiview_embedding_methods}

Here we describe three different methods for learning multiview
user embeddings.  Each of these multiview embedding methods are
evaluated against each other at the tasks described in
section \ref{sec:mv_twitter_users:experiment_description}.

\subsection{MAXVAR-GCCA}

We use Generalized Canonical Correlation Analysis (\gcca)
\parencite{carroll1968generalization} to learn a single embedding
from multiple views.  \gcca{} finds $G$, $U_i$ that minimize:

\begin{eqnarray}
\argmin_{G,U_i} \sum_{i} \left\|G - X_i U_i \right\|_F^2 ~~~~~~
\mathrm{s.t.} \ G^{T} G = I
\end{eqnarray}
where $X_i \in \mathbb{R}^{n \times d_i}$ corresponds to the data matrix for
the $i$th view, $U_i \in \mathbb{R}^{d_i \times k}$ maps from the latent space to observed view $i$, and $G \in \mathbb{R}^{n \times k}$  contains all user representations
\parencite{van2006}.

\subsubsection{Weighted GCCA}

Since each view may be more or less helpful for a downstream task, we do not want to treat each view equally in learning
a single embedding. Instead, we weigh each view differently in the objective:

\begin{eqnarray}
\argmin_{G,U_i}\! \sum_{i}\! w_i\! \left\|G - X_i U_i \right\|_F^2 ~
\mathrm{s.t.} \ G^{T} G = I, w_i \ge 0 \!\!
\end{eqnarray}

where $w_i$ explicitly expresses the importance of the $i$th view in determining the joint embedding.  The columns of $G$ are the eigenvectors of
$\sum_{i} w_i X_i (X_i^{T} X_i)^{-1} X_i^{T}$, and the solution for $U_i = (X_i^{T} X_i)^{-1} X_i^{T} G$.  In our experiments,
we use the approach of \textcite{rastogi2015} to learn $G$ and $U_i$, since it is more memory-efficient than decomposing the sum of projection matrices.

We also consider a minor modification of \gcca, where $G$ is scaled by the
square-root of the singular values of $\sum_i w_i X_i X_i^{T}$ (\gccasv). 
This is inspired by previous work showing that scaling each feature
of multiview embeddings by the singular values of the data matrix can
improve performance at downstream tasks such as image or caption retrieval
\parencite{mroueh2015}.  Note that if we only consider a single view, $X_1$,
with weight $w_1=1$, then the solution
to \gccasv{} is identical to the PCA solution for data matrix $X_1$, without
mean-centering.

\subsection{SUMCOR-GCCA}
\label{subsec:mv_twitter_users:sumcor_gcca}

In addition to the {\sc MAXVAR}-\gcca{} objective, we also consider another
generalization of CCA to more than two views: {\sc SUMCOR}-\gcca.
The {\sc SUMCOR}-\gcca{} problem is given in Equation
\ref{eq:mv_twitter_users:sumcor_gcca}:

\begin{align}
\label{eq:mv_twitter_users:sumcor_gcca}
\begin{split}
  \argmax_{\forall i \in [V], U_{i} \in \mathbb{R}^{m_i \times k}} \sum_{i=1}^{V} \sum_{j \neq i} Tr[U_i^{T} X_i^{T} X_j U_j] 
    \\ \text{subject to} \ \forall i \in [V], U_i^{T} X_i^{T} X_i U_i = I^{k}
\end{split},
\end{align}

where $V$ is the number of views and $U_i$ are the canonical weights
for view $i$.  {\sc SUMCOR}-\gcca{} seeks to find mappings that maximize the
sum of total correlation captured between every pair of views while
ensuring that the canonical variates for each view are orthonormal
as in CCA.  This differs from the {\sc MAXVAR}-GCCA objective in two ways:
(1) {\sc SUMCOR}-\gcca{} requires no nuisance variable, $G$, to ensure
views are mapped close to each other.  The orthonormality of projected
views is ensured by the hard constraints in the objective.
(2) The {\sc SUMCOR}-\gcca{} problem seeks to maximize the sum of correlations
between each pair of views.
The {\sc MAXVAR} formulation instead seeks to maximize the maximum eigenvalue
of the correlation matrix between all pairs of views \parencite{kettenring1971canonical}.

Jointly solving for all $U_i$ is difficult, so we run the Large-scale generalized
CCA (\lascca) algorithm \parencite{fu2016efficient} for a fixed number of iterations (100)
to solve for the mappings for each view. \lascca{} proceeds by
maximizing the {\sc SUMCOR}-\gcca{} objective with respect to each
$U_i$ round-robin, holding all other view mappings fixed. We consider this
multiview objective
for three reasons: (1) It allows us to compare if a slightly different
multiview objective yields similarly-performing embeddings to those learned
to maximize the {\sc MAXVAR}-GCCA. (2) We can assess how
performant the learned embeddings are as a function of \lascca{} epochs
devoted to solving the GCCA problem.  (3) Although \lascca{} does
not guarantee an optimal solution, the algorithm is designed to scale
well when the input views are very high-dimensional and sparse, avoiding
keeping low-rank approximations to the sum of projection matrices as in
multiview LSA.  This allows \lascca{} to learn multiview embeddings
directly from, for example, a bag-of-words in all of a user's tweets.

\subsubsection{Robust \lascca{} Algorithm}

\begin{algorithm}
  \caption{Rank-$k$ robust \lascca{}}
  \label{alg:mv_twitter_users:lascca}
  \begin{algorithmic}[1]
    \REQUIRE $\{ X_i, G_i, \redden{K_i} \}_{i=1}^{V}$ \Comment{Observations, auxiliary variates, and masks for each view}
    \REQUIRE $T$ \Comment{No. of epochs}
    \FOR{$i \gets 1 \ldots V$}
     \STATE $\redden{\mathcal{K}_i = \sum_{j=1, j \neq i}^{V} K_j}$ \Comment{No. of non-zero views per example}
     \STATE $\redden{\mathbb{K}_i = \mathbbm{1}[\mathcal{K}_i] K_i}$ \Comment{Indicator of data in view $i$ \emph{and} at least one other view}
    \ENDFOR
    \FOR{$t \gets 1 \ldots T$}
     \FOR{$i \gets 1 \ldots V$}
      \STATE $H_i \gets \texttt{H}_{\texttt{compute}}(\{X_i\}_{i=1}^{V}, \{G_j\}_{j=1, j \neq i}^{V}, \redden{\mathcal{K}_i, \mathbb{K}_i})$ \Comment{See Algorithm \ref{alg:mv_twitter_users:h_compute}}
      \STATE $U^{'}_i S_i V^{'}_i \gets H_i  $ \Comment{Singular value decomposition of $H_i$}
      \STATE $G_i \gets U^{'}_i V^{'}_i$
      \STATE $\hat{U}_i \gets \argmin_{U} \| \redden{\mathbb{K}_i} (X_i U - G_i ) \|_2^{F}$
     \ENDFOR
    \ENDFOR
   \STATE 
   \STATE \textbf{return} $(\{G_i\}_{i=1}^{V}, \{\hat{U}_i\}_{i=1}^{V})$
\end{algorithmic}
\end{algorithm}

\begin{algorithm}
  \caption{$\texttt{H}_{\texttt{compute}}$ subroutine for \lascca}
  \label{alg:mv_twitter_users:h_compute}
  \begin{algorithmic}[1]
    \REQUIRE $i$ \Comment{View to calculate $H$ for}
    \REQUIRE $\{X_j\}_{j=1}^{V}, \{G_j\}_{j=1, j \neq i}^{V}, \redden{\mathcal{K}_i, \{\mathbb{K}_j\}_{j=1}^{V}}$ \Comment{Observations, auxiliary variates, and masking matrices for each view}
    \FOR{$j \gets 1 \ldots V$}
     \IF{$j \neq i$}
      \STATE $\hat{R}_j \gets \argmin_{R} \| \redden{\mathbb{K}_j} (X_j R - G_j)\|_2^{F}$
      \STATE $C_j \gets \redden{\mathbb{K}_j} X_j \hat{R}_j$
     \ENDIF
    \ENDFOR
    \STATE $P_i \gets \redden{\frac{V}{\mathcal{K}_i}} \sum_{j =1, j \neq i}^{V} C_j$
    \STATE $\hat{E}_i \gets \argmin_{E} \| \redden{\mathbb{K}_i} (X_i E - P_i) \|_2^{F}$
    \STATE $H_i \gets \redden{\mathbb{K}_i} X_i \hat{E}_i$
   \STATE \textbf{return} $H_i$
\end{algorithmic}
\end{algorithm}

The \lascca{} algorithm is shown in Algorithm
\ref{alg:mv_twitter_users:lascca}, and relies on the subroutine
$\texttt{H}_{\texttt{Compute}}$ in Algorithm
\ref{alg:mv_twitter_users:h_compute}. In order to support our Twitter
data, we modified the original \lascca{} algorithm to ignore views with
missing data similar to the modification of multiview LSA.  The terms
that differentiate this algorithm from
the \lascca{} algorithm presented in \textcite{fu2016efficient} are
highlighted in red.  \lascca{} is more computationally efficient
than standard GCCA algorithms with many, high-dimensional views. For
example, \textcite{rastogi2015} requires (at best) an SVD of a
dense $n \times (Vk)$ matrix to solve the {\sc MAXVAR} \gcca{} objective,
where $k$ determines the low-rank approximation of each view's data
matrix. \lascca{} only involves solving a series of (possibly sparse)
linear least squares problems, and computing SVDs of dense
$n \times k$ data matrices which can be performed in $O(n k^2)$ time.

Robust LasCCA ignores views where no features are active,
avoiding unnecessarily entraining projected views toward zero.
$\mathbb{K}_i$ is an $n \times n$ diagonal matrix that masks examples
where either (1) view $i$ has no active features or (2) \emph{all
views but} $i$ have no active features.  $\mathcal{K}_i$ encodes
the number of active views that are not view $i$, for each example.
This is used to rescale the rows of $P_i$, to account for examples
that are missing data from views.

\begin{figure}
\begin{center}
\includegraphics[width=0.7\textwidth]{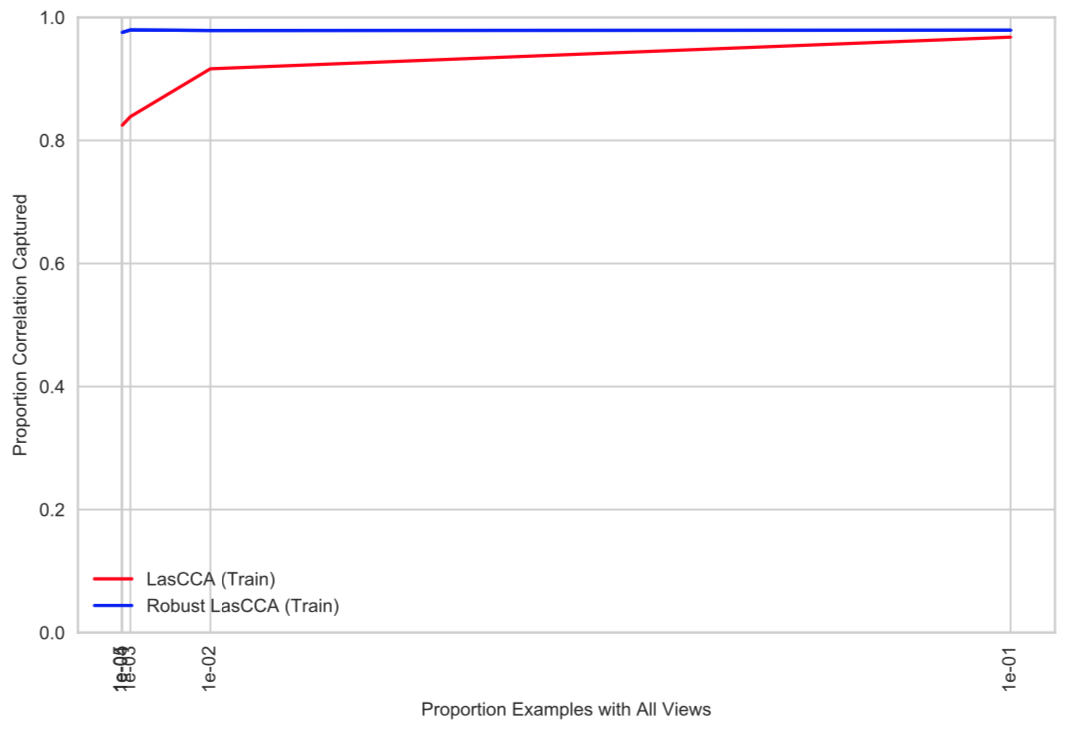}
\end{center}
\caption{Proportion of train correlation captured by vanilla and
  robust \lascca{} after 5 epochs of training, learning $k = 10$
  canonical variates. Proportion of examples where data from
  all-views-but-one are missing is listed along the x-axis.
  The leftmost point corresponds to a dataset where only 10 out of
  $10^5$ examples have active features in all views. Proportion
  correlation captured of 1.0 is optimal.}
\label{fig:mv_twitter_users:robust_lascca_validation}
\end{figure}

We verified the correctness of robust \lascca{} in the face of missing
views by recovering directions
of maximal correlation from a synthetic dataset. Data was generated by
first creating $N = 10^{5}$ example latent feature vectors of
dimension $F = 100$, with a randomly selected five features active in
each example, values sampled from a unit Gaussian. $F \times F$ sparse
maps from latent to observed views (three views total) were generated
with 10\% non-zero values, values also drawn from unit Gaussians. A
missingness parameter $\rho \in [0,1]$ was varied, such that $\rho$
proportion of examples contained active features in one and only
one view, and $(1 - \rho)$ proportion of examples contained active
features in all views. Figure
\ref{fig:mv_twitter_users:robust_lascca_validation} shows that
this robust extension of \lascca{} achieves near optimal proportion
correlation captured on train, regardless of how many examples are
missing data. Vanilla \lascca{} is sensitive to these missing view
examples, artifically forcing the view 1 projection toward zero when
these examples should be ignored.

\section{Experiment Description}
\label{sec:mv_twitter_users:experiment_description}

We selected three prediction tasks to evaluate the
effectiveness of the multi-view user embeddings: user
engagement prediction, friend recommendation and
demographic characteristics inference. Our focus is to
show the performance of multiview embeddings 
compared to other representations, not on building the
best system for a given task.

\subsection{Learning Embedding Details}

\gcca{} embeddings were learned over combinations of the views in Subsection \ref{subsec:mv_twitter_users:user_views}.  When available, we also consider \gccawnet{}, where in
addition to the four text views, we also include the follower and friend network views used by \bowpcawnet.
For computational efficiency, each of these views was first reduced in dimensionality
by projecting its \bow{} TF-IDF-weighted representation to a 1000-dimensional vector through PCA.\footnote{
We excluded count vectors from the \gcca{} experiments for computational efficiency since they performed
similarly to TF-IDF representations in initial experiments.
} We add
an identity matrix scaled by a small amount of regularization, $10^{-8}$, to the
per-view covariance matrices before inverting, for numerical stability, and use
the formulation of \gcca{} reported in \textcite{van2006}, which ignores rows with
missing data (some users had no data in the mention tweet view and some users accounts
were private).  We tune the weighting of each view $i$,
$w_i \in \{0.0, 0.25, 1.0\}$, discriminatively for each task, although the
\gcca{} objective is unsupervised once the $w_i$ are fixed (weighting swept over only for
linear \gcca{} embeddings).

When learning deep GCCA (\dgcca) and \lascca{} embeddings, we do
not apply any
view-weighting\footnote{Although it is not difficult to imagine altering the \dgcca{}
  and \lascca{} objectives to per-view loss weighting.}.  For \lascca{}, we consider
embeddings learned over the following sets of views: \{ego text, friend network\}, all four
text views, and all views (all text views along with two network views).  We run the \lascca{}
algorithm for a fixed 100 epochs and a maximum of 20 for solving linear least squares subproblems\footnote{}

When we compare representations in the following tasks, we sweep over embedding
width in $\{10, 20, 50, 100, 200, 300, 400, 500, 1000\}$ for all methods.
We also consider concatenations of vectors for every possible subset of
views: singletons, pairs, triples, and all views for the \bowpca{} baseline.

\subsubsection*{Deep GCCA Details}

We trained 40 different \dgcca{} model architectures, each with identical architectures across all text and network views, where the width of the hidden and output layers, $c_{1}$ and $c_{2}$, for each view
are drawn uniformly from $[10,1000]$, and the auxiliary representation width $r$ is drawn
uniformly from $[10,c_{2}]$\footnote{We chose to restrict ourselves to a single hidden layer with
non-linear activation and identical architectures for each view, so as to avoid a fishing
expedition.  If \dgcca{} is appropriate for learning
Twitter user representations, then we should be able to find a good architecture with little exploration.}.
All networks used ReLUs as activation functions, and were
optimized with Adam \parencite{kingma2014adam} for 200 epochs\footnote{From preliminary
experiments, we found that Adam pushed down reconstruction
error more quickly than SGD with momentum, and that ReLUs were easier to optimize than
sigmoid activations.}.
Networks were trained on 90\% of 102,328 Twitter users, with 10\% of users used as a
tuning set to estimate heldout reconstruction error for early stopping.
We report development and test results for the best performing model for each
downstream task development set.  Learning rate was set to $10^{-4}$ with an L1 and L2 regularization constants of $0.01$ and $0.001$ for all weights.  This setting of regularization constants led to low reconstruction error in preliminary experiments.

\subsection{User Engagement Prediction}
\label{subsec:mv_twitter_users:hashtag_results}

The goal of user engagement prediction is to determine which topics a user will likely tweet about, using the hashtags they mention as a proxy.
This task is similar to hashtag recommendation for a tweet based on its contents
\parencite{kywe2012,she2014,zangerle2013}. 
\textcite{purohit2011} presented a supervised task to predict if a hashtag would appear in a tweet using features from
the user's network, previous tweets, and the tweet's content. 

We selected the 400 most frequently used hashtags in messages authored by our users and which first appeared in March 2015,
randomly and evenly dividing them into development and test sets.
We held out the first 10 users who tweeted each hashtag as exemplars of users that would use the hashtag in the future.
We ranked all other users by the cosine distance of their embedding to the average embedding of these 10 users.
Since embeddings are learned on data pre-March 2015, the hashtags cannot impact the learned representations.
Performance is measured using precision and recall at $k$, as well as mean reciprocal rank (MRR), where a user is marked as correct
if they used the hashtag.  Note that this task is different than that reported in~\textcite{purohit2011}, since we are making recommendations
at the level of users, not tweets.

\subsection{Friend Recommendation}

The goal of friend recommendation/link prediction is to recommend/predict other accounts for a user to follow \parencite{liben2007}.

We selected the 500 most popular accounts -- which we call celebrities -- followed by our users, randomly, and evenly divided them into dev and test
sets.  We randomly select 10 users who follow each celebrity and rank all other users by cosine distance to the average of these 10 representations. The
tweets of selected celebrities are removed during embedding training so as not to influence the learned
representations. We use the same evaluation as user engagement prediction, where a user is marked as correct if they follow the given celebrity.

For both user engagement prediction and friend recommendation we z-score normalize each feature, subtracting off the mean
and scaling each feature independently to have unit variance, before computing cosine similarity.  We select the
approach and whether to z-score normalize based on the development set performance.

\subsection{Demographic Prediction}

Our final task is to infer the demographic characteristics of a user \parencite{zamal2012homophily,chen2015}.

We use the dataset from \textcite{volkova2014,volkova2015thesis}
which annotates 383 users for age (old/young), 383 for gender
(male/female), and 396 political affiliation (republican/democrat), with balanced classes.  Predicting each characteristic is a binary supervised prediction task.
Each set is partitioned into 10 folds, with two folds held out for test, and the other eight for tuning via cross-fold validation.
The provided dataset contained tweets from each user, mentioned users, friends and follower networks.  It did not contain the actual social
networks for these users, so we did not evaluate \collab, \bowpcawnet, or \gccawnet{}
at these prediction tasks.

Each feature for feature set was z-score normalized before being passed to a linear-kernel SVM where we swept over $10^{-4}, \ldots, 10^{4}$ for the penalty on the error term, $C$.

\section{Results}
\label{sec:mv_twitter_users:results}

\subsection{User Engagement Prediction}
\label{subsec:mv_twitter_users:engagement_results}

\begin{table}
  \tiny
\begin{center}
\begin{tabular}{|l|c|ccc|ccc|c|}
\hline
\bf Model & \bf Dim & \bf P@1 & \bf P@100 & \bf P@1000 & \bf R@1 & \bf R@100 & \bf R@1000 & \bf MRR \\ \hline
\bow    & 20000 & 0.03/0.045 & 0.021/0.014 & 0.009/0.005 & 0.001/0.002 & 0.075/0.053 & 0.241/0.157 & 0.006/0.006 \\
\bowpca & 500   & 0.050/0.060 & 0.024/0.024 & 0.011/0.008 & 0.002/0.003 & 0.079/0.080 & 0.312/0.290 & 0.007/0.009 \\
\collab & NA    & 0.02/0.015 & 0.013/0.014 & 0.006/0.006 & 0.001/0.000 & 0.042/0.047 & 0.159/0.201 & 0.004/0.004 \\
\bowpcawnet & 300 & 0.020/0.025 & 0.019/0.019 & 0.010/0.008 & 0.001/0.002 & 0.068/0.074 & 0.293/0.299 & 0.006/0.006 \\
\wordtovec & 100 & 0.030/0.025 & 0.022/0.016 & 0.009/0.007 & 0.000/0.001 & 0.073/ 0.057 & 0.254/0.217 & 0.005/0.004 \\ \hline
\gcca[text]  & 100 & 0.080/0.060 & 0.027/0.024 & 0.012/0.009 & 0.004/0.002 & 0.095/0.093 & 0.357/0.325 & 0.011/0.008 \\
\gccasv & 500 & 0.090/0.060 & 0.030/0.026 & \bf 0.012/0.010 & 0.003/0.003 & 0.104/0.106 & 0.359/0.334 & 0.010/0.011 \\
\gccawnet & 200 & 0.065/0.060 & 0.031/0.027 & 0.013/0.009 & \bf 0.003/0.004 & 0.105/0.113 & 0.360/0.346 & 0.011/0.011 \\
\lascca[ego+friendnet] & 400 & 0.060/0.045 & 0.032/0.027 & \bf 0.013/0.010 & 0.003/0.002 & \bf 0.111/0.115 & 0.371/0.358 & 0.012/0.011 \\
\lascca[text] & 300 & \bf 0.075/0.080 & 0.029/0.025 & \bf 0.012/0.010 & \bf 0.003/0.004 & 0.102/0.101 & 0.360/0.330 & 0.010/0.011 \\
\lascca[all] & 500 & 0.045/0.050 & \bf 0.030/0.028 & \bf 0.012/0.010 & 0.001/0.002 & 0.108/0.114 & 0.378/0.368 & 0.009/0.010 \\
\dgcca[all] & 710 & \bf 0.070/0.080 & \bf 0.030/0.028 & \bf 0.013/0.010 & \bf 0.003/0.004 & 0.105/0.112 & \bf 0.385/0.373 & \bf 0.010/0.012 \\ \hline
\netpop & NA    & 0.000/0.000 & 0.001/0.000 & 0.001/0.001 & 0.000/0.000 & 0.001/0.002 & 0.012/0.012 & 0.000/0.000 \\
\rand   & NA    & 0.000/0.000 & 0.001/0.000 & 0.000/0.000 & 0.000/0.000 & 0.001/0.000 & 0.002/0.008 & 0.000/0.000 \\ \hline
\end{tabular}
\end{center}
\caption{ Macro performance at user engagement prediction on dev/test.
  Ranking of model performance was consistent across metrics. Precision is low since
  few users tweet a given hashtag. Values are bolded by best test performance according to
  each metric.  Simple reference ranking techniques (bottom): \netpop: a ranking of users
  by the size of their local network; \rand{} randomly ranks users.  The \emph{Dim}
  column is the dimensionality of the selected embedding.}
\label{tab:mv_twitter_users:engagement_results}
\end{table}

\begin{figure}
\centering
\begin{tabular}{cc}
\hspace*{-25pt} \includegraphics[width=0.35\linewidth]{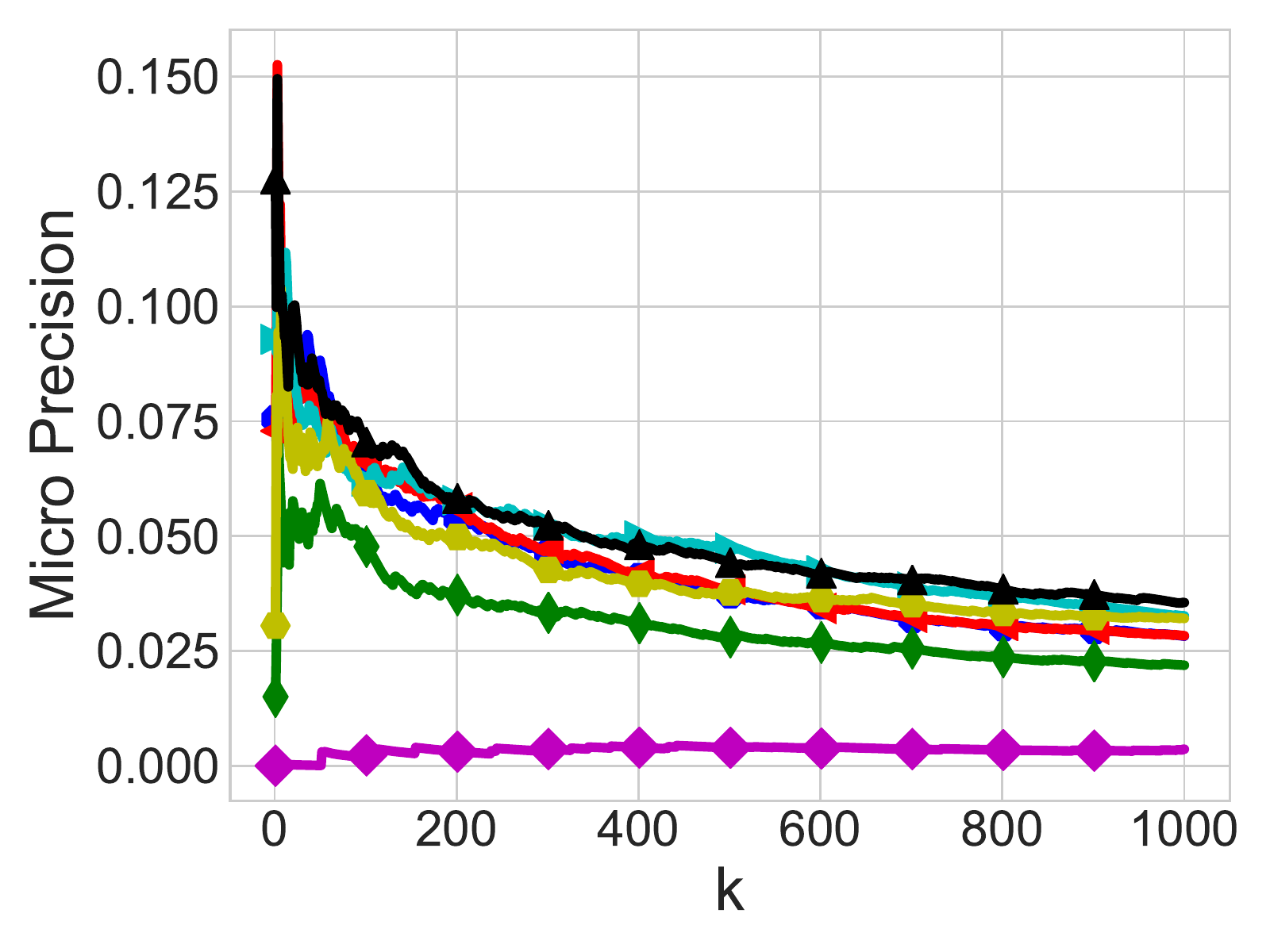} & 
\hspace*{-20pt} \includegraphics[width=0.35\linewidth]{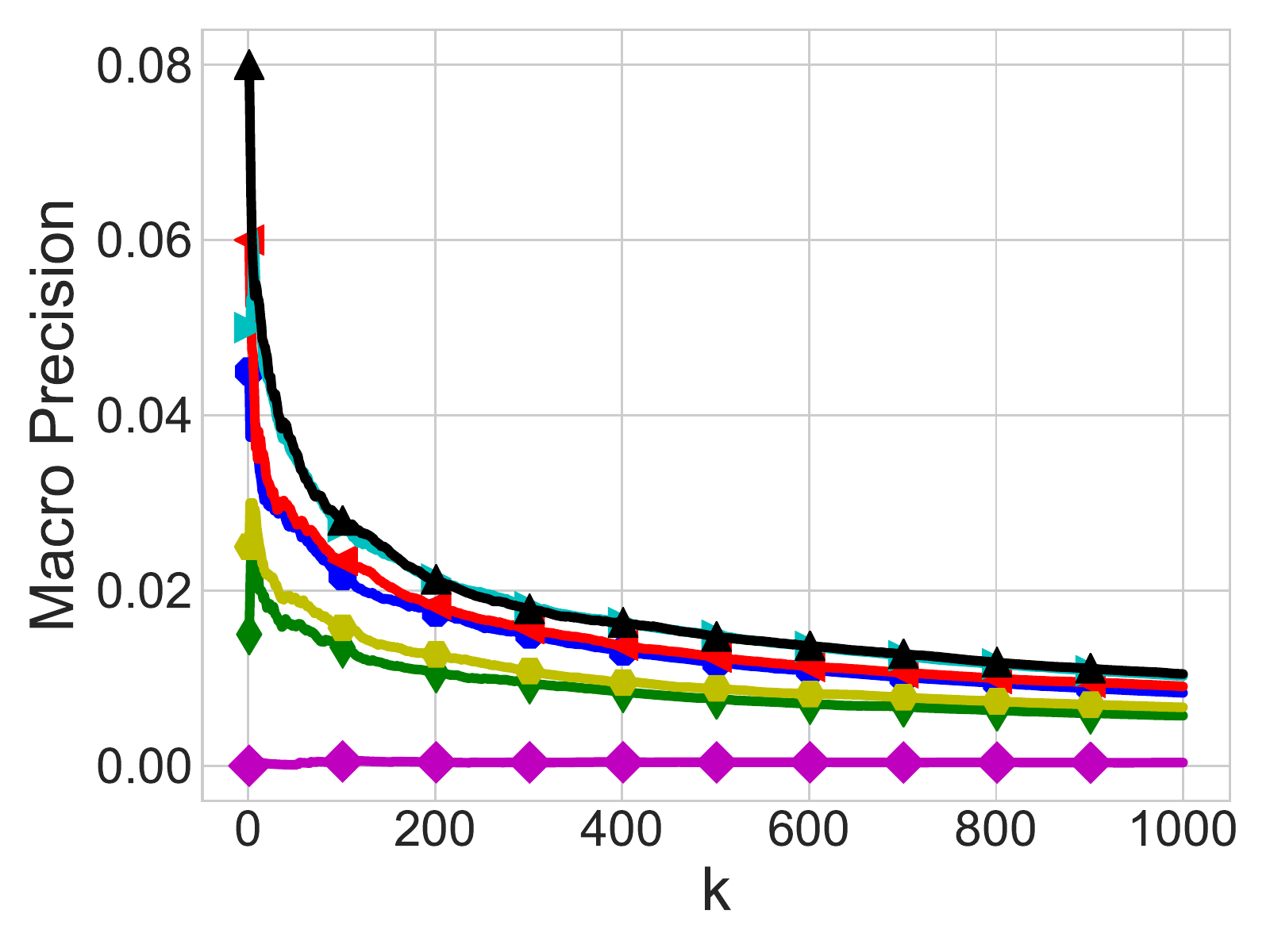} \\ 
\hspace*{-25pt} \includegraphics[width=0.35\linewidth]{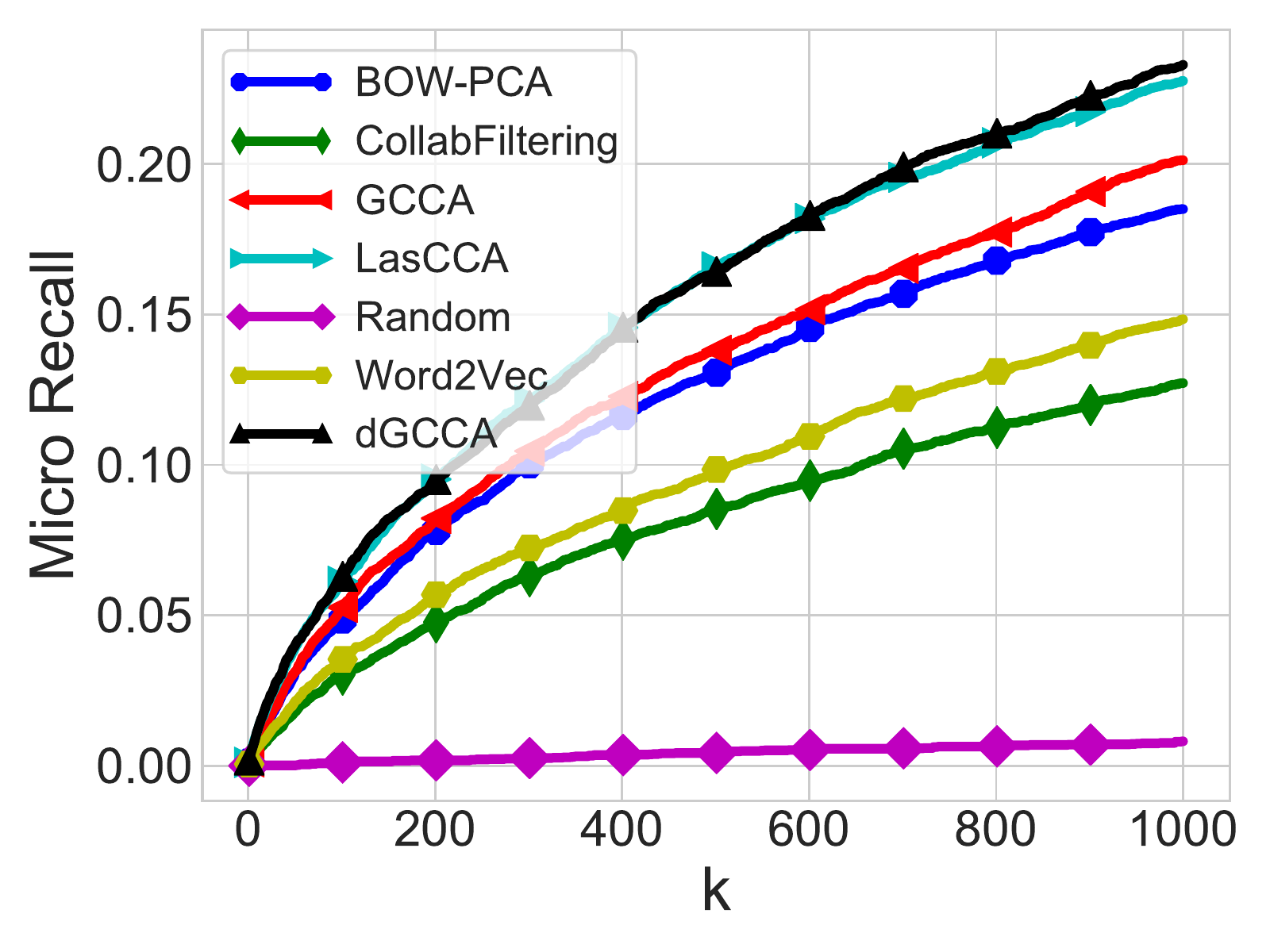} & 
\hspace*{-20pt} \includegraphics[width=0.35\linewidth]{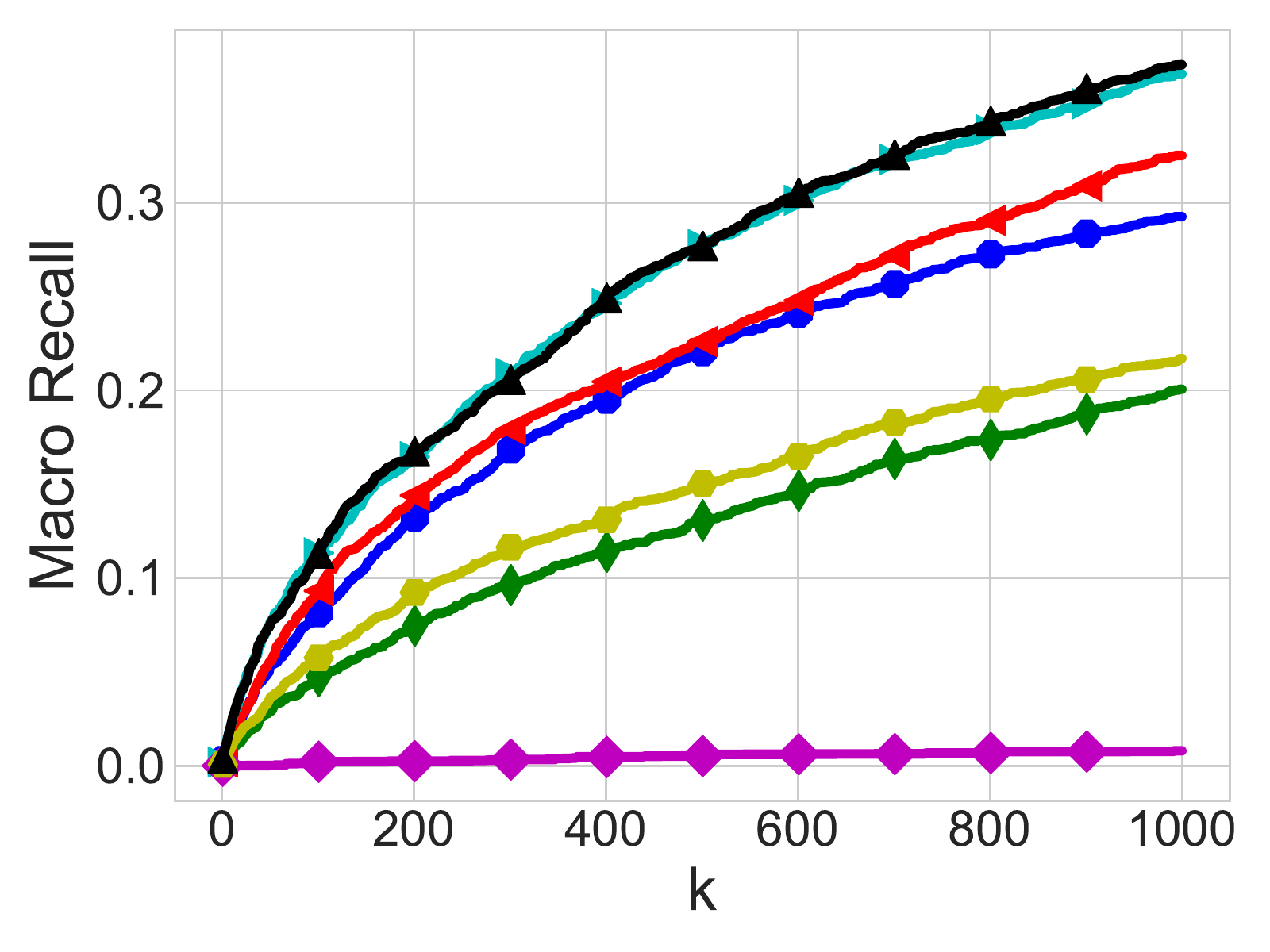}
\end{tabular}
\caption{The best performing approaches on user engagement prediction as a function
  of number of recommendations. The ordering of methods is consistent across $k$.  The plotted \lascca{} is learned over all views (\emph{[all]}).}
\label{fig:mv_twitter_users:engagement_results}
\end{figure}

\begin{figure}
\centering
\includegraphics[width=0.6\linewidth]{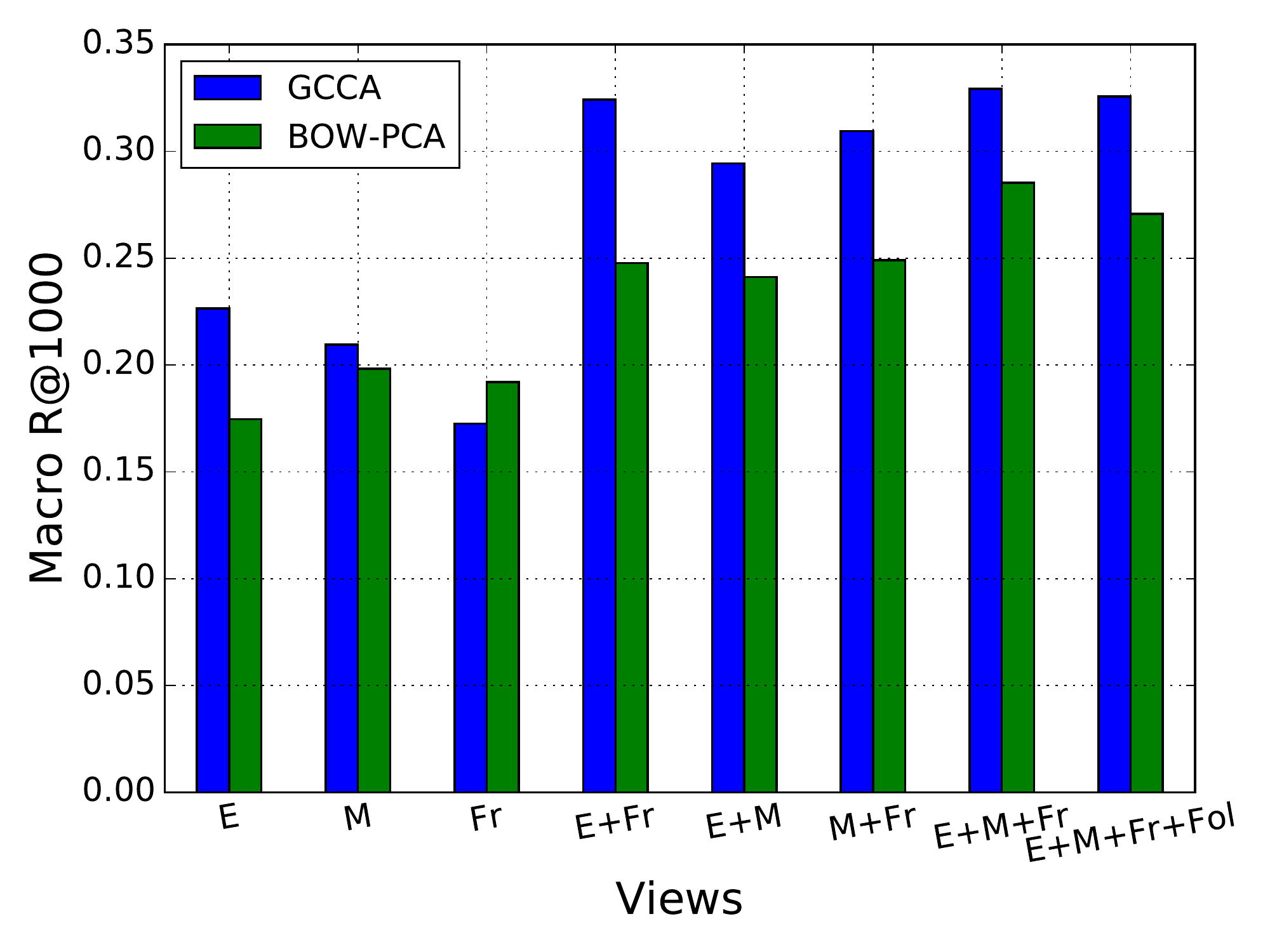}
\caption{Macro recall@1000 on user engagement prediction for different combinations of text views.
Each bar shows the best performing model swept over dimensionality.
\emph{E}: ego, \emph{M}: mention, \emph{Fr}: friend, \emph{Fol}: follower tweet views.}
\label{fig:mv_twitter_users:view_ablation}
\end{figure}

Table \ref{tab:mv_twitter_users:engagement_results} shows results for user engagement
prediction and 
Figure \ref{fig:mv_twitter_users:engagement_results} the precision and recall curves 
as a function of number of recommendations.
The multiview embeddings (\gcca, \dgcca, and \lascca) outperform the other baselines
according to precision and recall at 1000 as well as MRR (for all multiview embeddings except
\gcca{} learned over text views only).
Including network views (\gccawnet{} and \gccasv) improves the performance over just
considering text views.
The best performing \gcca{} setting placed weight 1 on the ego tweet view, mention view,
and friend view, while \bowpca{} concatenated these views, suggesting that these were the three
most important views but that \gcca{} was able to learn a better representation.  Figure
\ref{fig:mv_twitter_users:view_ablation} compares performance of different view
subsets for \gcca{} and \bowpca, showing that \gcca{}
uses information from multiple views more effectively for predicting user engagement.

There are a few other several points to note:
First is that \dgcca{} outperforms linear multiview methods according to recall at 1000
and MRR.  This is exciting because this task benefits from incorporating more
than just two views from Twitter users through linear multiview representation learning
methods.  These results suggest that a nonlinear transformation of the input views can yield
additional gains in performance.  In addition, \gcca{} models sweep over every
possible weighting of views with weights in $\{0, 0.25, 1.0\}$.
\gcca{} has a distinct advantage in that the model is allowed to discriminatively
weight views to maximize downstream performance.  The fact that
\dgcca{} is able to outperform \gcca{} at hashtag recommendation is encouraging, since
\gcca{} has much more freedom to discard uninformative views, whereas the \dgcca{}
objective forces networks to minimize reconstruction error equally across all views.

In addition, the \lascca{} embeddings learned on all views, also unweighted, perform
almost as well as \dgcca{}.  This suggests that linear multiview representation learning
methods may learn similarly effective embeddings as nonlinear ones, only with a slightly
different \gcca{} formulation.  However, it is not clear why the {\sc SUMCOR}
objective would be more appropriate than the {\sc MAXVAR} generalized CCA objective
for learning embeddings geared towards this task.
It also further underscores the fact that multiview techniques seem to be more appropriate
than single-view for learning embeddings effective at user engagement prediction.

\subsubsection*{Effect of \lascca{} Solution Quality on User Engagement Prediction}

\begin{figure}
  \centering
  
    \centering\includegraphics[width=0.31\linewidth,page=1]{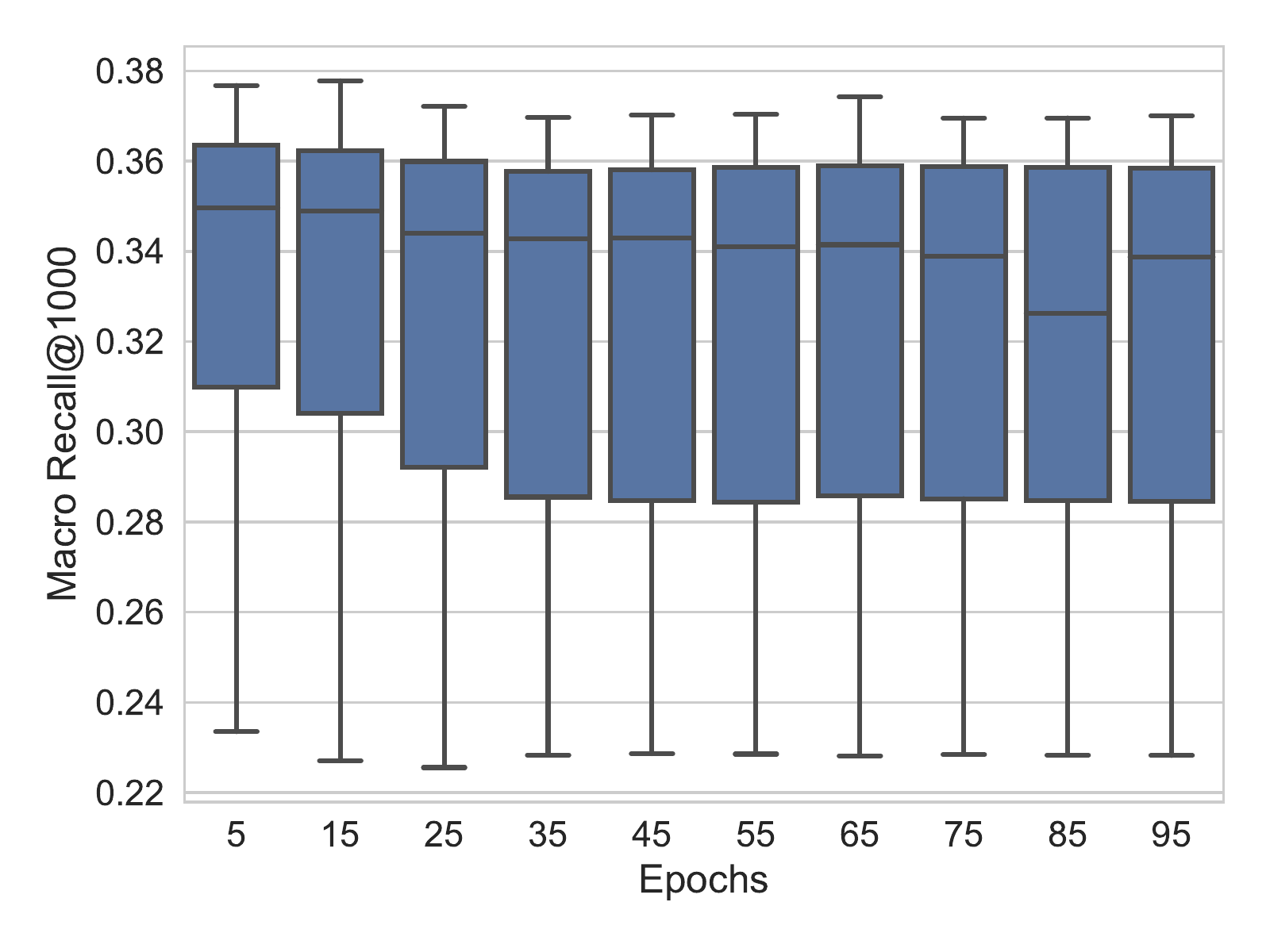}
    \centering\includegraphics[width=0.31\linewidth,page=2]{images/mv_twitter_users/lascca_hashtag_plots.pdf}
    \centering\includegraphics[width=0.31\linewidth,page=3]{images/mv_twitter_users/lascca_hashtag_plots.pdf}
  \caption{Development macro recall at 1000 recommendations for \lascca{} embeddings at the user engagement task.  Boxplots collapse performance across a full sweep of \emph{(left)} number of \lascca{} epochs, \emph{(center)} view subsets, and \emph{(right)} embedding width.}
  \label{fig:mv_twitter_users:lascca_downstream_analysis}
\end{figure}

\removed{
The \lascca{} algorithm relies on a linear least-squares solver to solve for two
intermediate variables\footnote{See lines 3 and 8 in Algorithm
  \ref{alg:mv_twitter_users:h_compute}} and
ultimately the linear maps from observed to shared space, $\hat{U}_i$.
In practice we use a fixed number of conjugate gradient (CG) iterations to solve these
least-squares problems, 20, implemented in the SciPy
library\footnote{\url{https://scipy.org/scipylib/}}.
How well do we need to solve these subproblems to achieve good downstream performance?
To test this, we varied the maximum number of CG iterations per least-squares problem in
$\{1, 5, 20\}$, holding the number of full \lascca{} iterations fixed to 100.  The SciPy
implementation we used stopped early when the norm of the residual $10^{-5}$ as small as
the norm of the target matrix, and the regression coefficients were warm-started with
the previous iteration's solution (making this more likely as optimization continued),
so we did not necessarily require the maximum number of CG iterations each pass.

It is encouraging that downstream performance
is completely insensitive to number of CG iterations (Figure
\ref{fig:mv_twitter_users:lascca_downstream_analysis} (a) and (b)).  Only performing one CG
iteration for least-squares subproblem speeds up learning embeddings by a factor of XX.X
while not affecting user engagement or friend recommendation performance.
In contrast, downstream performance is strongly influenced by which embedding width
we choose (Figures \ref{fig:mv_twitter_users:lascca_downstream_analysis} (c) and (d)).
}

\lascca{} is an iterative algorithm for solving the {\sc SUMCOR}-GCCA problem.
In practice we learned embeddings with a fixed 100 epochs for all embedding widths.
How many times must we iterate to achieve good downstream performance?
To test this, we varied the number of epochs training each \lascca{} embedding by
increments of 5 up to 100.  We then examined the final downstream performance at
user engagement prediction as a function of how many \lascca{} epochs
were taken to learn an embedding as well as other training parameters such as
embedding width and which views to apply \lascca{} too.

It is encouraging that performance at hashtag recommendation
is completely insensitive to number of epochs (Figure
\ref{fig:mv_twitter_users:lascca_downstream_analysis}, left).
In contrast, downstream performance is most influenced by
which embedding width we choose (Figure
\ref{fig:mv_twitter_users:lascca_downstream_analysis}, right), although
we also find that \lascca{} user embeddings learned over network views improve over
just text views (center), echoing what we see when learning \gcca{} embeddings.

\subsection{Friend Recommendation}
\label{subsec:mv_twitter_users:friend_results}

\begin{table}
\tiny
\begin{center}
\begin{tabular}{|l|l|ccc|ccc|c|}
\hline
\bf Model & \bf Dim & \bf P@1 & \bf P@100 & \bf P@1000 & \bf R@1 & \bf R@100 & \bf R@1000 & \bf MRR \\ \hline
\bow & 20000 & 0.164/0.268 & 0.164/0.232 & 0.133/0.153 & 0.000/0.000 & 0.005/0.007 & 0.043/0.048 & 0.000/0.001 \\
\bowpca & 20 & 0.480/0.500 & 0.415/0.421 & 0.311/0.314 & 0.000/0.000 & 0.014/0.014 & 0.101/0.102 & 0.001/0.001 \\
\collab & NA    & 0.672/0.680 & 0.575/0.582 & 0.406/0.420 & 0.000/0.000 & 0.019/0.019 & 0.131/0.132 & \bf 0.002/0.002 \\
\bf \bowpcawnet & 500 & \bf 0.720/0.736 & \bf 0.596/0.601 & \bf 0.445/0.439 & 0.000/0.000 & \bf 0.020/0.020 & \bf 0.149/0.147 & \bf 0.002/0.002 \\
\wordtovec & 200 & 0.436/0.340 & 0.344/0.320 & 0.260/0.249 & 0.000/0.000 & 0.011/0.010 & 0.084/0.080 & 0.001/0.001 \\ \hline
\gcca   & 50 & 0.484/0.472 & 0.370/0.381 & 0.269/0.276 & 0.000/0.000 & 0.012/0.013 & 0.089/0.091 & 0.001/0.001 \\
\bf \gccasv   & 500 & \bf 0.720/0.736 & \bf 0.596/0.601 & \bf 0.445/0.439 & 0.000/0.000 & \bf 0.020/0.020 & \bf 0.149/0.147 & \bf 0.002/0.002 \\
\gccawnet & 20 & 0.520/0.544 & 0.481/0.475 & 0.376/0.364 & 0.000/0.000 & 0.016/0.016 & 0.123/0.120 & 0.001/0.001 \\
\lascca[ego+friendnet] & 50 & 0.352/0.326 & 0.328/0.302 & 0.235/0.239 & \bf 0.000/0.001 & 0.010/0.010 & 0.077/0.078 & 0.001/0.001 \\
\lascca[text] & 50 & 0.352/0.412 & 0.223/0.318 & 0.244/0.250 & 0.000/0.000 & 0.010/0.010 & 0.079/0.080 & 0.001/0.001 \\
\lascca[all] & 50 & 0.420/0.448 & 0.335/0.342 & 0.260/0.263 & 0.000/0.000 & 0.011/0.011 & 0.085/0.086 & 0.001/0.001 \\
\dgcca[all] & 185 & 0.512/0.544 & 0.400/0.411 & 0.297/0.302 & 0.000/0.000 & 0.013/0.014 & 0.099/0.100 & 0.001/0.001 \\ \hline
\netpop & NA    & 0.180/0.176 & 0.025/0.026 & 0.033/0.035 & 0.000/0.000 & 0.001/0.001 & 0.009/0.010 & 0.000/0.000 \\
\rand   & NA    & 0.072/0.136 & 0.040/0.041 & 0.034/0.036 & 0.000/0.000 & 0.001/0.001 & 0.010/0.010 & 0.000/0.000 \\ \hline
\end{tabular}
\end{center}
\caption{ Macro performance for friend recommendation.  Performance of \bowpcawnet{} and \gccasv{} are
identical since the view weighting for \gccasv{} only selected solely the friend view.  Thus, these methods learned identical user embeddings.}
\label{tab:mv_twitter_users:friend_results}
\end{table}

\begin{figure}
\centering
\begin{tabular}{cc}
\hspace*{-25pt} \includegraphics[width=0.35\linewidth]{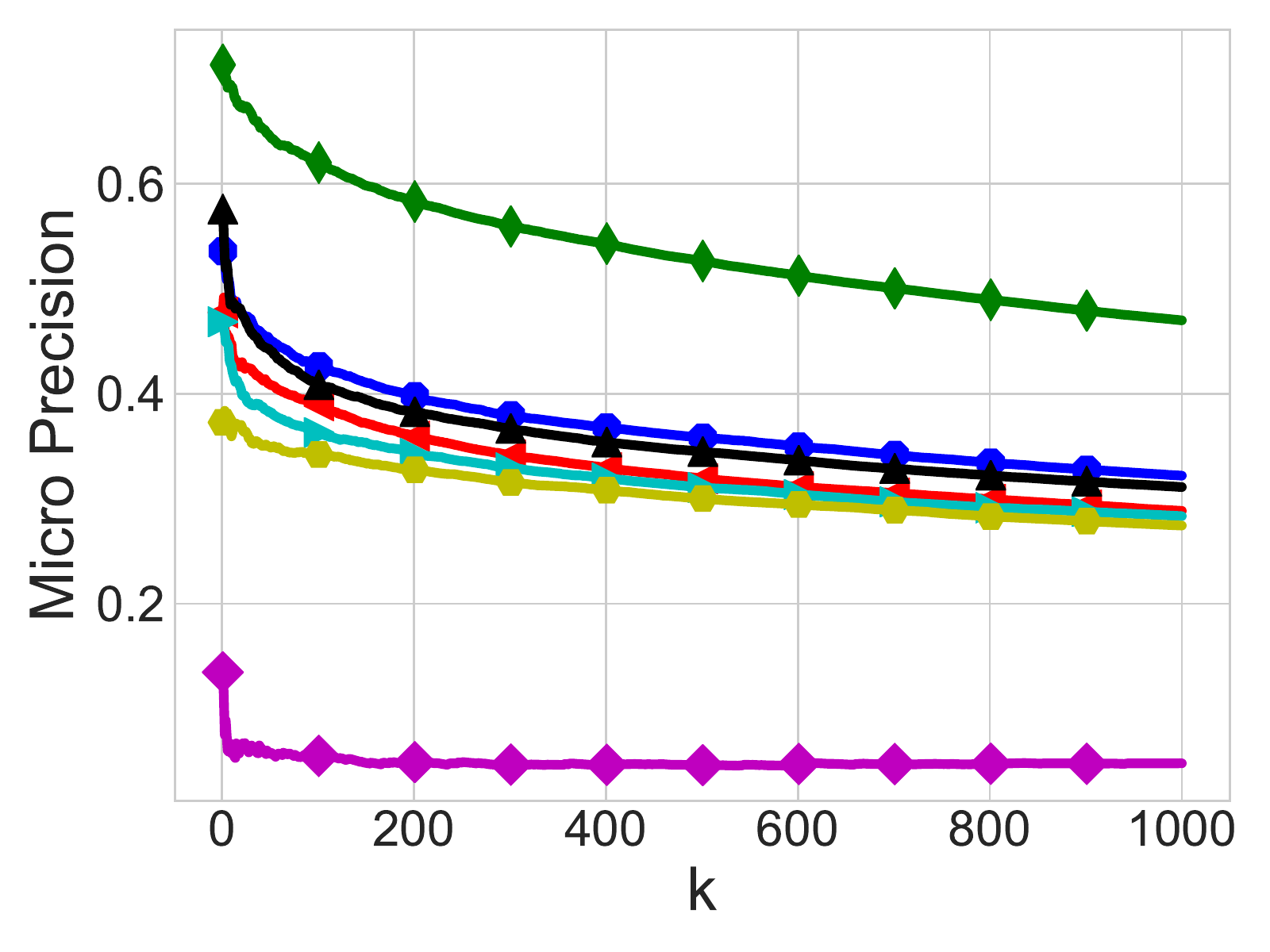} & 
\hspace*{-20pt}\includegraphics[width=0.35\linewidth]{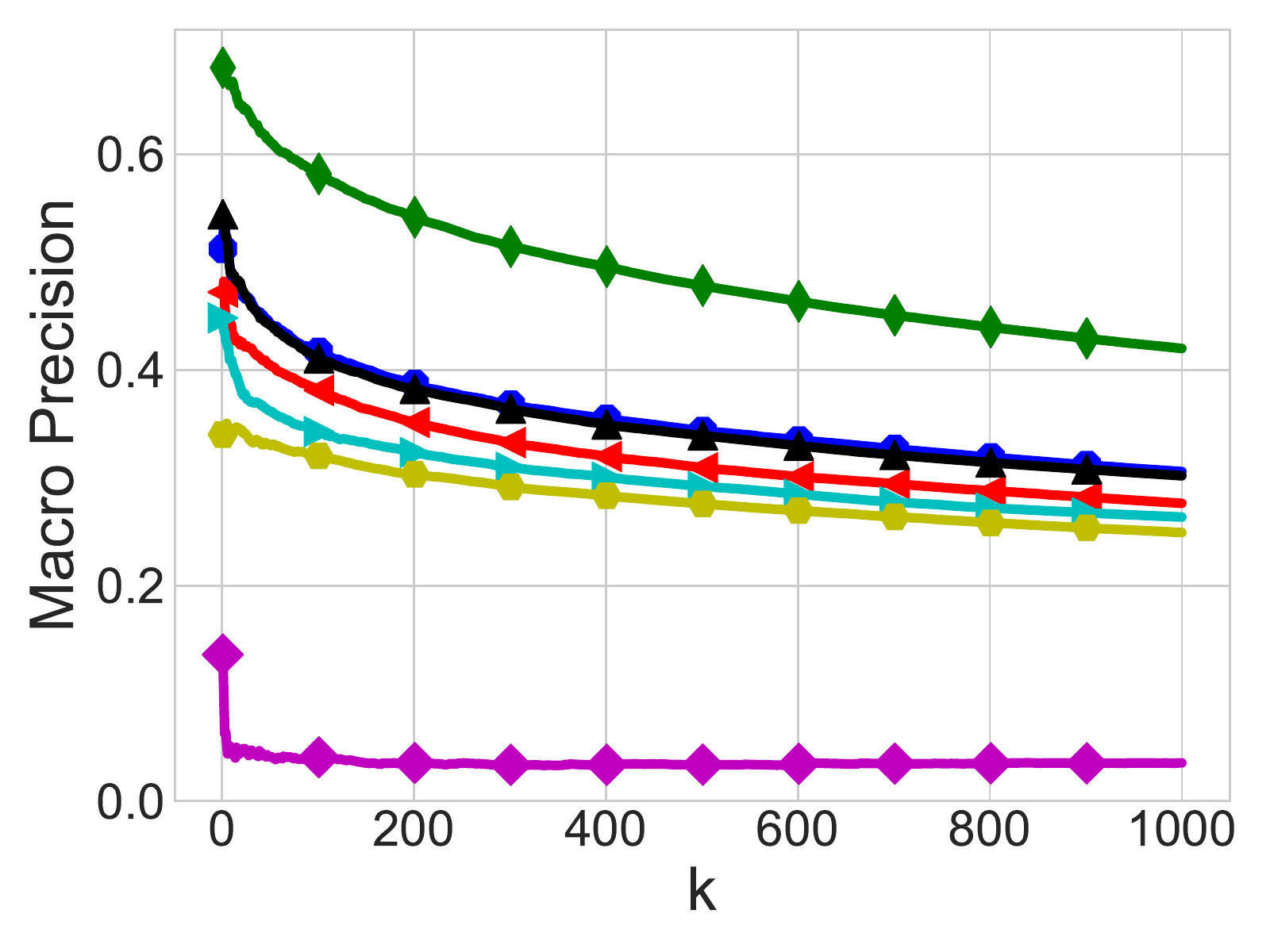} \\ 
\hspace*{-25pt} \includegraphics[width=0.35\linewidth]{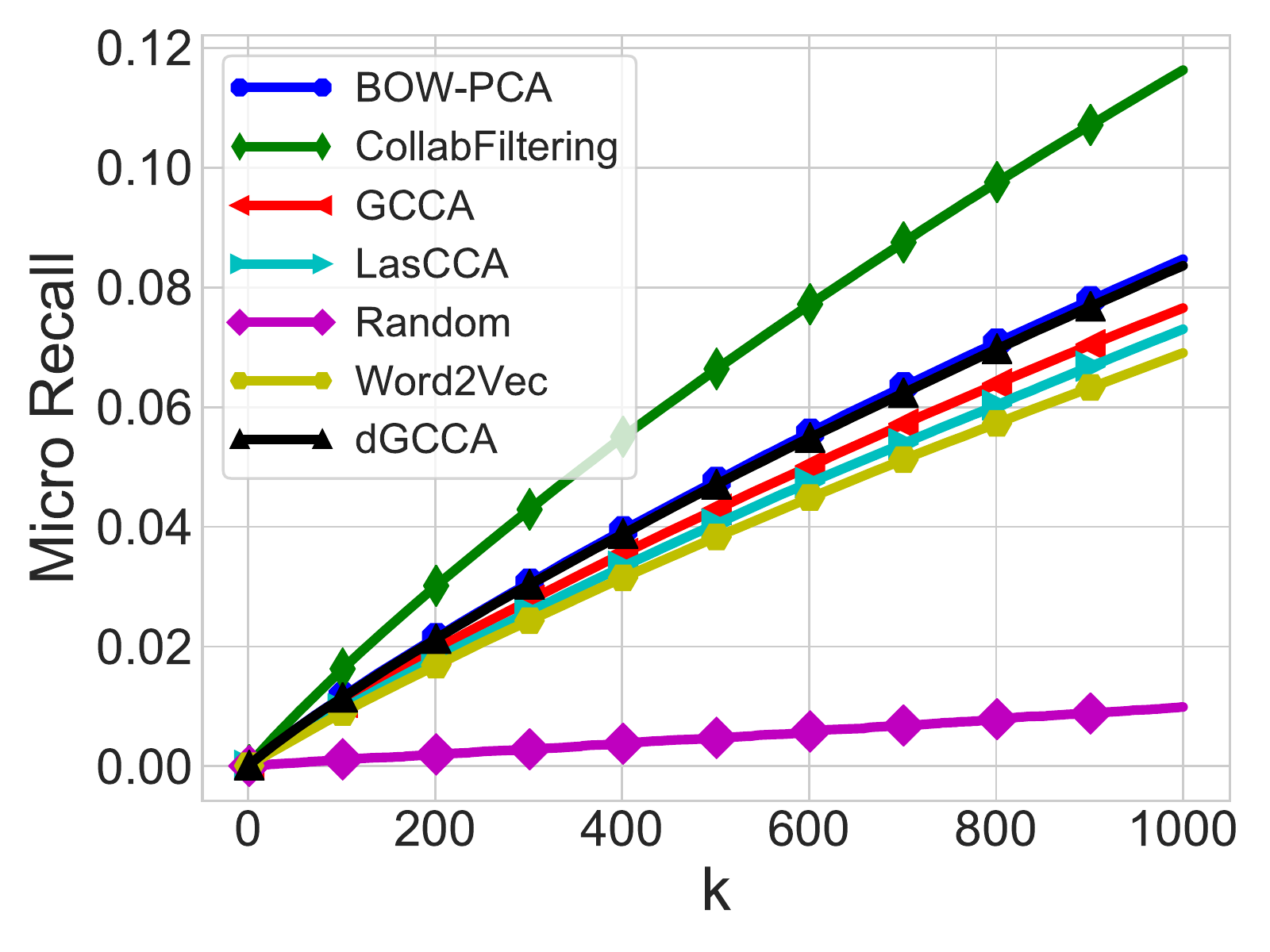} & 
\hspace*{-20pt} \includegraphics[width=0.35\linewidth]{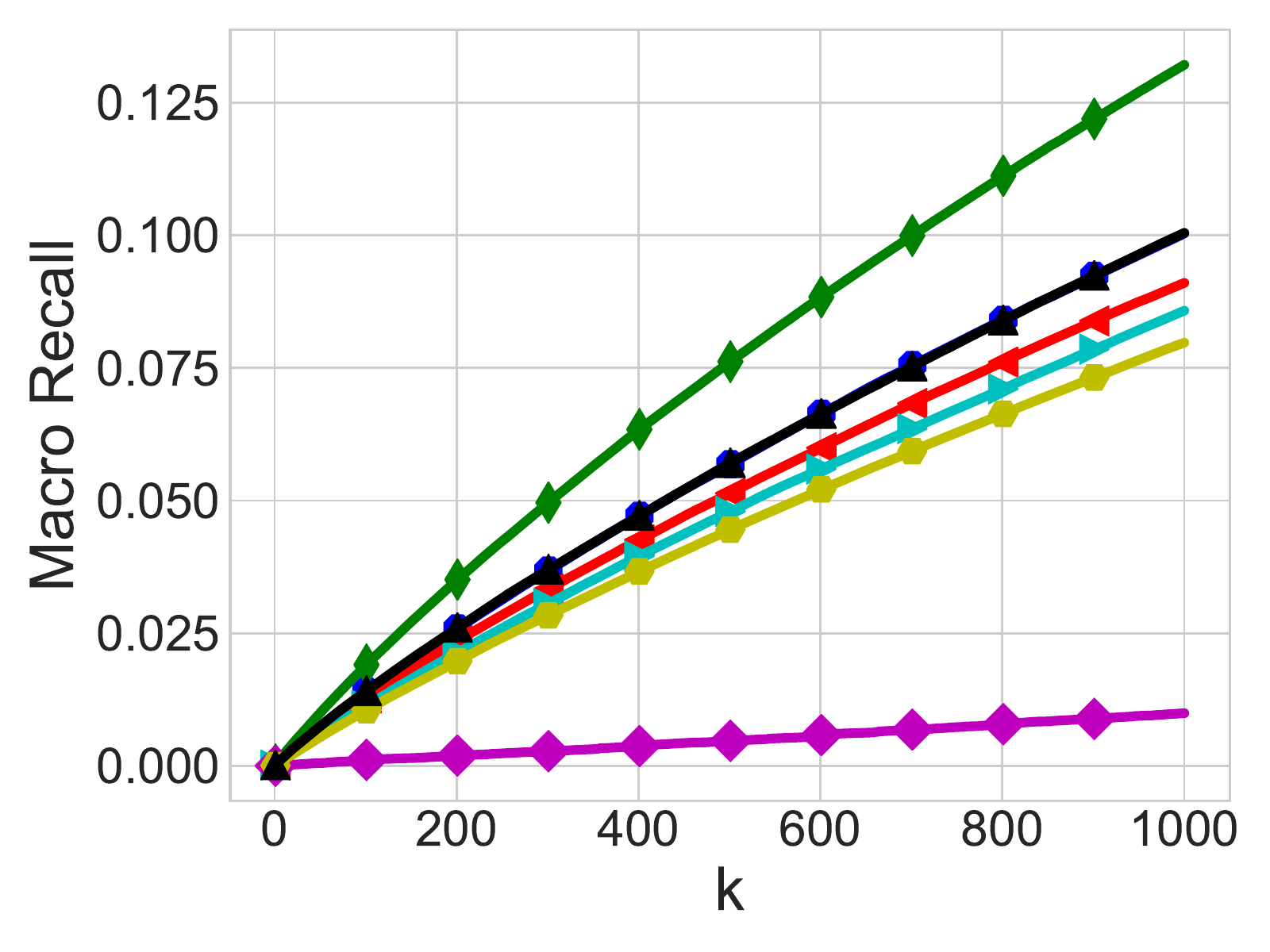}
\end{tabular}
\caption{Performance of user embeddings at friend recommendation as a function of number of recommendations.}
\label{fig:mv_twitter_users:friend_results}
\end{figure}

Table \ref{tab:mv_twitter_users:friend_results} shows results for friend prediction and Figure \ref{fig:mv_twitter_users:friend_results} similarly shows that performance differences
between approaches are consistent across $k$ (number of recommendations.)
Adding network views to \gcca, \gccawnet, improves performance, although it cannot contend with \collab{} or \bowpcawnet, although \gccasv{} is able to meet the performance of \bowpcawnet.
The best \gcca{} placed non-zero weight on the friend tweets view, and \gccawnet{} only
places weight on the friend network view; the other views were not
informative.  \bowpca{} and \wordtovec{} only used the friend tweet view.
This suggests that the friend view is the most important for this task, and multiview techniques cannot exploit additional views to improve performance.
\gccasv{} performs identically to \gccawnet{} since it only placed weight on the friend network view, learning identical embeddings to \gccawnet.

Since only the friend network view was useful for learning
representations for friend recommendation, it is unsurprising that
\dgcca{} when applied to all views cannot compete with \gcca{} representations
learned on the single useful friend network
view\footnote{The performance of WGCCA suffers compared to PCA because whitening the
  friend network data ignores the fact that the spectrum of the decays quickly with a long
  tail -- the first few principal components made up a large portion of the variance in
  the data, but it was also important to compare users based on other components.}.  The
same holds for embeddings learned by \lascca{} over several unweighted views.

\subsection{Demographic Prediction}
\label{subsec:mv_twitter_users:demographic_results}

\begin{table}
\begin{center}
\begin{tabular}{|l|c|c|c|}
\hline
\bf Model & \bf age & \bf gender & \bf politics \\ \hline
\bow  & 0.771/0.740 & 0.723/0.662 & 0.934/0.975 \\
\bowpca & 0.784/0.649 & 0.719/0.662 & 0.908/0.900 \\
\bowpcaandbow & 0.767/0.688 & 0.660/0.714 & \bf 0.937/0.9875 \\
\wordtovec & \bf 0.790/0.753 & \bf 0.777/0.766 & 0.927/0.938 \\ \hline
\gcca   & 0.725/0.740 & 0.742/0.714 & 0.899/0.8125 \\
\gccaandbow & 0.764/0.727 & 0.657/0.701 & 0.940/0.9625 \\
\gccasv     & 0.709/0.636 & 0.699/0.714 & 0.871/0.850 \\
\gccasvandbow & 0.761/0.688 & 0.647/0.675 & 0.937/0.9625 \\
\lascca[text] & 0.689/0.662 & 0.712/0.662 & 0.883/0.838 \\ 
\lascca[text] + \emph{BOW} & 0.754/0.662 & 0.666/0.649 & 0.931/ 0.950 \\
\dgcca & 0.735/0.727 & 0.699/0.649 & 0.845/0.800 \\ 
\dgccaandbow & 0.771/0.649 & 0.673/0.610 & 0.931/0.950\\ \hline
\end{tabular}
\end{center}
\caption{ \label{tab:mv_twitter_users:demographic_results} Average CV/test accuracy for
  inferring demographic characteristics given different feature sets.}
\end{table}

Table \ref{tab:mv_twitter_users:demographic_results} shows the average cross-fold
validation and test accuracy on the demographic prediction task.  \emph{+ BOW} indicates that
\bow{} features were concatenated to the embeddings as an additional feature set for the
classifier.  The wide variation in performance is due to the small size of the datasets,
thus it's hard to draw many conclusions (the average development performance of all models
are within one standard deviation of each other).
However, \wordtovec{} surpasses other representations in two out of three datasets, and
including a TF-IDF weighted bag of words features tends to improve the generalization
performance of most classifiers.

It is difficult to compare the performance of the methods we evaluate here to that
reported in previous work \parencite{zamal2012homophily}.  This is because they report cross-fold
validation accuracy (not test), they consider a wider range of hand-engineered features,
different subsets of networks, radial basis function kernels for SVM, and find
that accuracy varies wildly across different feature sets.  They report cross-fold
validation accuracy ranging from 0.619 to 0.805 for
predicting age, 0.560 to 0.802 for gender, and 0.725 to 0.932 for politics.

\subsection{Evaluating User Cluster Coherence}
\label{subsec:mv_twitter_users:qual_analysis}

Although we evaluated embeddings on several quantitative tasks,
these experiments do not tell us which embedding type best captures intuitive notions
of user groups, similar to how word embeddings have been shown to cluster words with
similar meaning or syntactic properties together in embedding space.
In order to evaluate how well different embeddings captured human notions of
types of people or user groups, we performed the following experiment.

We considered three different types of 500-dimensional user embeddings: \bowpca{} only
on ego text (\bowpca\emph{[ego]}), \bowpca{} on the concatenation of all views
(\bowpca\emph{[all]}), and \dgcca{} on all
views\footnote{We consider \dgcca{} to represent multiview embeddings in general, because
  it was the best performing multiview method at user engagement prediction.}.
For each embedding type, we fit a 50-cluster Gaussian mixture model with diagonal covariance
matrix, for 100 expectation maximization iterations.  For all users in our data, we assigned
them to the most probable Gaussian according to the mixture model, and we
selected the five most probable users assigned to that cluster under the
Gaussian distribution.  We ensured that
each of these accounts were still active by querying their user
summary page (\texttt{https://twitter.com/intent/user?user\_id=\$\{USER\_ID\}}) and excluded
the cluster from further analysis if we could not find five active users within the closest
10 users to the Gaussian centroid.  This yielded a total of 144 clusters: 50 clusters
for \bowpca\emph{[ego]} and \bowpca\emph{[all]} each, and 44 clusters for \dgcca.

\subsubsection{Experiment}

We used these cluster exemplars to construct an intruder detection task to submit to
Amazon Mechanical
Turk\footnote{This experiment was submitted in July 2018, over three years after the data
  used to learn embeddings were collected.}.  For each cluster,
we presented the subject with links to four of the five exemplar
users' Twitter summaries\footnote{\url{https://twitter.com/intent/user?user_id=[USER_ID]}},
along with an intruder user sampled uniformly at random from another cluster's exemplars.
The order of users was randomized for each HIT and the subject was asked to complete two tasks:

\begin{enumerate}
  \item Given only the information provided on the users' summary
    pages (their most recent tweets, user text description, and profile image),
    identify which user is the most different from the other four.
  \item Describe in your own words, and as specifically as possible, what the
    other four users have in common. 
\end{enumerate}

\begin{figure}
  \centering
  \fbox{\includegraphics[width=1.0\linewidth]{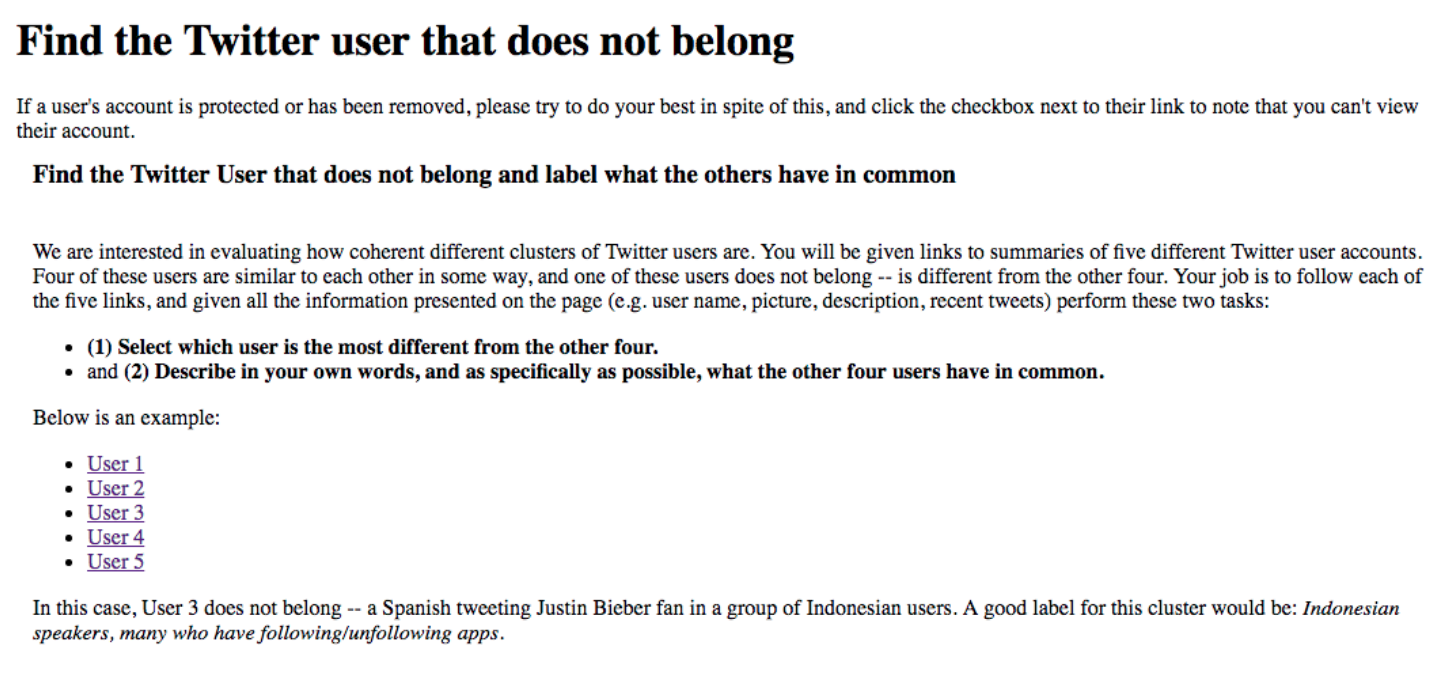}}
  \caption{Mechanical Turk instructions for the user cluster intruder detection task.}
  \label{fig:mv_twitter_users:mturk_cluster_hit_instructions}
\end{figure}

\begin{figure}
  \centering
  \fbox{\includegraphics[width=1.0\linewidth]{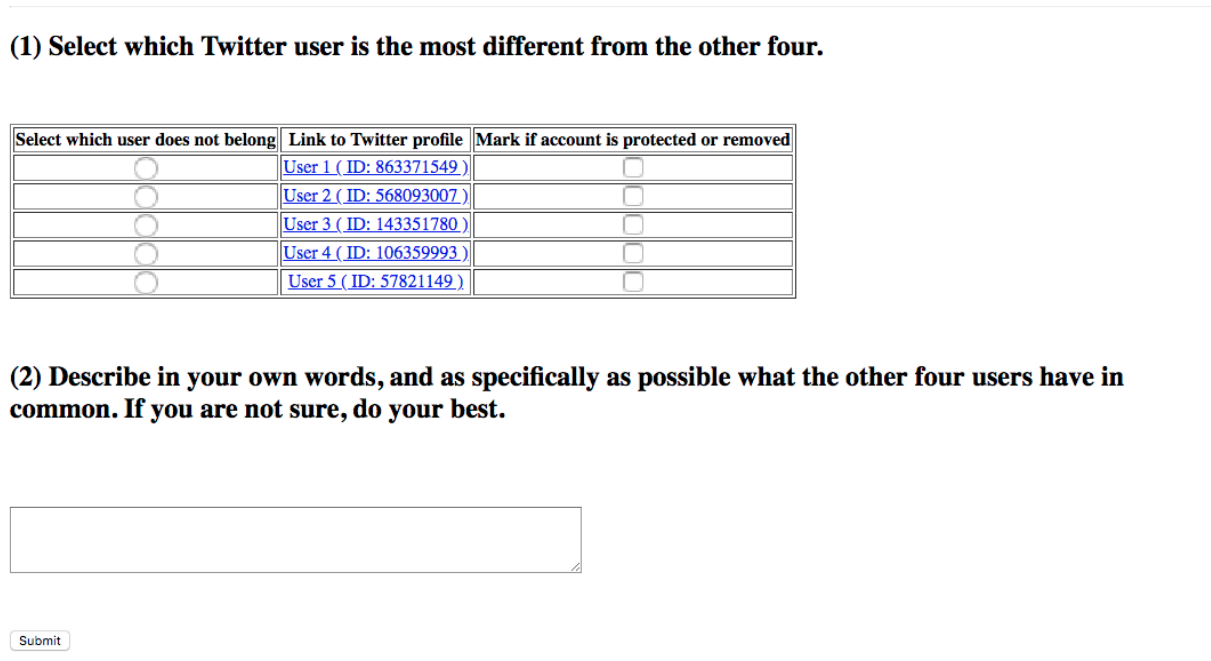}}
  \caption{Example assignment for the user cluster intruder detection HIT.
           The user ID links point the subject to a Twitter user's summary page.
           }
  \label{fig:mv_twitter_users:mturk_cluster_hit_prompt}
\end{figure}

Screenshots of the Mechanical Turk instructions and a sample HIT are presented in
Figures \ref{fig:mv_twitter_users:mturk_cluster_hit_instructions} and
\ref{fig:mv_twitter_users:mturk_cluster_hit_prompt}.

We treat the intruder detection task as a proxy for user cluster coherence -- how similar
are users belonging to the same cluster.
If it is easier for a subject to spot which user does not belong, that suggests that the
other users share an easily identifiable, common property.  This task was inspired by work
in evaluating the quality of topics learned by a topic model, specifically the
\emph{word intrusion} task described in \textcite{ChangEtAl09}.  In addition, we were
able to use this task to quickly collect cluster labels for qualitative analysis,
without influence from our own biases.

Each cluster was labelled by three unique subjects and we compare embedding types by
accuracy at the intruder detection task, averaged over all annotations.

\subsubsection{Results}

\begin{figure}
  \centering
  \includegraphics[width=0.6\linewidth,page=1]{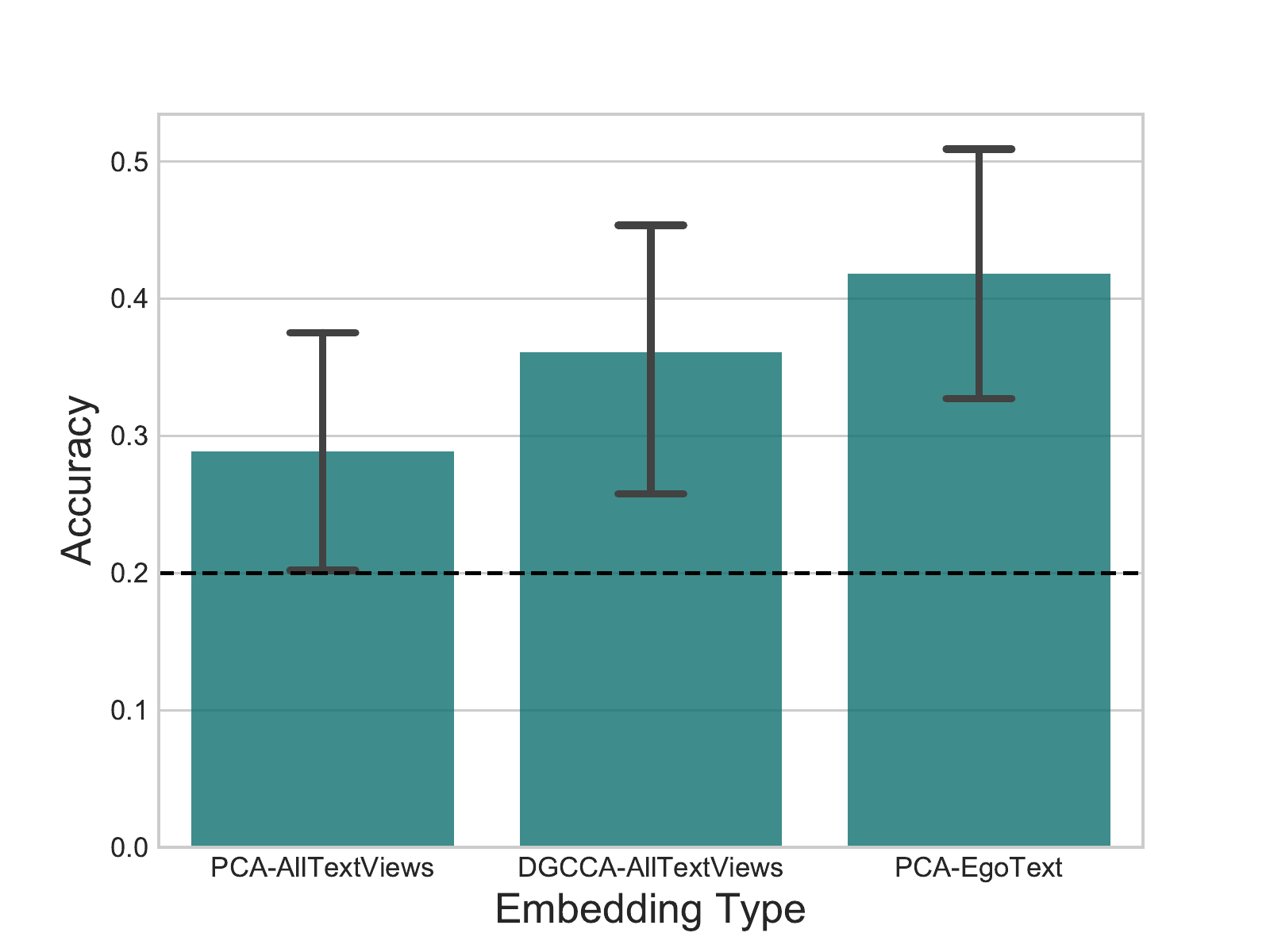}
  \caption{Average Turker accuracy at selecting the intruder out of a cluster of five
    users.  The horizontal line marks performance of random guessing (20\%).  95\%
    confidence interval bars are generated by 10,000 bootstrap samples with
    resampling of the same size as the original sample.  Each bar corresponds to
    a different type of embedding from which clusters were induced.}
  \label{fig:mv_twitter_users:mturk_cluster_intruder_accuracy}
\end{figure}

We omitted a single subject's HITs from analysis as they completed a large number of HITs
very quickly, performed only slightly above random chance (23\% accuracy), and labelled
clusters uninformatively (e.g. ``They are all the same'' or ``'posts are in English'').
After removing this user, we calculated accuracy over a total of
311 annotations (110 for \bowpca\emph{[ego]}, 104 for \bowpca\emph{[all]}, and 97 for \dgcca). 
Surprisingly, subjects found the clusters from \bowpca\emph{[ego]} tended to be the most
coherent (Figure \ref{fig:mv_twitter_users:mturk_cluster_intruder_accuracy}).  

Although the confidence intervals estimated by bootstrap samples are wide, subjects were
able to detect the intruder statistically significantly more frequently than chance for all
embedding types according to a proportion z-test
($p=0.05$)\footnote{From the \texttt{statsmodels} python library:

\texttt{statsmodels.stats.proportion.proportions\_ztest}}.  The \bowpca\emph{[ego]}
embeddings resulted in statistically significantly higher accuracy than PCA on all views
according to this same test ($p=0.05$).

One reason why subjects better detected intruders in the \bowpca\emph{[ego]} clusters was
likely because of the information they were allowed to act on: a short summary of the Twitter
user.  Grouping
users together by the frequent words they post is a simple cue for someone to latch onto.
These are exactly the sorts of features that methods that only consider the ego text view
will try to preserve.  However, we
fidn it interesting that \dgcca{} clusters are more coherent than \bowpca{[all]}.  Although
less coherent than an ego text embedding, this suggests that multiview
representation learning methods yield more ``natural'' user embeddings when
consolidating multiple types of input behavior.

Appendix \ref{app:user_clusters_mturk} contains an exhaustive list of labels
assigned to each cluster along with a few examples of Twitter users belonging
to the same cluster.  Many of these clusters were assigned vague labels
(``They all speak English'', ``none''), which speaks to the difficulty of this
task.  Not only are the user clusters noisy and subjects are given
scant information in the Twitter summary, but the user embeddings were learned
\emph{over three years before} the HIT was conducted. 

\paragraph*{Preprocessing Considerations}

In this chapter we na\"{i}vely preprocessed the text views by removing stop
words and restricting the vocabulary size to the 20,000 most frequent token
types.  Because of this, the user representations we learn in this chapter
sometimes captured user behavior that would be considered noise in most
downstream tasks.

\begin{figure}
\begin{center}
\includegraphics[width=0.6\textwidth]{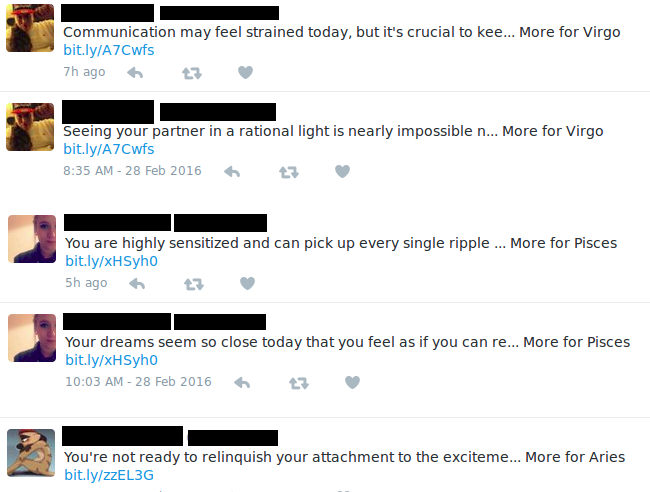}
\end{center}
\caption{Tweets from exemplar users from an ``astrology app'' cluster.  Members of
  this cluster belonged to a range of astrological signs and the only discernible
  feature shared between them were automated posts generated by the app.
  We intentionally obfuscated the users' names for their privacy.}
\label{fig:mv_twitter_users:astrology_cluster}
\end{figure}

\removed{
Figure \ref{fig:mv_twitter_users:astrology_cluster} shows exemplar tweets
from one such cluster learned from \gcca{} embeddings.
}


The most salient examples of this were clusters of users who registered for
the same Twitter app.  Astrological sign apps are particularly popular, and some
automatically post tweets associated with the user's astrological sign.
Figure \ref{fig:mv_twitter_users:astrology_cluster} shows exemplar tweets
from one such cluster learned from \gcca{} embeddings.  This
cluster mixes users with different signs suggesting that user representations
generalize to those who subscribed to this particular astrology app, rather than
homing in on repetition of tokens for one astrological sign.
Another cluster included users who subscribed to follower-tracking apps
that automatically tweet about changes in their follower network.
Although we focus on evaluating different \emph{methods} of
learning user representations, this underscores just how important
data quality and preprocessing are when applying these methods to
real-world data.


\section{Summary}

This chapter shows how unsupervised user embeddings can be learned from multiple
views of Twitter user behavior.   We find that although embeddings learned on
friending behavior alone are the most predictive of other friends a user may have,
multiview embeddings learned over views of both what the ego user posts and their
friending behavior better capture which hashtags they are likely to use in
the future.  Although subjects found embeddings
learned only on ego text to yield more coherent user clusters than multiview
embeddings, multiview user embedding clusters were more coherent than those learned
by applying a single-view dimensionality reduction technique to all views.

\cleardoublepage

\chapter{User-Conditioned Topic Models}
\label{chap:user_conditioned_topicmodels}

Chapter \ref{chap:mv_twitter_users} described different unsupervised
methods for learning social media user embeddings, and primarily evaluated
these embeddings intrinsically -- by how well they capture similar topic posting or
friending behavior.  In this chapter we present a non-traditional application of
user features, showing how they can be used to improve topic modeling of
social media text.

This chapter highlights the breadth of applications that can benefit from
user features.  Latent Dirichlet Allocation (\lda) is the most common topic model
applied by social scientists to uncover themes in large corpora.  We show that
opting to fit a supervised topic model with user features as supervision
can lead to improved model fit and better guide the topics that are learned.
We also present a new topic model,
deep Dirichlet Multinomial Regression (\dsprite), that can better make use of
highy-dimensional and only distantly related features, improving upon a previous
supervised topic model, Dirichlet Multinomial Regression (\dmr).

Section \ref{sec:user_conditioned_topicmodels:supervised_topic_models} gives
background on upstream supervised topic models: their basic generative story,
how one fits these models by collapsing Gibbs sampling, and how they compare to unsupservised topic models.
Section \ref{sec:user_conditioned_topicmodels:ddmr} describes \dsprite{} and
analyzes which types of corpora it excels at modeling by synthetic data experiments.
Section \ref{sec:user_conditioned_topicmodels:ddmr_evaluation}
evaluates \dsprite{} against other unsupervised and supervised models on
three datasets: a collection of New York Times articles, Amazon product reviews, and
Reddit messages.  Section \ref{sec:user_conditioned_topicmodels:twitter_data} finally applies \dsprite{} to modeling three public policy-related Twitter datasets using features derived
from inferred user location as supervision.
The Twitter distant user feature supervision experiments were published
as \textcite{benton2016collective}, a long paper in AAAI 2016, and also appear in
chapter 6 of Michael J. Paul's Ph.D. thesis \parencite{paul2015dissertation}.
The \dsprite{} model definition and evaluations were published as a long paper in
NAACL 2018 \parencite{benton2018deep}.

\section{Background: Supervised Topic Models}
\label{sec:user_conditioned_topicmodels:supervised_topic_models}

Social media has proved invaluable for research in social and health sciences,
including sociolinguistics \parencite{Eisenstein:2011:DSA:2002472.2002641},
political science \parencite{OConnorEtAl10}, and public health \parencite{paul2011you}.
A common theme is the use of topic models \parencite{BleiEtAl03},
which, by identifying major themes in a corpus, 
summarize the content of large text collections. 
Topic models have been applied to characterize tweets \parencite{ramage:characterizing},
blog posts and comments \parencite{YanoEtAl09,PaulGirju09a},
and other short texts \parencite{PhanEtAl08}.


%
Latent Dirichlet Allocation (\lda) is a fully unsupervised generative model, which may
have limited utility when trying to learn topics that capture the opinions of document authors.
Supervised topic models offer one option for guiding topics and improving model fit.  Supervised topic models come in many flavors, such as predicting labels for each document, e.g., supervised LDA \parencite{mcauliffe2008supervised}; 
modeling tags associated with each document, e.g., labeled LDA \parencite{Ramage:2009} or tagLDA \parencite{Zhu:2006nr}; 
placing priors over topic-word distributions \parencite{Jagarlamudi:2012lr,paul2013drug}; or interactive feedback from the user \parencite{hu2014}.
Using the terminology of \textcite{DMR},
these models can be classified as either ``Upstream'' or ``Downstream'',
referring to whether this supervision is assumed to be generated before or after
the text in the generative stories. The supervised models we consider in this
chapter are upstream models with document-level supervision, in particular
Dirichlet Multinomial Regression (\dmr).


\subsection{\dmr{} Generative Story}
\label{subsec:user_conditioned_topicmodels:supmodels_generative_story}

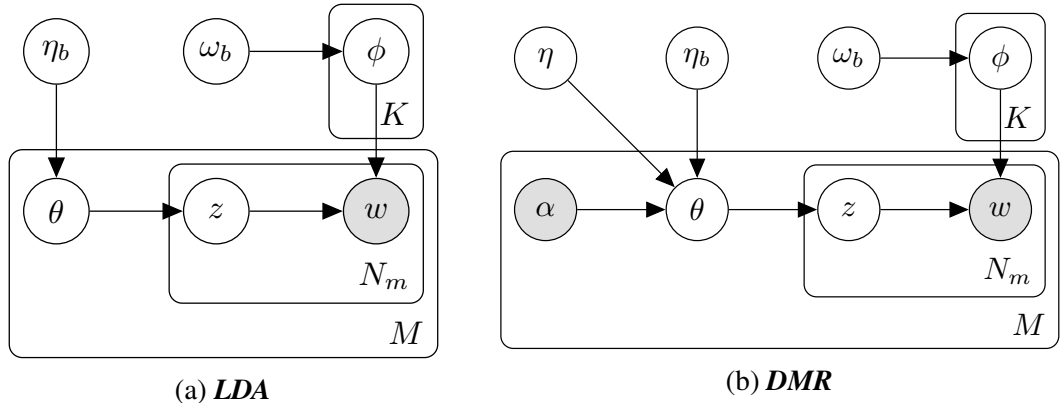
\begin{figure}
\centering
\begin{subfigure}{0.42\textwidth}
\resizebox{1.0\textwidth}{!}{
\begin{tikzpicture}[node distance=0.5cm]
  
  \node[obs]                    (w)        {$w$} ; %
  \node[latent, left=of w]      (z)        {$z$} ; %
  \node[latent, left=of z]      (theta)    {$\theta$}; %
  \node[latent, above=of theta] (bk)       {$\eta_{b}$} ; %
  
  
  \node[latent, above=of w]   (phi)   {$\phi$} ; %
  \node[latent, left=of phi]   (bv)    {$\omega_{b}$} ; %
  
  
  \edge {bk} {theta} ;
  \edge {bv} {phi} ;
  \edge {theta} {z} ;
  \edge {phi,z} {w} ;
  
  \plate {plateWord} { %
    (w)(z) %
  } {$N_m$}; %
  \plate {plateDoc} { %
    (plateWord) %
    (theta) %
  } {$M$}; %
  \plate {plateTopic} { %
    (phi) %
  } {$K$}; %
\end{tikzpicture}
}
\caption{\textbf{\lda}}
\label{fig:user_conditioned_topicmodels:lda_plate}
\end{subfigure}
\hfill
 \begin{subfigure}{0.54\textwidth}
\resizebox{1.0\textwidth}{!}{
\begin{tikzpicture}[node distance=0.5cm]
  
  \node[obs]                    (w)        {$w$} ; %
  \node[latent, left=of w]      (z)        {$z$} ; %
  \node[latent, left=of z]      (theta)    {$\theta$}; %
  \node[obs, left=of theta]  (alpha)    {$\alpha$} ; %
  \node[latent, above=of theta] (bk)       {$\eta_b$} ; %
  
  \node[latent, above=of alpha] (eta) {$\eta$} ; %
  
  \node[latent, above=of w]   (phi)   {$\phi$} ; %
  \node[latent, left=of phi]   (bv)    {$\omega_b$} ; %
  
  
  \edge {bk,alpha,eta} {theta} ;
  \edge {bv} {phi} ;
  \edge {theta} {z} ;
  \edge {phi,z} {w} ;
  
  \plate {plateWord} { %
    (w)(z) %
  } {$N_m$}; %
  \plate {plateDoc} { %
    (plateWord) %
    (theta)(alpha) %
  } {$M$}; %
  \plate {plateTopic} { %
    (phi) %
  } {$K$}; %
\end{tikzpicture}
}
\caption{\textbf{\dmr}}
\label{fig:user_conditioned_topicmodels:upstream_plate}
\end{subfigure}

\vspace{-2ex}
\caption{\label{fig:user_conditioned_topicmodels:plate}
  Graphical model of \lda{} (left) and \dmr{} (right) in plate notation.
  The key difference between these topic models is that \dmr{} includes
  document-dependent features, $\alpha$, that affect the document-topic
  prior through log-linear weights, $\eta$, shared across all documents.
  \lda{} conversely shares the same document-topic prior for all documents.
  }
\end{figure}

\ref{fig:user_conditioned_topicmodels:plate} illustrates the generative story
of \lda{} and \dmr{} in plate notation.
In upstream topic models, supervision influences the {\em priors} over topic
distributions in documents.
In \dmr{} this is done by parameterizing the document-topic Dirichlet prior as
a log-linear function of the document labels $\bm{\alpha}$ and regression
coefficients $\bm{\eta}$.
Under a \dmr{} topic model (the basic upstream model in our experiments),
each document has its own $\textrm{Dirichlet}(\tilde{\theta}_m)$ prior, with
$\tilde{\theta}_{mk}$$ =$$ \exp(\eta_{bk} + \alpha_m^{T} \eta_{k})$,
where $\alpha_m$ is the supervision feature vector of the $m^{\text{th}}$ document,
$\eta_k$ is the $k^{\text{th}}$ topic's feature coefficients,
and $\eta_{bk}$ is a bias term for topic $k$ (intercept).
For positive $\eta_k^{(i)}$, the prior for topic $k$ in document $m$ will
increase as $\alpha_m^{(i)}$ increases, while negative $\eta_k^{(i)}$ will
decrease the prior weight.

Figure \ref{fig:upstream_generative_story} provides the 
generative story for \dmr, with the portions that differ from \lda{} in red.

\begin{figure}
\small
\begin{enumerate}[itemsep=0pt, topsep=0pt, partopsep=0pt]
\item For each document $m$:
	\begin{enumerate}[itemsep=0pt, topsep=0pt, partopsep=0pt]
  \item $\tilde{\theta}_{mk} \leftarrow \exp(\eta_{bk})$, for each topic $k$
 	\item \color{red}{ $\tilde{\theta}_{mk} \leftarrow \tilde{\theta}_{mk} * \exp(\alpha_m^{T} \eta_{k})$, for each topic $k$}
	\item $\theta_{m} \sim \textrm{Dirichlet}(\tilde{\theta}_m)$
	\end{enumerate}
\item For each topic $k$:
	\begin{enumerate}[itemsep=0pt, topsep=0pt, partopsep=0pt]
	\item $\tilde{\phi}_{kv} = \exp(\omega_{bv})$
	\item $\phi_{k} \sim \textrm{Dirichlet}(\tilde{\phi}_k)$
	\end{enumerate}
\item For each token $n$ in each document $m$:
	\begin{enumerate}[itemsep=0pt, topsep=0pt, partopsep=0pt]
	\item Sample topic index $z_{mn} \sim \theta_m$
	\item Sample word token $w_{mn} \sim \phi_{z_{mn}}$
	\end{enumerate}
\end{enumerate}
\caption{\label{fig:upstream_generative_story}Generative story for \dmr.
  Differences between \lda{} and \dmr{} are written in red.}
\label{fig:user_conditioned_topicmodels:generative_story}
\end{figure}

\subsection{Fitting Topic Models}
\label{subsec:user_conditioned_topicmodels:sup_models_fitting_models}

The experiments described in this chapter use a collapsed Gibbs sampler with
regularized hyperparameter updates to infer topic model parameters.  Methods such
as variational expectation maximization are
also possible, but we only fit models by Gibbs sampling updates due to its simplicity
of implementation and applicability to all the topic model architectures we consider.

The inference procedure for each model involves alternating between one iteration
of collapsed Gibbs sampling (sampling each token's topic assignment) and one
iteration of gradient
ascent for the parameters ${\eta_{b}}$ (bias vector in the document-topic prior),
$\omega_{b}$ (bias in the topic-word prior), and $\eta$ (weights determining how document
supervision influences the document-topic prior).

\subsubsection*{Gibbs Sampling}

The Gibbs sampling step involves sampling $z_{mn}$, each topic assignment, in turn
for every word in the corpus, $w_{mn}$, where $m$ is the document index and $n$
is the word index within a document.  Each topic assignment is drawn conditioned
on all previous topic assignments as well as the document-topic and topic-word
priors, $Dirichlet(\widetilde{\theta}_m)$ and $Dirichlet(\widetilde{\phi})$
respectively.  Formally, the probability that $k$ is sampled as the current topic
assignment is proportional to:

\begin{equation}
\begin{aligned}
  p(z_{mn} = k \vert \{ z \ \text{s.t.} \  z \neq z_{mn} \} , \widetilde{\theta}_m, \widetilde{\phi} ) \propto \\
  ( C(m, k) + \widetilde{\theta}_{mk} ) ( \frac{ C(w_{mn}, k) + \widetilde{\phi}_{(w_{mn}})}{ \sum_{v} C(v, k) + \widetilde{\phi}_v } )
\end{aligned}
\label{eq:user_conditioned_topic_models:topic_samples}
\end{equation}

where $C(m, k)$ is the number of times topic $k$ was sampled in document $m$ (excluding
the word we are currently sampling for) and $C(w_{mn}, k)$ is the number of times topic $k$
was sampled for word $w_{mn}$ \parencite{paul2015dissertationbg}.  The counts are aggregated over all topic assignments except
the current word being sampled.  The term on the right-hand side can be converted to a
probability by normalizing by the sum of unnormalized topic sampling probabilities:

\begin{align}
  \label{eq:user_conditioned_topic_models:topic_samples_partition_fn}
  Z & = \sum_{k=1}^{K} p(z_{mn} = k \vert \{ z \ \text{s.t.}\  z \neq z_{mn} \} , \widetilde{\theta}_m, \widetilde{\phi} )
\end{align}

This Gibbs sampling step is the same for both unsupervised \lda{} as well as supervised
models like \dmr{} -- the only difference between these two models is how
$\widetilde{\phi}$ and $\widetilde{\theta}$ are parameterized
(Figure \ref{fig:user_conditioned_topicmodels:generative_story}).

\subsubsection*{Hyperparameter Updates}

The hyperparameters that parameterize the document-topic prior are learned by
first-order methods.  We first calculate the gradient of the joint log-likelihood of both
the observed words and sampled topics with respect
to the prior hyperparameters, and then update the hyperparameters along a descent
direction\footnote{In practice we do not use a fixed learning rate, choose the exact
  gradient as a descent direction, but instead use
  adaptive learning rate methods to update the prior hyperparameters:
  AdaDelta \parencite{zeiler2012adadelta} or AdaGrad \parencite{adagrad}.}.




The partial derivative of an upstream topic model's
log-likelihood with respect to the document-topic prior
$\widetilde{\theta}_m$:

\begin{equation}
\begin{aligned}
  \frac{\delta \log \mathcal{L}( z \vert \widetilde{\theta}_m)}{\delta \widetilde{\theta}_{mk}} = &  \psi(C(m, k) + \widetilde{\theta}_{mk}) - \psi(\widetilde{\theta}_{mk}) \ + \\
  & \psi(\sum_{k'=1}^{K} \widetilde{\theta}_{mk}) - \psi(\sum_{k'=1}^{K} C(m, k') + \widetilde{\theta}_{mk})
\end{aligned}
\label{eq:user_conditioned_topic_models:theta_update}
\end{equation}


where $k$ is a topic index and $\psi$ is the
\emph{digamma} function, the derivative of the natural
logarithm of the gamma function (a generalization of
factorial to complex and real numbers). The partial
derivative with respect to $\widetilde{\phi}$ is:

\begin{equation}
\begin{aligned}
  \frac{\delta \mathcal{L}(w \vert z, \widetilde{\phi})}{\delta \widetilde{\phi}_w} = & \sum_{k=1}^{K} \psi (C(w, k) + \widetilde{\phi}_{w}) - \psi (\widetilde{\phi}_{w}) \ + \\
  & \psi (\sum_{w=1}^{W} \widetilde{\phi}_{w'}) - \psi (\sum_{w=1}^{W} C(w', k) + \widetilde{\phi}_{w'})
\end{aligned}
\label{eq:user_conditioned_topic_models:phi_update}
\end{equation}


where $w$ and $w'$ are word indices.
If $\widetilde{\theta}_m$ is parameterized as
$exp(\eta_b + \eta^T \alpha_m)$ (as in \dmr), we can simply
apply the chain rule to solve for the partial derivative
with respect to the prior hyperparameters $\eta_b$ and $\eta$.
The same goes for the topic-word parameters $\omega_b$.

In practice we also include a small amount of $\ell_2$ regularization on the gradient
term.  This is necessary to prevent hyperparameter weights from growing far too large,
overfitting to the current topic samples.

\removed{
Variational approaches have computational benefits in that no random
sampling is required: variational updates are completely deterministic,
and avoid the problem of waiting for your Monte Carlo chain to mix.
However,
until recently the variational updates needed to be derived by hand for every new topic
model architecture.  Although recent work by \citetemp{ranganath2014black} has eliminated
much of the work in deriving variational updates for new models, it is still not
clear whether the variational lower bound is sufficiently tight.  In any case,
Gibbs sampling with hyperparameter gradient updates is an optimization scheme that
can be applied with little customization to any \dmr{} model as well as deep
\dmr.
}

\subsubsection*{Calculating Model Fit}
Supervised classifiers are typically evaluated according to predictive
performance on some heldout data, unobserved during training.
Topic models (and unsupervised models in general) are trickier
to evaluate since the quality of a topic model is ultimately
determined by how coherent or interpretable the learned topics are
according to the human reviewing them.  Needless to say, human-judged
interpretability is not a scalable measure of model quality.  It
cannot be easily used for model selection: how many topics should my
model learn; what kind of document supervision should I condition the
model on?  It also cannot be used to decide when the optimization
algorithm has converged to a good solution.  Instead,
we use heldout perplexity in many of our experiments to decide
when a model has converged and which model to select.

Perplexity is just the exponentiated average negative
log probability of the corpus under the model:

\begin{align}
  Perplexity(w \vert z, \widetilde{\theta}, \widetilde{\phi}) = & exp \left( \frac{- \sum_{m=1}^{M} \sum_{n=1}^{N_m} \log p(w_{mn} \vert z_{mn}, \widetilde{\theta}_m, \widetilde{\phi}) }{\sum_{m=1}^{M} N_m} \right)
\end{align}

where $N_m$ is the number of words in document $m$.  Perplexity can be interpreted
as encoding how ``confused'' the topic model is on average for each token in the corpus.
A topic model with lower perplexity is better at predicting which
words are likely to occur in a document than one with higher perplexity (assigning higher
average log-likelihood to words in the corpus).

Heldout perplexity is computed by only aggregating
document-topic and topic-word counts from every \emph{other} token
in the corpus, and evaluating perplexity on the remaining
heldout tokens.  This corresponds to the ``document completion''
evaluation method as described in \parencite{wallach2009evaluation},
where instead of holding out the words in the second half of
a document, every other token is held out after shuffling the words
within a document\footnote{Word ordering within a document
  is of no consequence to the probabilistic topic models we consider,
  since they assume that each word is generated independently
  of all other words in a document (given the current document-topic
  distribution).  We shuffle the tokens within each document before
  topic sampling to ensure that ordering effects do not influence
  the word distributions between training and heldout tokens.}.
The counts $C(m, k)$, $C(w, k)$ are computed only over training token
samples.



\section{Deep Dirichlet Multinomial Regression (\dsprite)}
\label{sec:user_conditioned_topicmodels:ddmr}

\paragraph*{Problems with \dmr}

Document collections are often accompanied by metadata and
annotations, such as a book's author, an article's topic descriptor
tags, images associated with a product review,
or structured patient information associated with clinical records.
These document-level annotations 
provide additional supervision for guiding topic model learning.
\dmr{} is an upstream topic model with a particularly attractive method for incorporating arbitrary document features.
Rather than defining specific random variables in the graphical model for each new document feature,
\dmr{} treats the
document annotations as features in a log-linear model.
By making no assumptions on model structure of new random variables, \dmr{} is flexible to incorporating different types of features. 

Despite this flexibility, \dmr{} models are typically restricted to a small number of document features. 
Several reasons account for this restriction: (1) Many text corpora only have a small number of document-level features;
(2) Model hyperparameters become less interpretable as the dimensionality grows; and (3) \dmr{} is liable to overfit the hyperparameters
when the dimensionality of document features is high. In practice, applications of \dmr{} are limited to settings with a small
number of features, or where the analyst selects a few meaningful features by hand.

\paragraph*{Proposal: Neuralize the Prior}

One solution to addressing this restriction is to learn low dimensional
representations of document
metadata before conditioning \dmr{} on them.
Neural networks have shown wide-spread success at learning generalizable representations, often
obviating the need for hand designed features \parencite{coll2008}.  A prime example is word embedding features
in natural language processing, which supplant traditional lexical features
\parencite{brown1992class,mikolov2013distributed,pennington2014glove}.  Jointly learning networks that construct feature representations
along with the parameters of a standard NLP model has become a common approach.
For example, \textcite{yu2015combining} used a tensor decomposition to jointly learn features from both word
embeddings and traditional NLP features, along with the parameters of
a relation extraction model. Additionally, neural networks can handle a variety of data types including text, images, and general
metadata features. This makes them appropriate tools for
addressing dimensionality reduction in \dmr.

{\bf Deep} Dirichlet Multinomial Regression (\dsprite) is a model that
extends \dmr{} by introducing a deep neural network that learns
a transformation of the input metadata into features used to form
the document-topic prior.
Whereas \dmr{} parameterizes the document-topic priors as a
log-linear function of document features,
\dsprite{} jointly learns a feature representation for each document along with a log-linear
function that best captures the distribution over topics. Since the
function mapping document features to topic prior is a neural network, we can jointly optimize the topic model
and the neural network parameters by gradient ascent and back-propagation.


\subsection{Model}
\label{subsec:user_conditioned_topicmodels:ddmr_model}

\dsprite{} extends \dmr{} by replacing the document
supervision (vector), $\alpha$, in the document-topic
Dirichlet prior with a supervision embedding learned by a
function $f$ mapping arbitrary document supervision to a
real-valued vector, $\alpha' = f(\alpha)$.
For simplicity we make no assumptions on the type of this
function, only that it can be optimized to minimize a cost
on its output by gradient ascent. In practice,
we define this function as a neural network, where the
architecture of this network is informed by the type of
document supervision,
e.g. a convolutional neural network for images.
We use neural networks since they are expressive, generalize
well to unseen data, and can be jointly trained using
straightforward gradient ascent with back-propagation.

The generative story for \dsprite{} is as follows:

\begin{enumerate}
  \item Representation function $f \in \mathbb{R}^{F} \rightarrow \mathbb{R}^{K}$
  \item Topic-word prior parameters: $\omega_b \in \mathbb{R}^{V}$
  \item For each document $m$ with features $\alpha_m \in \mathbb{R}^{F}$, generate document prior:
  \begin{enumerate}
    \item $\widetilde{\theta}_{m} = exp( f( \alpha_m ) )$
    \item $\theta_m \sim Dirichlet(\widetilde{\theta}_m)$
  \end{enumerate}
  \item For each topic $k$, generate word distribution:
  \begin{enumerate}
    \item $\phi_k \sim Dirichlet(exp(\omega_b)$
  \end{enumerate}
  \item For each token generate corpus:
  \begin{enumerate}
    \item Topic (unobserved): $z_{mn} \sim \theta_m$
    \item Word  (observed): $w_{mn} \sim \phi_{z_{mn}}$
  \end{enumerate}
\end{enumerate}

\begin{figure}
  \center
\includegraphics[width=0.8\columnwidth,trim={5cm 6cm 0 0.5cm},clip]{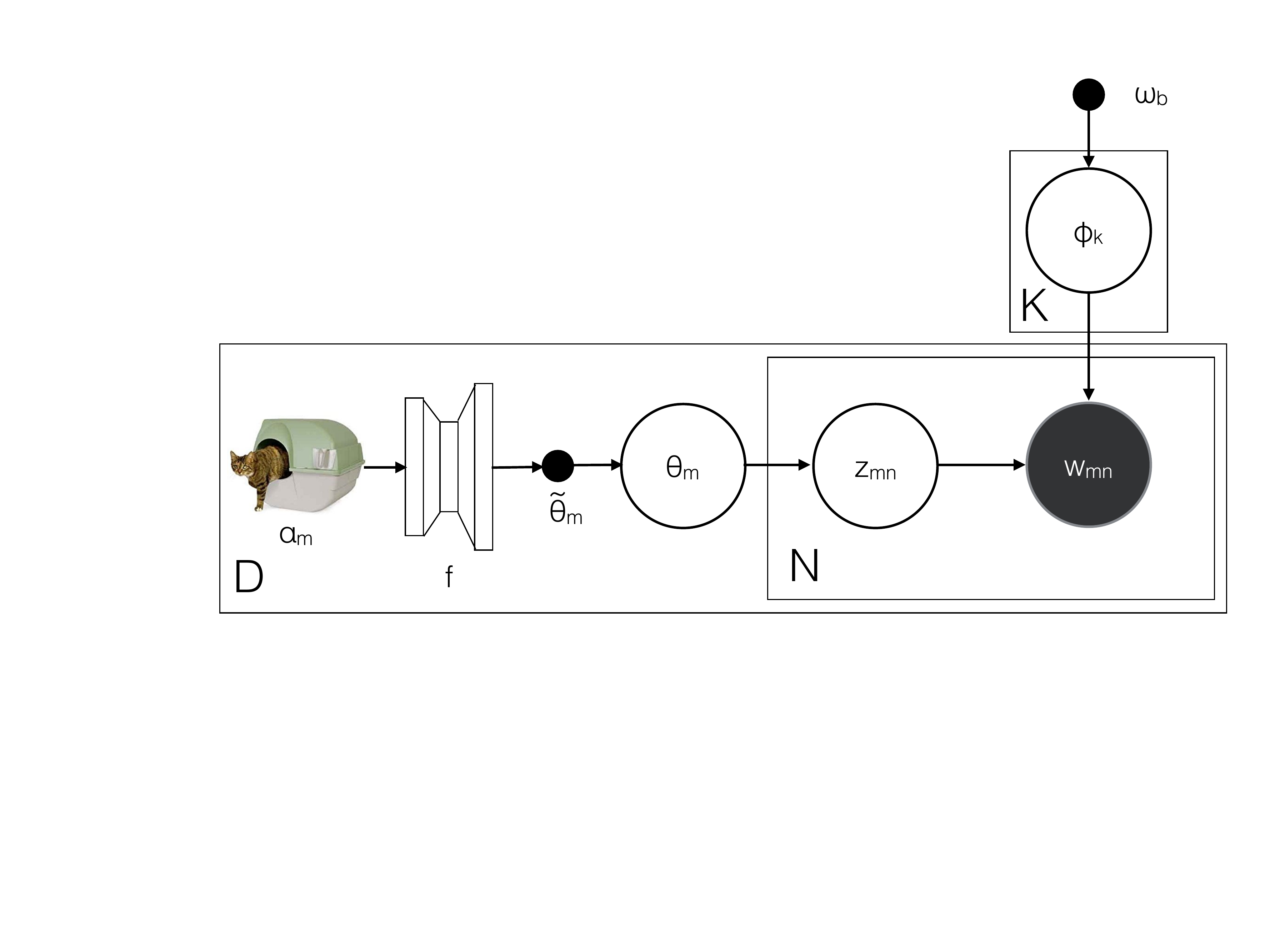}
\caption{Plate diagram of \dsprite{}.  $f$ is depicted as a
  feedforward fully-connected network, and the document features
  are given by an image -- in this case a picture of a cat.
}
\label{fig:user_conditioned_topic_models:ddmr_plate_diagram}
\end{figure}

where $V$ is the vocabulary size and $K$ are the number of topics. In practice, the document features need
not be restricted to fixed-length feature vectors, e.g. $f$ may be an RNN that maps from a sequence of characters to a fixed length vector in $\mathbb{R}^{k}$.
\dmr{} is a special case of \dsprite{} with the choice of a linear function for $f$.  
Figure \ref{fig:user_conditioned_topic_models:ddmr_plate_diagram} displays the graphical model diagram for \dsprite{}.

\subsection{Synthetic Experiments}
\label{subsec:user_conditioned_topicmodels:ddmr_synthetic}

Our intuition in developing \dsprite{} was that if the
document-level supervision is very high-dimensional
but lies on a low-dimensional manifold, then expressing
the supervision with respect to its position on this manifold
will avoid overfitting a topic model to the training corpus.
\dmr{} does not perform any such dimensionality
reduction and thus may be liable to overfitting when
the source of supervision is high-dimensional; a neural
prior topic model that learns an appropriate embedding
of the supervision will not be as susceptible.  We constructed a
synthetic dataset to determine what sort of corpora are more appropriate
to model with \dsprite{} rather than \dmr.

\subsubsection*{Data Generation}

Algorithm \ref{alg:user_conditioned_topicmodels:synth_data_story}
displays pseudocode for how the synthetic corpus
was generated.  10,000 documents were generated
with 50 tokens per document according to the generative
story of a \dsprite{} model where $f$ was defined as a
single-hidden-layer feedforward neural network with
5-dimensional hidden layer (sigmoid activation function) and
a 100-dimensional output layer (softmax activation function).
Each feature of the 100-dimensional document prior prefers
20\%, 4 out of a total of 20 topics, on average, where each
topic is sampled from a sparse Dirichlet prior over the
vocabulary.
The initialization of prior weights, $\eta$, in
the \dsprite{} model as well the the supervision for each
document is outlined in Algorithm
\ref{alg:user_conditioned_topicmodels:synth_sup_story}.

The observed document supervision was chosen such that it favored
one feature in the 100-dimensional ``true'' feature space.
We chose to generate document supervision this way, because
this ensured that the ``true'' supervision for
each document was sparse, preferring a small subset of topics,
and was therefore more interpretable.
The features are adjusted to this end by a technique similar
to the neural network visualization technique
in \textcite{simonyan2013deep} or image perturbation
in Deep Dream\footnote{Demo: \url{https://deepdreamgenerator.com} ;
  code: \url{https://github.com/google/deepdream}}. In this work, the
input features are adjusted
via gradient descent in order to better match the target
output layer activations (encoding which topics are preferred
by the current document).  We generate corpora where
the supervision noise variance $\epsilon$ varied from 0.01 to 10.0,
and the observed supervision dimensionality $d_{\text{i}}$ varied from
10 to 10,000 when generating corpora.  This was to gauge how
sensitive \dmr{} was to noisy or high-dimensional supervision
compared to \dsprite.

\begin{algorithm}
  \caption{Generate synthetic corpus from \dsprite{} model}
  \label{alg:user_conditioned_topicmodels:synth_data_story}
\begin{algorithmic}[1]
   \REQUIRE $\epsilon$, $d_{i}$ \Comment{Variance on supervision noise, supervision dimensionality}
   \STATE $M \gets 10^{4}$ \Comment{No. of documents}
   \STATE $N_m \gets 50$ \Comment{No. of tokens/document}
   \STATE $K \gets 20$ \Comment{No. of topics}
   \STATE $V \gets 100$ \Comment{Vocabulary size}
   \STATE $d_{o} \gets 100$ \Comment{``True'' supervision width}
   \STATE $d_{h} \gets 5$ \Comment{Hidden layer width}
   \STATE $\alpha_{i}, f \gets GenDocSup(M, d_{i}, d_{h}, d_{o})$ \Comment{Initialize prior network and supervision (Alg \ref{alg:user_conditioned_topicmodels:synth_sup_story})}
   \STATE $\alpha_{o} \gets f(\alpha_{i})$
   \STATE $\alpha_{i} \gets \alpha_i + Normal^{M \times d_i}(0.0, \epsilon)$
   \STATE $\delta \sim Bernoulli^{d_{o} \times K}(0.2)$ \Comment{Each true feature prefers roughly 20\% of topics}
   \FOR{$k \gets 1 \ldots K$}
    \STATE $\phi_k \sim Dirichlet^{V}(0.1)$ \Comment{Topic-word distribution drawn from sparse symmetric Dirichlet prior}
   \ENDFOR
   \STATE $\widetilde{\theta} \gets \alpha_o^{T} \delta$ \Comment{Document-topic prior}
   \STATE $D_{\text{index}}, W_{\text{index}} \gets [], []$ \Comment{Containers to store corpus}
   \FOR {$m \gets 1 \ldots M$}
    \STATE $\theta_m \sim Dirichlet(\widetilde{\theta}_m)$
    \FOR {$n \gets 1 \ldots N$}
     \STATE $D_{\text{index}} \gets D_{\text{index}} + [m]$
     \STATE $z \sim Multinomial(\theta_m)$ \Comment{Sample topic}
     \STATE $w \sim Multinomial(\phi_z)$ \Comment{Sample word}
     \STATE $W_{\text{index}} \gets W_{\text{index}} + [w]$
    \ENDFOR
   \ENDFOR
   \STATE \textbf{return} $(D_{\text{index}}, W_{\text{index}}, \alpha_{i})$
\end{algorithmic}
\end{algorithm}


\begin{algorithm}
  \caption{Generate supervision and \dsprite{} prior network.}
  \label{alg:user_conditioned_topicmodels:synth_sup_story}
\begin{algorithmic}[1]
   \REQUIRE $M, d_i, d_h, d_o, \epsilon$ \Comment{See Alg \ref{alg:user_conditioned_topicmodels:synth_data_story} for argument description.}
   \STATE $\alpha_i \sim Normal^{D \times d_i}(0.0, \epsilon) $ \Comment{Initialize supervision from mean-zero Gaussian}
   \FOR{$m \gets 1 \ldots M$}
    \STATE $j \sim Uniform(\{1, 2, \ldots, d_o\})$ \Comment{Sample ``true'' one-hot supervision} 
    \STATE $\alpha_{o}^{m} \gets e_j$
   \ENDFOR
   \STATE $W_i \sim Normal^{d_i \times d_h}(0.0, 1.0)$
   \STATE $W_h \sim Normal^{d_h \times d_o}(0.0, 1.0)$
   \STATE $f \gets (\alpha) \mapsto \sigma ( \sigma ( \alpha W_i ) W_h )$ \Comment{Initialize prior network}
   \STATE $\mathcal{L} = - \sum_{m=1}^{M} \sum_{j=1}^{d_o} \alpha_{o}^{m,j} \log f(\alpha_{i}^{m})_j$ \Comment{categorical cross-entropy loss}
   \FOR {$i \gets 1 \ldots 500$}
    \STATE $\alpha_i \gets \alpha_i - \frac{\delta \mathcal{L}}{\delta \alpha_i}$ \Comment{Update input features to match output layer activation}
   \ENDFOR
   \STATE \textbf{return} $(\alpha_i, f)$
\end{algorithmic}
\end{algorithm}

\subsubsection*{Model Fit to Synthetic Corpora}

For each corpus, we fit three 20-topic models: \lda, \dmr, and
\dsprite{}.  The \dsprite{} model had an identical prior architecture
to the generating model, but with randomly initialized weights.
We fit models by the procedure described in Section
\ref{subsec:user_conditioned_topicmodels:sup_models_fitting_models}
and evaluated model fit by heldout perplexity after 1,000 Gibbs sampling
iterations\footnote{Gradient updates were performed after a burnin
  of 100 iterations. Prior hyperparameters were updated with adaptive
  learning rate algorithm Adadelta, with a base step size of
  $\eta = 0.5$ and $\rho = 0.95$.}.

\begin{figure}
  \centering
    \includegraphics[width=0.7\textwidth]{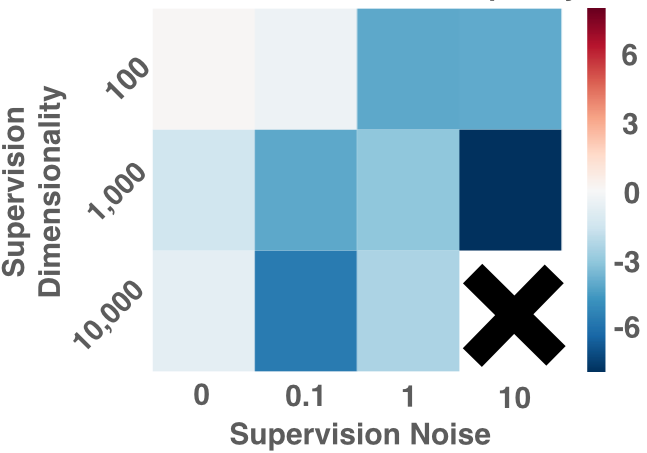}
    \caption{ Difference between \dsprite{} and \dmr{} heldout
      perplexity for different synthetic corpora (varying supervision
      dimensionality and Gaussian noise).  Bluer cells mean that \dsprite{}
      achieved lower perplexity than \dmr.  The case where \dmr{}
      hyperparameter optimization failed is marked by an ``X''.}
    \label{fig:user_conditioned_topicmodels:synth_heldout_ppl_diff}
\end{figure}

Figure \ref{fig:user_conditioned_topicmodels:synth_heldout_ppl_diff}
displays the absolute difference in heldout perplexity between
\dmr{} and \dsprite{} as a function of supervision dimensionality
and noise. In the case when $d_i = 10,000$ and $\epsilon = 10.0$,
\dmr{} failed to train properly.  NaNs were introduced in the
gradient when performing hyperparameter optimization and the model
could not be fit whereas \dsprite{} did not exhibit this problem during
training.  \dsprite{} always achieves at least as good model fit as \dmr{}
for the other synthetic datasets, with the gap between the two supervised
models widening as the observed supervision both grows in dimensionality and
becomes noisier.

\begin{figure}
  \centering
    \includegraphics[width=0.7\textwidth]{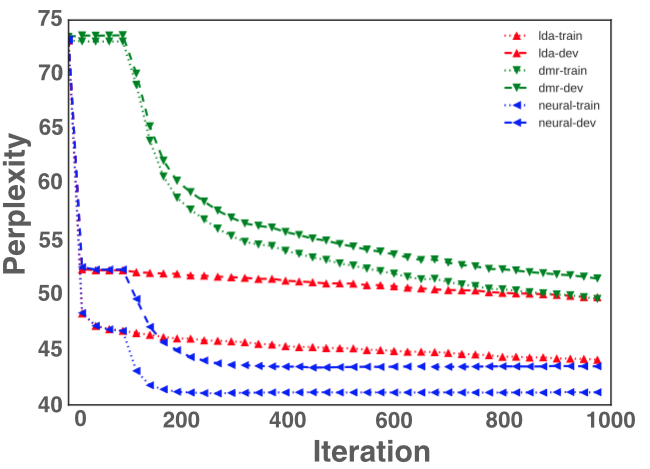}
    \caption{ Training and heldout perplexity training
      curves for synthetic corpus generated with $d_i = 1000$ and
      $\epsilon = 1.0$ for each model.  Training perplexity is marked by
      dotted lines, heldout by wider dashed lines.  Models --
      \lda: green, \dmr: red, \dsprite: blue.  The steep drop
      in perplexity after 100 iterations marks the end of burn-in
      and when hyperparameter optimization begins.}
    \label{fig:user_conditioned_topicmodels:synth_learning_curves}
\end{figure}

Figure \ref{fig:user_conditioned_topicmodels:synth_learning_curves}
displays training curves for each model with both train and heldout
perplexity.  For this corpus, \dmr{} actually \emph{underperforms} \lda,
meaning that the noisy supervision was leading the document-topic
priors astray.  \dsprite{}, on the other hand, can exploit the noisy
supervision to achieve a much lower perplexity.
\dsprite{} achieves no worse heldout perplexity
than \dmr{} across all corpora, excelling when noise is
high and supervision is wide.
This suggests that \dsprite{} is a promising model
for using high-dimensional, noisy supervision such as
user features to improve topic model fit.

\section{\dsprite{} Evaluation}
\label{sec:user_conditioned_topicmodels:ddmr_evaluation}

We explore the flexibility of \dsprite{} by considering three
different datasets that include different types of metadata
associated with each document. We first describe the documents
and metadata associated with each dataset and then the criteria
by which we evaluate topic models.

\subsection{Data}
\label{subsec:user_conditioned_topicmodels:ddmr_data}

All datasets were preprocessed similarly.  Article text was
tokenized by non-alphanumeric characters and numerals were
replaced by a special number token and infrequent word types were
excluded from the corpora, although the number of word types kept
varies slightly between corpora.

\subsubsection*{New York Times}

The New York Times Annotated Corpus \parencite{sandhaus2008new} contains
articles with extensive metadata used for indexing
by the newspaper. For supervision, we used the ``descriptor''
tags associated with each article assigned by archivists. 
These tags reflect the topic of an article, as well as organizations
or people mentioned in the article.  We selected all articles
published in 1998, and kept those tags that were associated with
at least 3 articles in that year -- 2424 unique tags. 20 of the
200 most frequent tags were held out from training for validation
purposes: \{\emph{ ``education and schools'',
  ``law and legislation'', ``advertising'',
``budgets and budgeting'', ``freedom and human rights'', ``telephones and telecommunications'', ``bombs and explosives'',
  ``sexual harassment'', ``reform and reorganization'', ``teachers and school employees'', ``tests and testing'', ``futures and options trading'', ``boxing'', ``firearms'', ``company reports'', ``embargoes and economic sanctions'', ``hospitals'', ``states (us)'', ``bridge (card game)'',} and \emph{``auctions''}\}. Articles contained an average of 2.1 tags each,
with 738 articles not containing any of these tags. Tags were
represented using a one-hot encoding to use for supervision.

Words occurring in more than 40\% of documents were removed, and only the 15,000 most frequent types were retained.  This resulted in a total of 89,397 articles with an average length of 158 tokens per article.

\subsubsection*{Amazon Product Reviews}

The Amazon product reviews corpus \parencite{mcauley2016addressing}
contains reviews of products as well as images of the product. 
We sampled 100,000 Amazon product reviews: 20,000 reviews sampled uniformly from the
\emph{Musical Instruments}, \emph{Patio, Lawn, \& Garden}, \emph{Grocery \& Gourmet Food},
\emph{Automotive}, and \emph{Pet Supplies} product categories.  We hypothesize that knowing
information about the product's appearance will indicate which words appear in the review,
especially for product images occurring in these categories.  66 of the reviews we sampled
contained only highly infrequent tokens, and were therefore removed from our data, leaving 99,934
product reviews.  Articles were preprocessed identically to the New York Times data.

We include images as supervision by passing each product's image through the Caffe
convolutional neural network reference model, trained to predict ImageNet object
categories\footnote{Features used directly from \url{http://jmcauley.ucsd.edu/data/amazon/}}.
We then extract the 4096-dimensional second fully-connected layer from this network to
use as document supervision.
Using these features as supervision in a \dsprite{} model with a feedforward network prior
is similar to fine-tuning a pretrained CNN to
predict a new set of labels.  Since the Caffe reference model is already trained on a large
corpus of images, we chose to fine-tune only the final layers so as to learn a transformation of
the already learned representation.

\subsubsection*{Reddit Messages}

We finally constructed a corpus of online text by selecting a sample of Reddit posts
made in January 2016.  A standard stop
list was used to remove frequent function words and we restricted the vocabulary to the 30,000
most frequent types.  We restricted posts made to subreddits, collections of topically-related
threads, with at least ten comments in this month (26,830 subreddits), and made by users with
at least five comments across these subreddits (total of 1,351,283 million users).  We then
sampled 10,000 users uniformly at random and used all their comments as a corpus, for a total
of 389,234 comments over 7,866 subreddits (document length mean: 16.3, median: 9)
We considered a one-hot encoding of the subreddit ID a comment belonged to as supervision.

This corpus differs from the others in two ways. First, Reddit documents
are very short, which presents a challenge for topic models that rely on 
detecting correlations in token use within a document. Second, the
Reddit metadata that may be useful for topic modeling is necessarily
high-dimensional (e.g. subreddit identity, a proxy for topical
content), so we believed that \dmr{} will likely have trouble exploiting it.

\subsection{Experiment Description}
\label{subsec:user_conditioned_topicmodels:ddmr_experiments}

We used the same procedure to fit topic models on each dataset.
Hyperparameter gradient updates were performed after a burnin period
of 100 Gibbs sampling iterations.
Hyperparameters were updated with the adaptive learning rate
algorithm Adadelta with a tuned base learning rate and fixed
$\rho=0.95$\footnote{We found
  this adaptive learning rate algorithm improved model fit in many
  fewer iterations than gradient descent with tuned step size and
  decay rate for all models.}.  All models were
trained for a maximum of 15,000 epochs, with early stopping if
heldout perplexity showed no improvements after 200 epochs (evaluated once every 20 epochs). 

We 
used single-hidden-layer multi-layer perceptrons (MLPs), with
rectified linear unit
(ReLU) activations on the hidden layer, and linear activation on the
output layer for the \dsprite{} neural prior architecture.  We sampled
three architectures for each dataset, by
drawing layer widths independently
at random from $[10, 500]$, and also included two architectures
with $(50, 10)$ and $(100, 50)$,
\emph{(hidden, output)} layers \footnote{We included these two very narrow architectures to ensure that some architecture learned a small feature representation, generalizing better when features are very noisy or only provide a weak signal for topic modeling.  We restricted ourselves to only train \dsprite{} models with single-hidden-layer MLPs in the priors to limit our search space.}.
We compare the performance of \dsprite{} to \dmr{} trained on the same
feature set as well as \lda.

For the New York Times dataset, we also compare \dsprite{} to \dmr{} trained on features
after applying principal components analysis (PCA) to reduce the dimensionality of
descriptor feature supervision, sweeping over PCA projection width in
$\{10, 50, 100, 250, 500, 1000\}$.  Comparing performance of \dsprite{} to PCA-reduced \dmr{}
tests two modeling choices.  First, it tests the hypothesis that explicitly learning
a representation for document annotations to maximize data likelihood produces a
``better-fit'' topic model than learning this annotation representation in unsupervised
fashion -- a two-step process.  It also lets us determine if a linear dimensionality
reduction technique is sufficient to learning a good feature representation for topic
modeling, as opposed to learning a non-linear transformation of the document supervision.
Note that we cannot apply PCA to reduce the dimensionality for subreddit id in the
Reddit data, since these are one-hot features.

\paragraph*{Model Selection}

Documents in each dataset were partitioned into ten equally-sized folds.  Model training
parameters of $\ell_1$ and $\ell_2$ regularization penalties on feature weights for \dmr{} and \dsprite{}
and the base learning rate for each model class were tuned to minimize heldout perplexity on the
first fold.  These were tuned \emph{independently for each model}, with number
of topics fixed to 10, and \dsprite{} architecture fixed to narrow layer widths $(50, 10)$.
Model selection was based on the macro-averaged performance on the next
eight folds, and we report performance on the remaining fold.  We selected models separately
for each evaluation metric. For \dsprite{}, model selection amounts to selecting the
document prior architecture, and for \dmr{} with PCA-reduced feature supervision, model selection
involved selecting the PCA projection width.

\subsection{Evaluation}
\label{subsec:user_conditioned_topicmodels:ddmr_model_evaluation}

Each model was evaluated according to heldout (1) perplexity,
(2) topic coherence by normalized pointwise mutual information (NPMI)
\parencite{lau2014machine}, and (3) a dataset-specific predictive task.  We
finally collect user preferences for topics learned by each model.
These are all typical approaches to evaluating topic models
\parencite{paul2015dissertation}.

NPMI computes an automatic measure of topic quality: the sum of pointwise
mutual information between pairs of the $m$ most likely words normalized
by the negative log probability of
each pair jointly occurring within a document (Equation \ref{eq:user_conditioned_topicmodels:npmi}).  A topic with a large NPMI score is one whose most probable words tend to occur in
the same documents more frequently than chance.
We calculated this topic quality metric on the top 20 most probable words in each topic,
and averaged over the most coherent 1, 5, 10, and all learned topics.  However, models
were selected to only maximize average NPMI over all topics.

\begin{align}
  {\sc NPMI} = & \sum_{i=1}^{m} \sum_{j=i+1}^{m} \frac{ \log \frac{P(w_i, w_j))}{P(w_i) P(w_j)} }{ - \log P(w_i, w_j) }
  \label{eq:user_conditioned_topicmodels:npmi}
\end{align}

For the prediction tasks, we used the sampled topic distribution associated with a
document, averaged over
the last 100 iterations, as features to predict a document label.
For New York Times articles we predicted 10 of the 200 most frequent descriptor tags restricting to
articles with exactly one of these descriptors.  For Amazon, we predicted the product category a document
belonged to (one of five), and for Reddit we predicted a heldout set of document subreddit IDs.  In the case of
Reddit, these heldout subreddits were 10 out of the 100 most prevalent in our data, and were held out just as in
the New York Times prediction task.  SVM models were fit on inferred topic distribution features and were
then evaluated according to accuracy, F1-score, and area under the ROC curve.  The SVM slack parameter was
tuned by 4-fold cross-validation on 60\% of the documents, and evaluated on the remaining 40\%.

\begin{figure}
\begin{center}
  \includegraphics[width=0.75\columnwidth]{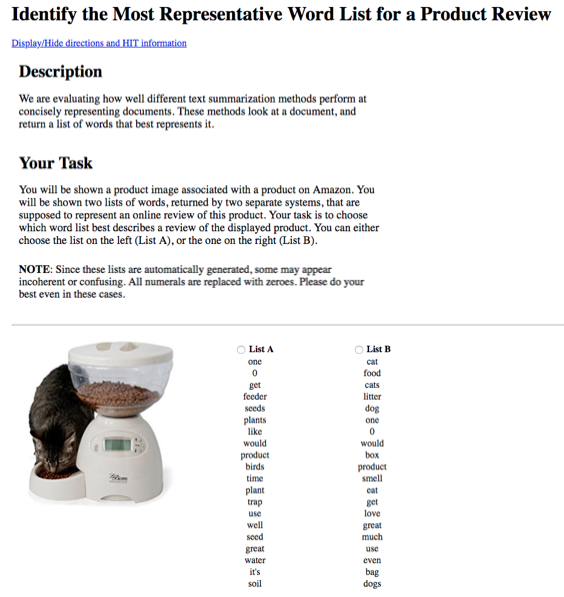}
  \caption{Screenshot of the topic quality judgment HIT.  Here we elicit which of two
    topics humans believe is more likely for an Amazon product with the displayed image
    (a cat feeder).}
\label{fig:user_conditioned_topicmodels:mturk_scrshot}
\end{center}
\end{figure}

We also collected human topic judgments using Amazon Mechanical Turk \parencite{callison-burch:10}.
Each subject was presented with a human-readable version of the features used for supervision.
For New York Times articles we showed the descriptor tags, for Amazon the product image,
and for Reddit the name, title, and public description of the subreddit.  We showed 
the top twenty words for the most probable topic sampled for the document with those features, as
learned by two different models.  One topic was learned by \dsprite{} and the other was either
learned
by either \lda{} or \dmr.  The topics presented were from the 200-topic model architecture that
maximized NPMI on
development folds.  Annotators were asked
``\emph{to choose which word list best describes a document} \ldots''
with the displayed features.  The topic learned by \dsprite{} was shuffled to lie on
either the right or left for each Human Intelligence Task
(HIT).
An example HIT for the Amazon data is shown in Figure \ref{fig:user_conditioned_topicmodels:mturk_scrshot}.

We obtained judgments on 1,000 documents for each dataset and each model evaluation
pair -- 6,000 documents in all. This task can be difficult for many of the features,
which may be unclear (e.g. descriptor tags without context) or difficult to interpret (e.g.
images of unfamiliar automotive parts). We chose to not present the document text
as well, since we did not want subjects to evaluate topic quality based on token
overlap with the actual document.

\subsubsection{Model Fit}
\label{subsubsec:user_conditioned_topicmodels:ddmr_model_fit}

\begin{table}
\begin{center}
\begin{tabular}{ll||p{1.0cm}p{0.56cm}|p{1.0cm}p{0.56cm}|p{1.0cm}p{0.56cm}}
\hline
 $Z$ & Model & \multicolumn{2}{c}{NYT} & \multicolumn{2}{c}{Amazon} & \multicolumn{2}{c}{Reddit} \\  \hline
\multirow{4}{*}{10}  & \lda &     3429 & (5) &     2300 & (7) &     3811 & (15) \\
    & \dmr &     \textbf{3385} & (6)  &     2475 & (9) &    3753 & (10) \\
    & \dmrpca &  3417 & (8) &   &    &    &   \\
    & \dsprite &    3395 & (7)  &     \textbf{2272} & (68) &   \textbf{3624} & (13) \\
\hline \multirow{4}{*}{20}  & \lda &     3081 & (6) &     2275 & (7) &  3695 & (19) \\
    & \dmr &     3018 & (4)  &     2556 & (48) &     3650 & (8) \\
    & \dmrpca &  3082 & (8) &   &    &  &     \\
    & \dsprite &    \textbf{3023} & (7) &     \textbf{2222} & (7) &     \textbf{3581} & (16) \\
\hline \multirow{4}{*}{50}  & \lda &     2766 & (8) &     2269 & (9) &     3695 & (17) \\
    & \dmr &      2797 & (34)  &     2407 & (20) &     3640 & (40) \\
    & \dmrpca &  2773 & (9) &   &    &    &   \\
    & \dsprite &    \textbf{2657} &  (8)  &     \textbf{2197} & (13) &     \textbf{3597} & (17) \\
\hline \multirow{4}{*}{100} & \lda &     2618 & (8) &     2246 & (10) &     3676 & (19) \\
    & \dmr &     2491 & (27) &     2410 & (75) &   3832 & (30) \\
    & \dmrpca &  2644 & (52)  &   &    &   &    \\
    & \dsprite &   \textbf{2433} & (10)  &     \textbf{2215} & (6) &    \textbf{3642} & (18) \\
\hline \multirow{4}{*}{200} & \lda &     2513 & (8) &     2217 & (7) &     3653 & (19) \\
    & \dmr &    2630 & (13) &     2480 & (65) &    3909 & (15) \\
    & \dmrpca & 2525 & (14) &   &    &   &    \\
    & \dsprite &    \textbf{2394} & (9)  &     \textbf{2214} & (12) &  \textbf{3587} & (11) \\
\hline
\end{tabular}
\caption{Test fold heldout perplexity for each dataset and model for number of topics $Z$.  Standard error of mean heldout perplexity over all cross-validation folds in parentheses.}
\label{tab:user_conditioned_topicmodels:all_ppl}
\end{center}
\end{table}

\dsprite{} achieves lower perplexity than \lda{} or \dmr{} for most combinations of
number of topics and dataset (Table \ref{tab:user_conditioned_topicmodels:all_ppl}).  It is striking that \dmr{}
achieves higher perplexity than \lda{} in many of these conditions.  This is particularly
true for the Amazon dataset, where \dmr{} consistently lags behind \lda.
\emph{Supervision alone does not improve topic model fit if it is too high-dimensional for learning}.
Perplexity is higher on the Reddit data for all models due to both a larger vocabulary size and
shorter documents.

It is also worth noting that finding a low-dimensional linear projection of the supervision
features with PCA does not improve model fit as well as \dsprite.
\dsprite{} benefits both from joint learning to maximize corpus log-likelihood and possibly by
the flexibility of learning non-linear projection (through the hidden layer ReLU activations).

\begin{figure*}
  \begin{center}
    \begin{subfigure}{\linewidth}
    \includegraphics[width=0.32\textwidth,page=1]{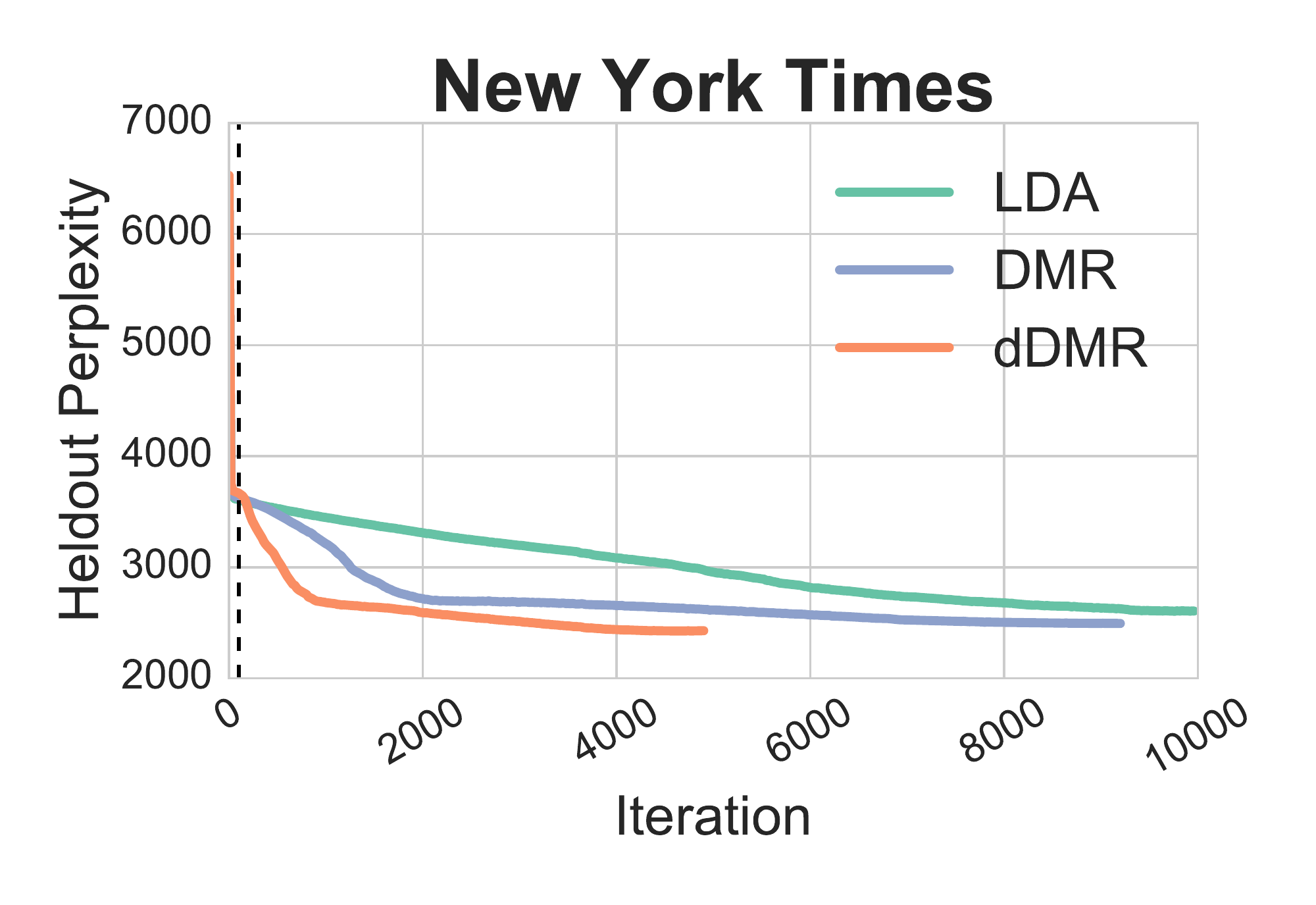} \hfill
    \includegraphics[width=0.32\textwidth,page=1]{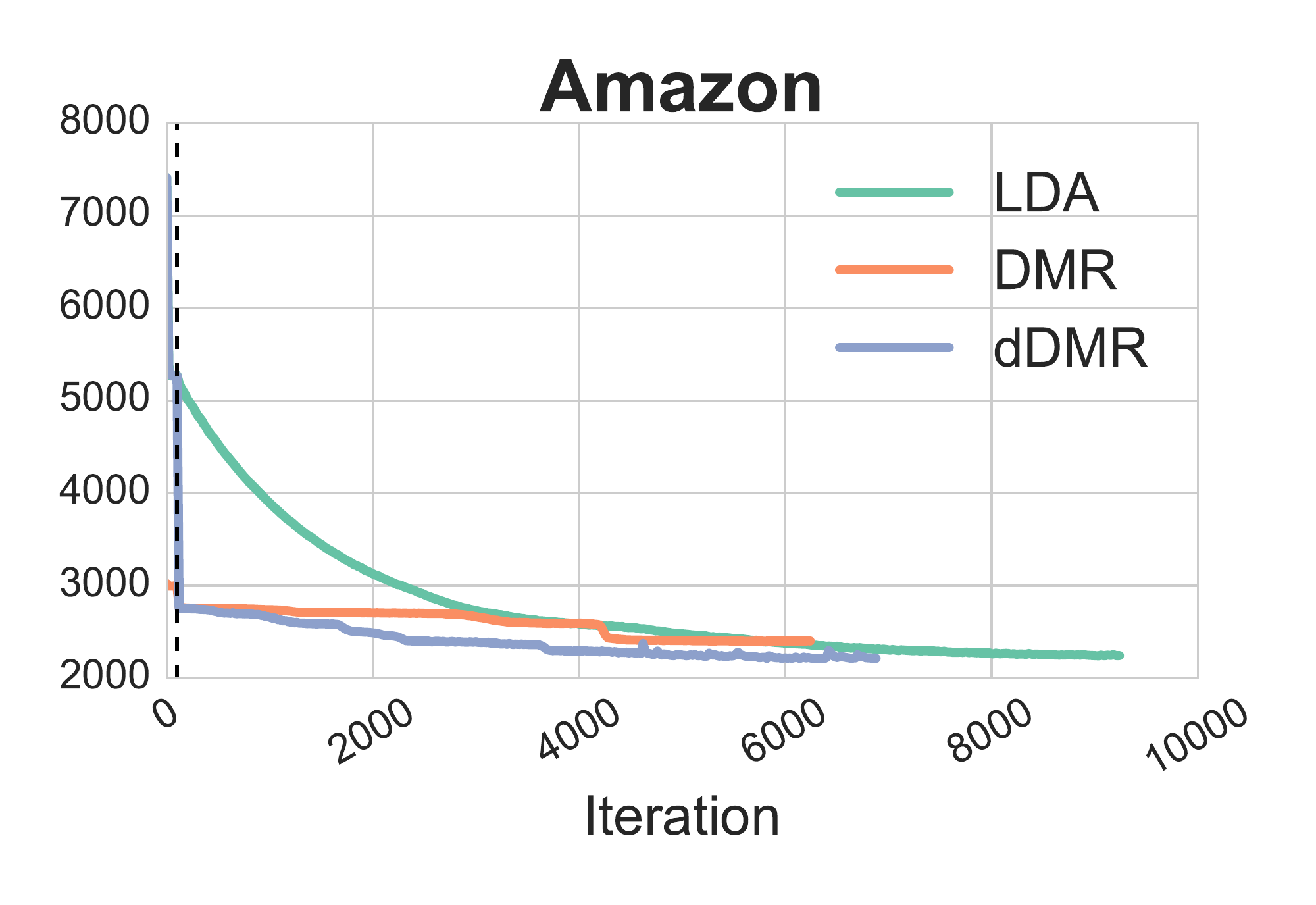} \hfill
    \includegraphics[width=0.32\textwidth,page=1]{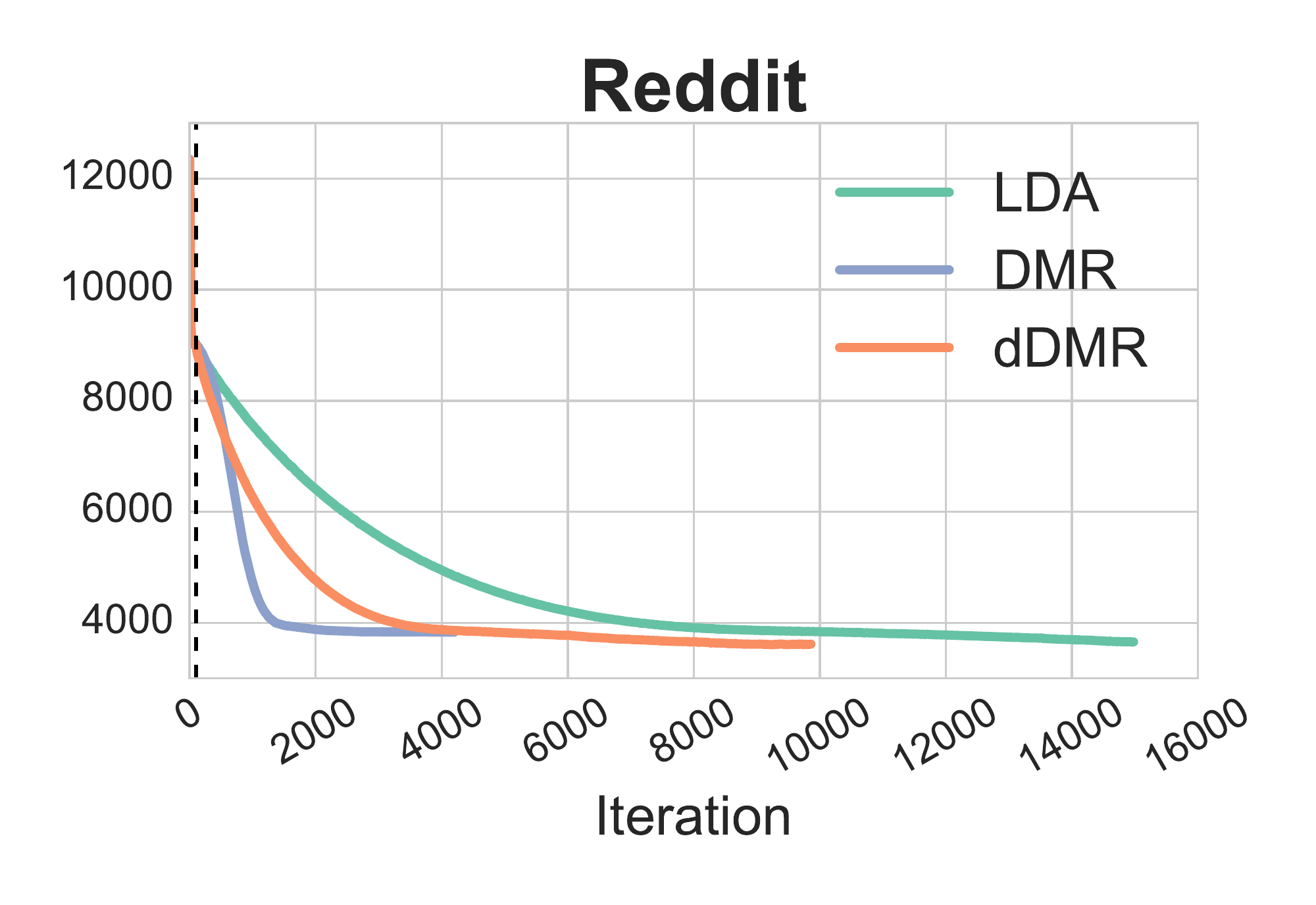} \hfill
    \end{subfigure}
    \caption{Heldout perplexity as a function of iteration for lowest-perplexity models
      with $Z=100$.  The vertical dashed line indicates the end of the burn-in period
      and when hyperparameter optimization begins.}
\label{fig:user_conditioned_topicmodels:ppl_learning_curves}
  \end{center}
\end{figure*}

Another striking result is the difference in speed of convergence between the supervised models and \lda{} (Figure
\ref{fig:user_conditioned_topicmodels:ppl_learning_curves}).  Even supervision that provides a weak
signal for topic modeling, such as Amazon product image features, can speed convergence over \lda{}.
In certain cases (Figure \ref{fig:user_conditioned_topicmodels:ppl_learning_curves} left), training \dsprite{} for 1,000 iterations
results in a lower perplexity model than \lda{} trained for over 10,000 iterations.

In terms of actual run time, parallelization of model training differs between supervised
models and \lda. Gradient updates necessary for learning the representation can be trivially
distributed across multiple cores using optimized linear algebra libraries (e.g. BLAS),
mitigating the additional cost incurred by hyperparameter updates in supervised models.
In contrast, the Gibbs sampling iterations can also be parallelized, but not as easily,
ultimately making resampling topics the most expensive step in model training.  Because of
this, the potential difference in runtime for a single iteration between
\dsprite{} and \lda{} is small, with the former converging in far fewer iterations.  The time taken per iteration by \dmr{} or \dsprite{} was at most twice as long as
\lda{} across all experiments.

\paragraph*{Sensitivity to Learning Parameters}

\begin{figure*}
  \begin{center}
    \includegraphics[width=0.96\textwidth,page=1]{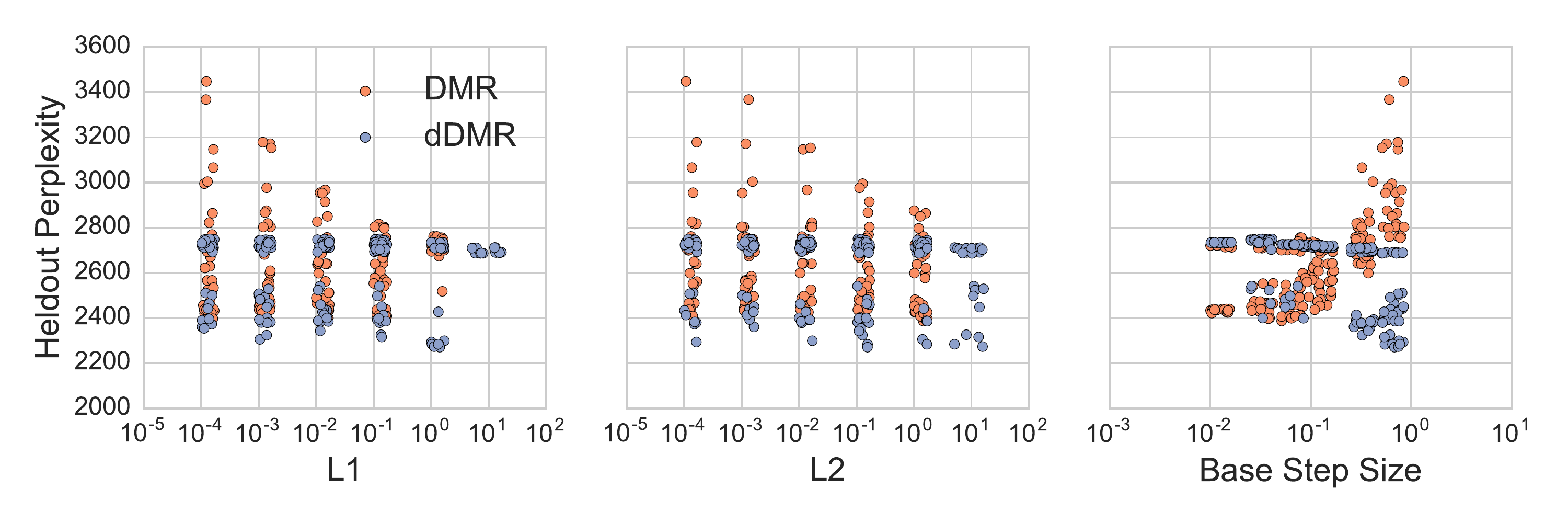} \hfill
    \caption{Heldout perplexity on the Amazon data tuning fold  for \dmr{} (orange) and \dsprite{} (purple) with a (50, 10) layer architecture as a function of training parameters: $\ell_1$, $\ell_2$ feature weight regularization, and base learning rate.  All models were trained for a fixed 5,000 iterations with horizontal jitter added to each point.}
  \label{fig:user_conditioned_topicmodels:tune_sensitivity}
  \end{center}
\end{figure*}

Also, \dsprite{} performance is much less sensitive to training parameters relative to \dmr.
While \dmr{} requires heavy $\ell_1$ and $\ell_2$ regularization and a very small step size to achieve low
heldout perplexity, \dsprite{} is relatively insensitive to the penalty on regularization
and benefits from a higher base learning rate (Figure \ref{fig:user_conditioned_topicmodels:tune_sensitivity}).
We found that \emph{\dsprite{} is easier to tune than \dmr}, requiring less exploration of
the training parameters.  This is also corroborated by higher variance in perplexity
achieved by \dmr{} across different cross-validation folds (Table \ref{tab:user_conditioned_topicmodels:all_ppl}).

\subsubsection{Topic Quality}
\label{subsubsec:user_conditioned_topicmodels:ddmr_topic_quality}

\begin{table}
  \small
  \begin{center}
\begin{tabular}{ll||rrrc|rrrr|rrrr}
\hline
 & & \multicolumn{4}{c}{New York Times} & \multicolumn{4}{c}{Amazon} & \multicolumn{4}{c}{Reddit} \\ \hline
$Z$ & Model & 1 & 5 & 10 & Overall & & \multicolumn{2}{c}{$\ldots$} &  & & \multicolumn{2}{c}{$\ldots$} &  \\ \hline
\hline
\multirow{4}{*}{10} & \lda &  52 &  49 &  43 &       43 &  25 &  23 &  20 &       20 & {\bf 125} &  {\bf 82} &  {\bf 56} & {\bf 56} \\
      & \dmr &  53 &  50 &  42 & 42 &  {\bf 58} &  {\bf 43} &  {\bf 31} & {\bf 31} &  43 &  35 &  30 &       30 \\
      & \dmrpca &  {\bf 63} &  {\bf 53} &  {\bf 45} &       {\bf 45} &  &  &  &       &  &  &  &       \\
      & \dsprite &  57 &  51 &  44 &       44 &  24 &  21 &  19 &       19 & 109 &  62 &  46 &       46 \\
\hline
\multirow{4}{*}{20} & \lda &  62 &  59 &  54 &       45 &  27 &  25 &  23 &       20 & {\bf 121} &  {\bf 87} &  {\bf 59} & {\bf 42} \\
      & \dmr &  63 &  60 &  56 &       45 &  66 &  56 &  {\bf 53} &       {\bf 43} &  81 &  49 &  41 &       34 \\
      & \dmrpca &  {\bf 76} &  {\bf 61} &  {\bf 57} &       {\bf 47} &  &  &  &       &  &  &  &       \\
      & \dsprite &  69 &  60 &  55 &       45 &  {\bf 97} &  {\bf 61} &  {\bf 53} &       40 & 109 &  66 &  49 &       38 \\
\hline
\multirow{4}{*}{50} & \lda &  80 &  66 &  62 &       44 &  30 &  27 &  25 &       20 & {\bf 135} &  {\bf 96} &  {\bf 64} &       34 \\
      & \dmr &  80 &  {\bf 67} &  {\bf 63} &       {\bf 46} & {\bf 136} &  {\bf 81} &  {\bf 73} &       {\bf 58} &  51 &  46 &  41 &       33 \\
      & \dmrpca &  {\bf 82} &  {\bf 67} &  63 &       45 &  &  &  &       &  &  &  &       \\
      & \dsprite &  76 &  65 &  61 &       45 &  71 &  65 &  62 &       44 & 121 &  74 &  54 &       {\bf 36} \\
\hline
\multirow{4}{*}{100} & \lda &  77 &  71 &  66 &       40 &  58 &  34 &  30 &       20 & 135 &  74 &  54 &       31 \\
      & \dmr &  {\bf 80} &  {\bf 74} &  70 &       {\bf 45} & {\bf 147} &  {\bf 83} &  {\bf 75} &       {\bf 59} & 111 &  67 &  50 &       {\bf 34} \\
      & \dmrpca & 79	& 69	& {\bf 75}	& {\bf 45} &  &  &  &       &  &  &  &       \\
      & \dsprite &  77 &  73 &  68 &       44 &  68 &  67 &  66 &       55 & {\bf 135} &  {\bf 78} &  {\bf 55} &       31 \\
\hline
\multirow{4}{*}{200} & \lda &  78 &  74 &  70 &       36 &  60 &  39 &  34 &       18 & {\bf 135} & {\bf 100} &  {\bf 67} &       29 \\
      & \dmr & 91 & {\bf 76} & 80 & 42 &  69 &  67 &  67 &       {\bf 61} & 132 &  84 &  59 & {\bf 32} \\
      & \dmrpca & {\bf 94} & {\bf 76} & {\bf 81} & 42 &  &  &  &       &  &  &  &       \\
      & \dsprite & 78 & 70 & 66 & {\bf 45} & {\bf 85} &  {\bf 73} &  {\bf 69} &       39 & {\bf 135} &  87 &  61 & 30 \\
\hline
\end{tabular}
\caption{Top-1, 5, 10, and overall topic NPMI across all datasets.  Models that maximized overall NPMI across dev folds were chosen and the best-performing model is in bold.}
\label{tab:user_conditioned_topicmodels:all_npmi}
\end{center}
\end{table}

Results for the automatic topic quality evaluation, NPMI, are mixed across datasets.
In many cases, \lda{} and \dmr{}
score highly according to NPMI, despite achieving higher heldout perplexity than
\dsprite{} (Table \ref{tab:user_conditioned_topicmodels:all_npmi}).  This may not be surprising as previous work
has found that perplexity does not correlate well with human judgments of topic
coherence \parencite{lau2014machine}.

However, in the Mechanical Turk evaluation, subjects found that \dsprite-learned topics
are more
representative of document annotations than \dmr{} (Table \ref{tab:user_conditioned_topicmodels:all_mturk_results}).
While subjects only statistically significantly favored \dsprite{} models over \lda{}
on the Reddit data, they favored \dsprite{} topics over \lda{} by a small margin
across all datasets, and statistically significantyl preferred \dsprite{} topics
over \dmr{} on two of the three datasets.
This is contrary to the model rankings according to NPMI, which predict that \dmr{} topics would be preferable.

\begin{table}
\begin{center}
\begin{tabular}{l|l|l}
\hline
 &  \lda & \dmr \\ \hline
New York Times & 51.1\% & 51.9\% \\  
Amazon & 51.9\% & 61.4\%$^{*}$ \\  
Reddit & 55.5\%$^{*}$ & 57.6\%$^{*}$ \\ 
\hline
\end{tabular}
\caption{\% HITs where humans considered \dsprite{} topics to be more representative
  of document supervision than the competing model.  $*$ denotes statistical significance
  according to a one-tailed binomial test at the $p=0.05$ level.}
\label{tab:user_conditioned_topicmodels:all_mturk_results}
\end{center}
\end{table}

\subsubsection{Predictive Performance}
\label{subsubsec:user_conditioned_topicmodels:ddmr_pred_performance}

\begin{table}
\footnotesize
\begin{center}
\begin{tabular}{ll||rrr|rrr|rrr}
  \hline
& & \multicolumn{3}{c}{New York Times} & \multicolumn{3}{c}{Amazon} & \multicolumn{3}{c}{Reddit} \\ \hline
$Z$ & Model & F1 & Accuracy & AUC &  & $\ldots$ & & & $\ldots$ & \\ \hline
\hline
\multirow{4}{*}{10} & \lda &   0.208 & {\bf 0.380} & 0.767 &   {\bf 0.662} & {\bf 0.667} & {\bf 0.891} &   0.130 & 0.276 & 0.565 \\
      & \dmr &   0.236 & 0.367 & 0.781 &   0.311 & 0.407 & 0.619 &   0.092 & 0.229 & {\bf 0.597} \\
      & \dmrpca &   {\bf 0.280} & 0.347 & 0.758 &      &    &    &      &    &    \\
      & \dsprite &   0.154 & 0.347 & {\bf 0.790} &   0.608 & 0.656 & 0.864 &   {\bf 0.170} & {\bf 0.300} & 0.596 \\
\cline{1-11}
\multirow{4}{*}{20} & \lda &   0.315 & 0.463 & 0.784 &   0.657 & 0.659 & 0.887 &   {\bf 0.121} & 0.258 & {\bf 0.579} \\
      & \dmr &   0.319 & 0.477 & 0.805 &   0.294 & 0.405 & 0.647 &   0.057 & 0.245 & 0.520 \\
      & \dmrpca &   0.343 & {\bf 0.540} & {\bf 0.831} &      &    &    &      &    &    \\
      & \dsprite &   {\bf 0.424} & 0.523 & 0.797 &   {\bf 0.706} & {\bf 0.711} & {\bf 0.911} &   0.071 & {\bf 0.274} & 0.566 \\
\cline{1-11}
\multirow{4}{*}{50} & \lda &   0.455 & 0.613 & 0.849 &   0.630 & 0.634 & 0.870 &   0.131 & 0.199 & 0.542 \\
      & \dmr &   0.478 & 0.650 & 0.877 &   0.396 & 0.499 & 0.619 &   {\bf 0.145} & 0.261 & {\bf 0.580} \\
      & \dmrpca &   0.505 & {\bf 0.667} & {\bf 0.887} &      &    &    &      &    &    \\
      & \dsprite &   {\bf 0.507} & 0.657 & 0.856 &   {\bf 0.716} & {\bf 0.726} & {\bf 0.916} &   0.118 & {\bf 0.272} & 0.551 \\
\cline{1-11}
\multirow{3}{*}{100} & \lda &   0.531 & 0.657 & 0.874 &   0.646 & 0.649 & 0.874 &   0.148 & 0.201 & 0.538 \\
      & \dmr &   0.552 & 0.683 & 0.898 &   0.392 & 0.463 & 0.688 &   0.107 & 0.233 & 0.512 \\
      & \dmrpca &  {\bf 0.602}  & {\bf 0.687} & {\bf 0.917} &      &    &    &      &    &    \\
      & \dsprite &   0.514 & 0.653 & 0.893 &   {\bf 0.650} & {\bf 0.660} & {\bf 0.893} &   {\bf 0.172} & {\bf 0.316} & {\bf 0.614} \\
\hline
\multirow{3}{*}{200} & \lda &   0.566 & 0.683 & 0.903 &   0.646 & 0.651 & 0.882 &   0.111 & 0.227 & 0.517 \\
      & \dmr &  0.576 & 0.670 &  {\bf 0.917} &   0.288 & 0.401 & 0.697 &   0.089 & 0.229 & 0.499 \\
      & \dmrpca & {\bf 0.648}  & {\bf 0.762} & 0.915 &  &    &    &      &    &    \\
      & \dsprite &   0.605 & 0.730 & 0.903 &   {\bf 0.716} & {\bf 0.721} & {\bf 0.909} &   {\bf 0.198} & {\bf 0.323} & {\bf 0.580} \\
\hline
\end{tabular}
\caption{Top F-score, accuracy, and AUC on prediction tasks for all \dsprite{} evaluation datasets.}
\label{tab:user_conditioned_topicmodels:all_prediction}
\end{center}
\end{table}

Finally, we consider the utility of the learned topic distributions for downstream prediction
tasks, a common use of topic models.  Although token perplexity is a standard measure
of topic model fit, it has no direct relationship with how topic models are typically used:
to identify consistent themes or reduce the dimensionality of a document corpus.
We found that features based on topic distributions from \dsprite{} outperform \lda{} and \dmr{}
on the Amazon and Reddit data when the number of topics fit is large, although they fail
to outperform \dmr{} on New York Times (Table \ref{tab:user_conditioned_topicmodels:all_prediction}).
Heldout perplexity is strongly correlated with predictive performance, with a Pearson correlation
coefficient, $\rho=0.898$ between F1-score and heldout perplexity on the Amazon
data.  This strong correlation is likely due to the
tight relationship between words used in product reviews and product category: a model
that assigns high likelihood to a words in a product review corpus should also be informative
of the product categories.  Prior work showed that upstream supervised topic models, such
as \dmr, learn topic distributions that are effective at downstream prediction tasks
\parencite{benton2016collective}.  We find that topic distributions learned by \dsprite{}
improve over \dmr{} in certain cases, particularly as the number of topics increases.


\subsubsection{Qualitative Results}
\label{subsubsec:user_conditioned_topicmodels:ddmr_qual_results}

We also qualitatively explored the product image representations \dmr{} and \dsprite{} learned on the Amazon data.
To do so, we computed and normalized the prior document distribution for a sample of documents learned by the lowest perplexity \dmr{} and \dsprite{} 200-topic models:

\begin{align}
  p(k \vert m) & = \frac{\widetilde{\theta}_m}{\sum_{k=1}^{Z} \widetilde{\theta}_{m,k}}
\end{align}


This is the prior probability of sampling topic $k$
conditioned on the features for document $m$ (before seeing
any words in the document).  We then marginalize over topics
to yield the conditional probability of a word $w$ given document $m$: $p(w \vert m) = \sum_{k=1}^{Z} p(w \vert k) p(k \vert m)$.

Table \ref{tab:user_conditioned_topicmodels:image_words} contains a sample of these probable words
given document supervision.  We find that \dsprite{} identifies words likely
to appear in a review of the product pictured.  However, some images lead \dsprite{}
down a garden path.
For example, a bottle of ``Turtle Food'' should not be associated with words for human consumables
like ``coffee'' and ``chocolate'', despite the container resembling some of these products.
However, the image-specific document priors \dmr{} learned are not as sensitive to the actual
product image as those learned by \dsprite.  The prior conditional probabilities $p(w \vert m)$
for ``Turtle Food'', ``Slushy Magic Cup'', and ``Rawhide Dog Bones'' product images are all
ranked identically by \dmr.

\begin{table}
\small
\begin{center}
  \begin{tabular}{|M{1.7cm}|M{2.0cm}|M{4.35cm}|M{4.35cm}|}
    \hline
{\small \bf Image} & {\small \bf Item} & {\small \bf \dsprite{} Probable Words} & {\small \bf \dmr{} Probable Words}  \\ \hline \hline
\adjustbox{center}{\includegraphics[height=2.05cm]{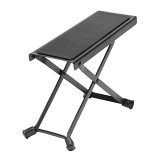}} & {\centering \bf Guitar Foot Rest } &
{\bf grill} easy cover well fit {\bf mower} fits {\bf job} {\bf gas} {\bf hose} light {\bf heavy} {\bf easily} {\bf stand} {\bf back} nice works {\bf use} {\bf enough} {\bf pressure}
 &
fit easy well works {\bf car} light {\bf sound} {\bf quality} {\bf work} {\bf guitar} {\bf would} {\bf 0000} cover nice {\bf looks} {\bf bought} {\bf install} {\bf battery} {\bf 00} fits
\\ \hline 
\adjustbox{center}{\includegraphics[height=2.05cm]{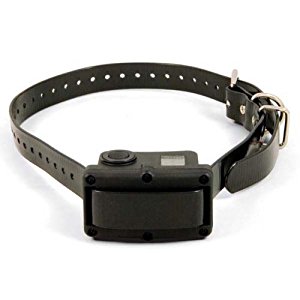}} & {\centering \bf Bark Collar} &
fit battery 0000 light install car sound easy work {\bf unit} {\bf amp} 00 {\bf lights} {\bf mic} {\bf power} works {\bf 000} {\bf took} {\bf replace} {\bf installed} &
fit easy {\bf well} works car light work {\bf quality} sound {\bf would} {\bf guitar} 0000 {\bf cover} {\bf nice} {\bf bought} {\bf looks} install battery 00 {\bf fits}
\\ \hline 
\adjustbox{center}{\includegraphics[height=2.05cm]{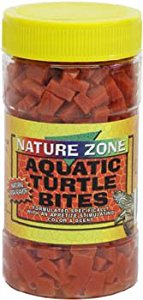}} & {\centering \bf Turtle Food } & taste coffee flavor food like love cat tea product tried dog eat chocolate {\bf litter} cats {\bf good} {\bf best} {\bf bag} sugar loves
 & taste coffee dog like love flavor food cat product tea cats tried {\bf water} {\bf dogs} loves eat chocolate {\bf toy} {\bf mix} sugar
\\ \hline 
\adjustbox{center}{\includegraphics[height=2.05cm]{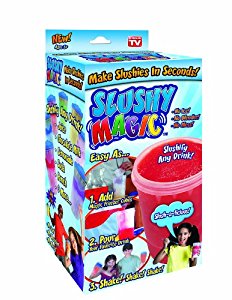}} & {\centering \bf Slushy Magic Cup } & food taste cat coffee flavor love like dog tea {\bf litter} cats eat tried product chocolate loves {\bf bag} good {\bf best} {\bf smell} &
taste coffee dog like love flavor food cat product tea cats tried {\bf water} {\bf dogs} loves eat chocolate {\bf toy} {\bf mix} good
\\ \hline 
\adjustbox{center}{\includegraphics[height=2.05cm]{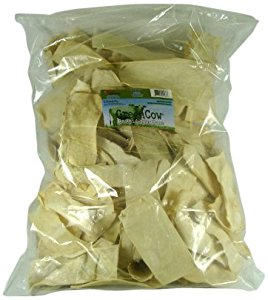}} & {\centering \bf Rawhide Dog Bones} & food cat dog cats {\bf litter} dogs loves love product {\bf smell} eat {\bf box} tried {\bf pet} {\bf bag} {\bf hair} taste {\bf vet} like {\bf seeds}  &
taste {\bf coffee} dog like love {\bf flavor} food cat product {\bf tea} cats tried {\bf water} dogs loves eat {\bf chocolate} {\bf toy} {\bf mix} {\bf good}
\\ \hline 
\adjustbox{center}{\includegraphics[height=2.05cm]{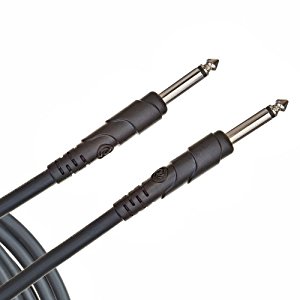}} & {\centering \bf Instrument Cable } & sound {\bf amp} guitar {\bf mic} {\bf pedal} {\bf sounds} {\bf price} {\bf volume} quality {\bf cable} {\bf great} {\bf bass} {\bf microphone} {\bf strings} music {\bf play} {\bf recording} {\bf 000} {\bf tone} unit &
sound guitar {\bf fit} {\bf easy} {\bf well} {\bf 0000} {\bf works} {\bf car} quality {\bf light} music {\bf cover} {\bf work} {\bf one} {\bf set} {\bf nice} {\bf looks} {\bf 00} {\bf install} unit
\\ \hline 
\end{tabular}
\caption{Top twenty words associated with each of the product images -- learned by
  \dsprite{} vs. \dmr{} ($Z=200$).  These images were drawn at random from the
  Amazon corpus (no cherry-picking involved).  Word lists were generated by marginalizing
  over the prior topic distribution associated with that image and then normalizing each word's
  probability by subtracting off its mean marginal probability \emph{across all images
  in the corpus}.  This is done to avoid displaying highly frequent words.  Words that differ
  between each model's ranked list are in bold.
}
\label{tab:user_conditioned_topicmodels:image_words}
\end{center}
\end{table}

\section{Application: Predicting Policy Surveys with Twitter Data}
\label{sec:user_conditioned_topicmodels:twitter_data}

In section \ref{sec:user_conditioned_topicmodels:ddmr}, we presented a
new supervised topic model that is more resilient to noisy supervision
than \dmr.  In this section we apply \dmr{} and \dsprite{} to three different
Twitter public policy opinion datasets, comparing how models conditioned on inferred
user location features fare against supervised models trained with distant
demographics and policy-relevant features.

\subsection{Motivation}

One goal of social media analytics is to complement or replace
traditional survey 
mechanisms \parencite{thacker1988,krosnick2005}. Traditional phone
surveys are both slow and expensive to run.
For example, the CDC's annual Behavioral Risk Factor Surveillance
System (BRFSS) is a health-related telephone survey
that collects health data by calling more than 400,000 Americans. 
The survey costs millions of dollars to run each year,
so adding new questions or obtaining finer-grained temporal
information can be prohibitive.

We consider three different public opinion data Twitter datasets.
Each of these datasets consists of tweets that are relevant to a
BRFSS survey question, an annual phone survey of hundreds of thousands
of American adults, chosen for its very large and geographically
widespread sample.  We selected the following three BRFSS questions:
the percentage of respondents in each U.S. state who (1) have
a firearm in their house (data from 2001, when the question was
last asked), (2) have had a flu shot in the past year (from 2013),
and (3) are current smokers (from 2013).

We would like to fit topic models to these data, and use the inferred topic
distribution to predict survey responses at the state level.  We consider
two classes of supervision for guiding supervised topic models: weak
author demographic and opinion supervision based on the inferred location
of the tweet author.
We also compare how predictive of BRFSS survey responses \dmr{} is to \dsprite{} when we
use a one-hot encoding of the author's inferred location as topic model supervision --
either at the state, county, or the city level.

\subsection{Datasets}
\label{subsec:user_conditioned_topicmodels:twitter_opinion_data}

We created three Twitter datasets
based on keyphrase filtering (Table
\ref{tab:user_conditioned_topicmodels:twitter_opinion_keywords}) with data 
collected from Dec.~2012 through Jan.~2015
to match tweets relevant to these three survey questions.
We selected 100,000 tweets uniformly at random for each dataset and geolocated them
to state/county using {\em Carmen} \parencite{carmen}.
Geolocation coverage is shown in Table \ref{tab:user_conditioned_topicmodels:twitter_opinion_data}.

\begin{table} 
\begin{center}
\begin{tabular}{|p{4.2cm}|p{3.6cm}|p{4.6cm}|}
\hline
{\bf Guns} & {\bf Vaccines} & {\bf Smoking} \\ \hline
gun, guns, second amendment, 2nd amendment, firearm, firearms &
vaccine, vaccines, vaccinate, vaccinated &
smoking, smoked, tobacco, cigarettes, cigarette, cigar, smoker, smokers \\
\hline
\end{tabular}
\end{center}
\caption{ \label{tab:user_conditioned_topicmodels:twitter_opinion_keywords} The keyphrases used to filter the BRFSS-related Twitter policy datasets.}
\end{table}

We consider the following sources of (distant) topic model supervision along with
one-hot author location indicators:

\vspace{-2ex}
\subsubsection{Survey}
This indirect supervision uses the values of the BRFSS survey responses that
we are trying to predict.
Tweets whose authors are resolved to a state are assigned the proportion of ``yes''
survey respondents within that state.
This setting reflects predicting the values for some states using data already available
from other states. This setting is especially relevant for BRFSS, since the survey is
run by each state with results collected and aggregated nationally.  Since not all states
run their surveys at the same time, BRFSS routinely has results available for some states
but not yet others.

\subsubsection{Census}
We also experimented with an alternative indirect type of supervision:
demographic information from the 2010 U.S. Census\footnote{ \url{http://www.census.gov/2010census/data/}}.
Demographic variables are correlated with the responses to the surveys we are trying to predict
\parencite{Hepburn2007-guns,King2012-tobacco,Gust2008-vaccines},
so we hypothesize that conditioning on demographic information may lead to more predictive
and interpretable topic models than no supervision at all.
This approach may be advantageous when domain-specific survey information is not readily available.

From the Census, we used the percentage of white residents per county as supervision for
tweets whose county could be resolved.
Although this feature is not directly related to the survey proportions we
are trying to predict,
it is sampled at a finer granularity than the state-level survey feature.
Proportion of tweets tagged with this feature are also included in Table
\ref{tab:user_conditioned_topicmodels:twitter_opinion_data}.  
In our experiments we consider these two
types of supervision in isolation to assess the usefulness of each class
of distant supervision.

\begin{table} 
\begin{center}
\begin{tabular}{|c|c|c|c|c|c|}
\hline
\bf Dataset & \bf Vocab & \bf State & \bf County & \bf City & \bf BRFSS\\ \hline
Guns & 12,358 & 29.7\% & 18.6\% & 16.7\% & Owns firearm \\
Vaccines & 13,451 & 23.6\% & 16.2\% & 14.8\% & Had flu shot \\
Smoking & 13,394 & 19.6\% & 12.8\% & 12.7\% & Current smoker \\
\hline
\end{tabular}
\end{center}
\caption{A summary of
  the three Twitter public policy datasets: size of the vocabulary,
  proportion of messages tagged at the state and county level, and the
  state-level survey question (BRFSS) asked.}
\label{tab:user_conditioned_topicmodels:twitter_opinion_data} 
\end{table}

\subsubsection{User Location Features}

In addition, we consider conditioned models on a one-hot encoding of location.  We
consider three different levels of granularity: {\em state}, {\em county}, and {\em city}. 
We restricted to only locations that were resolved in the United States,
treating tweets resolved to other countries as though they were not resolved at all.
As the surveys we are trying to predict are specific to American opinions, this ensured
that document-level features were restricted to those tweets that were more likely to come
from United State residents.  It also means that tweets that are tagged with a specific
location are a strict subset of those that were tagged by the state-level Survey feature.
Tweets that {\em Carmen} was unable to resolve were assigned a \texttt{NOT\_RESOLVED}
location feature, and finer granularity features backed off to the most
specific type of location resolved.

We consider these direct user location features since like the Census feature it is agnostic
to which survey question we are trying to predict.  However, unlike the Census feature, a topic
model conditioned
directly on location has more flexibility to learn which topics are more likely in that
specific location, rather than relying on a single, the proportion of white residents
in the county, as a proxy.

\subsection{Experiments}

We fit \dmr{} and \dsprite{} models conditioned on each feature
set, tuning for held-out perplexity and evaluated
its ability to predict the survey proportion for each state.
We also compared to an \lda{} model without any supervision.

The text was preprocessed by removing stop words and low-frequency words.
We also removed usernames, URLs, and non-alphanumeric tokens.
We applied z-score normalization to the BRFSS/Census values within each dataset, so that the mean value was 0.  
For tweets whose location could not be resolved, the Survey and Census document
supervision was set to 0.0, and the \texttt{NOT\_RESOLVED} one-hot location features
was active.

\paragraph*{Evaluation}

We evaluated the utility of topics as features for predicting the survey value
for each U.S. state, reflecting how well topics capture themes relevant to the survey question.
We inferred $\theta_m$ for each tweet and then
averaged these topic vectors over all tweets originating from each state, to
construct 50 feature vectors per model.  We used these features in a
regularized linear regression model. 
Average root mean-squared error (RMSE)
was computed using five-fold cross-validation:
80\% of the 50 U.S. states were used to train, 10\% to tune the 
$\ell_2$ regularization coefficient on the ridge regression model, and 10\% were used
for evaluation. In each fold, the topic models used supervision only for tweets from the
training set states, while the $\bm{\alpha}$ values were set to 0.0 (a neutral value)
for the held-out states.


For both perplexity and prediction performance, we sweep over number of topics in
$\{10, 25, 50, 100\}$ and report the best result.  
Results are averaged across five sampling runs to mitigate variation in performance due to
estimating model parameters by Gibbs sampling.

\paragraph*{Model Selection}

For tuning, we held out 10,000 tweets from the guns dataset and used the best learning parameters 
for all datasets.
We ran Spearmint \parencite{snoek2012} for 100 iterations to tune the learning parameters, running each sampler for 500 iterations.
We used Spearmint since it allowed us to automatically explore a large space of learning parameters quickly without resorting to brute-force grid search.  
Spearmint was used to tune the following learning parameters: the initial values for
$\bm{\omega_b}$ and $\bm{\eta_b}$, as well as $\ell_2$ regularization on
$\bm{\eta_b}$, $\bm{\omega_b}$, and $\bm{\eta}$.

Held-out perplexity is very sensitive to some parameters, such as initialization of $\bm{\eta_b}$ and $\bm{\omega_b}$,
while other parameters, such as the $\ell_2$ regularization on $\bm{\omega_b}$ had
little effect.  Once tuned, all models were trained for 2,000 iterations, using AdaGrad
with a master step size of 0.02, with no hyperparameter updates made in the first 200
iterations.

\subsubsection{Replication: Comparing \dmr{} to \dsprite{}}
\label{subsubsec:user_conditioned_topicmodels:experiment_description_ddmr_replication}

One crucial detail is that the initial set of experiments with \dmr{}
conditioned on Survey and Census features were run using a Java
package, \texttt{sprite}\footnote{\url{https://bitbucket.org/adrianbenton/sprite/}}, that
implemented Sprite topic models -- a class of upstream topic models with structured
priors \parencite{sprite}.
We attempted to replicate these experiments with a Python 3.5 library that supports defining
and training \dsprite{} models with feedforward neural network
priors, \texttt{deep-dmr}\footnote{\url{https://github.com/abenton/deep-dmr}}.  Relying
on \texttt{deep-dmr} was necessary as \texttt{sprite} does not support training
\dsprite{} models.

When replicating models in \texttt{deep-dmr}, we considered a different model
selection scheme due to the wide space of possible \dsprite{} models and time restrictions.
For each model class conditioned on feature type, we performed a grid search on the heldout
gun control tweets for $\ell_1$ and $\ell_2$ regularization constants in
$\{0.0, 10^{-4}, 10^{-2}, 10^{-1}\}$ and $\{10^{-4}, 10^{-2}, 10^{-1}, 10^{0}\}$, respectively.
These constants were then applied to
models trained on all datasets.  We also swept over base learning rate for each model class
in $\{10^{-2}, 10^{-1}, 10^{0}\}$, for Adadelta hyperparameter updates (this is the default
hyperparameter update algorithm in this package).
For all models, bias hyperparameters were initialized to $\bm{\eta_b}=-2$ and
$\bm{\omega_b}=-4$, corresponding to sparse initial Dirichlet priors.  


For \dsprite{} and each feature set, we swept over three single-hidden-layer architectures
with only linear activations for the document-topic prior: $\{[10, 5], [50, 10], [100, 50]\}$.
This amounts to a simple lookup embedding of the state, county, or city features.
For \dmr, we use each of the feature sets as supervision, but for \dsprite{} we
only consider state, county, or city indicator
features\footnote{Conditioning a \dsprite{} model with linear activations on a
  single feature offers no flexibility beyond \dmr{} on that feature.}.

\subsection{Results}

We first present the results on comparing \dmr{} conditioned on Survey and
Census features to an unsupervised topic model, \lda.  These experiments
were run using the \texttt{sprite} package.  We then present these experiments replicated
using \texttt{deep-dmr}, with the new model learning and selection criteria as described in
\ref{subsubsec:user_conditioned_topicmodels:experiment_description_ddmr_replication}.  We
compare perplexity and predictive performance of conditioning on location features under
this replication framework.

\subsubsection{Evaluating Survey and Census Features}

\begin{table}
\begin{center}
\small
\begin{tabular}{|l|l|cc|cc|cc|}
\hline
Features & Model & \multicolumn{2}{|c|}{Guns} & \multicolumn{2}{|c|}{Vaccines} & \multicolumn{2}{|c|}{Smoking}  \\ 
\hline
None & \lda{} & 17.44 & 2313 ($\pm 52$) & 8.67 & 2524 ($\pm 20$) & 4.50 & 2118 ($\pm 5$) \\
\hline
Survey & \dmr{} & 15.37 & {\bf 1529} ($\pm 12$) & 6.54 & {\bf 1552} ($\pm 11$) & \textbf{3.41} & {\bf 1375} ($\pm 6$) \\
Census & \dmr{}  & \textbf{11.51} & 1555 ($\pm 27$) & \textbf{5.15} & 1575 ($\pm 90$) & 3.42 & 1377 ($\pm 8$) \\ \hline
\end{tabular}
\end{center}
\caption{RMSE of the prediction task (left) and average perplexity (right) of topic models
  over each dataset, $\pm$ the standard deviation (learned under \texttt{sprite}).
  Perplexity is averaged over 5 sampling runs and RMSE is averaged over 5 folds of U.S.
  states.  As a benchmark, the RMSE on the prediction task using a bag-of-words model was
  11.50, 6.33, and 3.53 on the Guns, Vaccines, and Smoking data, respectively.
}
\label{tab:user_conditioned_topicmodels:results_pred_spearmintTuned}
\end{table}

Results from training models in \texttt{sprite} are shown in Table
\ref{tab:user_conditioned_topicmodels:results_pred_spearmintTuned}.
The important takeaway is that \dmr{} conditioned on indirect user features are more
predictive than LDA, an unsupervised model.
Not only do the supervised models substantially reduce prediction error, as might be
expected, but they also have substantially lower perplexity, and thus seem to be
topics that better represent the data.

The poor performance of \lda{} may be partially explained by the fact that Spearmint seems to overfit
\lda{} to the tuning set.  Other models attained a tuning set perplexity of between 1500 to 1600,
whereas \lda{} attained 1200.  To investigate this issue further, we separately ran experiments
with hand-tuned models, which gave us better held-out results for \lda, though still worse than
the supervised topic models (e.g., RMSE of 16.44 on the guns data).  Although Spearmint tuning is
not perfect, it is fair to all models.

For additional comparison, we experimented with a standard bag-of-words model, where features were
normalized counts across tweets from each state.  
This comparison is done to contextualize the magnitude of differences between models, even though
our primary goal is to compare different types of topic models.
We found that the bag-of-words results (provided in the caption of Table
\ref{tab:user_conditioned_topicmodels:results_pred_spearmintTuned}) are competitive with the best
topic model results.
However, topic models are often used for other advantages, e.g., interpretable models.

\dmr{} conditioned on Census features yielded worse predictive performance
than Survey-conditioned models on two of the three datasets, strangely enough.  We found
this surprising, but may be due to having an exceptionally small test set (test
performance averaged over 5 sets of 10 examples/states each).

\removed{
\paragraph*{Comparing Model Variants}

\removed{
\begin{table} 
\begin{center}
\small
\begin{tabular}{|c|c|cc|cc|}
\hline
\bf Model Class 1 & \bf Model Class 2 & \multicolumn{2}{|c|}{\bf Prediction (MSE)} & \multicolumn{2}{|c|}{\bf Perplexity} \\ \hline
Census & Survey & 1516 (0.032) & -0.15 (0.882) & 1266 (0.003) & 1.07 (0.287) \\
\dmr{} & \dsprite{} & X (X.X) & X.X (X.X) & X (X.X) & X.X (X.X) \\
State & Survey & X (X.X) & X.X (X.X) & X (X.X) & X.X (X.X) \\
County & Census & X (X.X) & X.X (X.X) & X (X.X) & X.X (X.X) \\
\hline
\end{tabular}
\end{center}
\vspace{-1.5ex}
\caption{ \label{tab:user_conditioned_topicmodels:twitter_model_significance} Performance comparison for different model/feature classes.  The first set of numbers in each cell is the Wilcoxon signed-rank statistic and corresponding p-value.  The second set is the paired t-test statistic and corresponding p-value.
A positive sign of the t-test statistic indicates that Model Class 1 has higher prediction error or perplexity than Model Class 2.
}
\end{table}
}

We measured the significance of the differences in performance of \dmr{} conditioned on
Census vs. Survey features according to (i) a Wilcoxon signed-rank test, and (ii) a paired t-test.
Model results were paired within each dataset and fold (for the prediction task, since mean
squared error varied from fold-to-fold).

Although conditioning on either Survey versus Census data improve over \lda{} and obtain
similar results, the Survey-conditioned models generally perform better.  The
Survey-conditioned model performance at prediction and heldout perplexity are both
significantly better than Census-conditioned models at the $p=0.05$ level, but only under the
Wilcoxon signed-rank test ($p=0.032$ for prediction and $p=0.003$ for perplexity)
not a paired t-test ($p=0.882$ for prediction and $p=0.287$ for perplexity).
The strong results using only Census data suggest that our methods can still have utility in
scenarios when there is limited survey data available, but for which the survey questions
have demographic correlates.
}

\paragraph*{Qualitative Inspection}

\begin{table}
\begin{center}
\small
\begin{tabular}{|cc|cc|cc|}
\hline
\multicolumn{2}{|c|}{Guns} & \multicolumn{2}{|c|}{Vaccines} & \multicolumn{2}{|c|}{Smoking}  \\ 
\hline
\bf $r=-1.04$ & \bf $r=0.43$ & \bf $r=-0.25$ & \bf $r=1.07$ & \bf $r=-0.62$ & \bf $r=1.04$ \\
\hline
gun & guns & ebola & truth & smoking & \#cigar \\
mass & people & trial & autism & quit & \#nowsmoking \\
shootings & human & vaccines & outbreak & stop & \#cigars \\
call & get & promising & science & smokers & cigar \\
laws & would & experimental & know & \#quitsmoking & james \\
democrats & take & early & connection & best & new \\
years & one & first & via & new & thank \\
since & away & results & knows & help & beautiful \\
australia & safe & hint & \#mhealth & \#smoking & \#habanos \\
1996 & use & safety & \#tetanus & please & \#cigarlovers \\
\hline
\end{tabular}
\end{center}
\vspace{-2ex}
\caption{Sample topics for
the \dmr{} model supervised with the survey feature.  A topic with a
strongly negative as well as a strongly positive $\eta$ value was chosen for
each dataset.  Positive value indicates that the tweet originates from a state with
many ``yes'' respondents to the survey question.}
\label{tab:user_conditioned_topicmodels:sample_topics_dmr}
\end{table}

Table \ref{tab:user_conditioned_topicmodels:sample_topics_dmr} displays example topics learned
by \dmr{} conditioned on Survey features.
For example, a topic about the results of the ebola vaccine trials is negatively correlated
with vaccine refusal,
while a topic about the connection between vaccines and autism is positively correlated
with vaccine refusal.  We did not observe noticeable qualitative differences in
topics learned by the different models, with an exception of \lda, where the topics
tended to contain more general words and fewer hashtags than topics learned by
the supervised models.

\paragraph*{Use Case: Predicting Support for Gun Restrictions}

\begin{table}
\begin{center}
\small
\begin{tabular}{|l|l|c|c|}
\hline
Features & Model &  RMSE (2001 Y included) &  RMSE (2001 Y omitted) \\ 
\hline
None & No model & 7.26 & 7.59 \\
& Bag of words & 5.16 & 7.31 \\
& \LDA{} & 6.40 & 7.59 \\
\hline
Survey & \dmr{} & \bf 5.11 & \bf 5.48 \\
\hline
\end{tabular}
\end{center}
\vspace{-2ex}
\caption{ RMSE when predicting proportion respondents opposing
  universal background checks with topic distribution features.  We experimented with
  (left) and without (right) including the 2001 proportion households with a firearm
  survey data as an additional feature.  ``\emph{No model}'' is the regression where
  we predict using only the 2001 proportion of households with a firearm.}
\label{tab:user_conditioned_topicmodels:predict_ubcs}
\end{table}

We ran an additional experiment to consider the setting of predicting a new survey with
limited available data.
We chose the subject of requiring universal background checks for firearm
purchases, a topic of intense interest in the U.S. in 2013 due to political events.
Despite the national interest in this topic, telephone surveys were only conducted for
less than half of U.S. states.
We identified 22 individual state polls  in 2013 that determined the proportion of
respondents that opposed universal background checks.  15 of the states were polled
by Public Policy Polling, while the remaining 7 states were polled by Bellwether Research,
Nelson A. Rockefeller Research, DHM Research, Nielsen Brothers, Repass \& Partners, or Quinnipiac University.  
We take this as a real-world example of our intended setting: a topic of interest where
resources limited the availability of surveys.


We used a topic model trained with data from the universal background check (UBC) survey
question as features for predicting the state values for the UBC surveys.
As in the previous experiments, we used topic features in a linear regression model, sweeping over $\ell_2$ regularization constants and number of topics, 
and we report test performance of the best-performing settings on the tuning set.  
We evaluated the model using five-fold cross-validation on the 22 states.

Additionally, we sought to utilize data from a previous, topically-related survey: the
``Guns'' BRFSS survey used in the previous section, which measured the proportion of
households with a firearm, asked in 2001. While the survey asks a different question, and is
several years out of date, our hypothesis is that the results from the 2001 survey will be
correlated with the new survey, and thus will be a good predictor. 
We experimented with and without including the values of the 2001 BRFSS survey (which
is available for all 50 states) as an additional feature in the regression model.

\begin{figure}
  \centering
    \includegraphics[width=0.85\columnwidth]{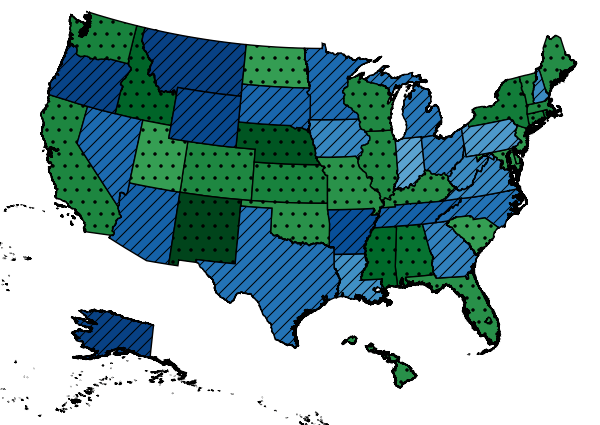}
\vspace{-2ex}
  \caption{Predictions from the \dmr{} model trained on
    the proportion opposed to universal background checks.  The 22 blue
    states hatched with lines were in the model's training set,
    while we have no survey data for the 28 green, dotted states.
    Darker colors denote higher opposition to background checks.
    New Mexico is predicted to have the highest percent of
    respondents opposed (53\%), while Utah has
    the lowest predicted opposed (18\%).
  }
  \label{fig:user_conditioned_topicmodels:predict_ubcs}
\end{figure}

Table \ref{tab:user_conditioned_topicmodels:predict_ubcs} contains the
cross-validation test results.  We compared the supervised topic model performance to
LDA as well as a bag-of-words model.  To put the results in context, we also trained
regression models using only the 2001 BRFSS values as features
(``No model, 2001 Y included'') as well as a regression model with no features at
all, only an intercept (``No model, 2001 Y omitted'').

In general, models that use text features outperform the baseline using only data from the 2001 survey, showing that text information derived from social media can improve survey estimation, even when using topically-related historic data.
Moreover, the supervised \dmr{} model trained on the UBC survey data is significantly better than an unsupervised topic model (\lda) with $p=0.06$, under a paired t-test across folds.  
The difference between \dmr{} and the bag-of-words model is not significant ($p=0.16$), although the difference is larger in the setting where the 2001 survey data is omitted.
For the Public Policy Polling surveys used to build the UBC data, the margin of error ranged from 2.9\% (more than 1000 polled) to 4.4\% (500 polled).  An RMSE of 5.1 is approximately equivalent to a 10\% margin of error at the 95\% confidence level, equivalent to polling roughly 100 people.

Finally, we trained the \dmr{} regression model (with 2001 BRFSS features)
on all 22 states, and used this model to make predictions 
of opposition to universal background checks 
for the remaining 28 states.
The predictions are shown in Figure
\ref{fig:user_conditioned_topicmodels:predict_ubcs}.  We generated
similar plots for \dsprite{} models conditioned on state and county
features.




\subsubsection{Conditioning on Location Features}

\begin{table}
\begin{center}
\small
\begin{tabular}{|l|l|cc|cc|cc|}
\hline
Features & Model & \multicolumn{2}{|c|}{Guns} & \multicolumn{2}{|c|}{Vaccines} & \multicolumn{2}{|c|}{Smoking}  \\ 
\hline
None & \lda{} & 9.72 & 1341 & 6.89 &   2192 & 3.81 & 1608 \\
\hline
Survey & \dmr{} & 9.76 & 1356 & 7.71 &  2216 & 5.24 & 1621 \\
Census & \dmr{}  & 12.11 & 1369 & 6.90 &  2223 & 3.64 & 1611 \\ \hline
State & \dmr{}  & {\bf 8.26} & 1370 & 6.89 &  2220  & 3.91 & 1608  \\
State & \dsprite{}  & 10.30 & 1350  & 8.00 & 2185 & 3.60 & 1624  \\
County & \dmr{}  & 11.75 & 1380  & 6.59 & {\bf 2183}  & 3.75 & 1616  \\
County & \dsprite{}  & 11.75 & 1366  & 6.17 & 2193  & 3.75 & 1613  \\
City & \dmr{}  & 10.35 &  1366  & 7.40 &  2185  & 3.55 & 1608  \\
City & \dsprite{}  & 10.81 & {\bf 1340}  & {\bf 5.75} & 2194  & {\bf 3.47} & {\bf 1605}  \\
\hline
\end{tabular}
\end{center}
\caption{RMSE of the prediction task (left) and average perplexity (right) of topic models
  over each dataset as replicated in \texttt{deep-dmr}.
  \emph{State}, \emph{County}, and \emph{City} are models trained with a one-hot
  encoding of the author's inferred state, county, or city.
}
\label{tab:user_conditioned_topicmodels:results_twitter_pred_replicated}
\end{table}

Table \ref{tab:user_conditioned_topicmodels:results_twitter_pred_replicated} contains the
performance of replicating the above perplexity and prediction evaluations with the model
training and selection criteria described in Section
\ref{subsubsec:user_conditioned_topicmodels:experiment_description_ddmr_replication}.  It also
includes performance of \dmr{} and \dsprite{} models conditioned on one-hot location features.

There most salient finding is that all models perform within a single standard deviation of
each other for all datasets and evaluation metrics.  This is different than what we had
observed originally where \lda{} was soundly beat by \dmr.  As mentioned above, \lda{} had
been likely performing worse since Spearmint overfit to the tuning set.

\paragraph*{Why is the Performance so Different?}

We took great pains to ensure that both \texttt{sprite} and \texttt{deep-dmr} optimized
models identically.  We made sure that initializing \lda{} under both both frameworks with
the same topic samples yielded identical training and heldout perplexity, and that they
achieved similar final heldout perplexity when learning over synthetic data.  In the process
of uncovering the difference between the original experiments and replications,
we noticed two small discrepancies between these implementations that were
subsequently resolved:

\begin{itemize}
  \item Treating every other token \emph{in the corpus} as heldout as opposed to every
    other token \emph{within each document}.  Since words are shuffled within each document
    as a preprocessing step, this did not affect heldout perplexity significantly.
  \item The bias hyperparameters were initialized to different values in each
    implementation: \texttt{deep-dmr} initialized them to
    $\bm{\eta_b}=-1$ and $\bm{\omega_b}=-2$.
\end{itemize}

The critical differences between model training and selection in the \texttt{sprite}-trained
models and the \texttt{deep-dmr} replicated models are as follows:

\begin{itemize}
  \item Hyperparameters updated with Adagrad (master learning rate fixed to $0.02$) $\rightarrow$ Adadelta (tuned master learning rate).
  \item Spearmint-tuned model selection $\rightarrow$ Grid search for training parameters for each model class
  \item Swept for bias hyperparameter initialization $\rightarrow$ Fixed to $\bm{\eta_b}=-2$ and $\bm{\omega_b}=-4$ for all models.
\end{itemize}

Although these \textbf{should be} relatively minor choices, they clearly had a
profound impact on the quality of topic models that were learned.

\section{Summary}
\label{sec:user-conditioned_topicmodels:summary}

This chapter presents a non-traditional application of user features and
embeddings: conditioning the topic distribution in a supervised topic model on
user features.  We show that supervision at the author-level is
important for modeling short-text corpora such as collections of social media
messages.  Specifically, we show that modeling three different opinionated
Twitter datasets benefit from distant, carefully chosen user features
-- responses to state-wide polls and county-level demographic features from
the United States census.

Synthetic experiments show that \dsprite{} is most appropriate when one is given
very high-dimensional and noisy supervision.  Empirically we find that it achieves
significantly better model fit (according to heldout perplexity) than \dmr{} on
three datasets with high-dimensional supervision regardless of number of topics
learned.  Topic distributions inferred by \dsprite{} with only
location indicator features perform on par with models conditioned on carefully
selected survey and census features in three Twitter opinion datasets related to
guns, vaccines, and smoking.  However, the performance of models heavily
depends on choices in model training and selection.

In light of our experiments, we encourage topic model practitioners to consider
fitting supervised topic models with document-level user features instead of \lda{}
when exploring new corpora.
Even distant or high-dimensional supervision helps improve the model fit and topic
quality, especially when choosing a \dsprite{} model.  Regardless, Gibbs
samplers converge far faster for models superivsed by user features than
unsupervised -- a very practical reason to opt for fitting \dsprite{} models over
\lda.

\cleardoublepage

\chapter{Multitask User Features for Mental Condition Prediction}
\label{chap:mtl_mentalhealth}

In Chapter \ref{chap:mv_twitter_users} we showed how multiview user embeddings
can be learned from different views of user behavior and can then be used to predict
hashtag use or friending behavior.  In Chapter \ref{chap:user_conditioned_topicmodels}
we showed that user-level features can even be used to speed topic
model convergence and improve topic model fit.  Although making accurate predictions of
who will friend whom is valuable to
social media platform engineers and fitting topic models more quickly is valuable to
social scientists understanding large text datasets, they are not strong examples of
how user features can directly improve people's lives.  In this chapter, we
show how to train stronger neural classifiers to predict a Twitter user's risk for suicide
as well as other mental health conditions solely from their tweets.  We do this by fitting
classifiers in the multitask learning (\MTL) framework where we consider predicting
multiple mental conditions a user has along with their gender as auxiliary tasks.

Section \ref{sec:mtl_mentalhealth:motivation} describes the motivation for building mental
condition classifiers from user tweets: why would we want to predict someone's mental
condition from social media and how might this save lives?  Section
\ref{sec:mtl_mentalhealth:architecture} describes the types of neural classifier
architectures we consider: basic logistic regression, single-task feedforward, and
\MTL{} feedforward architectures.
Section \ref{sec:mtl_mentalhealth:data} describes the training and evaluation dataset, a
collection of users with self-reported mental condition along with age and gender-matched
control users.  Section \ref{sec:mtl_mentalhealth:experiments} describes the experiment
protocol and how model hyperparameters were chosen, a crucial step in any careful
comparison of model classes.
Section \ref{sec:mtl_mentalhealth:results} ends by presenting model performance at
mental condition identification and an ablation analysis of which user features make the
most beneficial auxiliary tasks.
The content for this chapter is drawn from
\textcite{benton2017multitask}, a long paper in EACL 2017, and the majority of
experiments were performed as part of the 2016 JSALT workshop.

\section{Motivation}
\label{sec:mtl_mentalhealth:motivation}

Suicide is one of the leading causes of death worldwide, and over 90\% of individuals who die by suicide experience mental health conditions.\footnote{\url{https://www.nami.org/Learn-More/Mental-Health-Conditions/Related-Conditions/Suicide\#sthash.dMAhrKTU.dpuf}} However, detecting the risk of suicide, as well as monitoring the effects of related mental health conditions, is challenging.  Traditional methods rely on both self-reports and impressions formed during short sessions with a clinical expert, but it is often unclear when suicide is a risk in particular.\footnote{Communication with clinicians at the 2016 JSALT workshop \parencite{ws16ehr}.} Consequently, conditions leading to preventable suicides are often not adequately addressed.

Automated monitoring and risk assessment of patients' language has the potential to complement traditional assessment methods, providing objective measurements to motivate further care and additional support for people with difficulties related to mental health. This paves the way toward verifying the need for additional care with insurance coverage, for example, as well as offering direct benefits to clinicians and patients.

We explore some of these possibilities in the mental health space using {\it written social media text} that people with different mental health conditions are already producing.  Uncovering methods that work with such text provides the opportunity to help people with different mental health conditions by leveraging a data source they are already contributing to.

Social media text carries implicit information about the author, which has been modeled in natural language processing (NLP) to predict author characteristics such as \emph{age} \parencite{Goswami:ea:2009stylometric,Rosenthal:ea:2011age,Nguyen:ea:14}, 
\emph{gender} \parencite{Sarawgi:ea:2011gender,Ciot:ea:2013,Liu:ea:2013,volkova2015,hovy2015demographic}, \emph{personality} \parencite{schwartz2013toward,volkova2014inferring,plank2015personality,park:ea:2015,preotiuc-pietro:ea:2015personality}, 
and \emph{occupation} \parencite{preotiuc2015analysis}. 
Similar text signals have been effectively used to predict mental health conditions such as \emph{depression} \parencite{dechoudhury2013predicting,W15-1204,W14-3214}, \emph{suicidal ideation} \parencite{W16-0311,Y15-1064}, \emph{schizophrenia} \parencite{W15-1202} or \emph{post-traumatic stress disorder (PTSD)} \parencite{W15-1206}.

However, these studies typically model each condition in isolation, which misses the opportunity to model coinciding influence factors.  Tasks with underlying commonalities (e.g., part-of-speech tagging, parsing, and NER) have been shown to benefit from multi-task learning (\MTL{}), as the learning implicitly leverages interactions between them ~\parencite{caruana1993multitask,sutton:ea:2007,rush2010dual,collobert:ea:2011,sogaard2016deep}. Suicide risk and related mental health conditions are therefore good candidates for modeling in a multi-task framework.  

In this chapter we apply multi-task learning for detecting suicide risk and mental health conditions. The tasks in our model include the user mental health conditions of \emph{neuroatypicality} (i.e. having an atypical mental condition) and \emph{suicide attempt}, as well as the related mental health conditions of \emph{anxiety}, \emph{depression}, \emph{eating disorder}, \emph{panic attacks}, \emph{schizophrenia}, \emph{bipolar disorder}, and \emph{post-traumatic stress disorder (PTSD)}, and we explore the effect of task selection on model performance. We additionally include the effect of modeling a user demographic feature, \emph{gender}, which has been shown to improve accuracy in tasks using social media text \parencite{Volkova:ea:2013exploring,hovy2015demographic}.

Predicting suicide risk and several mental health conditions jointly opens the possibility for the model to leverage a shared representation for conditions that frequently occur together, a phenomenon known as \emph{comorbidity}. Further including gender reflects the fact that gender differences are found in the patterns of mental health \parencite{WHO}, which may help to sharpen the model.  The \MTL{} framework we propose allows such shared information across predictions and enables the inclusion of several loss functions with a common shared underlying representation.  This approach is flexible enough to extend to factors other than the ones shown here, provided suitable data.

We find that choosing tasks that are prerequisites or related to the main task is critical for
learning a strong model, similar to findings in \textcite{caruana1996algorithms}.  We further
find that including gender as an auxiliary task improves accuracy across a variety of
conditions, including suicide risk. The best-performing model from our experiments demonstrates
that multi-task learning is a promising new direction in automated assessment of mental health
and suicide risk, with possible application to the clinical domain.

\subsection{Findings}
\begin{enumerate}
\itemsep-0.5em
  \item We demonstrate the utility of \MTL{} in predicting mental health conditions from social user text -- a notoriously difficult task \parencite{coppersmith2015adhd,W15-1204} -- with potential application to detecting suicide risk.
  \item We explore the influence of task selection on prediction performance, including the effect of gender.
  \item We show how to model tasks with a large number of positive examples to improve the prediction accuracy of tasks with a small number of positive examples.
  \item We compare the \MTL{} model against a single-task model with the same number of parameters, which directly evaluates the multi-task learning approach.
  \item The proposed \MTL{} model increases the True Positive Rate at 10\% false alarms by up to 9.7\% absolute (for anxiety), a result with direct impact for clinical applications.
\end{enumerate}



\section{Model Architecture}
\label{sec:mtl_mentalhealth:architecture}

A neural multi-task architecture opens the possibility of leveraging commonalities and differences between mental conditions.  Previous work \parencite{collobert:ea:2011,caruana1996algorithms,caruana1993multitask} has indicated that such an architecture allows for sharing parameters across tasks, and can be beneficial when there is 
varying degrees of annotation across tasks.
This makes MTL particularly compelling in light of mental health comorbidity, and given that different conditions have different amounts of associated data. 

Previous \MTL{} approaches 
have shown considerable improvements over single task models, and the arguments are convincing: predicting multiple related tasks should allow us to exploit any correlations between the predictions.  However, in much of this work, an \MTL{} model is only one possible explanation for improved accuracy.  Another more salient factor has frequently been overlooked: The difference in the expressivity of the model class, i.e., neural architectures vs.~discriminative or generative models, and critically, differences in the number of parameters for comparable models.  Some comparisons might therefore have inadvertently compared apples to oranges.

In the interest of examining the effect of \MTL{} specifically, we compare the multi-task predictions to models with equal expressivity. We evaluate the performance of a standard logistic regression model (a standard approach to text-classification problems), a multilayer perceptron single-task learning (\STL) model, and a neural \MTL{} model, the latter two with equal numbers of parameters.
This ensures a fair comparison by decoupling the unique regularization of \MTL{} from the dimensionality-reduction aspect of deep architectures in general.

\begin{figure}
    \begin{center}
	\includegraphics[trim=0cm 0.5cm 2.0cm 2.0cm, width=1.0\textwidth]{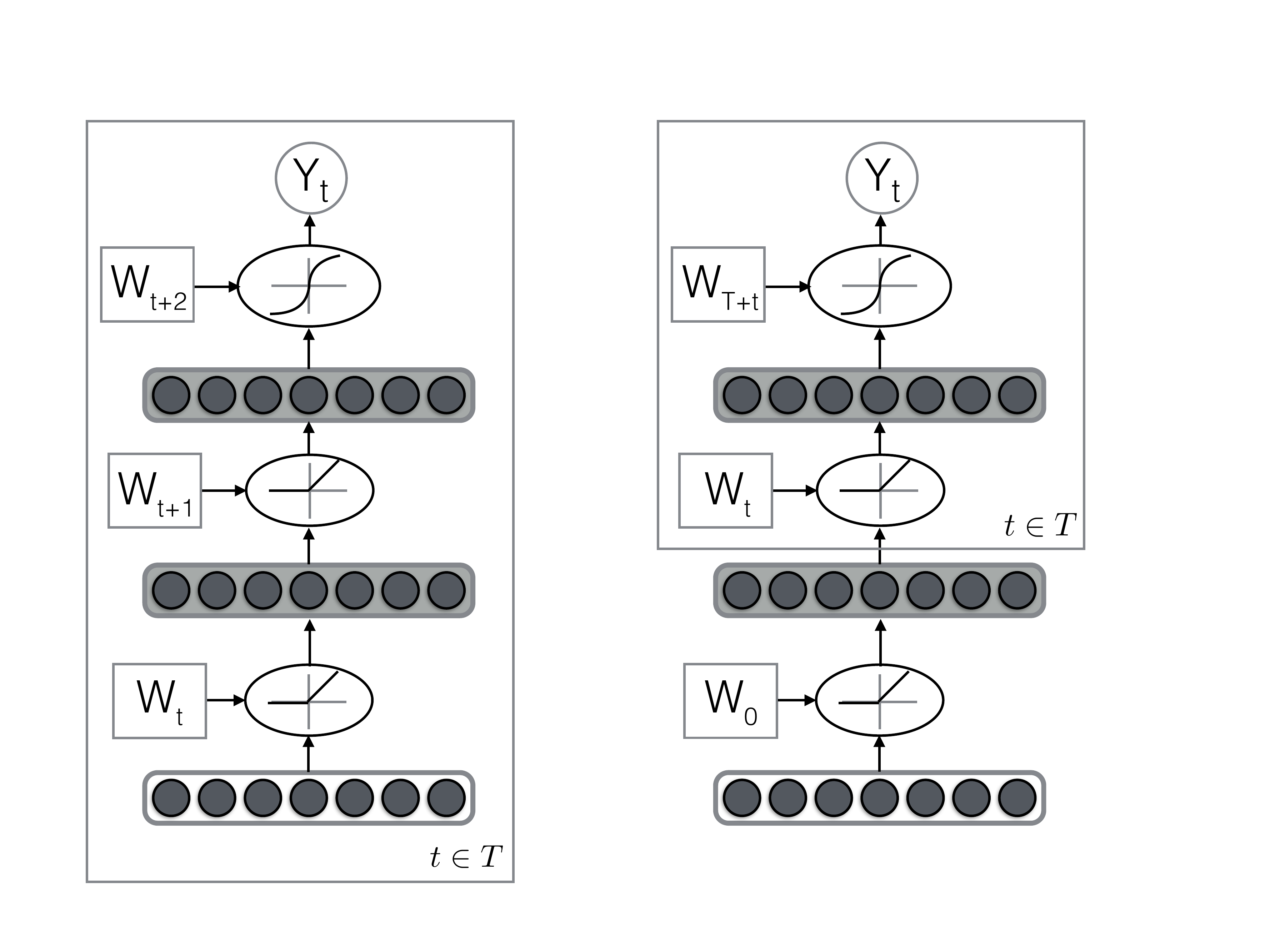} \\
    {\bf Single-task} \hspace{3.25em} {\bf Multi-task}
    \end{center}
	\caption{\STL{} model in plate notation (left): weights trained independently for each task $t$ (e.g., anxiety, depression) of the $T$ tasks.  MTL model (right): shared weights trained jointly for all tasks, with task-specific hidden layers.  Curves in ovals represent the type of activation used at each layer (rectified linear unit or sigmoid). Hidden layers are shaded.}
\label{fig:mtl_mentalhealth:model}
\end{figure}

The neural models we evaluate come in two forms.  The first, depicted in plate notation on the left in Figure \ref{fig:mtl_mentalhealth:model} are the \STL{} models.  These are feedforward networks with two hidden layers, trained independently to predict each task.  On the right in Figure \ref{fig:mtl_mentalhealth:model} is the \MTL{} model, where the first hidden layer from the bottom is shared between all tasks.  An additional per-task hidden layer is used to give the model flexibility to map from the task-agnostic representation to a task-specific one.  Each hidden layer uses a rectified linear unit as non-linearity.  The output layer uses a logistic non-linearity, since all tasks are binary predictions. 
The \MTL{} model can easily be extended to a stack of shared hidden layers,
allowing for a more complicated mapping from input to shared space.\footnote{We tried training a 4-shared-layer \MTL{} model to predict targets on a separate dataset, but did not see any gains over the standard 1-shared-layer \MTL{} model in our application.  Different classification tasks require different selections of neural architecture model depth.}

As noted in \textcite{collobert:ea:2011}, MTL benefits from mini-batch training, which both allows optimization to jump out of poor local optima, and more stochastic gradient steps in a fixed amount of time \parencite{bottou2012}.
We create mini-batches by sampling uniformly from the users in our data, where each user has some subset of the conditions we are trying to predict, and may or may not be annotated with gender.  At each mini-batch gradient step, we update weights for all tasks simultaneously.  This not only allows for randomization and faster convergence, it also provides a speed-up over the task selection process reported in earlier work \parencite{collobert:ea:2011}.

Another advantage of this setup is that we do not need complete information for every instance: learning can proceed with asynchronous updates, dependent on what the data in each batch has been annotated for, while sharing representations throughout.  This effectively learns a joint model with a common representation for several different tasks, allowing the use of several ``disjoint'' data sets, some with limited annotated instances.

\section{Data}
\label{sec:mtl_mentalhealth:data}

\renewcommand{\arraystretch}{1.25}
\begin{table*}
\scriptsize
\begin{tabular}{r||r|r|r|r|r|r|r|r|r||r|r|}
\cline{2-12}
& \rot{\sc neurotypical}	 & \rot{\sc anxiety}	 & \rot{\sc depression}	 & \rot{\sc suicide attempt}	 & \rot{\sc eating}	 & \rot{\sc schizophrenia}	 & \rot{\sc panic}	 & \rot{\sc PTSD}	 & \rot{\sc bipolar}  & \rot{\sc labeled male} & \rot{\sc labeled female} \\\hline
{\sc neurotypical}	 & 4820	 & 	\multicolumn{8}{c||}{~}  & - & - \\\cline{2-3}\cline{11-12}
{\sc anxiety}	 & 0	 & 2407	 	 & 	\multicolumn{7}{c||}{~}& 47 & 184\\\cline{2-4}\cline{11-12}
{\sc depression}	 & 0	 & 1148	 & 1400	  & 	\multicolumn{6}{c||}{~} & 54 & 158\\\cline{2-5}\cline{11-12}
{\sc suicide attempt}	 & 0	 & 45	 & 149	 & 1208	& 	 \multicolumn{5}{c||}{~}& 186 & 532\\\cline{2-6}\cline{11-12}
{\sc eating}	 & 0	 & 64	 & 133	 & 45	 & 749	 & 	 \multicolumn{4}{c||}{~}& 6 & 85 \\\cline{2-7}\cline{11-12}
{\sc schizophrenia}	 & 0	 & 18	 & 41	 & 2	 & 8	 & 349	 & \multicolumn{3}{c||}{~}& 2 & 4 \\\cline{2-8}\cline{11-12}
{\sc panic}	 & 0	 & 136	 & 73	 & 4	 & 2	 & 4	 & 263	 & 	\multicolumn{2}{c||}{~} &2 & 18 \\\cline{2-9}\cline{11-12}
{\sc PTSD}	 & 0	 & 143	 & 96	 & 14	 & 16	 & 14	 & 22	 & 191		&& 8 & 26 \\\cline{2-10}\cline{11-12}
{\sc bipolar} & 0	 & 149	 & 120	 & 22	 & 22	 & 49	 & 14	 & 25	 & 234	& 10 & 39 \\\cline{2-12}
\end{tabular}
\caption{Frequency and comorbidity across mental health conditions.}
\label{mtl_mentalhealth:tab:data}
\end{table*}

We train models on a union of multiple Twitter user datasets: 1) users identified as having anxiety, bipolar disorder, depression, panic disorder, eating disorder, PTSD, or schizophrenia \parencite{coppersmith2015adhd}, 2) those who had attempted suicide \parencite{coppersmith2015quantifying}, and 3) those identified as having either depression or PTSD from the 2015 Computational Linguistics and Clinical Psychology Workshop shared task \parencite{W15-1204}, along with neurotypical gender-matched controls (Twitter users not identified as having a mental condition).  Users were identified as having one of these conditions if they stated explicitly they were diagnosed with this condition on Twitter (verified by a human annotator), and the data was pre-processed to remove direction indications of the condition.  \textcite{coppersmith2015quantifying} describes how self-identifying tweets were stripped from the data as a preprocessing step. For a subset of 1,101 users, we also manually-annotate gender.  The final dataset contains 9,611 users in total, with an average of 3,521 tweets per user.  The number of users with each condition is included in Table \ref{mtl_mentalhealth:tab:data}. Users in this joined dataset may be tagged with multiple conditions, thus the counts in this table do not sum to the total number of users.

We use the entire Twitter history of each user as input to the model, and split it into character 1-to-5-grams, which have been shown to generaliize better than words for many Twitter text classification tasks \parencite{mcnamee2004character,coppersmith2015adhd}.  For instance, a character n-gram representation of a document is less sensitive to typographical errors than token n-gram features -- although a single mistyped character will yield an entirely different token, the misspelled word will share most of its character unigram features with the correctly spelled word. We compute the relative frequency of the 5,000 most frequent $n$-gram features for $n \in \{1,2,3,4,5\}$ in our data, and then feed this as input to all models.  This input representation is common to all models, allowing for fair comparison.

\section{Experiments}
\label{sec:mtl_mentalhealth:experiments}

Our task is to predict suicide attempt and mental conditions for each of the users in these data. 
We evaluate three classes of models: baseline logistic regression over character $n$-gram features (\LR), feed-forward multilayer perceptrons trained to predict each task separately (\STL), and feed-forward multi-task models trained to predict a set of conditions simultaneously (\MTL).  We experiment with a feed-forward network against independent logistic regression models as a way to directly test the hypothesis that neural classifiers can improve mental condition prediction, particularly when regularized with \MTL.

We also perform ablation experiments to see which subsets of tasks help us learn an \MTL{} model that predicts a particular mental condition best.  For all experiments, data were divided into five equal-sized folds, three for training, one for tuning, and one for test (we report performance on this fold).

All our models are implemented in Keras\footnote{\url{http://keras.io/}} with Theano backend and GPU support.  We train the models for a total of up to 15,000 epochs, using mini-batches of 256 eaxmples each. Training time on all five training folds ranged from one to eight hours on a machine with Tesla K40M.

\subsection{Evaluation Setup} 
\label{subsec:mtl_mentalhealth:evaluation_setup}

In clinical settings, we are interested in minimizing the number of false positives, i.e., incorrect diagnoses, which can cause undue stress to the patient.  We are thus interested in bounding this quantity. To evaluate the performance, we plot the false positive rate (FPR) against the true positive rate (TPR). This gives us a receiver operating characteristic (ROC) curve, allowing us to inspect the performance of each model on a specific task at any level of FPR.

While the ROC gives us a sense of how well a model performs at a fixed true positive rate, it makes it difficult to compare the individual tasks at a low false positive rate, which is also important for clinical application. We therefore report two more measures: the area under the ROC curve (AUC) and TPR performance at FPR=0.1 (TPR@FPR=0.1).  We do not compare our models to a majority baseline model, since this model would achieve an expected AUC of 0.5 for all tasks, and F-score and TPR@FPR=0.1 of 0 for all mental conditions -- users exhibiting a condition are the minority, meaning a majority baseline classifier would achieve zero recall.

\subsection{Optimization and Model Selection}
\label{subsec:mtl_mentalhealth:optimization}

\begin{figure*}
  \begin{center}
		\includegraphics[width=0.3\textwidth,trim={0 12.1cm 0.1cm 0},clip]{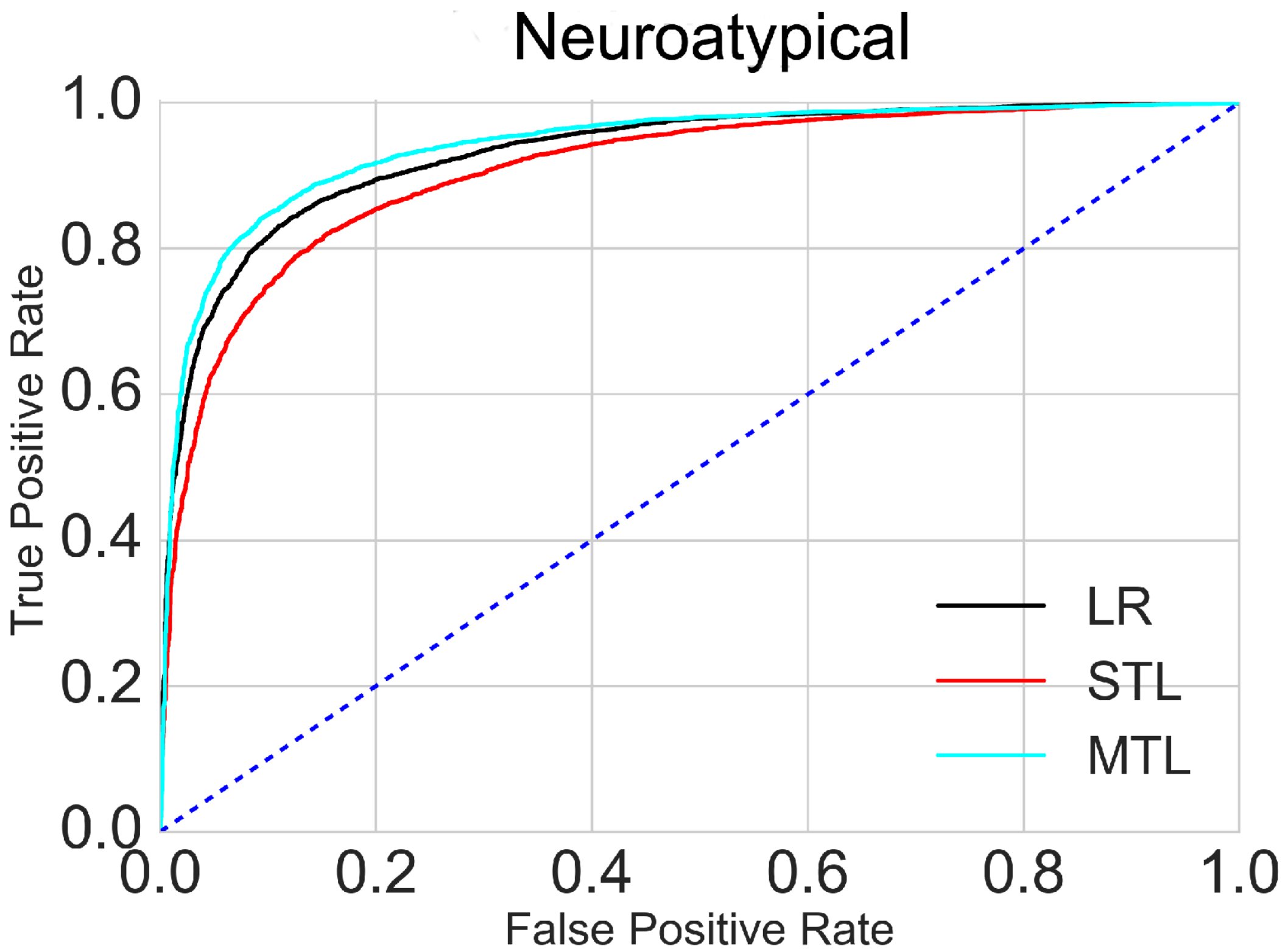} \hspace{1em} 		\includegraphics[width=0.3\textwidth,trim={0 12.25cm 0.1cm 0},clip]{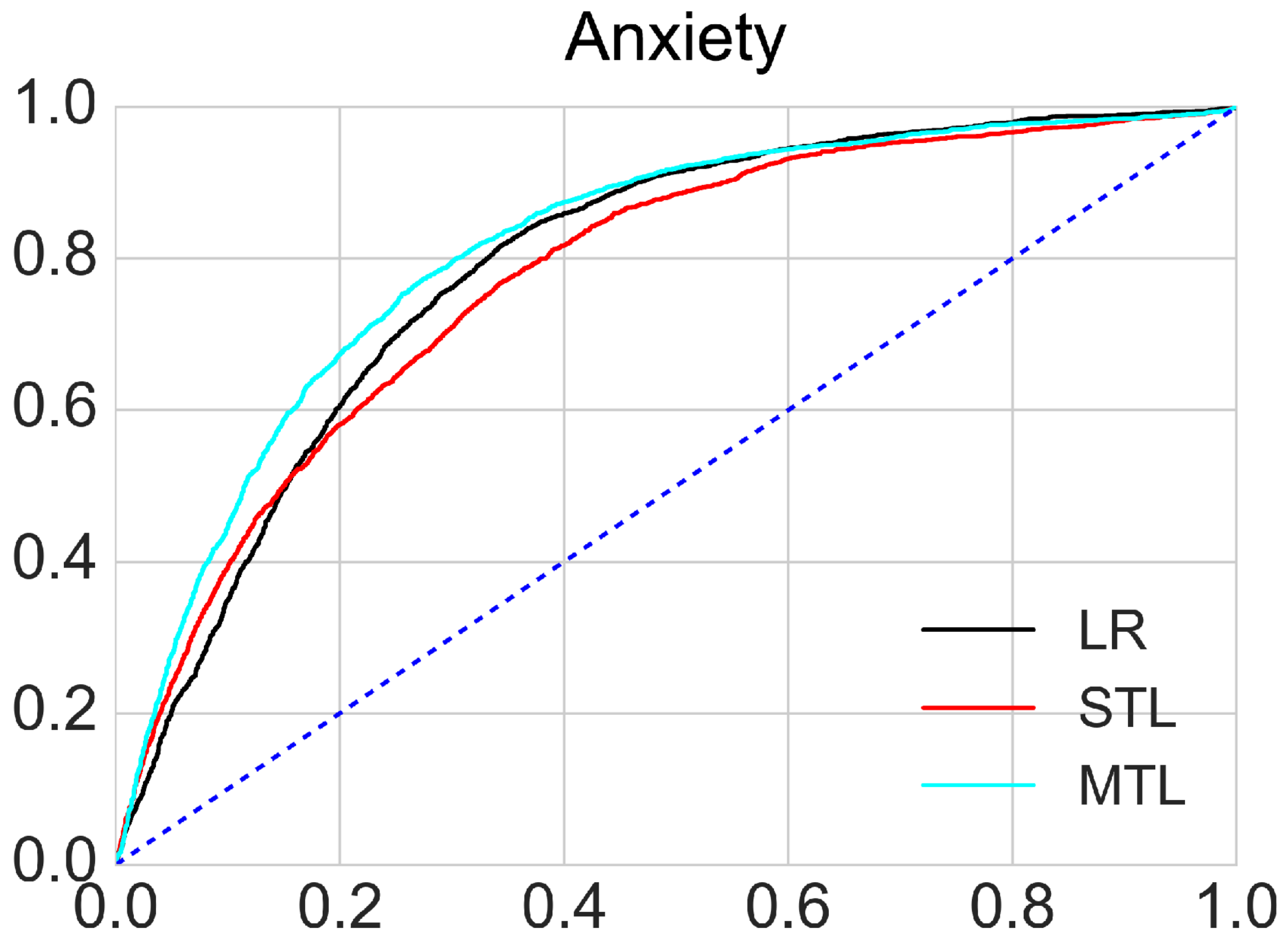}
\hspace{1em}		\includegraphics[width=0.3\textwidth,trim={0 12.25cm 0.1cm 0},clip]{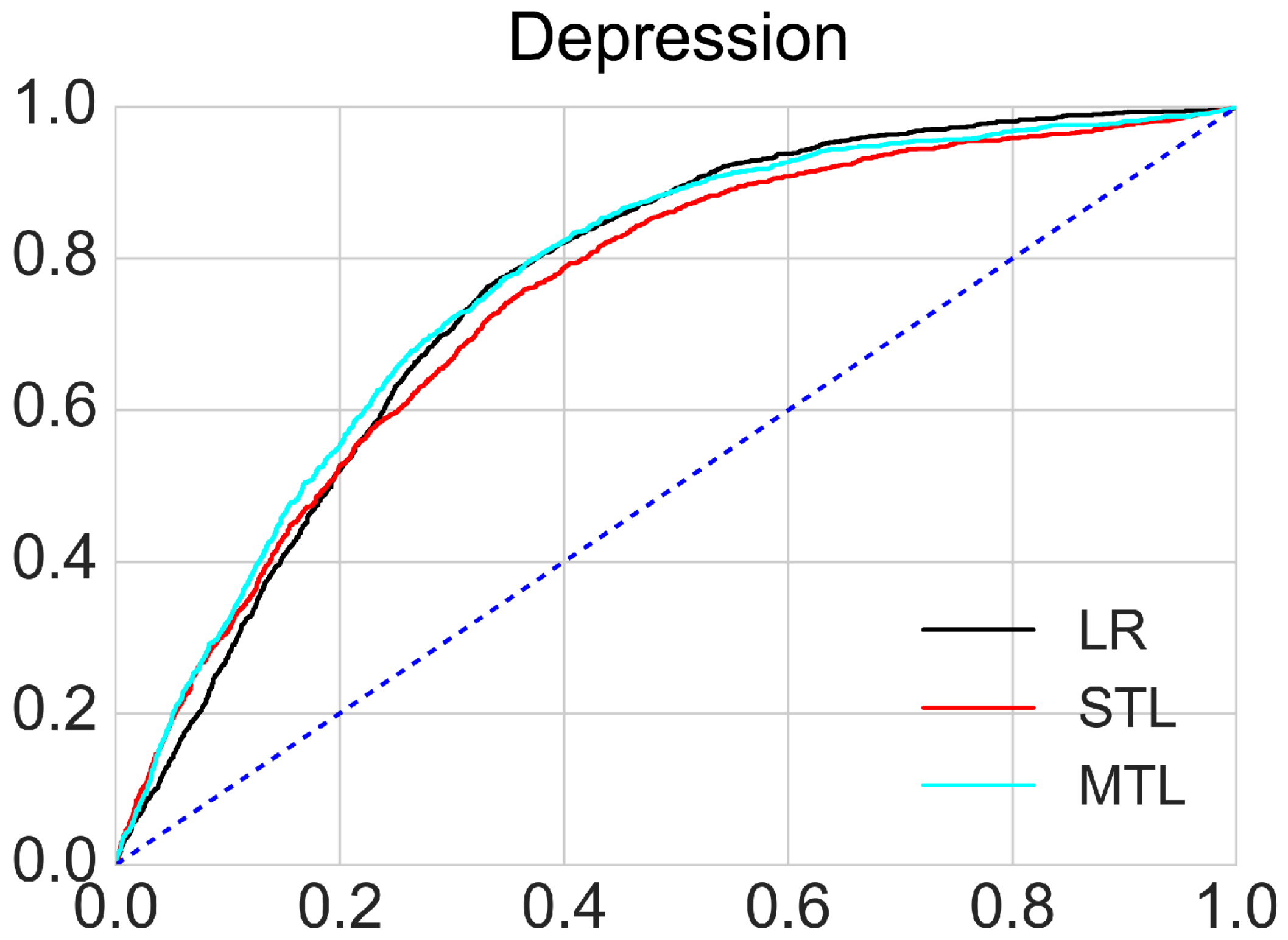}\\\vspace{1em}
		\includegraphics[width=0.3\textwidth,trim={0 12.25cm 0.1cm 0},clip]{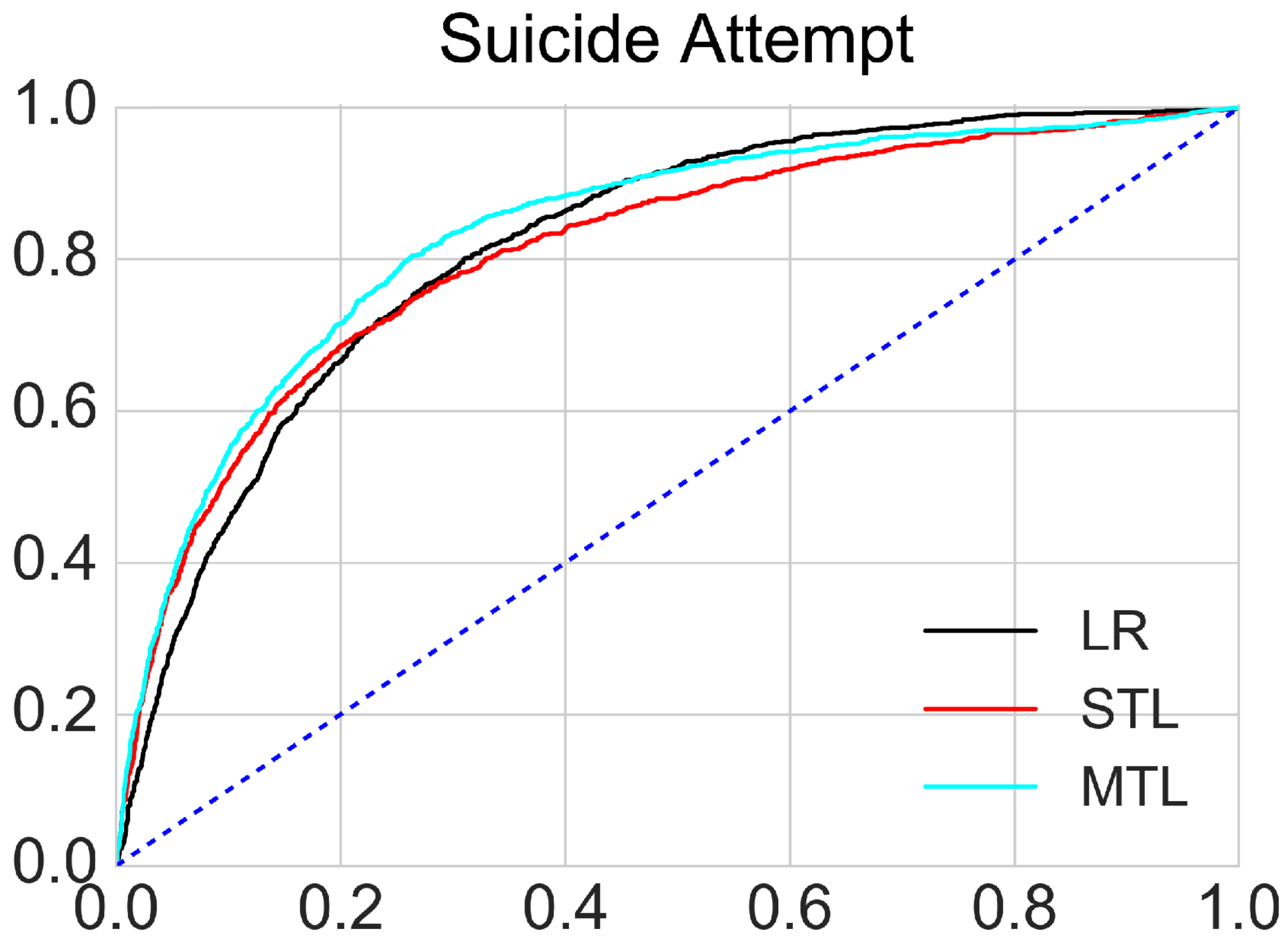}
        \hspace{1em}
		\includegraphics[width=0.3\textwidth,trim={0 12.25cm 0.1cm 0},clip]{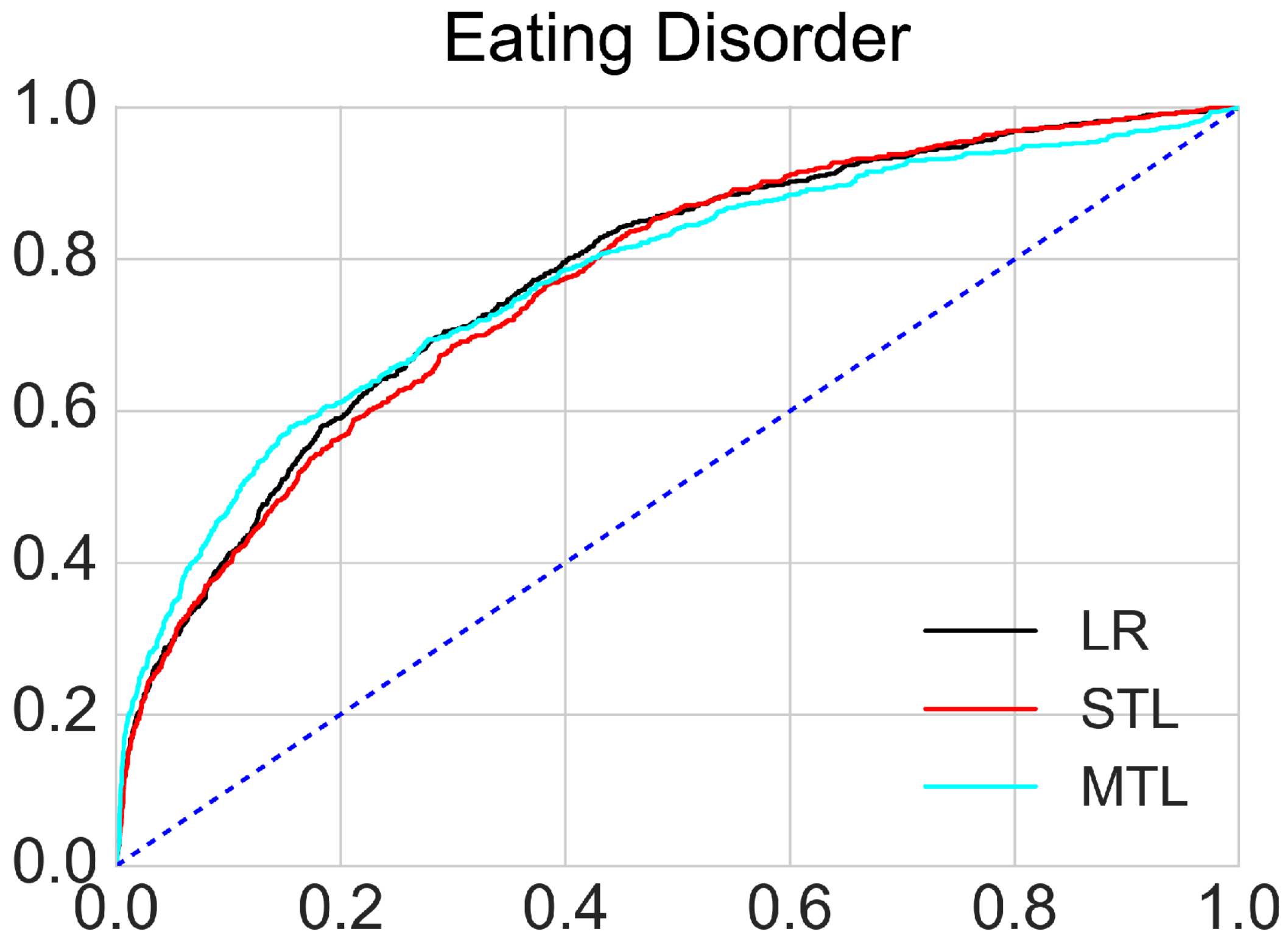}
        \hspace{1em}
		\includegraphics[width=0.3\textwidth,trim={0 12.25cm 0.1cm 0},clip]{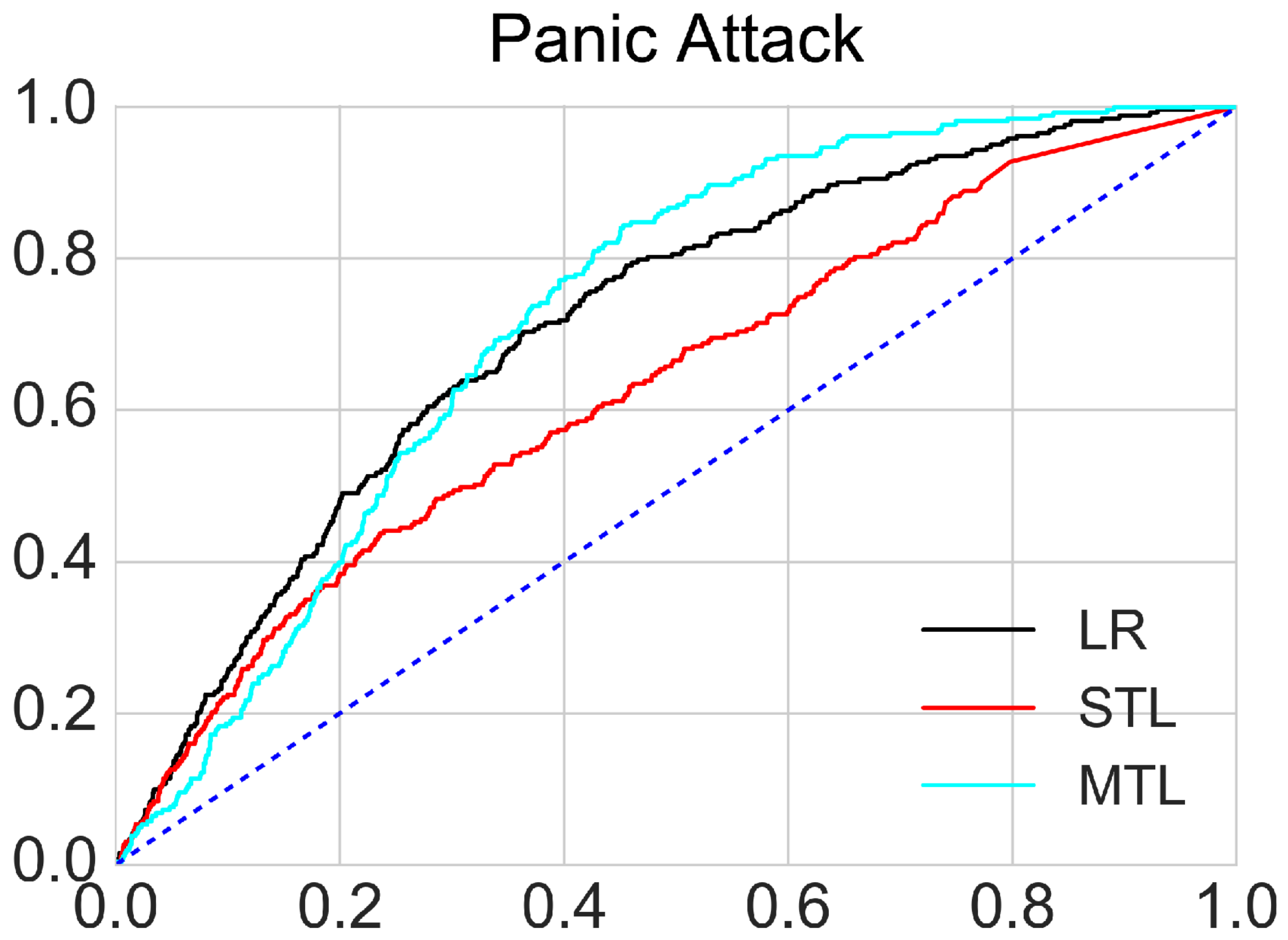}\\\vspace{1em}
		\includegraphics[width=0.3\textwidth,trim={0 12.25cm 0.1cm 0},clip]{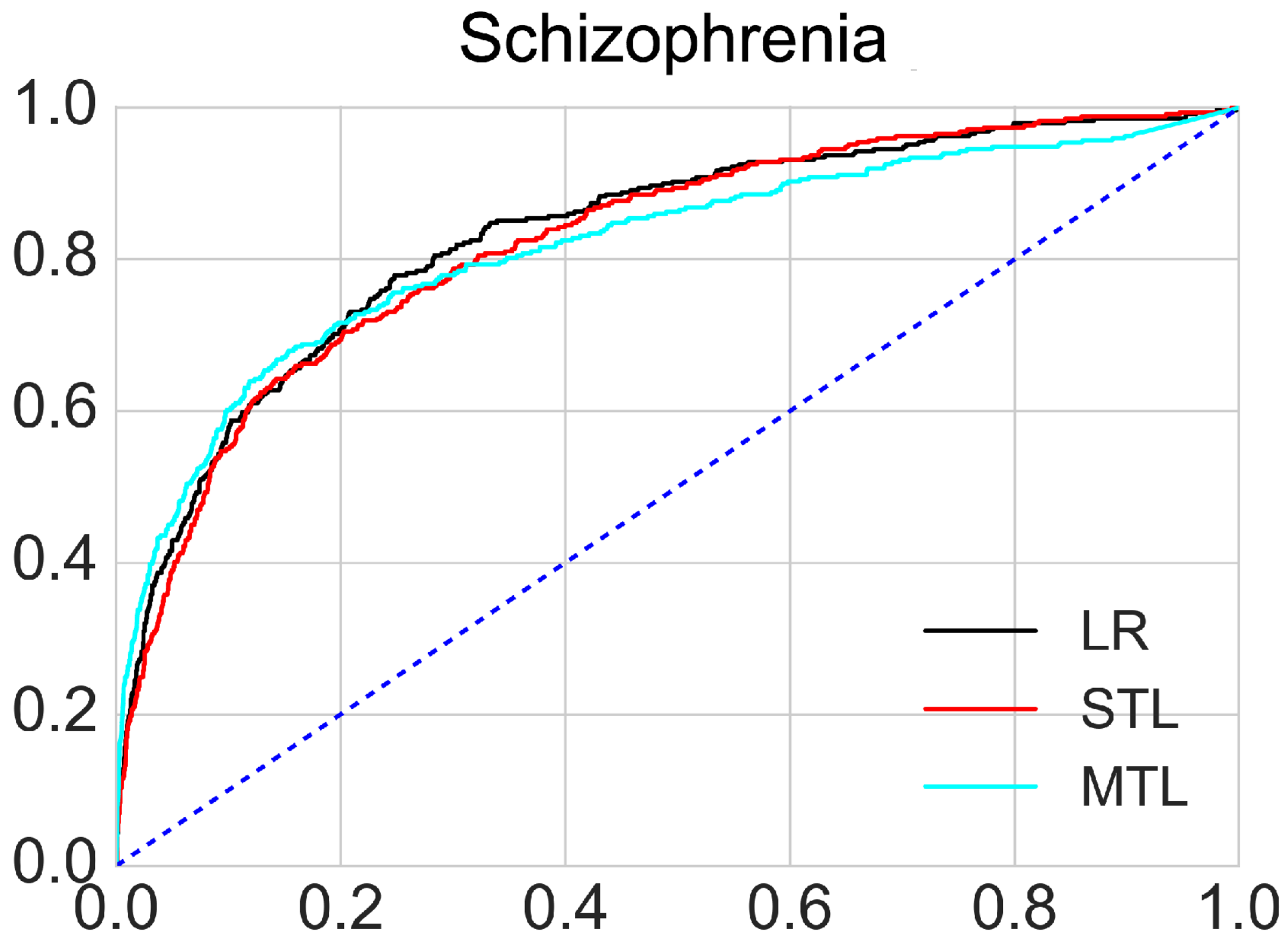}
        \hspace{1em}
		\includegraphics[width=0.3\textwidth,trim={0 12.25cm 0.1cm 0},clip]{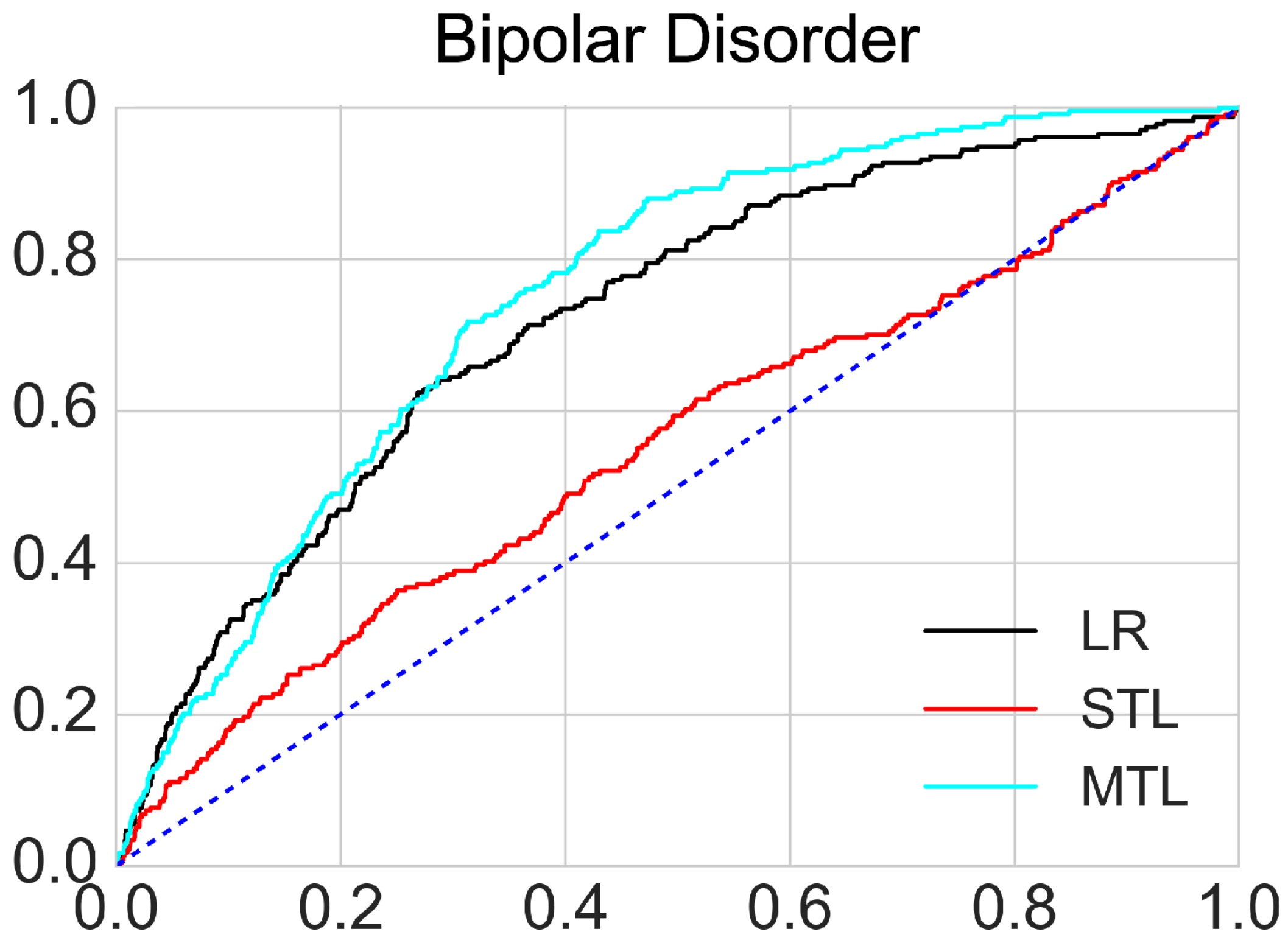}
        \hspace{1em}
		\includegraphics[width=0.3\textwidth,trim={0 12.25cm 0.1cm 0},clip]{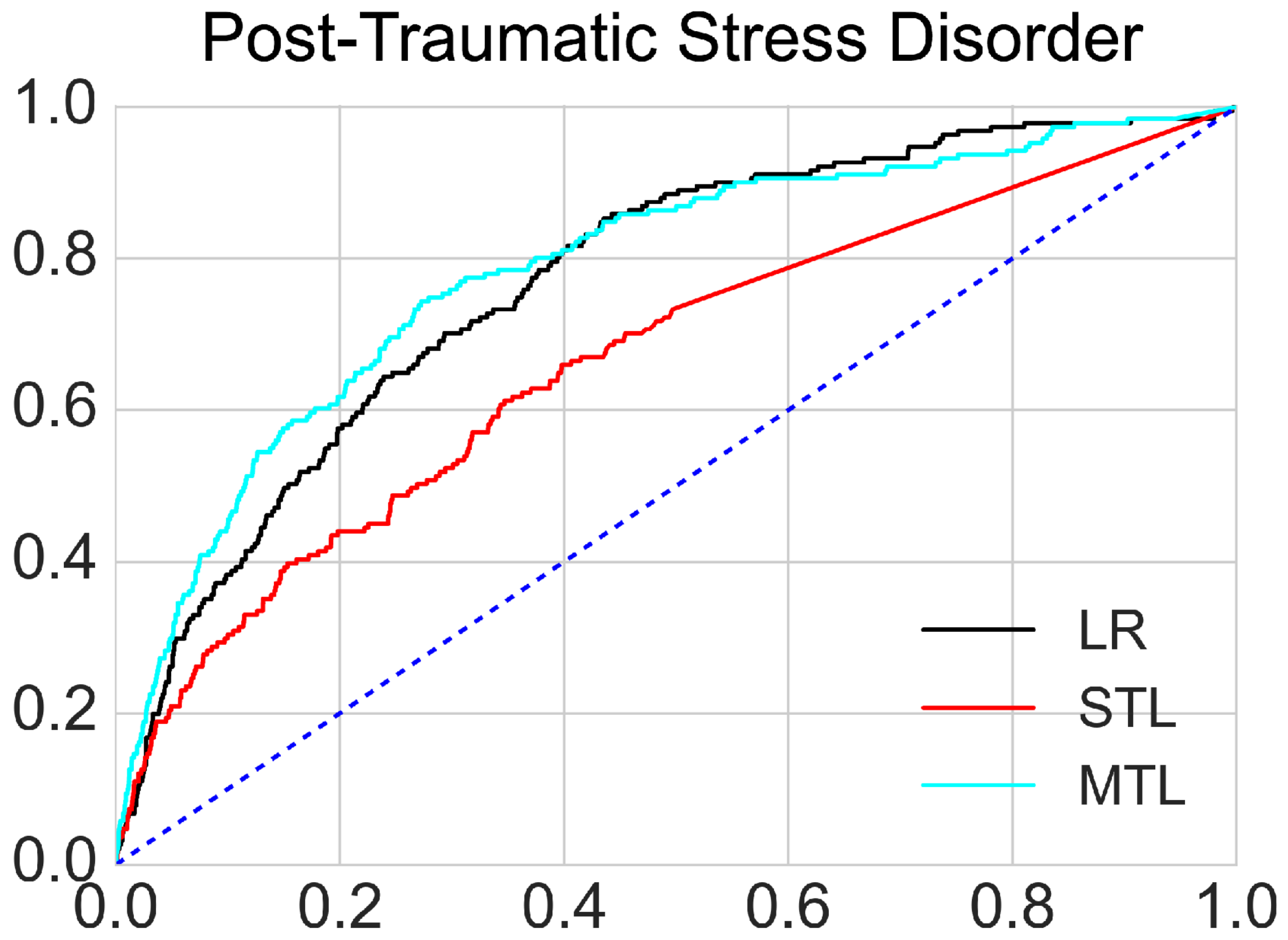}
		\captionof{figure}{ROC curves for predicting each mental health condition. The precision (diagnosed, correctly labeled) is on the $y$-axis, while the proportion of false alarms (control users mislabeled as having been diagnosed) is on the $x$-axis. Chance performance is indicated by the blue dotted diagonal line. \label{pic:mtl_mentalhealth:roc_curves}}
	\end{center}
\end{figure*}

Even in a relatively simple neural model, there are a number of hyperparameters that can (and have to) be tuned to achieve good performance. We perform a line search for every model we use, sweeping over $\ell_2$ regularization and hidden layer width.  We select the best model based on the development loss. Figure \ref{pic:mtl_mentalhealth:roc_curves} shows the performance on the corresponding test sets (plot smoothed by rolling mean of 10 for visibility).

In our experiments, we sweep over the $\ell_2$ regularization constant applied to all weights in $\{10^{-4}, 10^{-3}, 10^{-2}, 0.1, 0.5, 1.0, 5.0, 10.0\}$, and hidden layer width (same for all layers in the network) in $\{16, 32, 64, 128, 256, 512, 1024, 2048\}$.  We fix the mini-batch size to 256, and 0.05 dropout rate on the input layer.  Choosing a small mini-batch size and the model with lowest development loss helps to account for overfitting.

We train each model for 5,000 iterations, jointly updating all weights in our models.  After this initial joint training, we select each task separately, and only update the task-specific layers of weights independently for another 1,000 iterations (selecting the set of weights achieving lowest development loss for each task individually). Weights are updated using mini-batch Adagrad \parencite{adagrad} -- this converges more quickly than other optimization schemes we initially experimented with.  We evaluate the tuning loss every 10 epochs, and select the model with the lowest tuning loss.

\section{Results}
\label{sec:mtl_mentalhealth:results}

\begin{figure}
	\begin{center}
		\includegraphics[width=0.7\textwidth]{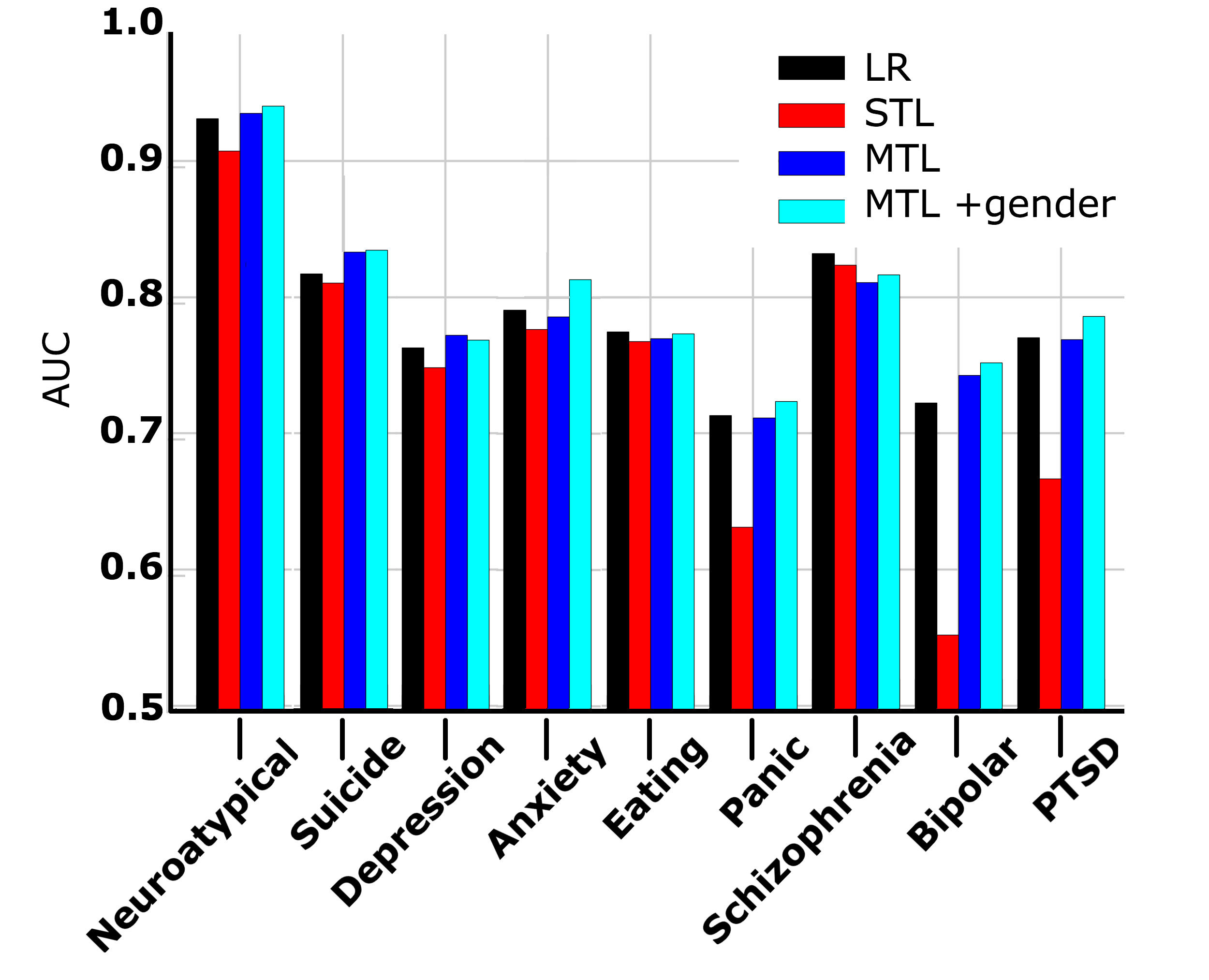}
		\captionof{figure}{AUC for different main mental health prediction tasks. \label{pic:mtl_mentalhealth:auc}}
	\end{center}
\end{figure}

\begin{figure}
	\begin{center}
		\includegraphics[width=0.7\textwidth]{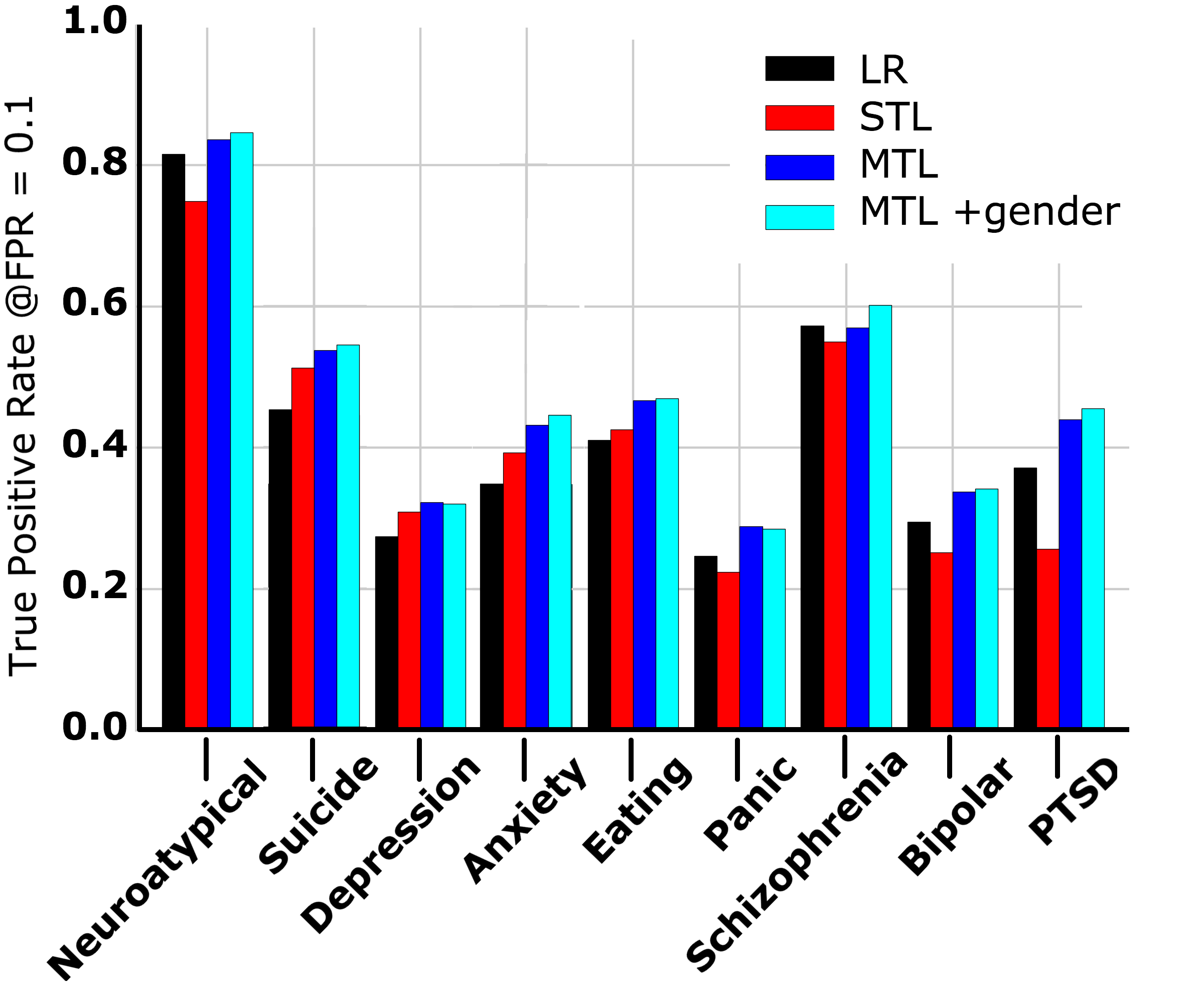}
		\captionof{figure}{TPR at 0.10 FPR for different main mental health prediction tasks. \label{pic:mtl_mentalhealth:tpr}}
	\end{center}
\end{figure}

\begin{figure*}[ht!]
  \begin{center}
		\includegraphics[width=0.3\textwidth,trim={0 12.25cm 0.1cm 0},clip]{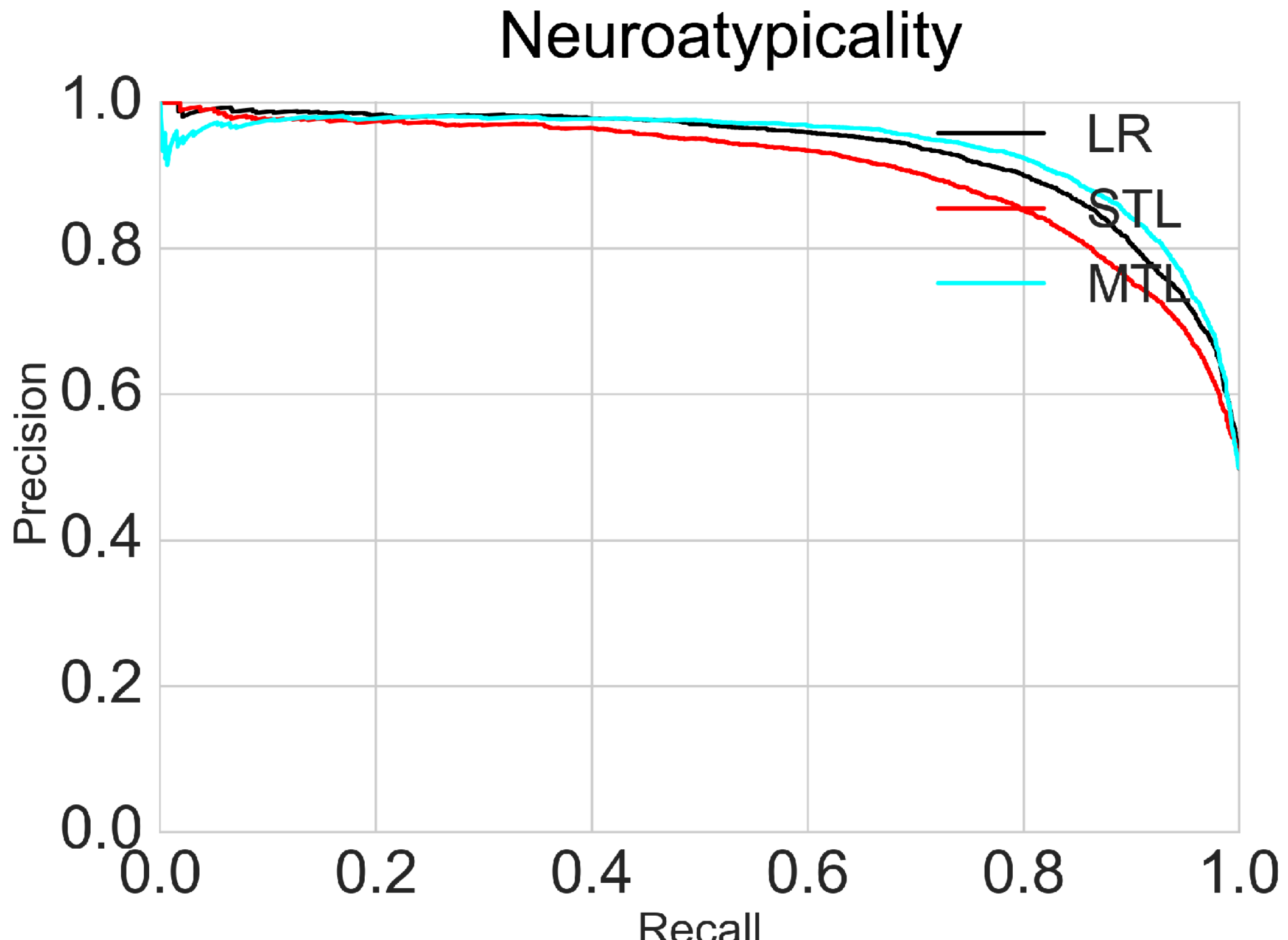}
\hspace{1em}	\includegraphics[width=0.3\textwidth,trim={0 12.25cm 0.1cm 0},clip]{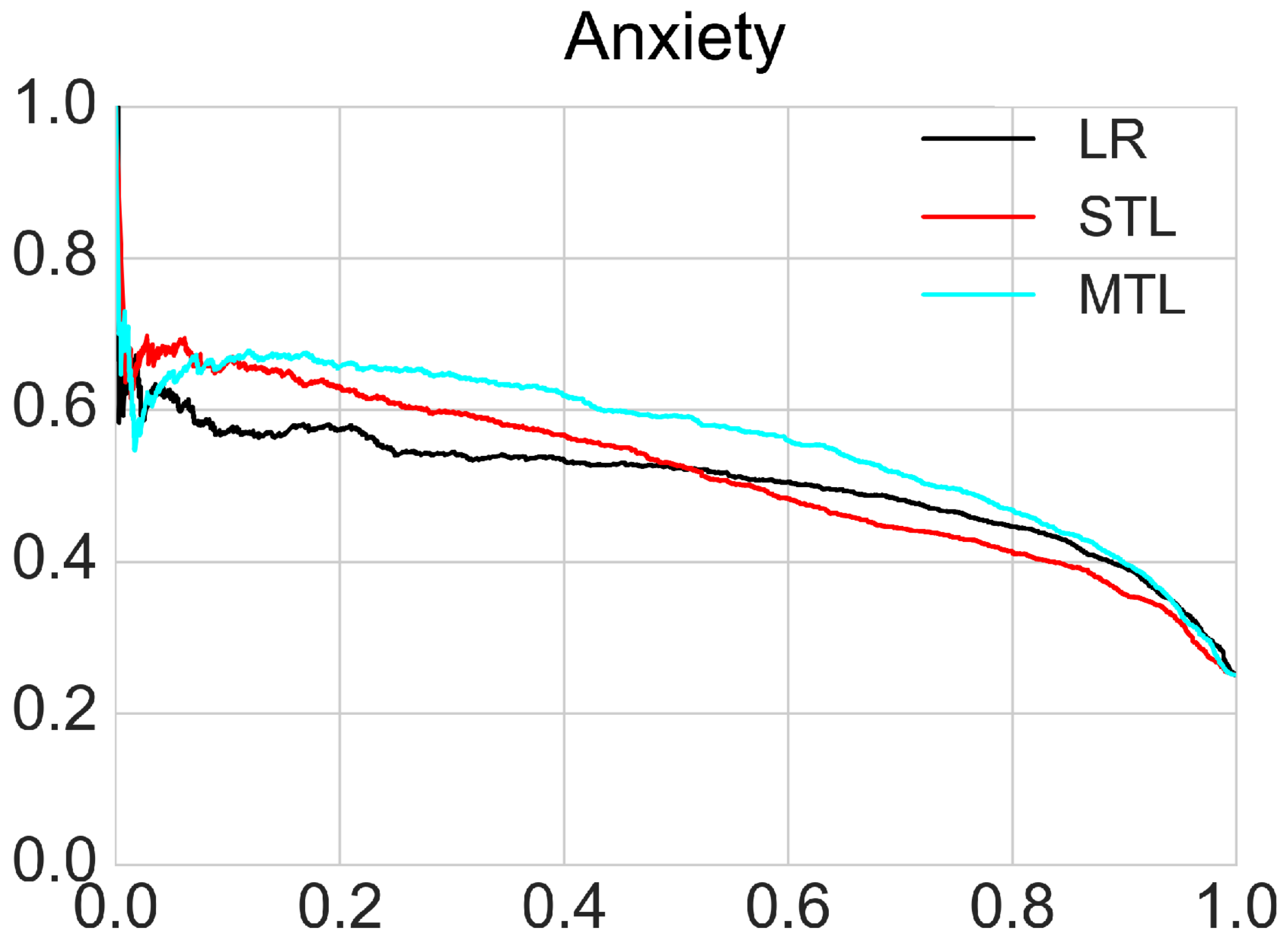}
\hspace{1em}	\includegraphics[width=0.3\textwidth,trim={0 12.25cm 0.1cm 0},clip]{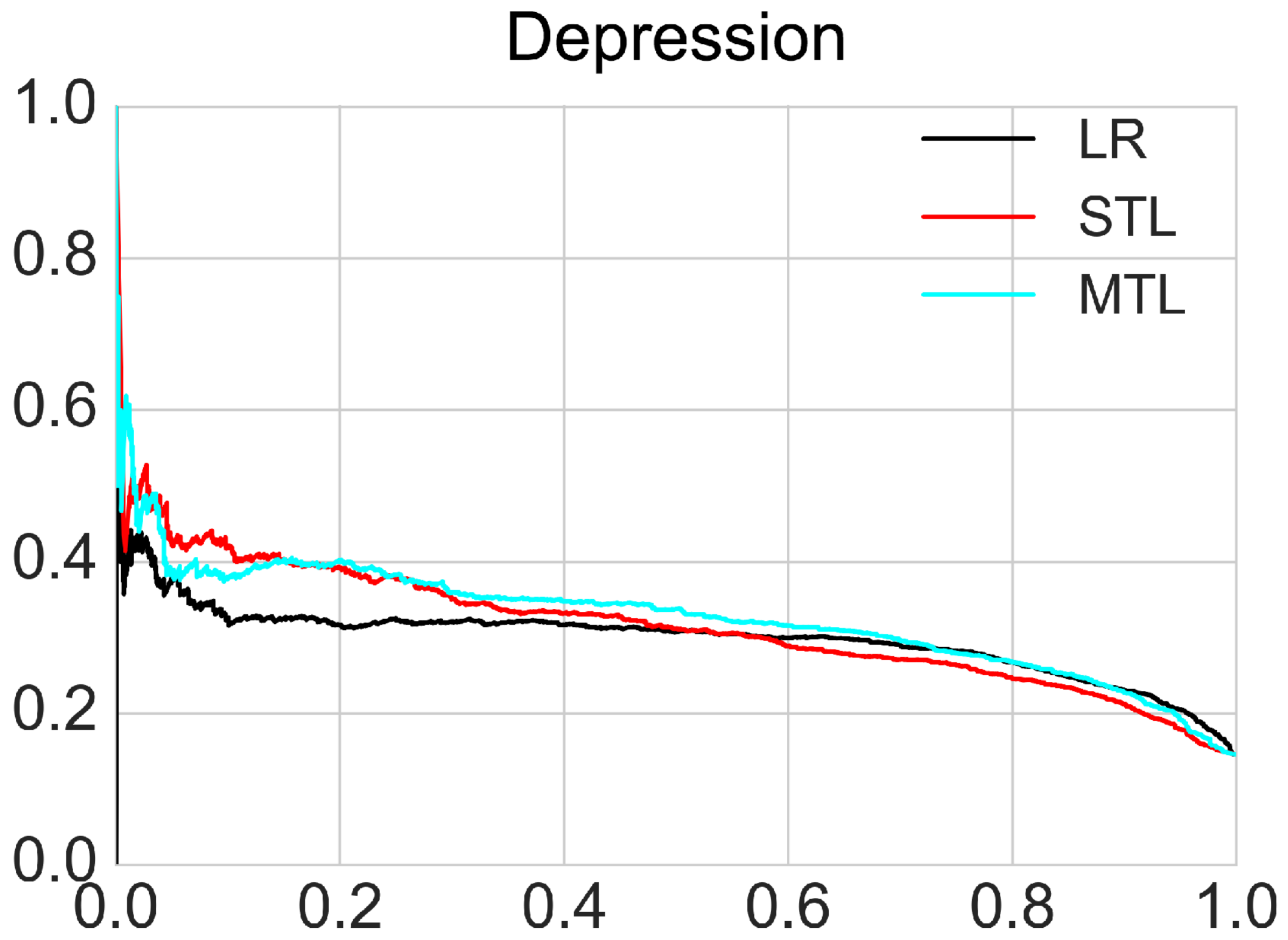} \\
\vspace{1em}
\includegraphics[width=0.3\textwidth,trim={0 12.25cm 0.1cm 0},clip]{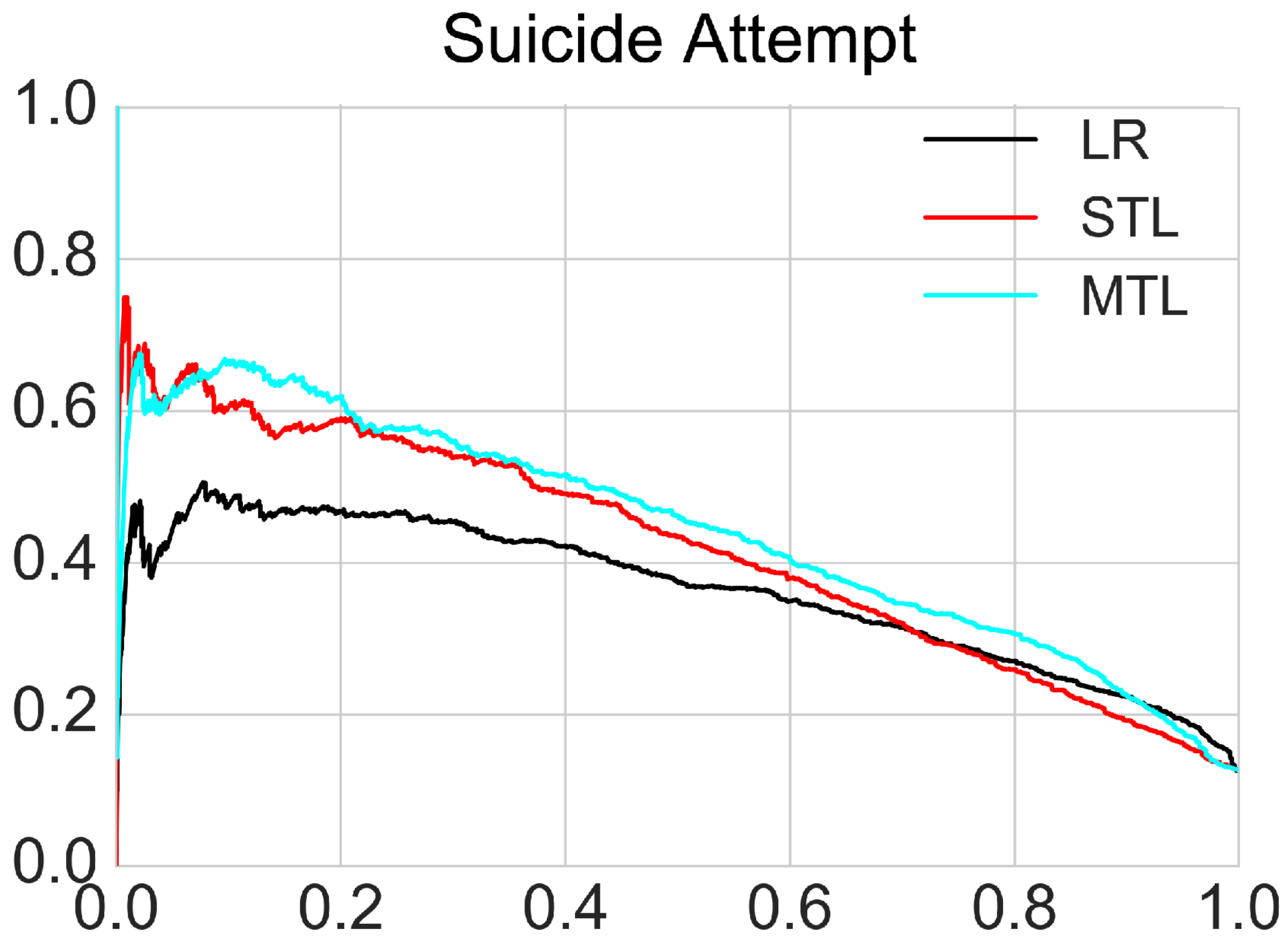}
	\hspace{1em}	\includegraphics[width=0.3\textwidth,trim={0 12.25cm 0.1cm 0},clip]{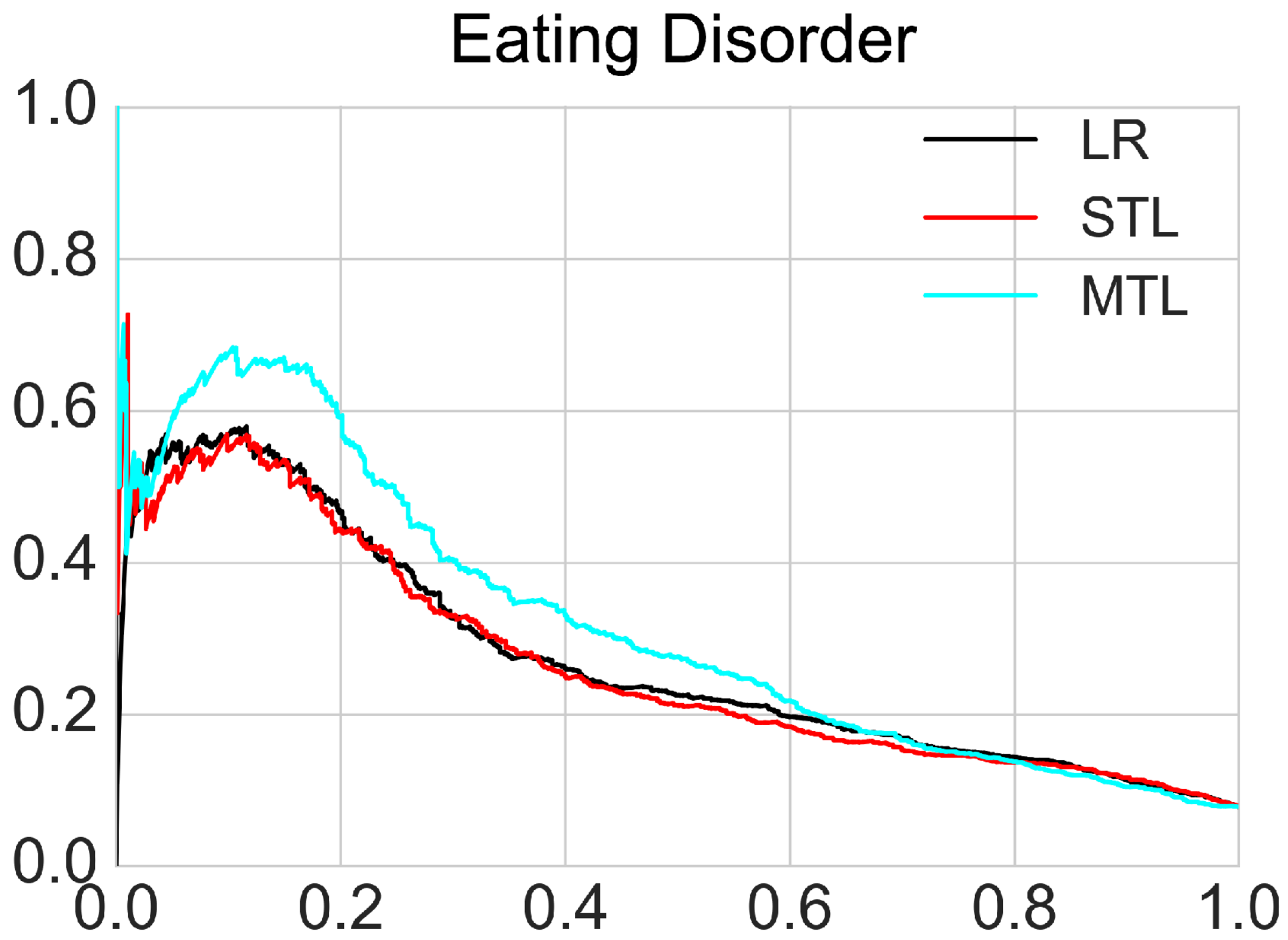}
	\hspace{1em}	\includegraphics[width=0.3\textwidth,trim={0 12.25cm 0.1cm 0},clip]{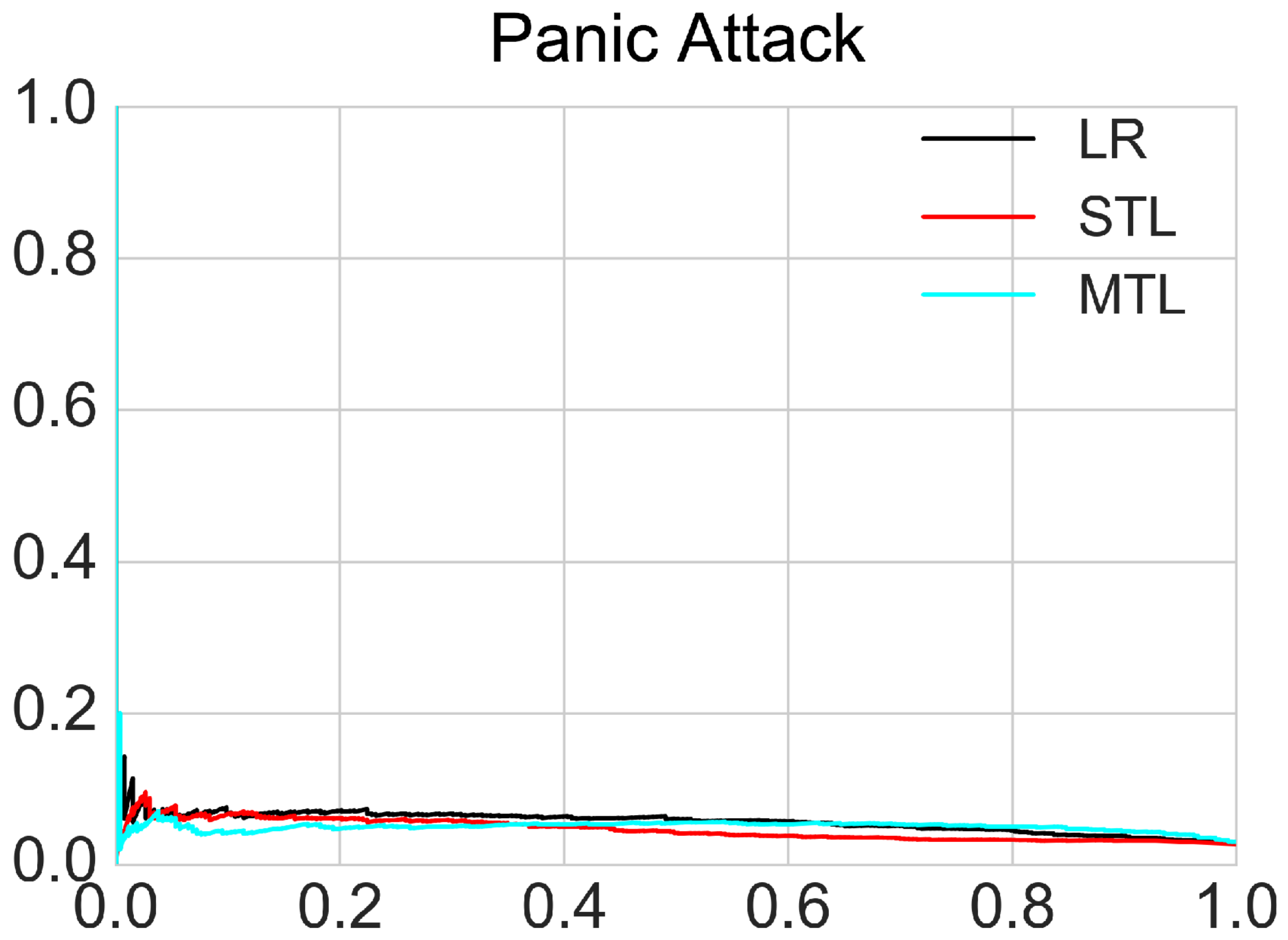}\\
        
\vspace{1em}
\includegraphics[width=0.3\textwidth,trim={0 12.25cm 0.1cm 0},clip]{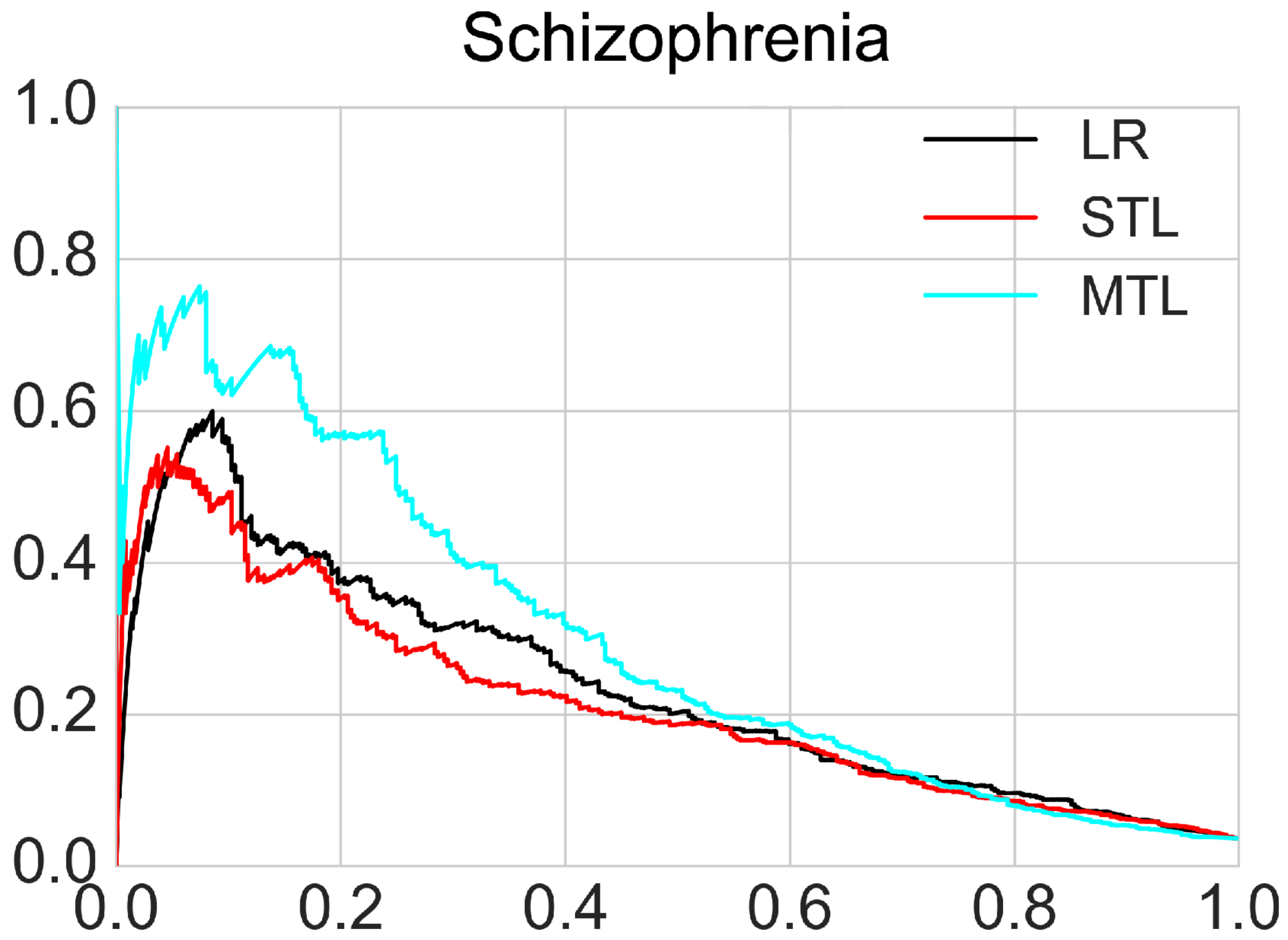} \hspace{1em}
		\includegraphics[width=0.3\textwidth,trim={0 12.25cm 0.1cm 0},clip]{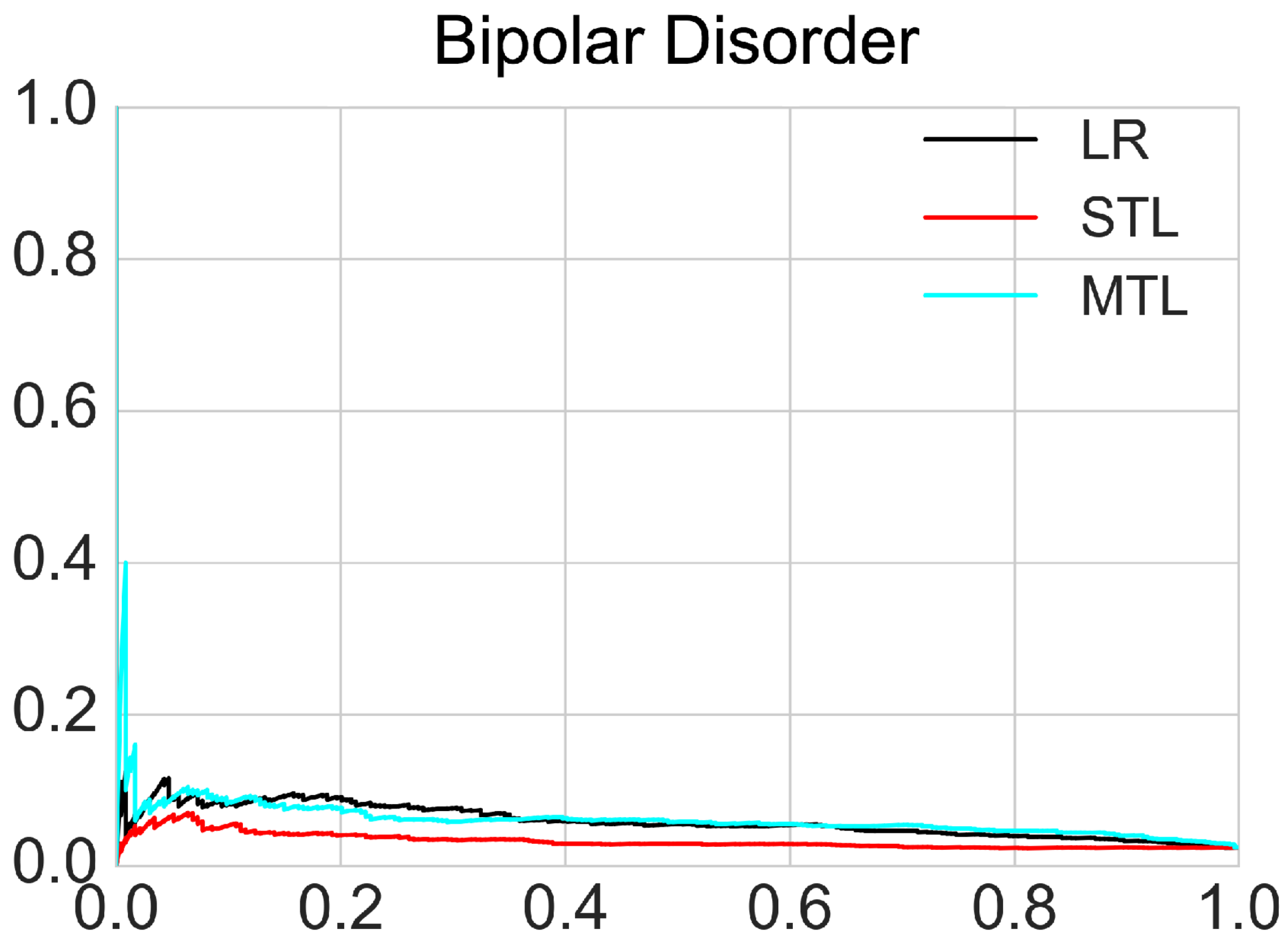}
      \hspace{1em}  \includegraphics[width=0.3\textwidth,trim={0 12.25cm 0.1cm 0},clip]{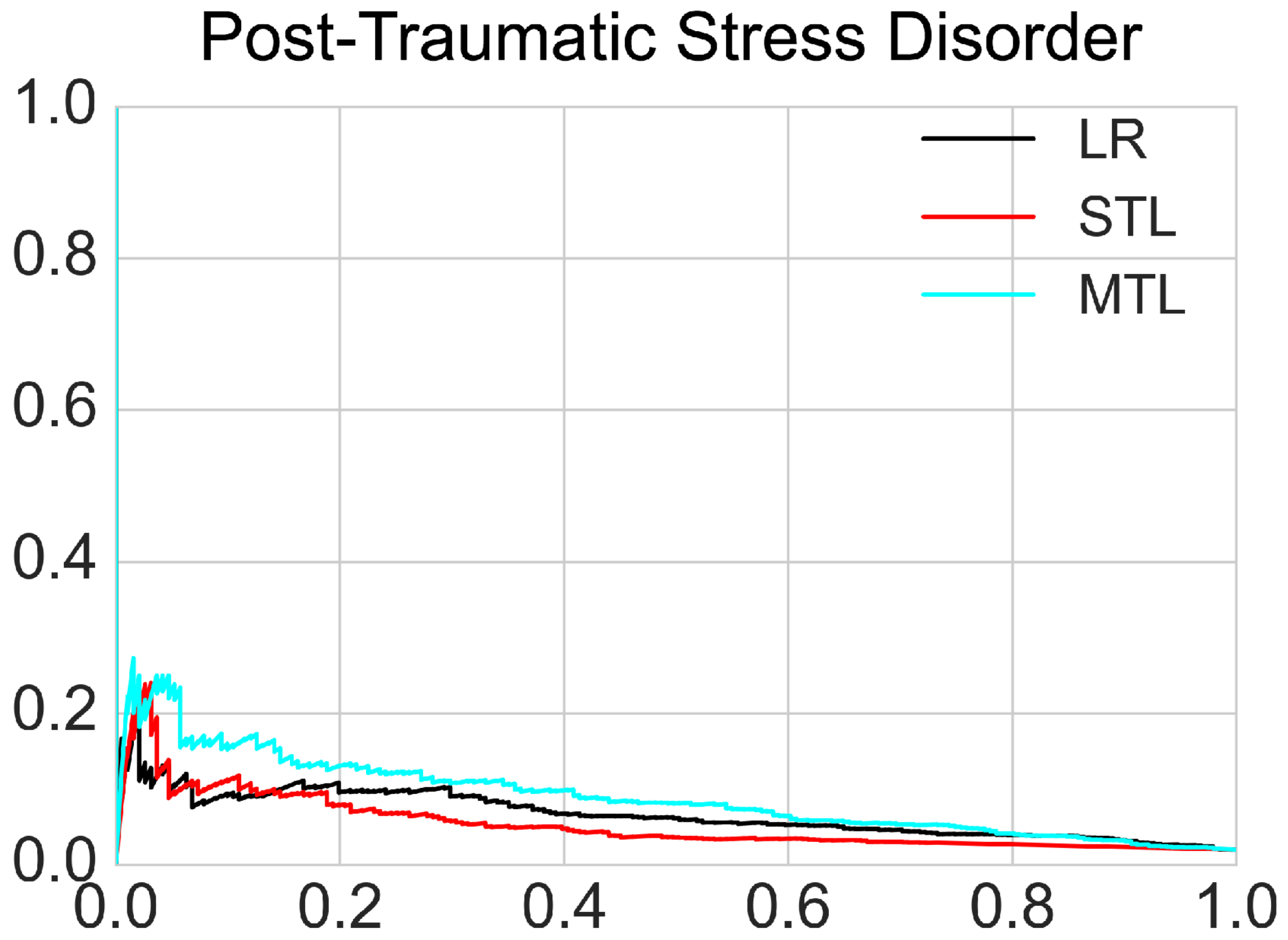}
		\captionof{figure}{Precision-recall curves for predicting each mental health condition.
      \label{pic:mtl_mentalhealth:precrec_curves}}
	\end{center}
\end{figure*}

Figure \ref{pic:mtl_mentalhealth:auc} shows the AUC-score of each model for each task separately, and Figure \ref{pic:mtl_mentalhealth:tpr} the true positive rate at a low false positive rate of 0.1.  Precision-recall curves for model/task are in Figure \ref{pic:mtl_mentalhealth:precrec_curves}.  \STL{} is a multilayer perceptron with two hidden layers (with a similar number of parameters as the proposed \MTL{} model).  The \MTL +gender and \MTL{} models predict all tasks simultaneously, but are only evaluated on the main respective task.


Both AUC and TPR (at FPR=0.1) demonstrate that single-task models do not perform nearly as well as multi-task models or logistic regression.  This is likely because the neural networks learned by \STL{} cannot be guided by the inductive bias provided by \MTL{} training.  Note, however, that \STL{} and \MTL{} are often perform comparably in terms of F1-score, where false positives and false negatives are equally weighted.  

Multi-task suicide predictions reach an AUC of 0.848, and predictions for anxiety and schizophrenia are not far behind (Figure \ref{pic:mtl_mentalhealth:auc}).  Interestingly however, schizophrenia stands out as being the only condition to be best predicted with a single-task model.  MTL models show improvements over STL and LR models for predicting suicide, neuroatypicality, depression, anxiety, panic, bipolar disorder, and PTSD.  The inclusion of gender in the MTL models leads to direct gains over an LR baseline in predicting anxiety disorders: anxiety, panic, and PTSD.

Figure \ref{pic:mtl_mentalhealth:tpr} illustrates the \emph{true positive rate} -- that is, how many cases of mental health conditions that we correctly predict -- given a low \emph{false positive rate} -- that is, a low rate of predicting people have mental health conditions when they do not.  This is particularly useful in clinical settings, where clinicians seek to minimize over-diagnosing, especially when false positives incur an unnecessary, great treatment and emotional cost.  In this setting, MTL leads to the best performance across the board, for all tasks under consideration: neuroatypicality, suicide, depression, anxiety, eating, panic, schizophrenia, bipolar disorder, and PTSD.  Including gender in MTL further improves performance for neuroatypicality, suicide, anxiety, schizophrenia, bipolar disorder, and PTSD.

\subsection{Comorbid Conditions Improve Prediction Accuracy}
\label{subsec:mtl_mentalhealth:comorbid}

We find that the prediction of the conditions with the least amount of data -- \emph{bipolar disorder} and \emph{PTSD} -- are significantly improved by having the model also predict comorbid conditions with substantially more data:  \emph{depression} and \emph{anxiety}.  We are able to increase the AUC for predicting PTSD to 0.786 by \MTL, from 0.770 by \LR, whereas \STL{} fails to perform as well with an AUC of 0.667.  Similarly for predicting bipolar disorder (\MTL:0.723, \LR:0.752, \STL:0.552) and panic attack (\MTL:0.724, \LR:0.713, \STL:0.631).

These differences in AUC are significant at $p=0.05$ according to bootstrap sampling tests with 5,000 samples.  The wide difference between MTL and STL can be explained in part by the increased feature set size -- MTL training may, in this case, provide a form of regularization that \STL{} cannot exploit.  Further, modeling the common mental health conditions with the most data (depression and anxiety) helps improve performance in predicting rarer conditions comorbid with these common health conditions.  This provides evidence that an MTL model can help in predicting elusive conditions by using large data for common conditions, and a small amount of data for more rare conditions.

\subsection{Utility of Author Demographic Features}
\label{subsec:mtl_mentalhealth:demographics}

Figures \ref{pic:mtl_mentalhealth:auc} and \ref{pic:mtl_mentalhealth:tpr} both suggest that adding an author's demographic feature, such as gender, as an auxiliary task leads to more predictive models, even though the difference is not statistically significant for most tasks. This is consistent with the findings in previous work \parencite{Volkova:ea:2013exploring,hovy2015demographic}.
Interestingly, though, the \MTL{} model is worse at predicting gender itself. While this could be a direct result of data sparsity (recall that we have only a small subset annotated for gender), which could be remedied by annotating additional users for gender, this appears unlikely given the other findings of our experiments, where \MTL{} helped in specifically these sparse scenarios. 

However, \textcite{caruana1996algorithms} notes that not all tasks benefit from a \MTL{} setting in the same way, and that some tasks serve purely auxiliary roles. Here, gender prediction does not benefit from including mental conditions, but guides \MTL{} models to better predict other mental health conditioned.
In other words, predicting gender is qualitatively different from predicting mental health conditions: it seems likely that the signals for anxiety are much more similar to the ones for depression than for, say, being male, and can therefore add to detecting depression. However, the distinction between certain conditions does not add information for the distinction of gender.
The effect may also be due to the fact that these data were constructed with inferred gender (used to match controls), so there might be a degree of noise in the data.

\subsection{Selecting User Features as Auxiliary Tasks}
\label{subsec:mtl_mentalhealth:choosing_tasks}

Although \MTL{} tends to dominate \STL{} in our experiments, it is not clear whether modeling several tasks provide a beneficial inductive bias in \MTL{} models in general, or if there exist specific subsets of auxiliary tasks that are most beneficial for predicting suicide risk and related mental health conditions.   We perform ablation experiments by training \MTL{} models on  a subset of auxiliary tasks, and prediction for a single main task.  We focus on four conditions to predict well: suicide attempt, anxiety, depression, and bipolar disorder.  For each main task, we vary the auxiliary tasks we train the \MTL{} model with.  Since considering all possible subsets of tasks is combinatorially infeasible, we selected the following task subsets as auxiliary:

\begin{itemize}
\itemsep-0.5em
\item \emph{all}: all mental conditions along with gender
\item \emph{all conds}: all mental conditions, no gender
\item \emph{neuro}: only neurotypicality
\item \emph{neuro+mood}: neurotypicality, depression, and bipolar disorder (mood disorders)
\item \emph{neuro+anx}: neurotypicality, anxiety, and panic attack (anxiety conditions)
\item \emph{neuro+targets}: neurotypicality, anxiety, depression, suicide attempt, bipolar disorder
\item \emph{none}: no auxiliary tasks, equivalent to \STL{}
\end{itemize}

Table \ref{table:mtl_mentalhealth:ablation} shows AUC for the four prediction tasks with different subsets of auxiliary tasks.  Statistically significant improvements over the respective LR baselines are denoted by superscript.  Restricting the auxiliary tasks to a small subset tends to hurt performance for most tasks, with exception to \emph{bipolar}, which benefits from the prediction of depression and suicide attempt. All main tasks achieve their best performance using the full set of additional tasks as auxiliary.  This suggests that the biases induced by predicting different kinds of mental conditions are mutually beneficial -- e.g., multi-task models that predict suicide attempt may also be good at predicting anxiety.

Based on these results, we find it useful to think of \MTL{} with user features 
as a framework to leverage auxiliary tasks as regularization to effectively combat data paucity and less-than-trustworthy labels.  As we have demonstrated, this may be particularly useful when predicting mental health conditions and suicide risk.

\renewcommand{\arraystretch}{1.25}
\begin{table}[t]
\small
\begin{center}
\begin{tabular}{@{}r||l|l|l|l|@{}}
\multirow{5}{5em}{\bf Auxiliary Tasks} & \multicolumn{4}{c}{\bf{Main Task}} \\\cline{2-5}
 & \multicolumn{4}{c}{\vspace{-.5em}} \\
& \rot{\sc anxiety} & \rot{\multirow{2}{*}{\sc bipolar}} & \rot{\multirow{2}{*}{\sc depression}} & \rot{\multirow{2}{3em}{\sc suicide attempt}}\\
 & \multicolumn{4}{c}{\vspace{-.5em}} \\\hline
 

\emph{all} & $0.813^{*\dagger}$ & $0.752^{*\dagger}$ & $0.769^{\dagger}$ & $0.835^{*\dagger}$ \\
\emph{all conds} & 0.786 & $0.743^{\dagger}$ & $0.772^{\dagger}$ & $0.833^{*\dagger}$ \\
\emph{neuro} & 0.763 & $0.740^{\dagger}$ & 0.759 & 0.797 \\
\emph{neuro+mood} & 0.756 & $0.742^{\dagger}$ & 0.761 & 0.804 \\
\emph{neuro+anx} & 0.770 & $0.744^{\dagger}$ & 0.746 & 0.792 \\
\emph{neuro+targets} & 0.750 & $0.747^{\dagger}$ & 0.764 & 0.817 \\
\hline 
\emph{none (\STL)} & 0.777 & 0.552 & 0.749 & 0.810 \\
\emph{\LR} & 0.791 & $0.723^{\dagger}$ & 0.763 & 0.817 \\
\end{tabular}
\end{center}
\caption{Test AUC when predicting \emph{Main Task} after multitask training to predict a subset of auxiliary tasks.  Significant improvement over \LR{} baseline at $p=0.05$ is denoted by $^*$, and over no auxiliary tasks (\STL) by $^\dagger$.}
\label{table:mtl_mentalhealth:ablation}
\end{table}

\subsection{Discussion}
\label{subsec:mtl_mentalhealth:discussion}

Our results indicate that an \MTL{} framework with user feature tasks can lead to significant gains over single-task models for predicting suicide risk and several mental health conditions.
We find benefit from predicting related mental conditions and demographic attributes simultaneously.


\removed{
\begin{figure}
    \begin{center}
	\includegraphics[draft,width=0.6\textwidth,page=1]{images/mtl_mentalhealth/training_curve_sgd_vs_adagrad.pdf} \\
    \end{center}
	\caption{}
\label{fig:mtl_mentalhealth:training_curve}
\end{figure}
}

We experimented with all the optimizers that Keras provides, and found that Adagrad seems to converge fastest to a good optimum, although all the adaptive learning rate optimizers (such as Adam, etc.) tend to converge quickly.  This indicates that the gradient is steeper along certain parameters than others.  Default stochastic gradient descent (SGD) was not able to converge as quickly, since it is not able to adaptively scale the learning rate for each parameter in the model -- taking too small steps in directions where the gradient is shallow, and too large steps where the gradient is steep.
We further note an interesting behavior: all of the adaptive learning rate optimizers yield a strange ``step-wise'' training loss learning curve, which hits a plateau, but then  drops after about 900 iterations, only to hit another plateau.
Obviously, we would prefer to have a smooth training loss curve.  We can indeed achieve this using SGD, but it takes much longer to converge than, for example, Adagrad.  This suggests that a well-tuned SGD would be the best optimizer for this problem, a step that would require some more experimentation and is left for future work.  

We also found that feature counts have a pronounced effect on the loss curves: relative feature frequencies yield models that are much easier to train than raw feature counts.  This of course is understandable, since feature counts will be sensitive to differences in raw number of tweets between users, whereas relative feature frequencies will be less sensitive.

Feature representations are therefore another area of optimization, e.g. different ranges of character $n$-grams ($n$ > 5).  We used character 1-to-5-grams, since we believe that these features generalize better to a new domain (e.g., Facebook) than word unigrams.  However, there is no fundamental reason not to choose longer character $n$-grams, other than time constraints in regenerating the data, and accounting for overfitting with proper regularization.

Initialization is a decisive factor in neural models, and \textcite{goldberg2015primer} recommends repeated restarts with differing initializations to find the optimal model. 
In an earlier experiment, we tried initializing an \MTL{} model (without task-specific hidden layers) with pretrained word2vec embeddings of unigrams trained on the Google News $n$-gram corpus. However, we did not notice an improvement in F-score. This could be due to the other factors, though, such as feature sparsity.

\begin{table}
\small
\begin{center}
\begin{tabular}{ r|c||r|c||r|c }
\textbf{Learning Rate} & \textbf{Loss} & \textbf{L2} & \textbf{Loss} & \textbf{Hidden Width} & \textbf{Loss} \\
 \hline
 $10^{-4}$ & 5.1 & $10^{-3}$ & 2.8 & 32 & 3.0 \\
 $5*10^{-4}$ & 2.9 & $5*10^{-3}$ & 2.8 & 64 & 3.0 \\
 $10^{-3}$ & 2.9 & $10^{-2}$ & 2.9 & 128 & 2.9 \\
 $5*10^{-3}$ & 2.4 & $5*10^{-2}$ & 3.1 & 256 & 2.9 \\
 $10^{-2}$ & 2.3 & 0.1 & 3.4 & 512 & 3.0 \\
 $5*10^{-2}$ & 2.2 & 0.5 & 4.6 & 1024 & 3.0\\
 0.1 & 20.2 & 1.0 & 4.9 & & \\
 \hline
\end{tabular}
\end{center}
\caption{Average development set loss over epochs 990-1000 of joint training on all tasks as a function of different learning parameters.  Models were optimized using Adagrad with hidden layer width 256 (aside for the rightmost column which sweeps over hidden layer width.).\label{tab:mtl_mentalhealth:sweep}}
\end{table}

Table \ref{tab:mtl_mentalhealth:sweep} shows parameters sweeps with hidden layer width 256, training the \MTL{} model on the social media data with character trigrams as input features.  The sweet spots in this table may be good starting points for training models in future work.

\subsection{Related Work}
\label{subsec:mtl_mentalhealth:related_work}

Some of the first works on MTL were motivated by medical risk prediction \parencite{caruana1996using}, and it is now being rediscovered for this purpose \parencite{lipton2016learning}. The latter use a long short-term memory (LSTM) structure to provide several medical diagnoses from health care features (yet no textual or demographic information), and find small, but probably not significant improvements over a structure similar to the \STL{} we use here.

The target in previous work was medical conditions as detected in patient records, not mental health conditions in social text.  The focus in this work has been on the possibility of predicting suicide attempt and other mental health conditions using social media text that a patient may already be writing, without requiring full diagnoses.

The framework proposed by~\textcite{collobert:ea:2011} allows for predicting any number of NLP tasks from a convolutional neural network (CNN) representation of the input text.  The model we present is much simpler: A feed-forward network with $n$-gram input layer, and we demonstrate how to constrain $n$-gram embeddings for clinical application.  Comparing with additional model architectures is possible, but distracts from the question of whether \MTL{} training with user features can improve mental condition prediction in this domain.  As we have shown, it can.

\section{Summary}
\label{sec:mtl_mentalhealth:summary}

In this chapter we showed that user mental health and gender features can be
used to learn more accurate suicide risk and mental health classifiers from
Twitter user text. Integrating user features as auxiliary tasks during training is clearly a
more effective way to integrate user features into a classifier than
treating them as predictors, since properties like user mental condition and gender
are not available at test time.  This shows that user features can improve machine
learning models that broadly improve public health.

Our results show that an \MTL{} model trained to predict all user mental health tasks performs significantly better than other models, reaching $0.846$ true positive rate for predicting neuroatypicality at a false positive rate of $0.1$ (TPR@FPR=0.1), and a TPR@FPR=0.1 of $0.559$ for predicting suicide risk.  Due to the nature of \MTL, we also find pronounced gains in detecting anxiety, PTSD, and bipolar disorder.  MTL predictions for anxiety, for example, reduce the error rate from a single-task model by up to $11.9$\%.


Our results also underscore the general challenge neural models face in defeating strong
linear models with scarce training data.  Logistic regression classifiers predict
a single mental condition more accurately than feedforward neural networks trained
on a single task.  It is only with the beneficial regularization of user demographic
and mental condition tasks and that neural networks outperform logistic regression.
This suggests that explicitly designing a neural architecture with the classification task
in mind can make the critical difference between under or overperforming a baseline
linear model.  In this case, an architecture of a ``forest'' of tasks corresponding to
correlated user demographic and mental condition comorbidities improved mental condition
prediction.


Whether user embeddings can act as useful auxiliary tasks for learning
mental health classifiers is still open.  However, they may be noisy
surrogates for user gender, age, and other demographic features, as evidenced by
the experiments in Section \ref{subsec:mv_twitter_users:demographic_results}.
Therefore, it is natural to assume that user embeddings would be useful
auxiliary targets in cases where predicting user demographic properties are
related to the main task. Chapter \ref{chap:mtl_stance} follows this line of
research by exploring whether user embeddings are beneficial auxiliary tasks in
an \MTL{} framework to improve tweet-level stance classifiers.

\cleardoublepage

\chapter{User Embeddings to Improve Tweet Stance Classification}
\label{chap:mtl_stance}

Chapter \ref{chap:mtl_mentalhealth} showed that ground truth user features -- mental condition
and demographic features -- help learn more accurate classifiers, specifically at
predicting the mental conditions a Twitter user has based on their character $n$-gram usage
in tweets they post. This was accomplished by training neural classifiers in a \MTL{}
framework where additional user conditions and gender were added as auxiliary tasks.
The question remains: can user embeddings take the place of ground truth user
features and also act as beneficial auxiliary tasks?
This chapter answers this question (in the affirmative!) for the domain of tweet
stance classification.  This is more evidence that semi-supervised training, predicting
user embeddings as an initial auxiliary task, can be used to improve a wide range of
tasks beyond predicting latent user features.

In this chapter we consider recurrent neural network (RNN) tweet-level stance classifiers,
and evaluate the efficacy different pretraining schemes.  We evaluate on two separate
datasets: 1) the hashtag-annotated Twitter gun control opinion dataset described in Chapter
\ref{chap:user_conditioned_topicmodels} and 2) the stance classification dataset
released as part of the SemEval 2016 6A shared task.  We show that user embeddings alone
are surprisingly effective at predicting gun control stance.  We then proceed to use the
author embeddings indirectly, as auxiliary tasks to pretrain the parameter-heavy RNN
stance classifiers.  We find that this pretraining improves stance classification
performance on average across the five domains in the SemEval shared task, although it
still performs on par or underperforms compared to linear classifiers trained on tweet
token $n$-gram features.

Section \ref{sec:mtl_stance:introduction} introduces the problem of stance classification.
Section \ref{sec:mtl_stance:datasets} then describes the different datasets used for
training stance classifiers as well as learning the user embeddings used
in pretraining.  Section \ref{sec:mtl_stance:experiments} discusses the experimental
setting and section \ref{sec:mtl_stance:models} describes the model architectures we
evaluate in detail.  Finally, section \ref{sec:mtl_stance:results}
presents performance of user embeddings for predicting stance alone, along with
the performance of RNNs.  This work was presented at W-NUT 2018
\parencite{benton2018using}.

\section{Introduction}
\label{sec:mtl_stance:introduction}

Social media analyses often rely on a tweet classification step to produce structured
data for analysis, 
including tasks such as sentiment \parencite{jiang2011target} and stance \parencite{mohammad2016semeval}
classification.
Common approaches feed the text of each message to a classifier, which predicts a label
based on the content of the tweet. However, many of these tasks benefit from knowledge
about the context of the message, especially since short messages can be difficult
to understand \parencite{aramaki2011twitter,collier2011syndromic,kwok2013locate}.
One of the best sources of context is the message author herself.
Consider the task of stance classification, where a system must identify the stance
towards a topic expressed in a tweet. Having access to the latent beliefs of the tweet's author
would provide a strong prior as to their expressed stance, e.g. general
political leanings provide a prior for their statement on
a divisive political issue. Therefore, we propose providing user level information
to classification systems to improve classification accuracy.

One of the challenges with accessing this type of information on social media users, and 
Twitter users in particular, is that it is not provided by the platform. While political
leanings may be helpful, they are not directly contained in metadata or user-provided
information.
Furthermore, it is unclear which categories of user information will best inform each
classification task.
While information about the user may be helpful in general, {\em what } information is
relevant to each task may be unknown.

We propose pretraining tweet stance classifiers to predict a user embedding
given the tweet text.  This is similar to multitask training of mental health
classifiers in Chapter \ref{chap:mtl_mentalhealth}, where ground truth binary user features
were used as auxiliary tasks, instead of embeddings.  Since
a deployed classifier will likely encounter many new users for which we do not have
embeddings, we use the user embeddings as a mechanism for pretraining the classification model
By pretraining model weights to be predictive of user embeddings, a classifier will be
able to generalize better on heldout data
after training on a task-specific dataset.  This pretraining can be performed
on a separate, unlabeled dataset of tweets and user embeddings and tends to improve
downstream task performance.  Although semi-supervised approaches to stance
classification are far from new, they have been implemented at the message-level --
predicting heldout hashtags from a tweet, for example \parencite{zarrella2016mitre}.
Our approach leverages additional user information that may not be contained in a
single message.

In this chapter, we evaluate our approach on two stance classification datasets: 1)
the SemEval 2016 task of stance classification \parencite{mohammad2016semeval}
and 2) the guns-related Twitter opinion data described in Section
\ref{subsec:user_conditioned_topicmodels:twitter_opinion_data}.
On both datasets we compare the benefit of pretraining neural stance classifiers
to predict different user embeddings derived from different types of online
user activity: an author's ego text user embedding, their friend network embedding,
and a multiview embedding of both of these views.  We also compare pretraining on
within-domain user embeddings vs. pretraining on the generic out-of-domain user
embeddings learned in Chapter \ref{chap:mv_twitter_users}.

\section{Stance Classification}
\label{sec:mtl_stance:stance_classification}

The popularity of sentiment classification is motivated in part by the utility of understanding
the opinions expressed by a large population \parencite{pang2008opinion}. Sentiment analysis of movie reviews \parencite{pang2002thumbs}
can produce overall ratings for a film, analysis of product reviews allow for better recommendations \parencite{blitzer2007biographies},
and analysis of opinions on important issues can serve as a form of public opinion polling \parencite{tumasjan2010predicting,bermingham2011using}.

Although similar to sentiment classification, stance classification concerns the
identification of an
author's position with respect to a given target \parencite{anand2011cats,murakami2010support}.
This is related to the task of targeted sentiment classification, in which both the sentiment and its target must be identified \parencite{somasundaran2009recognizing}. In the case of stance classification, we are given a fixed target, e.g.
a political issue, and want to predict the opinion of a piece of text
towards that issue.  While stance classification can be expressed as a complex set of opinions
and attitudes \parencite{rosenthal2017semeval}, we confine ourselves to the task of binary stance classification, in which we seek to determine if a
single message expresses support for or opposition to the given target (or neither). This definition was used in the 
SemEval 2016 task 6 stance classification task \parencite{mohammad2016semeval}.

A key observation behind stance classification is that the system is designed to uncover the
latent position held by the author of the message. While most work in this area seeks to infer
the author's position based only on the given message, other information 
about the author may be available to aid in the analysis of a message. Consider a user who frequently expresses liberal positions on a range of political topics. Even without observing any messages from the user about a specific liberal political candidate,
we can reasonably infer that the author would support the candidate. Therefore, when given a
message from this author whose target is the political candidate, our model should have a
strong prior to predict a positive label.

This type of information is readily available on social media platforms where we can observe multiple messages from a user,
as well as other behaviors such as sharing content, liking or promoting content, and the social network around the user.
Additionally, this type of contextual information is most needed in a social media setting. Unlike long form text common in 
sentiment analysis of articles or reviews, analysis of social media messages necessitates understanding short, informal text.
Context becomes even more important in a setting that is challenging for NLP algorithms to operate in.

How can we best make use of contextual information about the author? Several challenges present
themselves:

First, what contextual information is valuable to social media stance classifiers? We may have previous messages from the user, social network information, and a variety of other types of online behaviors. How can we best summarize a wide array of user behavior in an online platform into a single, concise representation?

We answer this question by exploring several representations of this context encoded a user embedding: a low dimensional representation of the user that can be used as features by the classification system. We include a multiview user embedding that is design to summarize multiple types of user information into a single embedding, learned in Chapter \ref{chap:mv_twitter_users}.

Second, how can we best use contextual information about the author in the learning process? Ideally we would be provided a learned user representation along with every message we were asked to classify. This is unrealistic. Learning user representations requires data to be collected for each user and computation time to process that data. Neither of these are available in many production settings, where millions of messages are streamed on a given topic. It is impractical to insist that additional information be collected for each user, new representations inferred, all while the consumer of a stance classifier waits for a label to be predicted for a single tweet.

Instead, we integrate user context in multitask learning setting, in a similar way to how
user gender was used as an auxiliary task to improve mental condition classification in
Chapter \ref{chap:mtl_mentalhealth}. We consider augmenting neural models with a pretraining
step that updates model weights according to an auxiliary objective function based on
available user representations. This pretraining step initializes the hidden layer weights of
the stance classification neural network so that the resulting model improves
even when observing only a single message at classification time.

Finally, while our focus is stance classification, this approach is applicable to a variety
of document classification tasks in which author information can provide important insights
in solving the classification problem.

\section{Models}
\label{sec:mtl_stance:models}

Our stance classification tasks focus on tweets: short snippets of informal text. We
rely on recurrent neural networks as a base classification model, as they have been effective classifiers for this type of data
\parencite{tang2015document,vosoughi2016tweet2vec,limsopatham2016bidirectional,yang2017attention}.

Our base classification model is a gated recurrent unit (GRU) recurrent neural
network classifier.
The GRU consumes the input text as a sequence of tokens and produces a 
sequence of final hidden state activations. Prediction is based on a
convex combination of these activations, where the combination weights are determined by
global dot-product attention using the
final hidden state as the query vector \parencite{luong2015effective}.  A final
softmax output layer predicts the stance class labels based
on the convex combination of hidden states. Input layer word embeddings are initialized
with GloVe embeddings pretrained on Twitter text \parencite{pennington2014glove}.

For this baseline model, the RNN is fit directly to the training set, without any
pretraining, i.e. training
maximizes the likelihood of class labels given the input tweet.
As in Chapters \ref{chap:user_conditioned_topicmodels} and \ref{chap:mtl_mentalhealth},
we have the option of exploring an entire
zoo of neural architectures.  This is however not the point of this thesis -- we want to
show how user features and embeddings can be used to improve downstream tasks;
indiscriminately exploring different architectures distracts from this point.

We now consider an enhancement to our base model that incorporates user embeddings.

\paragraph*{RNN Classifier with User Embedding Pretraining}

We augment the base RNN classifier with an 
additional final (output) layer to predict an auxiliary user embedding for the
tweet author.  
The objective function used for training this output layer depends on the type of
user embedding (described below). A single epoch is made over the pretraining set
before fitting to train.

In this case, the RNN must predict information about the tweet author
in the form of an $d$-dimensional user embedding based on the input tweet text.
If certain dimensions of the user embedding correlate
with different stances towards the given topic, the RNN will learn representations of the input that predict 
these dimensions, thereby initializing the RNN with good representations for determining stance.

The primary advantage of this semi-supervised setting is that it decouples the
stance classification annotated training set from a set of user embeddings.
It is not always possible have a dataset with stance-labeled tweets as well as user
embeddings for each tweet author (as is the case for our datasets). 
Instead, this setting allows us to utilize
a stance-annotated corpus, and separately create representations for a disjoint set of
pretraining users, even without knowing the identity of the authors of the stance-annotated
tweets.

\subsection{User Embedding Models}
\label{subsec:mtl_stance:user_embedding_types}

We explore pretraining on several different user embeddings. These methods capture both information from previous tweets by the user as well as social network features.

\paragraph*{Keyphrases} In some settings, we may have a set of important keyphrases that we believe to be correlated with the stance we are trying to predict. Knowing which phrases are most commonly used by an author may indicate the likely stance of that author to the given issue. We consider how an author has used keyphrases in previous tweets by computing a distribution over keyphrase mentions and treat this distribution as their user representation.

\paragraph*{Author Text} When a prespecified list of keyphrases is unknown, we include all words in the user representation. Rather than construct a high dimensional embedding -- one dimension for each type in the vocabulary -- we reduce the dimensionality by applying principal component analysis (PCA) to the TF-IDF-weighted user-word matrix based on tweets from authors (latent semantic analysis) \parencite{deerwester1990indexing}. We use the 
30,000 most frequent token types after stopword removal.

\paragraph*{Social Network} On social media platforms, people friend other users who share common beliefs 
\parencite{bakshy2015exposure}. These beliefs may extend to the target issue in stance classification. Therefore, a friend relationship can 
inform our priors about the stance held by a user. We construct an embedding based on the social network by creating an adjacency matrix of the 100,000 most frequent Twitter friends in our dataset (users whom the ego user follows). We construct a PCA embedding of the local friend network of the author.

\paragraph*{MultiView Representations} Finally, we consider a canonical
correlation analysis (CCA) multiview embedding over the content of the user's messages as well
as their social network\footnote{In actuality, we fit a \gcca{} model to these two views using the \texttt{wgcca} library: \url{https://github.com/abenton/wgcca}.  However, as \textcite{kettenring1971canonical} notes, when the number of views is two, this reduces to the same solution as CCA.  Thus we refer to it as CCA going forward.}. We project both the text and
friend network PCA embeddings described above, and take the mean projection of both views as
a user's embedding

We use a mean squared error loss to pretrain the RNN on these embeddings since they are all real-valued vectors.  When pretraining on a user's keyphrase distribution, we instead use a final softmax layer and minimize cross-entropy loss.

For embeddings that rely on content from the author, we collected the most recent 200
tweets posted by these authors using the Twitter REST
API\footnote{\url{https://api.twitter.com/1.1/statuses/user_timeline.json}}.  If the user
posted fewer than 200 public tweets, then we collected all of their tweets.
We constructed the social network by collecting the friends of users
as well\footnote{\url{https://api.twitter.com/1.1/friends/list.json}}.
We collected user tweets
and networks between May 5 and May 11, 2018.

We considered user embedding widths between 10 and 100 dimensions, but selected
dimensionality 50 based on an initial grid search to maximize cross validation (CV)
performance for the author text PCA embedding.

\subsection{Baseline Models}
\label{subsec:mtl_stance:alternative_models}

\begin{table}
  \small
  \begin{center}
  \begin{tabular}{ |m{2.5cm}|m{1cm}|m{8.5cm}| }
    \hline
        {\bf Topic} & {\bf Count} & {\bf Hashtags} \\ \hline
        Atheism &  38,667 & {\footnotesize \#jesus, \#atheist, \#bible, \#christ, \#god, \#lord, \#islam, \#atheists, \#religion, \#christian, \#christians, \#islamophobia, \#quran, \#christianity, \#atheism, \#judaism, \#allah, \#secularism, \#sacrilegesunday, \#secular, \#humanism, \#godless, \#athiest, \#faith, \#evolution, \#islamicstate, \#atheistrollcall, \#muhammad, \#muslim} \\ \hline
        Climate Change is a Real Concern & 12,417 & {\footnotesize  \#cop21, \#climatechange, \#climate, \#globalwarming, \#science, \#environment, \#climateaction, \#actonclimate, \#paris, \#energy, \#sustainability, \#water, \#renewables, \#nuclear, \#climatechangeisreal, \#solar, \#parisagreement, \#co2, \#coal, \#cop21paris, \#fracking, \#action2015, \#carbon, \#climatemarch, \#green, \#pollution, \#sdgs, \#agriculture, \#nature, \#vegan, \#earthtoparis, \#geoengineering, \#renewableenergy, \#junkscience, \#keystonexl, \#keepitintheground, \#oil, \#paris2015 } \\ \hline
        Feminist Movement &  2,534 & {\footnotesize \#feminismo, \#women, \#feminism, \#feminist, \#feminismus, \#equality, \#fem2, \#yesallwomen, \#feminisme, \#sexism, \#womenagainstfeminism, \#misogyny, \#feminismiscruelty, \#gender, \#genderequality, \#womensrights } \\ \hline
        Hillary Clinton & 734 & {\footnotesize \#demdebate, \#gopdebate, \#hillary, \#hillaryclinton, \#stophillary, \#whyimnotvotingforhillary } \\ \hline
        Legalization of Abortion & 1,854 & {\footnotesize \#prolife, \#abortion, \#standwithpp, \#waronwomen, \#1in3, \#defundpp, \#ppact } \\ \hline
        Topic Unclear & 19,481 & {\footnotesize  \#tcot, \#chat, \#pjnet, \#usa, \#trump, \#uniteblue, \#stoprush, \#uk, \#philippines, \#auspol, \#p2, \#africa, \#australia, \#india, \#2a, \#1a, \#nra, \#gunfail, \#gop, \#cdnpoli, \#ccot, \#makeamericagreatagain, \#isis, \#truth, \#obama, \#imwithhuck, \#teaparty, \#england, \#trumptrain, \#britain, \#health, \#tlot, \#gamergate, \#lgbt, \#foodsecurity, \#chine, \#aus, \#arctic, \#humanrights, \#popefrancis, \#blacklivesmatter, \#libcrib, \#feelthebern, \#france, \#world, \#gunsense, \#tntweeters, \#london, \#politics, \#rednationrising, \#bernie2016, \#tpp, \#votetrump2016, \#ndp, \#berlin, \#cruzcrew, \#trump2016, \#realdonaldtrump, \#china, \#donaldtrump, \#drought, \#potus, \#parisattacks, \#boycott, \#c51, \#syria, \#poverty, \#farm365, \#chemtrails } \\ \hline
  \end{tabular}
  \end{center}
  \caption{Hashtags used for hashtag prediction pretraining.  These were selected based on corpus
    frequency and hand-curated.  They are grouped by topic for presentation, with hashtags that
    could be relevant to multiple topics in ``Topic Unclear''.
    The second column contains the number of times hashtags associated with that topic occurred
    in the pretraining set.\label{tab:mtl_stance:mitre_hashtags}}
\end{table}

We compare our approach against the following two baseline models:

\paragraph*{Hashtag Prediction Pretraining}
As part of the SemEval 2016 task 6 tweet stance classification task, 
\textcite{zarrella2016mitre} submitted 
an RNN-LSTM classifier that used an auxiliary task of predicting the hashtag distribution
\emph{within} a tweet to pretrain their model.
There are a few key differences between our proposed method and this work.
Their approach is restricted to the stance classification dataset,
whereas we consider building representations of the user from context.
Additionally, their method is restrictive in that they are predicting
a task-specific set of hashtags, whereas user features/embeddings offer more flexibility in that they are not as strongly tied to a specific task.
However, we select this as a baseline for comparison because of how they utilize hashtags
within a tweet for semi-supervised training. We call this model {\tt RNN-content-hashtag}.

We evaluate a similar approach by identifying the 200 most frequent hashtags in
the SemEval-hashtag pretraining set (dataset describe below).
After removing non-topic hashtags (e.g. \#aww, \#pic),
we were left with 189 unique hashtags, with 32,792 tweets containing at
least one of these hashtags (Table \ref{tab:mtl_stance:mitre_hashtags}).
Pretraining was implemented by using a 189-dimensional softmax output layer to
predict held-out hashtag.

RNNs were trained by cross-entropy loss where only the most frequent hashtag was
considered to be the target.  RNNs were trained by cross-entropy loss where the
target distribution placed a weight of 1 on the most frequent hashtag, with all
other hashtags having weight of 0.
This is the identical training protocol used in \textcite{zarrella2016mitre}.

There are a few key differences between our proposed method and the MITRE submission.
First, the MITRE submission's pretraining regimen relies on predicting tweet-level features,
whereas
we are predicting user features.  Second, their method is restrictive in that they
are predicting a task-specific set of hashtags, whereas generic user features or embeddings
offer more flexibility in that they are not as strongly tied to a specific task.

\paragraph*{SVM Baseline}
We also reproduce a word and character n-gram linear support vector machine 
that uses word and character n-gram features.
This was the best performing method on average in the 2016 SemEval Task 6 shared task \parencite{mohammad2016semeval}.
We swept over the slack variable penalty coefficient to maximize macro-averaged
F1-score on held-out CV folds.

\section{Data}
\label{sec:mtl_stance:datasets}

\subsection{Stance Classification Datasets}
\label{subsec:mtl_stance:stance_data}

We consider two different tweet stance classification datasets, which in total provide
six domains of English language Twitter data.

\paragraph*{SemEval 2016 Task 6A (Tweet Stance Classification)}

This is a collection of 2,814 training and 1,249 test set tweets
that are about one of five politically-charged targets:
\emph{Atheism}, the \emph{Feminist Movement},
\emph{Climate Change is a Real Concern}, \emph{Legalization of Abortion}, or
\emph{Hillary Clinton}.
Given the text of a tweet and a target, models must classify
the tweet as either {\sc favor}, {\sc against} or {\sc neither} if the tweet does not express support or opposition
to the target topic. Many shared task participants struggled with this task, as it was especially difficult
due to imbalanced class sizes, small training sets, short examples, and tweets
where the target was not explicitly mentioned. See
\textcite{mohammad2016semeval} for a thorough description of this data.
We report model performance on the provided test set for each topic and perform
four-fold CV on the training set for model selection\footnote{CV folds were not
  released with these data.  Since our folds are different than other submissions
  to the shared task, there are likely differences in model selection.}.

\paragraph*{Guns}

We built the second stance dataset from the gun control opinion dataset described in Section
\ref{subsec:user_conditioned_topicmodels:twitter_opinion_data}.
Tweets were collected from the Twitter keyword streaming API starting in December 2012 and throughout 2013\footnote{\url{https://stream.twitter.com/1.1/statuses/filter.json}}.
The collection includes all tweets containing guns-related keyphrases, subject to rate limits.
We labeled tweets based on their stance towards gun control: {\sc favor} was supportive of gun control, {\sc against}
was supportive of gun rights. 
We automatically identified the stance to create labels based 
on commonly occurring hashtags that were clearly associated with one of these positions 
(see Table \ref{tab:mtl_stance:guns_kws} for a list of keywords and hashtags).
Tweets which contained hashtags from both sets or contained no stance-bearing hashtags
were excluded from our data.

We constructed stratified samples from 26,608 labeled
tweets in total.  Of these, we sampled 50, 100, 500, and 1,000
examples from each class, five times, to construct five small, balanced training
sets, and divided the remaining examples equally between development and
test sets in each case.  We then divided the remaining examples equally between
development and test sets in each case.
Model performance for each number of examples was macro-averaged over the five
training sets. The hashtags used to assign class labels were removed from the
training examples as a preprocessing step.

\begin{table}
  \small
  \begin{center}
  \begin{tabular}{ c|p{9cm} } 
    \hline
       {\bf Set Name} & {\bf Keyphrases/Hashtags} \\ \hline
       About Guns (General) & gun, guns, second amendment,
                    2nd amendment, firearm, firearms \\ \hline
       Favors &  \#gunsense, \#gunsensepatriot, \#votegunsense,
                 \#guncontrolnow, \#momsdemandaction, \#momsdemand,
                 \#demandaplan, \#nowaynra, \#gunskillpeople, \#gunviolence,
                 \#endgunviolence\\ \hline
       Against & \#gunrights, \#protect2a, \#molonlabe, \#molonlab,
                 \#noguncontrol, \#progun,\#nogunregistry,
                 \#votegunrights, \#firearmrights, \#gungrab,
                 \#gunfriendly \\
    \hline
  \end{tabular}
  \end{center}
  \caption{Keyphrases used to identify gun-related tweets along with hashtag sets used to
           label a tweet as \emph{Favors} or is \emph{Against} additional gun control
           legislation.\label{tab:mtl_stance:guns_kws}}
\end{table}

\subsection{User Embedding Datasets}
\label{subsec:mtl_stance:useremb_data}

We considered two unlabeled datasets as a source for constructing user embeddings for
model pretraining. Due to data limitations, we were unable to create all of our
embedding models for all available datasets. We describe below which embeddings were
created for which datasets.

\paragraph*{SemEval 2016 Related Users} 

\begin{table}
  \small
  \begin{center}
  \begin{tabular}{ m{4cm}|p{9cm} } 
    \hline
       {\bf Topic} & {\bf Example hashtags} \\ \hline
       Atheism & \emph{\#nomorereligions, \#godswill, \#atheism} \\ 
       Climate Change is a Concern & \emph{\#globalwarmingisahoax, \#climatechange} \\ 
       Feminist Movement & \emph{\#ineedfeminismbecaus, \#feminismisawful, \#feminism} \\ 
       Hillary Clinton & \emph{\#gohillary, \#whyiamnovotingforhillary, \#hillary2016} \\ 
       Legalization of Abortion & \emph{\#prochoice, \#praytoendabortion, \#plannedparenthood} \\ 
    \hline
  \end{tabular}
  \end{center}
  \caption{Subset of hashtags used in \textcite{mohammad2016semeval} to identify politically-relevant tweets.  We used this set of hashtags to build a pretraining set relevant to the stance classification task.\label{tab:mtl_stance:semeval2016_hashtags}}
\end{table}

The SemEval stance classification dataset does not contain tweet ids or user ids, so we are unable to determine
authors for these messages. Instead, we sought to create a collection of users whose tweets and online behavior
would be relevant to the five topics discussed in the SemEval corpus.

We selected query hashtags used in the shared task \parencite{mohammad2016semeval} and searched
for tweets
that included these hashtags in a large sample of the Twitter 1\% streaming API sample  
from 2015\footnote{\url{https://stream.twitter.com/1.1/statuses/sample.json}}.
Table \ref{tab:mtl_stance:semeval2016_hashtags} lists the example hashtags described in
\textcite{mohammad2016semeval} used to sample politically relevant tweets from
the Twitter stream.  This ensured that tweets were related to one of the targets in the
stance evaluation task, and were from authors discussing these topics in a similar time period.
We recorded the author of each of these tweets and then queried the Twitter API
to pull the tweet authors' most recent 200 tweets and local friends and followers network. 
We omitted tweets made by deleted and banned users
as well as those who had fewer than 50 tweets total returned by the API.
In total, we were able to obtain 79,367 tweets for 49,361 unique users,
and were able to pull network information for 38,337 of these users.

For this set of users, we constructed the {\bf Author Text} embedding (PCA representation of a TF-IDF-weighted bag of words from the user) as well as the {\bf Social Network} embedding (PCA representation of the friend adjacency matrix.)
For users with missing social network information, we replaced their network embedding with
the mean embedding over all other users.  This preprocessing was applied before learning
{\bf Multiview} embeddings over all users.

\paragraph*{General User Tweets}
Is it necessary for our pretraining set to be topically-related to the stance task we are
trying to improve, or can we consider a generic set of users?  To answer this question we
created a pretraining set of randomly sampled users, of over 102 thousand user learned in
Chapter \ref{chap:mv_twitter_users}, not specifically related to any of
our stance classification topics.
If these embeddings prove useful, it provides an attractive method whereby general
embeddings can be created
for users not specifically related to the stance classification topic.

Although there are many
potential user embeddings we could consider pretraining with, we only consider the
ego text, friend network, and a CCA embedding of these two views as user embeddings
for pretraining.  We selected these since the PCA ego text embedding is a clear baseline,
the friend network embedding was shown to be most effective at friend network prediction,
and a CCA representation of ego text and friend network was shown to outperform other
embeddings at hashtag prediction.  We avoided considering CCA embeddings of all subsets
of viwes to narrow the model search space.

To pretrain classifiers, we randomly selected three tweets that each user made in
March 2015 as pretraining tweets.  Embeddings were learned over tweets from January
and February 2015, a disjoint sample from the three randomly selected tweets from March.
This resulted in a pretraining set of 152,751 tweets for 61,959 unique users.

\paragraph*{Guns User Tweets}

We also kept 49,023 unlabeled guns tweets for pretraining on the gun control stance task,
using the distribution over {\em general} keyphrases that an author posted across
the pretraining set as the user embedding.  We pretrained on the ({\bf Author Text})
embedding of these tweets, along with a {\bf Social Network} embedding (network data collected
identically to above pretraining datasets).

\section{Model Training}
\label{sec:mtl_stance:experiments}

We preprocessed all tweets by lowercasing and tokenizing with a Twitter-specific
tokenizer \parencite{gimpel2011part}\footnote{\url{https://github.com/myleott/ark-twokenize-py}}. 
We replaced usernames with \texttt{$<$user$>$} and URLs with \texttt{$<$url$>$}.

For training on the SemEval dataset, we selected models based
on four-fold cross validation macro-averaged F1-score for {\sc Favor}
and {\sc Against} classes (the official evaluation metric for this task).  For
the gun dataset we select models based on average development set F1-score.
For SemEval, each classifier is trained independently for each
target. Reported test F1-score is averaged across each model fit on
CV folds.

All neural networks were trained by minibatch gradient descent with
Adam \parencite{kingma2014adam} with base step size 0.005, $\beta_1=0.99$, and $\beta_2=0.999$,
with minibatch size of 16 examples, and the weight updates were clipped
to have an l2-norm of $1.0$.  Models were trained for a minimum of 5 epochs with
early stopping after 3 epochs if held-out loss did not improve, and the loss per
example was weighted by the inverse class frequency of the
example\footnote{This improved performance for tasks with imbalanced class labels such as the ``Climate Change'' topic.}.

The neural model architecture was selected by performing a grid search over
hidden layer width
(\{25, 50, 100, 250, 500, 1000\}), dropout rate (\{0, 0.1, 0.25, 0.5\}),
word embedding width (\{25, 50, 100, 200\}), number of layers (\{1, 2, 3\}), and
RNN directionality (forward or bidirectional).
Architecture was selected to maximize cross-fold macro-averaged F1 on
the ``Feminist Movement'' topic with the GRU classifier without pretraining.  We
performed a separate grid search of architectures for the pretraining models.

\begin{table}
  \small
  \begin{center}
  \begin{tabular}{ c|c|c }
    \hline
        {\bf Hyperparameter} & {\bf Min} & {\bf Max} \\ \hline
        hidden unit width & 10 & 1000 \\
        dropout probability & 0.0 & 0.9 \\
        word embedding width & 25 & 200 \\
        number layers & 1 & 3 \\
        directionality & forward & bidirectional \\ \hline
  \end{tabular}
  \end{center}
  \caption{Grid search range for different architecture and training parameters.
    \label{tab:mtl_stance:grid_search}
  }
\end{table}

\begin{figure}
  \begin{subfigure}{\linewidth}
    \centering
    \includegraphics[clip,trim={1.5cm 0cm 1.5cm 0.25cm},width=0.48\textwidth,page=1]{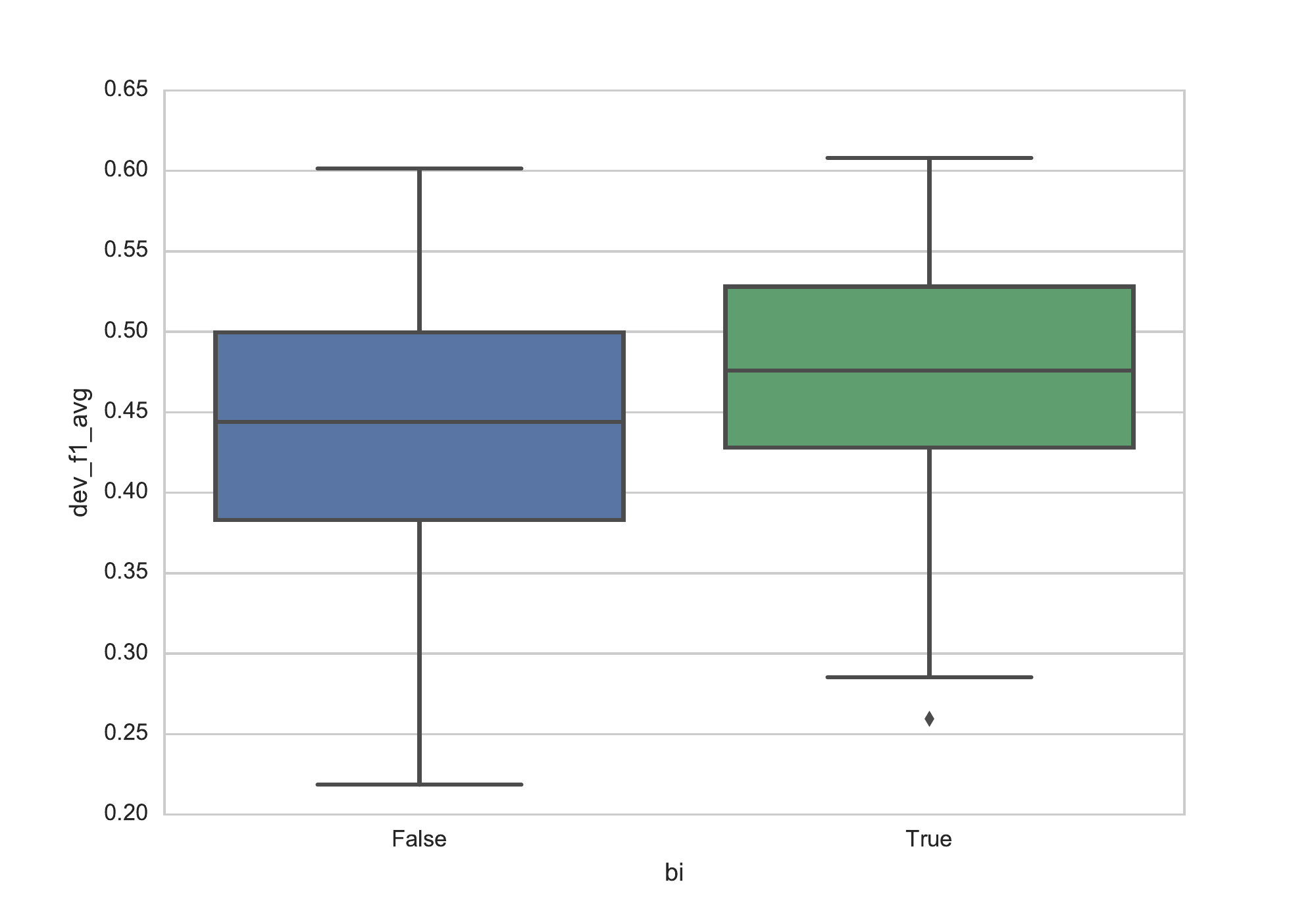}
    \includegraphics[clip,trim={1.5cm 0cm 1.5cm 0.25cm},width=0.48\linewidth,page=2]{images/mtl_stance/pretrain_stance_archsweep_plots.pdf}
  \end{subfigure} \\
  \begin{subfigure}{\linewidth}
    \centering
    \includegraphics[clip,trim={1.5cm 0cm 1.5cm 0.25cm},width=0.32\linewidth,page=3]{images/mtl_stance/pretrain_stance_archsweep_plots.pdf}
    \includegraphics[clip,trim={1.5cm 0cm 1.5cm 0.25cm},width=0.32\linewidth,page=4]{images/mtl_stance/pretrain_stance_archsweep_plots.pdf}
    \includegraphics[clip,trim={1.5cm 0cm 1.5cm 0.25cm},width=0.32\linewidth,page=5]{images/mtl_stance/pretrain_stance_archsweep_plots.pdf}
  \end{subfigure}
\caption{Boxplots of mean cross-fold F1-score as a function of different hyperparameters:
  RNN bi-directionality (upper left), number of layers (upper right), hidden layer width
  (lower left), embedding width (lower center), and dropout rate (lower right) for
  an {\bf Author Text}-pretrained RNN on the ``Feminism Movement'' stance classification task.
  \label{fig:mtl_stance:grid_search_f1}}
\end{figure}

Table \ref{tab:mtl_stance:grid_search} lists the range of hyperparameters swept by the
grid search.  Figure \ref{fig:mtl_stance:grid_search_f1} displays average cross-fold F1
for an RNN pretrained on predicting the ego text user embedding during grid search along with
on the SemEval 2016 hashtag-filtered pretraining set.

\section{Results and Discussion}
\label{sec:mtl_stance:results}

\subsection{SemEval 2016 Task 6A}
\label{subsec:mtl_stance:semeval_results}

\begin{table}
  \small
  \begin{center}
  \begin{tabular}{ c|c|c|c|c|c||c } 
    \hline
        {\bf Model} & \multicolumn{5}{c}{{\bf Target }} & \\ \hline
                    & Ath & Cli & Fem & Hil & Abo & Avg \\ \hline
        SVM & 61.2 & 41.4 & 57.7 & 52.0 & 59.1 & 54.3 \\ \hline \hline
        RNN & $54.0^{\triangledown}$ & 39.6 & $48.5^{\triangledown}$ & {\bf 53.5} & 58.6 & 50.8 \\ \hline
        RNN-content-hashtag & 53.4 & 41.0 & $48.4^{\triangledown}$ & 48.0 & 55.8 & 49.3 \\ \hline \hline
        RNN-hset & 58.2 & {\bf 44.5} & 51.2 & 50.9 & {\bf 60.2} & {\bf 53.0 } \\ \hline \hline
        RNN-text-hset & 58.2 & {\bf 44.5} & 51.2 & 50.9 & {\bf 60.2} & {\bf 53.0 } \\ \hline
        RNN-network-hset & 42.7 & 38.8 & 48.2 & 42.0 & 45.0 & 43.3 \\ \hline
        RNN-multiview-hset & {\bf 60.1} & 40.5 & 49.9 & 52.5 & 56.5 & 51.9 \\ \hline \hline
        RNN-genset & 56.7 & 41.9 & {\bf 54.4 } & 51.7 & 56.5 & 52.2 \\ \hline \hline
        RNN-text-genset & 56.7 & 38.2 & {\bf 54.4$^{\diamondsuit \spadesuit}$} & 51.7 & 56.5 & 51.5 \\ \hline
        RNN-network-genset & 54.6 & 41.4 & 47.8 & 50.5 & 50.6 & 49.0 \\ \hline
        RNN-multiview-genset & 57.3 & 41.9 & 52.1 & 50.4 & 54.4 & 51.2 \\ \hline
  \end{tabular}
  \end{center}
  \caption{Positive/negative class macro-averaged F1 model test performance at
    SemEval 2016 Task 6A.
    \emph{hset}: SemEval 2016
    hashtag pretrain set, \emph{genset}: general user pretrain set. The
    best-performing neural model is in bold.
    The \emph{RNN-genset} and \emph{RNN-hset} rows contain test performance
    if we select the pretraining embedding type (text, friend, or CCA) according to CV F1
    for each domain.
    The final column is the macro-averaged F1 across all domains. $^{\diamondsuit}$
    means model performance is statistically
    significantly better than a non-pretrained RNN according to a bootstrap sampling
    test (p=0.05, with 1000 iterations of sample size 250), $^{\triangledown}$ is worse
    than SVM, and $^{\spadesuit}$ is better than tweet-level hashtag prediction pretraining.}
  \label{tab:mtl_stance:semeval2016_predclass_user}
\end{table}

Table \ref{tab:mtl_stance:semeval2016_predclass_user} contains the performance
for each target in the SemEval 2016 stance classification task.

Considering the pretrained models versus the non-pre-trained RNN, 
pretraining improves in four out of five targets. Additionally, one of our models 
always beats the baseline of tweet-level hashtag distribution pre-training
({\tt RNN-content-hashtag}). Notably, while topic specific user embeddings
({\tt hsetpre}) improve over not pretraining in four out of five cases,
the generic user embeddings ({\tt genset}) improves in three out of five cases. This suggests
that even embeddings for generic users who don't necessarily discuss the topic of interest
can have value in model learning.

In terms of embedding type, embeddings built on the author text tended to be best,
but results were not clear.

The linear SVM baseline with word and character n-gram features outperforms neural models
in two out of five tasks, and perform the best on average.
This agrees with the submissions to the
SemEval 2016 6A stance classification task, where the baseline SVM model
outperformed all submissions on average -- several of which were neural models.

\subsection{Guns}
\label{subsec:mtl_stance:gun_control_results}

\begin{table}
  \begin{center}
  \begin{tabular}{ c|c|c|c|c }
    \hline
        {\bf Model} & \multicolumn{4}{c}{{\bf \# Train Examples }} \\ \hline
                    & 100 & 200 & 1000 & 2000 \\ \hline
        SVM & 79.2 & 81.1 & 85.9 & 87.4 \\ \hline \hline
        RNN & 72.2$^{\triangledown}$ & 79.0 & {\bf 84.0} & 85.3 \\ \hline
        RNN-keyphrase-gunset & 73.1$^{\triangledown}$  & 76.7 & 83.6 & {\bf 85.6} \\ \hline
        RNN-text-gunset & 72.2$^{\triangledown}$  & 79.0  & {\bf 84.0 } & 85.3  \\ \hline
        RNN-text-genset & 71.7$^{\triangledown}$  & 76.6 & 83.6 & 85.3 \\ \hline
        RNN-network-genset & 73.1$^{\triangledown}$  & 77.2 & 83.3 & 85.4 \\ \hline
        RNN-multiview-genset & {\bf 75.0 } & {\bf 79.1} & 83.9 & 85.4 \\ \hline
  \end{tabular}
  \end{center}
 
  \caption{Model test accuracy at predicting gun stance.
           RNNs were pre-trained on either the guns-related pre-training set ({\em gunset}) or
           the general user pre-training set ({\em genset}).  The best-performing neural model
           is bolded.  $^{\triangledown}$ indicates that the model performs significantly worse
           than the SVM baseline ($p <= 0.05$ according to a 1000-fold bootstrap test
           with sample size 250).}
  \label{tab:mtl_stance:guns_full_eval}
\end{table}

\begin{table}
  \begin{center}
  \begin{tabular}{ c|c|c|c|c } 
    \hline
        {\bf Model} & \multicolumn{4}{c}{ {\bf \# Train Examples} } \\ \hline
                    & 100 & 200 & 1000 & 2000 \\ \hline
        tweet & {\bf 79.2} & {\bf 81.1} & 85.9 & 87.4 \\ \hline
        user-text & 72.1$^{\triangledown}$ & 74.1$^{\triangledown}$ & 76.5$^{\triangledown}$ & 76.6$^{\triangledown}$ \\ \hline
        keyphrase & 52.2$^{\triangledown}$ & 50.8$^{\triangledown}$ & 51.0$^{\triangledown}$ & 51.8$^{\triangledown}$ \\ \hline
        tweet + user-text & {\bf 79.2$^{\clubsuit}$ } & {\bf 81.1$^{\clubsuit}$ } & {\bf 86.0$^{\clubsuit}$ } & {\bf 87.6$^{\clubsuit}$ } \\ \hline
        tweet + keyphrase & {\bf 79.2$^{\clubsuit}$ } & {\bf 81.1$^{\clubsuit}$ } & 85.9$^{\clubsuit}$ & 87.4$^{\clubsuit}$ \\ \hline
  \end{tabular}
  \end{center}
  \caption{\label{tab:mtl_stance:guns_predclass_user} Test accuracy of an SVM at
    predicting gun control stance
    based on guns-related keyphrase distribution ({\em keyphrase}), user's {\bf Author Text}
    embedding ({\em text}), and word and character n-gram features
    ({\em tweet}).  $^{\triangledown}$ encodes models significantly
    worse ($p=0.05$) than a tweet features-only SVM according to a bootstrap
    sampling test with sample size 250 and 1000 iterations, and $^{\clubsuit}$ means the
    feature set did significantly better than user-text-PCA.}
\end{table}

Using the guns dataset, we sought to understand how the amount of training data affected
the effectiveness of model pre-training.
Table \ref{tab:mtl_stance:guns_full_eval} show the accuracy for different models at varying amounts of training data.
As the amount of training data increases, so does model accuracy. Additionally, we tend to see larger
increases from pre-training with less training data overall. 
It is unclear which user embedding or pre-training set is most effective.  The CCA embedding
is most effective at improving the neural classifier, although the difference is not
statistically significant.  The difference may be related to the finding in Section
\ref{subsec:mv_twitter_users:hashtag_results}: that multiview embeddings are more predictive
of a person's future hashtag use than considering a single view.  

As with SemEval, the SVM always outperforms neural models, though the improvement is only statistically significant
in the smallest data setting.
Although we are unable to beat SVM models, the improvements we observe in RNN performance
after user embedding pre-training are promising.  Neural model architectures offer more
flexibility than SVMs, particularly linear-kernel, and we only consider a single model class
(recurrent networks with GRU hidden unit).  Further architecture exploration is necessary,
and user embedding pre-training will hopefully play a role in training state-of-the-art
stance classification models.

Since for the guns data we have an intersection between the annotated stance data
and the users for whom we have embeddings, we sought to understand how much information
may be contained in the embeddings relevant to stance classification.
Like above, we trained an SVM to predict the gun stance but instead of providing the tweet,
we alternately provided the tweet, one of the embeddings, or both together.
Higher prediction accuracy indicates that the input is more relevant, and helpful, in predicting stance.

Table \ref{tab:mtl_stance:guns_predclass_user} shows test accuracy for this task across different amounts of training data.
Unsurprisingly, the tweet content is more informative at predicting stance than the user embedding. However, the embeddings
did quite well, with the ``Author Text'' embedding coming close to performance of tweet text
in some cases. Providing both had no or a marginal improvement over tweet text alone.

\section{Summary}
\label{sec:mtl_stance:summary}

This chapter shows that pretraining on unsupervised user embeddings improves
tweet-level neural stance classifiers.  We find that author embedding pretraining
yields improvements on four out of five domains for the SemEval 2016 Task 6A tweet
stance classification task over a non-pretrained neural network, although benefits
are less discernible on the gun control stance classification dataset of tweets.

We expand on Chapter \ref{chap:mtl_mentalhealth} and show that
pretraining to predict unsupervised user embeddings also improves classifier
performance, in spite of not having gold user features.
\emph{This remains true even when pretraining on a completely
generic set of user embeddings, when the domains of the training and unlabeled
sets do not match.}  This suggests that the set of user embeddings compiled
in Chapter \ref{chap:mv_twitter_users} can be used as a generic target
to initialize tweet-level classifiers in multiple domains.
This chapter also successfully applies user feature semi-supervised training
to a much more modern and slippery collection of neural bells and whistles, a
RNN with GRU layers, and an attention mechanism across intermediate hidden
states.  Compared to the multilayer perceptrons trained in
the previous chapter, these stance classifiers are grotesquely baroque.

In Section \ref{subsec:mv_twitter_users:hashtag_results} we found that multiview
embeddings were more predictive of future hashtag use than other combinations of
user views.  This difference in performance on the SemEval shared task between
pretraining on ({\bf Author Text}, {\bf Social Network}) CCA user embedding features vs.
{\bf Author Text} alone is not statistically significant, although pretraining on
{\bf Author Text} performs better on average.  It is hard to make strong claims on
the merits of different embeddings since the stance classification datasets have
small evaluation sets.

However, it is clear that pretraining RNN stance classifiers
to predict a user's local friend network embeddings actively hurts final performance
-- {\bf Social Network} embeddings provide a malicious inductive bias, at least for stance
classification.  This highlights the findings in chapter \ref{chap:mv_twitter_users},
that different user embeddings are best suited for different downstream tasks.  Local friend
network embeddings can predict other friending behavior accurately, but they are less
effective at informing one's stance on politically-sensitive issues.

\cleardoublepage


\chapter{Conclusion}
\label{chap:conclusion}

This thesis explores representation learning techniques to learn
social media user embeddings and evaluates embeddings at improving
downstream task performance.  In the process we develop novel methods
to learn user embeddings and integrate them into existing models.
We conclude by summarizing the contributions of each chapter: whether
user embeddings are being learned (and if so how), how they are evaluated,
the tasks we try to improve with them, and any novel models described there.
In section \ref{sec:conclusion:contributions} we retrospectively summarize the main
contributions of this thesis.  In section \ref{sec:conclusion:ethics} we touch
on ethical considerations around social media data and the choices we made
in our research to respect the privacy of users.
We conclude in section \ref{sec:conclusion:future} with directions for
future research.

\paragraph*{Chapter \ref{chap:background}} begins by reviewing work on
applications of user features and then provides an overview
of relevant computational methods: canonical correlation analysis
(CCA)-based multiview representation learning methods, and the multitask
learning setting. Section \ref{sec:background:user_application_motivation}
is a review of work on inferring user demographic features and applications
that benefit from user features and embeddings.
Section \ref{sec:background:mv_rep_learning} is an overview of correlation-based
multiview representation learning methods covering how these models are fit and
the data assumptions that they make.
This section presents the derivation of the vanilla two-view
CCA solution, the derivation of many-view generalized CCA, and describes
existing extensions to these methods.  This section could stand as a primer on
correlation-based multiview representation learning.  Section
\ref{sec:background:mtl_neural} describes the multitask learning setting, the motivation
behind this framework, and describes (at a high level) how learn neural models are learned in this setting.

\paragraph*{Chapter \ref{chap:mv_twitter_users}} describes how a set of multiview
embeddings were learned for a general collection of over 100,000 English-speaking
Twitter users.  There are two main contributions in this chapter.

First, this chapter contains the methodological meat of how multiview user embeddings
are learned.  An extension of MAXVAR-GCCA is presented.  Weighted
GCCA (\wgcca) introduces scalar weights into the per-view GCCA loss which adjust the penalty
placed on failing to recover the latent embedding from features of a particular view.
The \wgcca{} per-view weights are tuned by the practitioner based on their belief of which
views should be best captured by the user embedding or explicitly tuned for downstream
task loss.

Second, this chapter includes experiments on using GCCA methods to learn user embeddings
from different views of Twitter user behavior/features including the text they post, who
they friend, who follows them, who they mention, and what sort of language their local
network uses.  The key takeaways from these analyses are: (1) multiview user
embeddings learned on ego text along with friending behavior are more predictive of
hashtag usage (and presumably word/topic usage) than simple combinations of each of
these views, (2) multiview user embeddings are not a panacea for every task, and
friend prediction is best captured solely by other friending behavior and (3) \dgcca{}
embeddings considering all user behavior views simultaneously, without discriminatively
tuning the loss weighting for each view can predict future hashtag usage better than a linear
\wgcca{} embedding with freedom to weight each view.  This suggests that relying on nonlinear
techniques to learn user embeddings is important, at least when predicting hashtag usage.

\paragraph*{Chapter \ref{chap:user_conditioned_topicmodels}} centers on an application
of using inferred user location features as supervision for topic models to improve
topic model fit on social media text.  Topic models are fit to three keyword-filtered
Twitter message corpora.  We present a new topic model, Deep Dirichlet
Multinomial Regression (\dsprite), which is better-suited to high-dimensional and noisy
document supervision than \dmr.  \dsprite{} is evaluated on a synthetic dataset as well as
a corpus of New York Times articles, Amazon product reviews, and Reddit posts.
Although this is presented as an application of using
user embeddings to improve topic modeling of social media documents, the neural
representation layer of \dsprite{} can be viewed as \emph{learning} user embeddings (when using
author-specific features are provided).  These user embeddings are explicitly trained to
be predictive of topic preference.

\paragraph*{Chapter \ref{chap:mtl_mentalhealth}} describes experiments on predicting
suicide risk and mental health conditions for Twitter users given their tweet history.
Here we show that training classifiers in a multitask learning framework, with a user's
other mental health conditions and gender as auxiliary tasks, learns stronger mental health
condition classifiers from Twitter user text than training classifiers independently.
This chapter presents the following findings: (1) multitask training for a user's mental
health improves performance over independently trained systems where
auxiliary tasks are considered other possible conditions, (2) predicting a demographic
feature like gender improves over a model that is only trained to predict mental conditions,
and (3) the more auxiliary mental health conditions that are predicted, the stronger the
learned classifier.  The main contribution is that multitask training
of classifiers to predict demographics, and other user features, learns stronger classifiers,
better exploiting comorbidities between mental conditions and correlation with demographics.

\paragraph*{Chapter \ref{chap:mtl_stance}} describes a similar application as chapter
\ref{chap:mtl_mentalhealth}, except instead of training a classifier to predict a
user's mental condition based on tweet history, we want to train an RNN
to predict the stance expressed in a tweet based purely on that tweet's text.  We
show that on the SemEval 2016 stance classification benchmark, pretraining classifiers
to predict user embedding features improves RNN performance.  It is even statistically
significantly better than pretraining classifiers to predict heldout hashtags from
within a tweet.  Semi-supervised pretraining is key for the same reason as multitask learning is
when predicting user mental health -- user features are hard to acquire at test time,
but useful when we already have a general collection of user embeddings and tweets to
pretrain classifier weights.

\section{Contributions}
\label{sec:conclusion:contributions}

This thesis applies multiview representation learning methods to learning user embeddings
and evaluates them against other traditional user representations on an array of downstream
tasks: improving topic model fit, predicting user demographic features, mental
health condition, hashtag usage, friending behavior, as well as improving
message-level stance classification on social media.
We hope researchers look to these experiments as a guide for determining which user
embeddings are most appropriate for their downstream task and draw inspiration for
injecting user embeddings into their own models.

The work presented in this thesis has resulted in three major contributions, as
evidenced by subsequently published research.

\subsection*{Expanding What Constitutes as Model Supervision}
In chapter \ref{chap:mv_twitter_users} we apply several novel variants of multiview
representation learning methods for learning social media user embeddings using
auxiliary user information as additional views. In chapter
\ref{chap:user_conditioned_topicmodels} we present a new supervised topic
model that can exploit high-dimensional, noisy supervision.  In chapters
\ref{chap:mtl_mentalhealth}
and \ref{chap:mtl_stance} we use neural multitask learning to improve classifier
generalization at social media prediction tasks by entraining model weights to
also be predictive of features associated with the tweet author.

Although these extensions belong to entirely different model classes
(correlation-based multiview learning methods, probabilistic topic models,
and supervised neural networks), they were all motivated by the need to
exploit the wealth of unstructured, auxiliary information around social media
users to improve downstream task performance.  This need is not specific to social
media data but pervades many applied machine learning domains, encouraging others to
arrive at similar solutions.  \textcite{card2018neural} developed a supervised
neural topic model that allows for metadata to appear as either a covariate or a
predicted variable in the model structure.  There is also a line of work that
integrates information from multiple user views to learn more robust user embeddings
\parencite{li2017joint,tao2017multiview,kursuncu2018contextualized,hazarika2018cascade}.
The models we present are all trying to extract value out of features that are
only distantly related to a task of interest.

By developing these models, we hope to widen the set of viable signals
that machine learning practitioners will consider when training
semi-supervised models.  For example, instead of considering using author
demographics as auxiliary tasks when training a multitask model, one may
just as well consider predicting prior tweeting history as an auxiliary task
(a feature that is readily accessible although higher dimensional).

\subsection*{Learning Social Media User Embeddings}
Although representing users as a vector of real numbers is not a new idea,
learning user embeddings and evaluating them as first-class objects is a new
contribution to social media
research.  In chapter \ref{chap:mv_twitter_users} we evaluate multiview user
embeddings at multiple tasks
in a similar way to how word embeddings have been subjected to a battery of
syntactic and semantic similarity tasks, as well as prediction tasks.
Subsequent research has also taken this tact and treats user embeddings as
first-class objects worth learning and evaluating in their own right.

For example, \textcite{xing2017incorporating} take inspiration from the
friend prediction task to evaluate their own user embeddings.  \textcite{li2017joint}
takes a multitask approach to learning user embeddings and evaluates the embeddings
according to how well they predict which text and other users they are likely
to agree with.  Although considering a supervised objective to learn user embeddings,
\textcite{kursuncu2018contextualized} also collapses features from each view into a
joint user embedding.  Multiview user embeddings have even been shown to
be predictive of sarcasm in author tweets \parencite{hazarika2018cascade}.

\subsection*{State-of-the-art for Social Media Mental Health Monitoring}
At the time of publishing, the neural MTL model presented in chapter
\ref{chap:mtl_mentalhealth} marked the state-of-the-art for
mental health inference from social media text.  \textcite{tran2017predicting}
reference this work as follows:

\begin{quote}
  There is also a quickly growing body of literature detailing machine learned
  models to predict mental health status based on social media data.  For a
  detailed analysis of the current state-of-the-art in this emerging domain,
  readers are encouraged to refer to the deep learning architecture by
  Benton et al. [6].
\end{quote}

Subsequent work has also taken semi-supervised approaches to monitor
mental health from users' social media data \parencite{yazdavar2017semi,zou2018multi}.
In particular, our finding that \emph{classification tasks with little labeled
  training data tend to benefit more from MTL than tasks with more training data}
is frequently cited as justification for the use of MTL \parencite{bingel2017identifying}
across many application domains: news headline popularity prediction
\parencite{hardt2018predicting}, information retrieval \parencite{salehi2018multitask},
medical concept normalization and recognition
\parencite{crichton2017neural,niu2018multi}, and NLP
\parencite{bjerva2017dissertation,schulz2018multi}.

\section{Ethical Considerations}
\label{sec:conclusion:ethics}

The work described in this thesis can lead to more robust social media systems,
benefiting the users whose data these models were tuned to.  At the same time,
it is important to recognize that although these analyses were purely
observational\footnote{We did not directly engage with the users whose data we
  collected or attempt to influence their behavior.}, the users' whose data
comprise our training sets may not be comfortable with being studied.  The
survival of social media research depends on the trust of the users whose data
are studied and betraying this trust will can have far-reaching effects\footnote{\url{https://www.nytimes.com/2018/04/10/us/politics/zuckerberg-facebook-senate-hearing.html}}.

There is a wide set of ethical concerns associated with doing social
media research, especially when dealing with users' physical and mental health.
These concerns are extensively covered by \textcite{mckee2013ethical},
\textcite{conway2014ethical}, and \textcite{ayers2018don}.
In \textcite{benton2017ethical} we condense these
into a set of maxims and describe ethical concerns that new social media health
researchers should be aware of.

In this thesis we made several explicit choices to respect the privacy of users
in our studies.  In Chapter \ref{chap:mv_twitter_users} we chose to only present
anecdotes of users that were obfuscated.  The user clusters in Appendix
\ref{app:user_clusters_mturk} do not include user IDs of cluster members, only a
bag of words associated with
exemplar users in that cluster.  We also made sure not to release tweets made
by these users as this
would explicitly break the Twitter REST API's terms of service (although we did
release the pre-trained user embeddings).

The work in Chapter \ref{chap:mtl_mentalhealth} is the most ethically treacherous.
In other chapters, user data is used to improve models that are only tangentially
related to these same users, or are used in innocuous tasks (e.g. improving topic
model fit, predicting hashtag use, and friending behavior).  Although the users
in the mental health prediction work made their accounts publicly available, they
probably did not expect to be enrolled in an observational study.  The stigmatism
surrounding mental illness prevented us from sharing their data or identities
with outside researchers.  We also explicitly chose to not release the models
we trained over their tweets, nor do we release an analysis of which features
were most influential in predicting mental health
conditions for fear that these character/word choice features would be used to
stigmatize others.

\section{Directions for Future Research}
\label{sec:conclusion:future}

\subsection{User Embedding Evaluation Suite}

Creating a benchmark evaluation suite set for user embeddings similar to word
similarity benchmarks for word embeddings
\parencite{rubenstein1965contextual,miller1991contextual,finkelstein2001placing,hill2015simlex}
would be a valuable addition to systematically comparing user embeddings.
Just like word embeddings, user embeddings can always be finetuned for a specific downstream
task, but it is not clear whether specific representation learning methods or data sources are
a better starting point for a wide array of tasks.  The evaluation in
chapter \ref{chap:mv_twitter_users} is a first step at constructing such a benchmark
suite: future hashtag, friend link, and demographic prediction.  However, there is
a night inexhaustible list of dimensions in which users can be similar to each other:
according to socioeconomic status, geographical origin, entertainment preference,
personality profiles, etc.  A suite that covers a diverse set of user
properties would better judge how useful a user embedding is.

\subsection{Scalable Multiview Representation Learning}

\paragraph*{Scaling Over Examples:} The MV-LSA approximate solution to MAXVAR-GCCA
presented in chapter \ref{chap:mv_twitter_users}
is a batch method that is difficult to scale up to very large datasets.  The barrier
to scalability arises from ensuring that
orthonormality constraints on the columns of $G$ are satisfied (columns span all
training examples).  Finding a GCCA embedding for billions of Facebook users would be
intractable under this algorithm.  Algorithms that solve CCA by updating projections based
on a single example at a time such as stochastic AppGrad \parencite{ma2015finding} or
stochastic CCA \parencite{arora2017stochastic} do not run into memory constraints as the
number of examples increases, but in practice may take far too long to converge to
a sufficiently low los solution.  Algorithms to scalably solve generalized CCA problems are
less mature than those for two-view CCA although there has been some work.  For example,
\textcite{fu2016efficient} gives an alternating optimization algorithm for the SUMCOR-GCCA
problem that decomposes the optimization as a sequence of linear least-squares problems
(the \lascca{} algorithm described in Section \ref{subsec:mv_twitter_users:sumcor_gcca}),
and \textcite{fu2017scalable} gives a similar algorithm for approximately solving a
regularized variant of MAXVAR-GCCA.

Neural architectures have also been proposed to
maximizing correlation between more than two views, but these architectures make additional
assumptions on which correlations are maximized between views.
For example, the Bridge Correlational Networks proposed in \textcite{rajendran2015bridge}
assume that one of the views is designated a ``pivot'' view whose representation all other
views are mapped close to.  On the other hand, generalized CCA formulations such as
SUMCORR-GCCA and MAXVAR-GCCA maximize a loss function dependent on correlation between all
pairs of views. Neural proposals also do not come with theoretical guarantees on solution
quality, and therefore may have trouble difficulty finding good user embedfdings in practice.

\paragraph*{Accounting for Noisy Views:} Certain example views will have higher variance
than others.  Imagine a user has only posted a
message once in their online life, but they have friended over 1000 other users.  The ego text
view for that user will be a noisier estimate of their true ego text distribution (when
they have produced infinite tweets) than their friend network view if they explicitly took
the time to vet every other user in the entire network as a potential friend.
Although the example-view binary mask in MV-LSA, $K$, explicitly ignores example views that are
missing data, it cannot tell how much to \emph{trust} each view.  \textcite{rastogi2015}
suggests a heuristic weighting of example views
in the section on ``Handling Missing Data''. This could be used to downweight views whose
estimates are noisier.
There has also been work on Bayesian online classifiers of user demographics
that gain confidence as more user information arrives, but it is unclear how to apply
these models in the multiview representation learning setting \parencite{volkova2015online}

\paragraph*{Scaling to Variable Feature Dimensionality:}
Another general problem in learning and updating multiview social media user embeddings is the
introduction of new social media members and new vocabulary as time goes on.
This will result in new features being introduced into the per-view feature vectors.
We are not familiar with work on extending the GCCA problem to cases where input feature
dimensionality may grow over time.  One possible direction is to incorporate new view
features into your model by refreshing the embeddings as a batch periodically.  This is
not a very satisfying solution and is expensive to apply at scale.  A second direction
would be to look to the nonlinear mappings in \dgcca{} to map new feature elements to the same
feature space.  Imagine for instance that one of our views is a representation of our local
friend network, but we take the mean of friend user description embeddings as our friend
network feature vector. Adding a new friend would only require embedding their
user description and integrating it into our view average. This is very much an open problem
and these solutions are not particularly satisfying.

\subsection{Interpretability of User Embeddings}

There has been work in interpreting the dimensions of textual embeddings
\parencite{li2015visualizing}.  One strength of distributed
user embeddings, their ability to encode many user features in a single dense vector,
is also their weakness -- user embeddings are opaque.  This technique
could be applied to user embeddings by calculating the distance between an input user
to prototypical users (e.g. man, woman, athlete, or a wine-snob).  Although this thesis
focuses on learning user embeddings to improve downstream tasks, interpretation of these
embeddings will be important in analyzing what they capture and convincing
engineers that they are worth including in production systems.
\cleardoublepage

\appendix

\cleardoublepage
\chapter{User Embedding Clusters}
\label{app:user_clusters_mturk}

Here we present the labels assigned by Mechanical Turk subjects to user clusters from
three different user embeddings.
Each cluster was represented by four user
exemplars (with one false exemplar/intruder), corresponding to a single Gaussian in a Gaussian
mixture model (subsection
\ref{subsec:mv_twitter_users:qual_analysis}).  Many of the cluster labels are vague, hinting
at the difficulty of this task.  For each cluster, we present the assigned labels
along with tokens from the four exemplar user tweets.

One important thing to be aware of is that different Gaussians capture more users than others.
This is likely because we assume all Gaussians have diagonal, but unconstrained covariance
matrices -- a single Gaussian can have very large variance and capture many users that do
not fit neatly into other, narrower Gaussians.  The \emph{Number Members} column in
the tables below contains the number of users belonging to each cluster (assigned highest
probability under mixture model).  Next to each label is a sign for whether the subject
correctly identified the intruder ($+$ for correct, $-$ for incorrect).

\pagebreak

\section*{PCA on Ego Text}

\begin{longtable}{|p{0.1\textwidth}p{0.15\textwidth}|p{0.35\textwidth}|p{0.4\textwidth}|}
 \hline
  \rowcolors{2}{gray!50}{white}
  \small
  \centering
  \bf Cluster Index &  \bf Number Members & \bf Labels & \bf User Text  \\ \hline
 1 &  1148 &  \emph{Tweeting personal thoughts and opinions} ( - ) \emph{user 5 does not belong tweeting snoop dogg} ( - ) \emph{different language} ( - ) &  \emph{snap chat instagram} \emph{bitch bitch guilty guilty} \emph{} \emph{always following back toke} \emph{} \\
 2 &  960 &  \emph{other four are women} ( - ) \emph{The other 4 are female} ( - ) &  \emph{nah} \emph{tanto como tanto cmo quiera} \emph{architecture art politics good day three time} \emph{} \emph{instagram askfm} \\
 3 &  374 &  \emph{Personal accounts talking about their personal lives and interests} ( - ) \emph{The other four seem less professional} ( - ) &  \emph{film production photographer olympic swimming online editor} \emph{words stuck mind alive i've learned seize day home returned} \emph{july} \emph{} \\
 4 &  208 &  \emph{The others seem like homebodies} ( + ) \emph{It doesn't look like everyone else have kids.} ( - ) &  \emph{proud mommy laila nicole dont fuckery dont ask past wont ask yurs kik \#teambi} \emph{face it's breathing air stop} \emph{dont followers} \emph{competition striving better woman yesterday} \\
 5 &  483 &  \emph{The other four are young people.} ( - ) \emph{The other four users are a younger age group} ( - ) \emph{They are younger people.} ( - ) &  \emph{loving life instagram} \emph{alexis nothing w/out god love help need forgive hurt pray leave ... romans} \emph{jesus freak instagram} \emph{boynton beach florida} \emph{christian wife mom beautiful girls humbled saving grace jesus} \\
 6 &  6057 &  \emph{other four are women} ( - ) \emph{The other four are young women} ( - ) \emph{Young women talking about their lives and interests} ( - ) &  \emph{} \emph{carpe diem} \emph{reality junkie !!! loves good crafty challenge} \emph{photography student} \\
 7 &  7097 &  \emph{@EvieMarieR is only professional account (Publicist @EdgePublicity.) The rest are personal accounts with no indication about their occupation} ( - ) \emph{People tweeting about personal, not professional, information} ( + ) &  \emph{} \emph{} \emph{baby world} \emph{everyone else already taken oscar wilde} \\
 8 &  1183 &  \emph{Male} ( - ) \emph{none} ( - ) &  \emph{self proclaimed coolest guy alive also i'm actor comedian new yorker enjoying} \emph{olvides nunca quien hizo bien pero que ese recuerdo} \emph{aficionado elche \#15} \emph{follow} \\
 9 &  188 &  \emph{Young people} ( + ) \emph{NONE} ( + ) &  \emph{make assessment based recent tweets ... don't} \emph{alphabetical order cats essex fella mine imac ipad iphone kids mine london movies music print publishing thfc travel} \emph{} \emph{aspire perspire inspire without god nothing} \\
 10 &  1374 &  \emph{English speakers} ( + ) \emph{none} ( - ) &  \emph{aos del lectora tributo lll amor que mereces ser} \emph{0/4 cute bitch every single one 1000 percent boyband tears} \emph{community activist \#hamont survivor realities poverty school hard knocks :-p ;-)} \emph{} \\
 11 &  92 &  \emph{other four are men} ( + ) \emph{user 5 missing Sagittarius} ( + ) \emph{They're all men.} ( + ) &  \emph{aint got mind \#father \#cardnation rip granny cora} \emph{asshole masters don't give fuck bachelors \#teamfollowback \#teamfreak \#teamokc} \emph{life short} \emph{simply organized mess chaos} \\
 12 &  10316 &  \emph{Young women talking about their lives} ( - ) \emph{They are all female twitter users.} ( + ) \emph{others are female} ( + ) &  \emph{don't get} \emph{sports physiologist love liverpool sports general \#jft96 \#ynwa} \emph{obsessing ugly animals trying act like know i'm} \emph{} \\
 13 &  4672 &  \emph{She's the only non-white person.} ( + ) \emph{The other 4 feel like they are into american culture} ( + ) &  \emph{} \emph{like sports stuff} \emph{} \emph{} \\
 14 &  820 &  \emph{the others are women} ( + ) &  \emph{fans teuku wisnu dan shireen akuntansi uir universitas islam riau} \emph{} \emph{i'm adventist} \emph{accounting university mks 011 september virgo simple \#viscabarca} \\
 15 &  7241 &  \emph{A politically active group of tweeters} ( - ) &  \emph{welcome lair ... it's filled cosplay cars princesses animals food sexy magic enjoy} \emph{like like \#tgod} \emph{} \emph{veteran south african profit organisation committed making human rights real live south africa} \\
 16 &  81 &  \emph{The others are men} ( - ) \emph{The other four are men} ( - ) &  \emph{creative} \emph{follow get follow back} \emph{proud directioner fun weird crazy loud chick ....... follow back 100} \emph{gunner @arsenal} \emph{guys follow new twitter yang twitter ini udah ngga pakai thankyou} \\
 17 &  2823 &  \emph{A bunch of young Americans tweeting about their lives in English} ( + ) &  \emph{} \emph{professional movie watcher} \emph{imitation sincerest form flattery} \emph{hello} \\
 18 &  3632 &  \emph{Communicating interests and personal information} ( + ) \emph{@petsarefriends is different as \r1. display picture of animals while all others have their own picture as DP\r2. most tweets are animal/pet based while all others are varied.} ( + ) &  \emph{dios track\&field boca juniors vida entera} \emph{i'm pet owner lover} \emph{tattoo artist hard knox tattoos central avenue yonkers check website} \emph{fewer 160 characters ask} \\
 19 &  135 &  \emph{Posting thoughts in text form} ( - ) \emph{@CentralFics is a website's official account. The other four are personal accounts of individuals.} ( - ) &  \emph{} \emph{proud directioner fun weird crazy loud chick ....... follow back 100} \emph{} \emph{hons fine art commonly known deedee lover thor criminal minds minions way life secret member expendables} \\
 20 &  1625 &  \emph{NONE} ( - ) \emph{1, 2, 4, and 5 all seem to be sports fans/tweet about sports. 3 doesn't.} ( + ) &  \emph{union local columbus .... lifelong buckeye fan ... usher ... pete rose belongs hof} \emph{} \emph{ironman finisher crossfitter veteran father kcco} \emph{love dallas cowboys till short live fun !!!} \\
 21 &  56 &  \emph{The other four seem a lot more mature based on their posts.} ( + ) \emph{They are all white people.} ( + ) \emph{@SarahYogiKAY is the only non-individual account. All other accounts are personal accounts while this belongs to an yoga class / institute} ( - ) &  \emph{one follow follow back cares \#teamaquarius \#teamidgaf !!!!} \emph{} \emph{every tweet cry help !!!} \emph{kay yoga brings songs games creative stories together teach kids ages joys yoga fun way learn mindful centered} \emph{} \\
 22 &  368 &  \emph{They're all individuals} ( + ) \emph{Young people} ( + ) &  \emph{} \emph{collins winter springs professionals strive reduce stress worry associated going dentist} \emph{} \\
 23 &  243 &  \emph{he's an athlete} ( + ) \emph{all are personal accounts} ( - ) &  \emph{fanbase indonesia share hottest news} \emph{i'm gentle man caring loving importantly i'm god's fear} \emph{pour another one} \emph{pianist singer artist luhan} \\
 24 &  4680 &  \emph{The others seem more happy with their lives} ( - ) \emph{The others are all women.} ( - ) &  \emph{nobody back start new beginning anyone start today make new ending !!!!} \emph{thoughts create reality keep positive stay lifted} \emph{sinner who's probably gonna sin lord forgive insta} \emph{crew --- bands --- rock roll ---} \emph{carson} \\
 25 &  1568 &  \emph{734477232 / @CornishAccounts is the only business account of a chartered accountant firm. The rest of the four are personal accounts of individuals.} ( + ) \emph{The other 4 are people not businesses} ( + ) &  \emph{chartered accountants business advisors gold xero partners cornwall xero accounting partner year 2015 south tweets} \emph{don't see things see} \emph{wife mommy three handsome boys seeking following god wonderful blessed life} \emph{written stars} \emph{junkie mad men addict @howardu reinventing status quo htown beyond} \\
 26 &  114 &  \emph{The first one is a website's twitter, the other 4 are personal accounts.} ( - ) \emph{NONE} ( + ) &  \emph{dance list ceo} \emph{entrepreneur technology evangelist} \emph{sedang dan menerima sesuatu istiqomah faculty nursing sky green} \emph{} \\
 27 &  454 &  \emph{The rest are not commercial accounts.} ( - ) \emph{none} ( - ) &  \emph{texan tarleton state crossfitter biology major justen} \emph{} \emph{casual youtube gamer looking bring comedy fun within call duty naruto game play love video games anime wwe mlp reading manga} \emph{holiday rental video tours web video marketing services rental owners agents extend market reach increase bookings} \emph{love hanging friends going amusement parks obsessed coasters love good time} \\
 28 &  319 &  \emph{The others are female Indonesians.} ( + ) \emph{Keeping track of followers via bots} ( + ) &  \emph{vida april 1993} \emph{sus ojos eran mova con saben muchas cosas} \emph{xliii spn pwt} \emph{jangan} \\
 29 &  1785 &  \emph{The other 4 are people not businesses} ( + ) \emph{Talking about a variety of issues but not focused on followers} ( - ) &  \emph{account berbagi informasi seputar teknik komputer politeknik negeri jakarta care share} \emph{\#team dreambig} \emph{} \emph{los das lluvia agua las calles cabeza} \\
 30 &  1721 &  \emph{Female} ( + ) \emph{NONE} ( + ) &  \emph{striving towards best self} \emph{laugh day you'll live happy life} \emph{god things} \emph{abigail sierra potter} \\
 31 &  1303 &  \emph{none} ( - ) \emph{Speak English} ( - ) &  \emph{tanner things possible} \emph{} \emph{bittersweet like caramel} \emph{he's broken cause ready everything} \\
 32 &  3420 &  \emph{The other profiles are all in foreign languages.} ( - ) \emph{Foreigners tweeting in their native languages, but not bots} ( - ) &  \emph{} \emph{} \emph{panggil aja yan yang juga boleh} \emph{anthony santos estudiante del \#19 mundo mil} \\
 33 &  7094 &  \emph{He has a much greater amount of followers, which is 1,282. Also, he is following so many more people, 1,700} ( - ) \emph{The others seem to be males} ( + ) &  \emph{rings like bell night wouldn't love love instagram} \emph{technology radio geek content controller northsound bauer media views} \emph{professional pool spa care csp afo cpo certifications} \emph{dad firefighter blogger rocker new block party !!!} \\
 34 &  510 &  \emph{The other four seem like real people, not bots} ( - ) \emph{Communicating with a world community, not just close friends} ( - ) \emph{They aren't afraid to show pictures of themselves in their profile pics.} ( - ) &  \emph{disfruta cada dia como ultimo} \emph{regular dude} \emph{give take world got takes} \emph{fsu} \\
 35 &  865 &  \emph{other four are young  women} ( - ) \emph{The other four are real people} ( + ) \emph{no human photo} ( - ) &  \emph{berks} \emph{summer 2013} \emph{real food recipes natural living side sass} \emph{ese momento ella beso mundo one direction justin bieber demi lovato jonas brothers cd9 5sos miley cyrus} \\
 36 &  3627 &  \emph{A group tweeting lots of business and/or promotional information} ( - ) \emph{the others speak the english language} ( + ) \emph{The other four involve people that speak english} ( + ) &  \emph{londoners} \emph{dad hobbit photographer project manager android enthusiast round gadget lover} \emph{account berbagi informasi seputar teknik komputer politeknik negeri jakarta care share} \emph{nyla marie \#proudmom} \\
 37 &  191 &  \emph{Posting daily horoscope links} ( + ) \emph{other four are women} ( - ) &  \emph{} \emph{lanta} \emph{ain't got bait takes hook \#teamvirgo \#teamsexy \#teamnosleep} \emph{welcome lair ... it's filled cosplay cars princesses animals food sexy magic enjoy} \\
 38 &  8309 &  \emph{none} ( - ) \emph{Young people} ( + ) &  \emph{sk8} \emph{} \emph{don't trip mind business} \emph{stay lane bahamas 242 r.i.p} \emph{law graduate aspiring solicitor} \\
 39 &  2318 &  \emph{Young people} ( + ) \emph{none} ( - ) &  \emph{} \emph{keperawatan nurse mahasiswa tingkat akhir} \emph{ambitious work hard play harder partner construction inc} \emph{snap acres} \\
 40 &  1139 &  \emph{the others are male} ( + ) \emph{A group of people tweeting about politics and general interests} ( - ) \emph{The other profiles seem to be involved in sports somehow (soccer)} ( + ) &  \emph{rings like bell night wouldn't love love instagram} \emph{} \emph{sports physiologist love liverpool sports general \#jft96 \#ynwa} \emph{\#afc @arsenal} \\
 41 &  681 &  \emph{A group of people tweeting personal thoughts and hobbies} ( - ) &  \emph{hit white ball around field sometimes} \emph{} \emph{good days i'm charming fuck multifandom} \emph{designer individualist infantryman} \\
 42 &  926 &  \emph{This one is not a full profile} ( - ) \emph{All seem to have an interest in popular music.} ( - ) \emph{the rest are in english} ( - ) &  \emph{una lugar donde usted puede perder sin perder} \emph{kind happened} \emph{life short unhappy smile darn} \emph{} \\
 43 &  1589 &  \emph{This user is in a foreign language, maybe Spanish and the others are in English.} ( - ) \emph{The others did ot say they like britney spears} ( + ) \emph{Personal accounts talking about their interests and lives} ( - ) &  \emph{tengo tantas cosas que decir que ahogo parece ser que dolor como tesoro} \emph{god sapiosexual ... love music frank ocean yamo pisces} \emph{sharing love local 1st gift card exclusively independent shops services near} \emph{amante vida hacer rer gente ser buen hijo hermano amigo gran fan @britneyspears} \\
 44 &  2446 &  \emph{All are young women, college age or somewhat older.} ( - ) \emph{the others seem like they are not very adventurous} ( - ) &  \emph{nyc} \emph{} \emph{positive} \emph{fab} \\
 45 &  1786 &  \emph{The other 4 are not black.} ( - ) \emph{The other four are young} ( + ) \emph{People sharing personal non-celeb related thoughts and quotes} ( - ) &  \emph{follower christ texan} \emph{paige baby|} \emph{finnish girl loves beauty stuff good music ice hockey} \emph{..... never said perfect nobody walkin earth ....} \\
 46 &  1673 &  \emph{They appear to be nothing more than bots} ( + ) \emph{NONE} ( - ) &  \emph{without struggle progress} \emph{handsome clever boy always liked boy true humorist} \emph{real hustler ontop game gang} \emph{smart gorgeous gurl gurl ain't gat tym type bio buh willing knw knw wah ryt \#team gemini \#peace} \\
 47 &  650 &  \emph{Foreign accounts talking about life in their native language} ( - ) \emph{different language} ( - ) \emph{They all have English as common but the user 5 in some other language.} ( - ) &  \emph{} \emph{instagram} \emph{hear birds summer breeze} \emph{inna maal yusra optimis sesuatu yang tapi tetap} \\
 48 &  702 &  \emph{User 4 is a man, the rest are women. Also, user 4 is the only one who mentions that he is a parent.} ( - ) \emph{NONE} ( - ) &  \emph{actions good thoughts positive energy speak louder judgmental words powerful tools use working toward better world} \emph{felicidad esta dentro uno ado nadie} \emph{texas beer gypsy time's person year 2006 opinions sales marketing manager red caboose winery} \emph{} \\
 49 &  593 &  \emph{business} ( + ) \emph{Personal accounts sharing opinions, mundane details, and interests} ( + ) &  \emph{} \emph{arte} \emph{} \emph{hey everyone god joy strength life stay faithful} \emph{} \\
 50 &  689 &  \emph{They have in common that they can understand foreign language and there is no English in their comments under the profile. They might not be native Americans.} ( + ) \emph{Foreign accounts tweeting in their native languages} ( + ) &  \emph{fotgrafo reportero del centro argentino estudiante perro} \emph{bad life bad day} \emph{desde 2011 hasta ely sueo importa quien quiera antes querido gira tanto} \emph{nyc life home sweet home glad back} \\ \hline
 \caption{ \textbf{PCA on ego text} -- embedding cluster labels.\label{tab:user_embedding_clusters:pca_ego_qual_clusters} }
 \end{longtable}

\pagebreak

\section*{PCA on All Views}

\begin{longtable}{|p{0.1\textwidth}p{0.15\textwidth}|p{0.35\textwidth}|p{0.4\textwidth}|}
 \hline
    \rowcolors{2}{gray!50}{white}
  \small
  \centering
  \bf Cluster Index &  \bf Number Members & \bf Labels & \bf User Text  \\ \hline
 1 &  847 &  \emph{They are all young people} ( - ) \emph{none} ( - ) &  \emph{aos del lectora tributo lll amor que mereces ser} \emph{} \emph{living dream osu student} \emph{} \\
 2 &  3563 &  \emph{The others are  more about american culture} ( - ) &  \emph{} \emph{boyes} \emph{} \emph{} \\
 3 &  925 &  \emph{The others are women} ( - ) \emph{Young women tweeting about female interests} ( - ) &  \emph{} \emph{} \emph{sabe como sempre fui uma pessoa ...} \emph{health travel enthusiast cisco zeal thoughts donate team savannah} \\
 4 &  1075 &  \emph{none} ( - ) \emph{They are all young, hip people} ( + ) &  \emph{\#teamfollowback .... tweet 5'0 earth chick named would tweet} \emph{die bury inside gucci store} \emph{live life like theres tomorrow regrets love everything cause never know last moment} \emph{} \\
 5 &  710 &  \emph{@\_AskMeIf\_IGAF appears to be a bot as most tweets are automated. The other four are posting own tweets} ( - ) \emph{The other four are less offensive} ( - ) \emph{Communicates interest in pop culture and personal tidbits} ( - ) &  \emph{snapchat} \emph{siempre todo ingeniero civil obras muy esto catalan corazon jaja} \emph{\#teamvirgo \#teamfollowback ... catch .!! cant love hoe make weak ...} \emph{7teen cal family friends anything} \emph{sarcastic} \\
 6 &  1937 &  \emph{The other accounts have more substance} ( - ) \emph{Tweets from follower counters and picture platforms} ( - ) &  \emph{} \emph{desde los con una locura una felicidad mundo redondo} \emph{pin} \emph{dreamer hopeless romantic poet ass} \\
 7 &  3069 &  \emph{The other 4 are adults who speak english} ( + ) \emph{Group sharing personal tidbits, unprofessionally} ( - ) &  \emph{southern seaside newcastle supporting teacher ict geography leader like chocolate travelling} \emph{architect} \emph{rise ummah} \emph{dalton bean} \\
 8 &  114 &  \emph{none} ( - ) \emph{Identical bots} ( + ) &  \emph{} \emph{everything i'm made everything} \emph{} \emph{} \\
 9 &  4435 &  \emph{none} ( - ) \emph{Foreign people tweeting in their native languages} ( + ) &  \emph{never don't mind thing} \emph{brooklyn based street artist} \emph{} \emph{charity worker owner little soapy secrets empire mother wife love life love work hard play harder amo mucho} \\
 10 &  1982 &  \emph{The other four seem like actual people} ( - ) \emph{These users type in English.} ( - ) \emph{Men talking about their lives and interests} ( - ) &  \emph{husband father friend student artist soon game design} \emph{porno star prev life give dey want make feel like dey need \#teamleo} \emph{bout one character middle finger} \emph{fanbase @mi\_christychibi keep calm support @cherrybelleindo 4ever admin bff} \\
 11 &  35 &  \emph{Women tweeting about personal issues and interests} ( - ) \emph{They are all girls} ( - ) &  \emph{} \emph{dire che come dio non dadi non credo nelle} \emph{theres certain happiness bein silly rediculous} \emph{time succes comes work dictionary} \\
 12 &  126 &  \emph{Scorpios interested in astrology} ( + ) \emph{The others did not post about soccer.} ( - ) &  \emph{} \emph{trust select} \emph{i'm filipina bitch} \emph{experience} \emph{} \\
 13 &  3029 &  \emph{@PauliiiTCA is the only one tweeting in a non-english language (spanish). The other four users are tweeting in english.} ( - ) \emph{the rest appear to be male} ( - ) &  \emph{ian mitchell phd hcpc swansea city afc wales national team performance psychologist} \emph{hakuna matata} \emph{i'm singer/songwriter unsigned artist also follow instagram} \emph{sexo son} \emph{taylor lautner @avrillavigne @coldplay @eminem} \\
 14 &  7191 &  \emph{They are all women} ( - ) \emph{none} ( + ) &  \emph{\#teamfollowback surgical tech student aspiring model/actress ultimate dreamer} \emph{} \emph{naked dude squash} \emph{libra} \\
 15 &  431 &  \emph{Personal accounts not tweeting bot stats} ( + ) \emph{The other accounts feel more authentic} ( + ) &  \emph{artist specializing drawing sculpture printing graphic design} \emph{ironic memer youtube star} \emph{never give fix mistakes keep stepping} \emph{star ambassador} \\
 16 &  4615 &  \emph{might be a business} ( - ) \emph{Males discussing their lives and opinions} ( - ) &  \emph{freedom isn't free} \emph{getting money} \emph{francophone instagram live life like boss} \emph{wedding prewedding service} \\
 17 &  102 &  \emph{Girls talking about music and makeup} ( - ) &  \emph{auntie future superstar forever} \emph{want maccies} \emph{vivir con fuerza locura libertad msica alma instagram} \emph{follow instagram put mind achieve anything} \\
 18 &  150 &  \emph{The one that doesn't belong is a family with kids and the others are young and probably single.} ( + ) \emph{the others are adults, these are children} ( - ) \emph{Young adults expressing their opinions and interestes} ( + ) &  \emph{motivation} \emph{christ follower husband father friend} \emph{bonito encontrar amor vida todos los das misma persona} \emph{quiero sonrer siempre lado kidrauhl} \\
 19 &  1103 &  \emph{The other profiles did not talk about band or sports} ( - ) \emph{@DenunaArifandi is the only user tweeting in non-english language. He is an Indonesian.} ( - ) &  \emph{free thinker wine drinker par t-t ime investor occasional putt sinker views retweet necessarily endorsement} \emph{follower christ husband beautiful wife daddy amazing kids student discipleship pastor coffee lover stl cards indy colts fan} \emph{love guns roses wwe big cubs fan huge fantasy football baseball player} \emph{loves laugh trys things} \\
 20 &  6867 &  \emph{the others are young} ( - ) \emph{The others are coaches or motivational people.} ( - ) &  \emph{worrying wont stop bad happenin stop good bein enjoyed ...} \emph{mac pro athlete \#broncos \#heatnation jerz} \emph{e-book blogger information take areas life average extraordinary} \emph{since 1912 helping people organizations achieve outstanding results professional daily lives} \emph{sr. media strengthening specialist \#journalist \#trainer} \\
 21 &  56 &  \emph{He is a male who doesn't use English. The others are English-speaking females.} ( - ) \emph{Young women tweeting about their lives} ( - ) \emph{The others show off less skin} ( - ) &  \emph{trinity middle storm allstars} \emph{\#directioner \#lovatic \#1dfamily \#mixer} \emph{everything beautiful time rip lcpl nick forever hearts} \emph{cupcakes baker music lover dancing queen pink princess marketing major runner wellness blogger lover laughter sports lover indian girl} \\
 22 &  1057 &  \emph{They are american.} ( - ) \emph{Communicating sports or pop culture interests} ( - ) &  \emph{goin put gay saying} \emph{taylor love pink frank ocean cheetah print jake family friends 412} \emph{} \emph{tamo} \\
 23 &  560 &  \emph{The other four seem more easy going and laid back} ( - ) \emph{The others seem well adjusted} ( - ) \emph{A group of foreigners tweeting in different languages} ( - ) &  \emph{born raised dot say what's mind alot times that's random shit} \emph{drag make always looking new opportunity adventure say} \emph{bachelor engineering dude} \emph{want prove words} \\
 24 &  368 &  \emph{2-5 are all people, while number one seems to be some kind of website.} ( - ) \emph{none} ( - ) &  \emph{vivir con fuerza locura libertad msica alma instagram} \emph{corse telecom cinma entrepreneur ces tweets que moi} \\
 25 &  1195 &  \emph{Speak English} ( + ) \emph{Women, possibly all Hispanic} ( - ) &  \emph{} \emph{menos como forma vida lado los realidad donde sea pero con verdadero hermanas} \emph{cosplayer full-time art history student otl} \emph{daily astrology vivian owen author lucky stars astrology bringing ancient star wisdom modern-day life featured writer} \\
 26 &  1348 &  \emph{Communicating information relevant to their person or business instead of bots} ( - ) \emph{The others re all individuals} ( - ) &  \emph{husband literature enthusiast sga bee gees dadakan baca horison sekali lagi seneng lari} \emph{don't live something you'll die nothing snap} \emph{hospital gato projeto que felina horas por dia} \emph{life short unhappy smile darn} \\
 27 &  80 &  \emph{They appear to be mostly bot posts} ( + ) \emph{none} ( + ) &  \emph{arsenal life} \emph{hallo} \emph{proud directioner fun weird crazy loud chick ....... follow back 100} \emph{better god i'm promise xii path putri} \\
 28 &  684 &  \emph{This person seems famous and the others don't} ( - ) \emph{The others seem like young people who haven't figured out who they want to be in life yet.} ( - ) &  \emph{saya ank dari} \emph{competition striving better woman yesterday} \emph{amikom lillywhite temen} \emph{work make giving} \\
 29 &  3733 &  \emph{Men tweeting about various issues besides music} ( - ) \emph{This person is very vague and the others describe themselves better} ( + ) &  \emph{...} \emph{books yarn that's life} \emph{rip shanell gone never forgotten love 4ever brothers otf} \emph{ces dames pote tel papillon} \\
 30 &  1801 &  \emph{Interested in pop culture and celebrities} ( - ) \emph{The other 4 seem more conservative and relaxed} ( + ) &  \emph{huge mma fan undercover ninja phat suit improving skillz daily day ninja appear without suit avenge} \emph{belieber forever belieber love justin drew bieber !!! download @shots tell friend} \emph{we're alive free conoces orgullosa justin drew bieber mallette demetria devonne lovato hart} \emph{enamor con} \\
 31 &  665 &  \emph{Looks like a fan account} ( + ) \emph{the others are women} ( - ) &  \emph{} \emph{poeta palabras tan del estudio salud vida ...} \emph{fanbase @mi\_christychibi keep calm support @cherrybelleindo 4ever admin bff} \emph{mirror lie shows what's inside} \\
 32 &  86 &  \emph{Personal accounts of guys tweeting about their lives} ( + ) &  \emph{hustle like starving going hard gotta eat} \emph{smiling happy gay guy trying humble best possible way smile face times} \emph{twitter tumblr blog fangirls guide everything hamdan bin mohammed bin rashid maktoum} \emph{don't fucks many catch solo dolo \#teamcapricorn} \\
 33 &  2528 &  \emph{This one is male and a foreign musician. The others are females and pretty average.} ( - ) \emph{the rest are women} ( - ) &  \emph{fearfully wonderfully made psalm 139:14 fitness coach blessed beyond belief diggin deeper everyday} \emph{canadian raised i'm competition --- dream chaser getter} \emph{idk idc bastille 1975} \emph{bangladeshi musician} \\
 34 &  975 &  \emph{A group of college girls from around the world} ( - ) \emph{The other four are women} ( - ) &  \emph{music life god} \emph{geronimo anderson jr.} \emph{typical cute chick international relations fisip universitas katolik bandung} \emph{manajemen usu impossible line aditya pratama} \\
 35 &  5832 &  \emph{Young women talking about their lives and female issues} ( - ) \emph{They are all female.} ( - ) &  \emph{founder club} \emph{live laugh love} \emph{bartender workaholic random follow follow back} \emph{twitter 0/4 0/5} \\
 36 &  81 &  \emph{none} ( - ) \emph{Male} ( + ) &  \emph{kay yoga brings songs games creative stories together teach kids ages joys yoga fun way learn mindful centered} \emph{par t-t ime home brewer full time beer geek} \emph{follower christ husband tech nerd awesomesauce} \emph{} \emph{occupied things designed brewed necessarily often overlapping hub universe} \\
 37 &  704 &  \emph{The other four seem much skinnier} ( + ) \emph{This one is the only one in English} ( + ) \emph{This guy is the only English speaking person.} ( + ) &  \emph{alvin} \emph{sederhana tapi} \emph{disfruta momento} \emph{springsteen fanatic pittsburgher penn stater wwe fan} \\
 38 &  1095 &  \emph{A foreign group of people tweeting in their native languages} ( - ) \emph{the others are women} ( + ) \emph{these four users are you} ( + ) &  \emph{allah art pharmacist} \emph{born sleep} \emph{stalker} \emph{} \\
 39 &  1265 &  \emph{NONE} ( - ) \emph{People likely looking for a job} ( - ) &  \emph{estudiante comunicacin audiovisual \#happiness tdcs} \emph{shortage fault found amid stars} \emph{ankara university school medicine} \emph{things important make it's mind mind control ig-} \\
 40 &  635 &  \emph{The other users aren't smoking in their profile pic.} ( + ) \emph{Personal anecdotes and information} ( - ) \emph{all others are English} ( - ) &  \emph{getting money} \emph{nwmsu omaha} \emph{things important make it's mind mind control ig-} \emph{careful hurt could ruin life} \\
 41 &  1667 &  \emph{Personal accounts talking about their interests and opinions} ( + ) \emph{based on the images and what I can see he is not from not caucasian} ( + ) \emph{The other accounts are much older} ( - ) &  \emph{madrid} \emph{\#ravensnation} \emph{yamaha banshee oregon dunes} \emph{} \\
 42 &  154 &  \emph{Foreigners in native languages talking to a wide audience} ( + ) \emph{The other four seem to live a more American-like culture} ( - ) &  \emph{sometimes fun bad things} \emph{ponta portugal realmadrid allah always sma negeri} \emph{enjoy moment still last} \emph{} \emph{comedian lovers} \\
 43 &  9230 &  \emph{They all have over 100 followers.} ( - ) \emph{this person seems like a comedy account while the others seem real} ( - ) &  \emph{texan tarleton state crossfitter biology major justen} \emph{always thinking one meal ahead} \emph{mig man bor uppsala .\_. snapchat lite shr och} \emph{nerd viking engineering student destroyer worlds} \\
 44 &  9907 &  \emph{Girls talking about female issues and their lives} ( - ) \emph{the rest aren't from Australia} ( + ) &  \emph{stop worrying take chance} \emph{heyy year old aussie life taking life comes trying let anything get high school yeah} \emph{} \emph{} \\
 45 &  3685 &  \emph{The other accounts are actual people} ( - ) \emph{Personal accounts tweeting issues personal to them} ( - ) &  \emph{integrated medicine center aiming holistic improvement health bring efficient alternative complementary treatments} \emph{city ward citizen helpers inc elder care compete remodeling services advisor} \emph{} \\
 46 &  2568 &  \emph{Tweeting mostly in text, not just pictures from another platform} ( - ) \emph{this looks like the only fan account} ( - ) &  \emph{fan account dez duron spreading word talented artist tea enthusiast dreams black white} \emph{rnb pizza lover| dancer singer} \emph{} \emph{} \\
 47 &  476 &  \emph{An interest in religion, particularly, Jesus} ( + ) \emph{Christians making mention of their faith} ( + ) &  \emph{may boast christ crucified alone son student sunday school teacher bible study leader} \emph{remain love john} \emph{} \emph{lover jesus president walk incorporation} \\
 48 &  1195 &  \emph{Men tweeting about non-professional lives and interests} ( - ) \emph{The other four seem older} ( - ) &  \emph{came drink milk kick ass i've finished milk one lab accident away super villian} \emph{} \emph{basketball writer china press man loves basketball \#lfc \#ynwa \#faith \#believe \#basketballneverstops} \emph{official fanbase account fitri} \\
 49 &  5880 &  \emph{Largely anonymous and impersonal, tweeting about specific interests} ( - ) &  \emph{belieber} \emph{like mauve} \emph{} \emph{} \\
 50 &  512 &  \emph{none} ( - ) &  \emph{} \emph{\#blessed} \emph{fans final fantasy kingdom classic} \emph{} \\
 \hline
 \caption{ \textbf{PCA on all views} -- embedding cluster labels.\label{tab:user_embedding_clusters:pca_all_qual_clusters} }
 \end{longtable}

\pagebreak

\section*{\dgcca{} on All Views}

\begin{longtable}{|p{0.1\textwidth}p{0.15\textwidth}|p{0.35\textwidth}|p{0.4\textwidth}|}
 \hline
  \small
  \centering
 \bf Cluster Index &  \bf Number Members & \bf Labels & \bf User Text  \\
 \hline
 1 &  48 &  \emph{the others are non business types} ( - ) \emph{A group of male sports fans} ( - ) &  \emph{huge cleveland sports fan i'm absolutely hilarious least} \emph{made hawaii giants 49ers} \emph{snap chat instagram} \emph{gomez class 2012 world cup 2014 dream come true \#mexico god family} \emph{saint mary's alumnus poker junky state farm insurance} \\
 3 &  57 &  \emph{the others seem to be more young} ( - ) \emph{Personal opinions on life and culture} ( + ) &  \emph{kinda shy unpopular kid lover music i'm going broadway knw i'll bak} \emph{\#weare} \emph{psu panda lover music lover fall boy hell yah !!!! maroon5 take concert} \emph{things christ strengthens -philippians 4:13} \\
 4 &  80 &  \emph{the other four are skinnier women} ( - ) \emph{Young women talking about love and partying} ( - ) &  \emph{hunters foxes} \emph{live live right enough} \emph{life short regrets} \emph{happily married mummy got perfect husband feel blessed crazy little life} \\
 5 &  32 &  \emph{this one appears to be a musician} ( - ) \emph{Foreign accounts tweeting in their native language} ( + ) \emph{the others are female} ( - ) &  \emph{fight achieve dreams} \emph{fearfully wonderfully made psalm 139:14 fitness coach blessed beyond belief diggin deeper everyday} \emph{rest peace boo baby sister always love} \emph{rhymes make sick} \\
 6 &  594 &  \emph{The others are individual people} ( - ) \emph{They are in one form or another involved with Christianity or a church.} ( + ) \emph{Christians expressing their faith} ( + ) &  \emph{generations christan church connecting people christ} \emph{christ follower husband father friend} \emph{nyc} \emph{queen's palace chosen enter i'm royal priesthood conqueror sick broke blessed friend jesus} \\
 7 &  112 &  \emph{Female-centric tweets and interests} ( - ) \emph{The others have more of a human touch} ( - ) &  \emph{non-profit providing services job seekers employers \#followback updates career events advice openings job fairs} \emph{currently starring reality show titled modern cinderella one girls search love shoes} \emph{super mom business owner pageant girl host cool chick radio} \emph{take inches single treatment instant body sculpting results natural without surgery} \\
 9 &  182 &  \emph{The others are young} ( - ) \emph{The other four are younger individuals in their 20s} ( - ) &  \emph{\#teamshady \#teamfollowback \#stan} \emph{live music moments nobody lives forever life really really short} \emph{instagram} \emph{female rebel native pride call orange grove high school senior} \\
 10 &  92889 &  \emph{Girls directly tweeting thoughts relevant to their personal lives} ( + ) \emph{The other accounts speak english} ( - ) &  \emph{gonna eat tots} \emph{beware suss} \emph{} \emph{} \\
 12 &  272 &  \emph{the other accounts dont seem to represent individual real people.} ( - ) \emph{There's one other business account, but for the most part the other four are personal and political while this one is promotional} ( - ) &  \emph{alberta top notch movers moving needs edmonton please contact website call} \emph{} \emph{work tirelessly provide clients high quality language branding solutions language problem follow back} \emph{engineer/ scholar love \#pakistan living kpk} \\
 13 &  322 &  \emph{Number 2 is japanese and only tweets in japanese. The others are tweet in English.} ( + ) \emph{none} ( + ) &  \emph{screaming guitar dark humor jack jameson recording engineer training hard shell soft heart} \emph{never regret mistake made one point it's exactly wanted ... bbm pin} \emph{thought toaster possessed put poptart caught fire} \\
 14 &  62 &  \emph{These users type in English.} ( - ) \emph{Personal accounts talking about their lives in a personal way} ( - ) &  \emph{\#warrior someday .... \#belieber \#mahomie} \emph{\#begreat} \emph{currently starring reality show titled modern cinderella one girls search love shoes} \emph{mejor sin duda fan del mejor dolo del mundo} \\
 15 &  148 &  \emph{The others seem like mature adults} ( + ) \emph{They're all women with very positive attitudes.} ( - ) \emph{A group of younger people tweeting mundane details about personal matters} ( + ) &  \emph{disneyland cast member attractions enthusiast} \emph{english studies student stirling uni} \emph{florida snapchat instagram} \emph{never get wet never expose bright light never ever feed midnight} \\
 16 &  130 &  \emph{Young people posting about their lives} ( - ) \emph{NONE} ( + ) &  \emph{official account pearl block report spam mentions} \emph{follow genesis \#sagittarius} \emph{florida snapchat instagram} \emph{} \\
 17 &  34 &  \emph{This user is male and all the others are female} ( + ) &  \emph{} \emph{rings like bell night wouldn't love love instagram} \emph{nigeria base rapper song writer composer cash flow entertainment ... booking contact} \emph{three} \\
 18 &  376 &  \emph{The others are women} ( - ) \emph{A group of girls tweeting random thoughts} ( - ) \emph{last four seem to be women} ( - ) &  \emph{love even though famous still humble boy canada proud \#belieber saw july 10th 2013} \emph{llamo cami encanta justin perfecto justin} \emph{years basketball hiphop jazz} \emph{i'm awesome} \\
 19 &  136 &  \emph{none} ( - ) \emph{The four users are all young males and seem to have stereotypical "male-related" interests like sports, gaming, etc.} ( - ) &  \emph{follow new account} \emph{football cooking sleeping life ... yeah playing xbox} \emph{miss understanding follow i'll follow back :)))} \emph{love guns roses wwe big cubs fan huge fantasy football baseball player} \\
 20 &  736 &  \emph{NONE} ( - ) \emph{They are all young people} ( - ) &  \emph{auntie future superstar forever} \emph{longtime fitness enthusiast big time lover certified personal trainer holistic nutritionist wannabe runner yogi self taught tea guru} \emph{photograper dreamer fashion media student fashion stylist} \emph{argentina alegra sueo gracias @donniewahlberg} \\
 21 &  37 &  \emph{the four other seem to be younger women, this one has a panda and is in korean.} ( - ) \emph{Girls talking about their lives and interests} ( - ) \emph{different language than the others} ( - ) &  \emph{mostly tweet pancakes disney like people let's friends} \emph{carpe diem} \emph{ball light reborn christ shine even darkest nights} \emph{rowan instagram snapchat} \\
 22 &  96 &  \emph{@AtikahLey appears to be bot as all her tweets are automated. The other four post their own tweets} ( + ) \emph{the others are women} ( - ) \emph{A group of people with bots to check their twitter stats and followers} ( - ) &  \emph{allah mbf instagram} \emph{} \emph{} \emph{love guys cooking nascar dale \#88} \\
 23 &  121 &  \emph{none} ( - ) \emph{Gay men} ( - ) &  \emph{amante msica gym lady gaga little monster born way} \emph{valley boy life play puck since giver get fuck way follow beauties} \emph{} \emph{} \\
 24 &  21 &  \emph{Female} ( - ) \emph{NONE} ( + ) &  \emph{don't like someone try walking mile shoes still don't like you're mile away shoes} \emph{jackson followed wesley stromberg followed kenneth lowery} \emph{} \emph{} \\
 25 &  228 &  \emph{business} ( - ) \emph{Socially conscious and inspiring accounts} ( - ) &  \emph{amo mis amigos familia buena msica feminista atea poco loca estudiante hincha racing} \emph{certified teach dr. tai chi health programs teacher various locales write column home page blog meditation motion} \emph{stay young wild one} \emph{hunters foxes} \\
 26 &  55 &  \emph{Seems to be the only English speaker.} ( + ) \emph{They are all foreign speaking individuals who live somewhere in the middle East.} ( + ) \emph{Foreign women tweeting in their native language} ( + ) &  \emph{gonna eat tots} \emph{10\% genius 90\% idiot} \emph{everlasting @siwon407 endless @allrisesilver 94l} \emph{admiring without desiring} \emph{empty ......} \\
 27 &  62 &  \emph{A group of English speaking sports and motivational quote fans} ( + ) \emph{He seems older than the others} ( - ) &  \emph{} \emph{would attempt knew could fail} \emph{} \emph{|19| you're favorite muggle fangirl band members youtubers} \\
 28 &  476 &  \emph{A group of females with inspirational/ health related quote and interests} ( + ) \emph{language} ( - ) &  \emph{oos letras profesional sarcasmo mierda} \emph{living life man} \emph{} \emph{either like don't care life} \\
 29 &  59 &  \emph{She doesnt look like shes smiling in her profile pictures like the others. The other 4 have better bios. She doesnt tweet in English at all while the others do} ( - ) \emph{language} ( - ) &  \emph{line path nida tanjung} \emph{striving towards best self} \emph{committed constituents life} \emph{percussion} \\
 30 &  69 &  \emph{none} ( - ) \emph{Real accounts, not fraud/bots} ( - ) &  \emph{lively disposition delighted anything ridiculous} \emph{beautiful life short live} \emph{full time \#vip \#bbc} \emph{positive mind positive vibes positive life} \emph{pain discipline better pain regret} \\
 31 &  2502 &  \emph{A diverse group of people tweeting about their diverse lives, not just one specific interest} ( + ) \emph{The other 4 are young} ( - ) &  \emph{conservative minded free thinker loves god guns country retired usaf snapchat kik} \emph{belum baik dan pengen jadi baik kalau pun udah baik pengen dan harus lebih baik} \emph{gaeilge teo tones} \emph{} \\
 32 &  286 &  \emph{Sharing personal thoughts, not follower information} ( - ) \emph{rest are women} ( - ) \emph{The other four are young women} ( - ) &  \emph{instragram .... vuelvo dormir ... soy tipo simple} \emph{ser feliz con locura} \emph{} \emph{mujer sentimental} \\
 34 &  57 &  \emph{Young women talking about their lives and views on the world} ( + ) \emph{most creative picture} ( + ) &  \emph{god family dacc nursing major dreams come true courage pursue instagram} \emph{} \emph{xxiv} \\
 35 &  91 &  \emph{only company} ( - ) \emph{The other four feel like real people and this account feels like its meant to push promotions} ( - ) \emph{People talking about pop culture and entertainment as opposed to personal lives} ( + ) &  \emph{} \emph{} \emph{breathe live carry} \emph{follow} \\
 37 &  205 &  \emph{NONE} ( - ) \emph{Men} ( + ) &  \emph{yid end come \#coys \#ukip} \emph{\#ttid \#coys} \emph{smart gorgeous gurl gurl ain't gat tym type bio buh willing knw knw wah ryt \#team gemini \#peace} \emph{i'm matty space cowboy billionaire liar ... instagram snapchat \#thfc \#coys} \\
 38 &  36 &  \emph{The others are non business accounts} ( + ) \emph{Personal accounts talking about their lives and views} ( + ) &  \emph{exceptional representation defending rights children interests times 877} \emph{worldwide dirtbag} \emph{\#23 \#bayarea \#dubnation} \emph{lively disposition delighted anything ridiculous} \\
 39 &  176 &  \emph{All seem to live in the United States of America.} ( - ) \emph{the others seem more happy with life} ( - ) \emph{The others are females with pretty basic female interests unlike the male with a political bent.} ( - ) &  \emph{changes wish see world mahatma ghandi} \emph{bad 90s boy bands diet coke} \emph{time succes comes work dictionary} \emph{tired called relationship} \\
 40 &  382 &  \emph{the others have a profile with a human touch} ( + ) &  \emph{capture dream make real} \emph{5sos fall boy screamau} \emph{directioner july 2010 niall harry louis} \emph{graduated high school currently attending year college love gaming enjoy making videos interacting subs that's} \\
 41 &  48 &  \emph{none} ( + ) \emph{The 4 profiles belong to girls, number 3 is a guy.} ( + ) &  \emph{} \emph{leaders always followers it's guy going walk} \emph{here's good times} \emph{keep smiling change someones day} \\
 43 &  141 &  \emph{The other four seem to be more conservative} ( - ) \emph{the other four are regular people, this appears to be a musician} ( - ) \emph{Foreign accounts tweeting in their native languages} ( + ) &  \emph{wife mom esthetician chocolate lover book reader embarrassing dancer slightly ocd someday traveler life lover always moments notice away crazy} \emph{} \emph{mature honeybee} \emph{art way run away without leaving home ask.fm:} \\
 44 &  41 &  \emph{The others are more into american culture} ( - ) \emph{Young men tweeting about personal lives and interests} ( - ) \emph{The other 4 use one language while this account uses at least two} ( - ) &  \emph{use smile change world don't let world change smile} \emph{panam} \emph{\#yrn} \emph{invisible family} \\
 45 &  43 &  \emph{A group of foreign people tweeting text and words, not just pictures} ( - ) \emph{The other 4 seem to be more dedicated to Hispanic culture} ( + ) &  \emph{avila tierra que sabe cantante guitarrista jornada completa} \emph{con arriba tengo corazn que llora por que} \emph{disfruta cada dia como ultimo} \emph{kismet roll dice raise stakes leave rest faith} \\
 46 &  223 &  \emph{none} ( + ) \emph{1, 3, 4, and 5 are girls. 2 is a guy.} ( + ) &  \emph{mexicano aos grammar nazi enamorado del oblivion las mujeres que} \emph{} \emph{idk idc bastille 1975} \emph{fab} \\
 47 &  38 &  \emph{The others seem like more independent individuals.} ( - ) &  \emph{} \emph{cant say something nice dont say nothing} \emph{student brother son athlete thinker} \emph{i'm awesome} \\
 48 &  177 &  \emph{Food related businesses} ( + ) \emph{They are all twitter accounts for adults or businesses.} ( + ) \emph{All promoting businesses.} ( + ) &  \emph{proud winners best product year 2013 food producer year 2012 hampshire life food drink awards} \emph{we're traditional microbrewery based heart kent tenterden} \emph{taekwondo dream try pray success} \emph{consultant publisher writer rock roll singer norwich ambassador norwich tourist guide} \\
 49 &  47 &  \emph{All of them have profile pictures of themself exvept the one picked} ( - ) &  \emph{} \emph{} \emph{} \emph{living dream snapchat} \\
 50 &  53 &  \emph{none} ( - ) \emph{English speakers} ( - ) &  \emph{add snapchat instagram kik telegram} \emph{mad cat lady} \emph{i'm half walrus half potato} \emph{walk new way hip hop} \\ \hline
 \caption{ \textbf{\dgcca{} on all views} -- embedding cluster labels.\label{tab:user_embedding_clusters:dgcca_qual_clusters} }
\end{longtable}

\printbibliography
\cleardoublepage

\chapter*{Vita}

Adrian Benton received the B.A. degree in Linguistics from the University of
Pennsylvania in 2008.  He received the M.S. degree in Computer Science from the
University of Pennsylvania in 2012, and enrolled in the Computer Science Ph.D.
program at Johns Hopkins University in 2013.
His research centers around applying machine learning techniques to analyze
social media data.

\end{document}